\documentclass[manuscript,screen,accept]{acmart}
\usepackage[utf8x]{inputenc}

\AtBeginDocument{%
  }

\setcopyright{acmcopyright}
\copyrightyear{2022}
\acmYear{2022}
\acmDOI{XXXXXXX.XXXXXXX}

\usepackage{array}
\usepackage{booktabs}
\newlength{\ColWidthNormal} 
\newlength{\ColWidthRowHeader} 

\newlength{\RuleOffsetLeft} 
\newlength{\RuleThicknessNormal} 

\newcolumntype{C}{>{\centering\arraybackslash\leavevmode}p{\ColWidthNormal}}
\newcolumntype{R}{>{\raggedleft\arraybackslash}p{\ColWidthNormal}}
\newcolumntype{L}{>{\raggedright\arraybackslash}p{\ColWidthRowHeader}}

\usepackage{soul}
\definecolor{beaublue}{rgb}{0.74, 0.83, 0.9}
\sethlcolor{beaublue}

\usepackage{amsmath}

\usepackage[caption=false]{subfig}

\usepackage{multirow}  
\usepackage{array}        

\usepackage{adjustbox}
\usepackage[graphicx]{realboxes}
\usepackage{rotating}

\begin{document}
\title{A Comprehensive Review on Autonomous Navigation}

\author{Saeid Nahavandi}
\email{saeid.nahavandi@deakin.edu.au}
\author{Roohallah Alizadehsani}
\email{r.alizadehsani@deakin.edu.au}

\author{Darius Nahavandi}
\email{darius.nahavandi@deakin.edu.au}

\author{Shady Mohamed}
\email{shady.mohamed@deakin.edu.au}

\author{Navid Mohajer}
\email{n.mohajer@deakin.edu.au}

\author{Mohammad Rokonuzzaman}
\email{m.pappu@deakin.edu.au}

\author{Ibrahim Hossain}
\email{i.hossain@deakin.edu.au}

\affiliation{%
  \institution{Institute for Intelligent Systems Research and Innovation (IISRI), Deakin University}
  \streetaddress{75 Pigdons road}
  \city{Geelong}
  \state{Victoria}
  \country{Australia}
  \postcode{3216}
}

\renewcommand{\shortauthors}{Nahavandi et al.}

\begin{abstract}
The field of autonomous mobile robots has undergone dramatic advancements over the past decades. Despite achieving important milestones, several challenges are yet to be addressed. Aggregating the achievements of the robotic community as survey papers is vital to keep the track of current state-of-the-art and the challenges that must be tackled in the future. This paper tries to provide a comprehensive review of autonomous mobile robots covering topics such as sensor types, mobile robot platforms, simulation tools, path planning and following, sensor fusion methods, obstacle avoidance, and SLAM. The urge to present a survey paper is twofold. First, autonomous navigation field evolves fast so writing survey papers regularly is crucial to keep the research community well-aware of the current status of this field. Second, deep learning methods have revolutionized many fields including autonomous navigation. Therefore, it is necessary to give an appropriate treatment of the role of deep learning in autonomous navigation as well which is covered in this paper. Future works and research gaps will also be discussed.
\end{abstract}

%

\keywords{Autonomous navigation, SLAM, Obstacle avoidance, Sensor fusion, Path planning, Robotic simulation}

\maketitle

\section{Introduction}
Robotics have impacted our lives in many ways. Backed up by technological advancements, robots have found their way in several application domains such as
medical \cite{di2021medical,barducci2019adaptive,li2018magnetically,sitti2018miniature},
military \cite{patil2020survey,ha2019robotic,packbot},
industrial \cite{huynh2020modelling,bader2020methodology,kuka-industial-robots,LevineWA15,yu2021roboassembly},
space \cite{nesnas2021autonomy,maurette2003mars,volpe1996rocky,maimone2006autonomous,bajracharya2008autonomy},
agricultural \cite{libin2008overview,525635,foglia2006agricultural,edan1995design,zhang2020high,malavazi2018lidar}, etc. Adding autonomous navigation ability to robot platforms boosts their performance significantly since they can reach wherever they are needed on their own. This motivation has driven researchers to push the autonomous navigation technology to its limits. Given the rich and fast-evolving literature on the topic of autonomous navigation, it is necessary to prepare literature surveys on regular basis. This way, experienced researchers, as well as newcomers, can get an insight into the current state-of-the-art in autonomous navigation which is the inspiration for this survey.

The journey toward autonomous mobile robots has been started with development of first-ever general-purpose mobile robot named Shakey \cite{nilsson1984shakey}. Developed by Artificial Intelligence Center of Stanford Research Institute, Shakey had the ability to reason about its own actions. Another notable mobile robot is CART \cite{adams1961remote,stanford-cart} which was capable of line following using a camera and a radio control link. The required processing was done on a mainframe computer which was connected to CART via the aforementioned radio link. There have also been attempts at the development of mobile robots for space missions. For example, Lunokhod \cite{kassel1971lunokhod,Lunokhod1} was developed by Soviet Union for moon exploration or the United States developed Viking Mars landers (VL-1, and VL-2) for measurement of atmospheric temperature, wind speed and direction, and pressure \cite{hess1977meteorological} as well as Mars rover Sojourner \cite{mishkin1998experiences}.

Autonomous navigation has made constant progress over the years. Autonomous mobile robots were developed to drive \cite{Dickmanns-car,keirsey1987autonomous,dickmanns2007dynamic,benz-autonomous-driving,pomerleau1994defense}, operate as medical assistance in hospitals \cite{nocks2007robot}, explore active volcanoes (Dante I and II) \cite{krotkov1994lunar,bares1999dante}, etc. The progress has even been boosted by holding international competitions focused on autonomous mobile robots. The DARPA grand challenge held in 2004 \cite{darpa-grand2004} was a turning point in the field of autonomous mobile robots. In this competition, autonomous cars were tasked to navigate a 240km off-road path without human intervention. Although none of the contenders managed to reach the finish line, valuable lessons were learned from the experienced failures. The challenge was repeated a year later and this time, five of the autonomous vehicles reached the finish line; among them, Stanley \cite{thrun2006stanley} from Stanford University claimed first place. In 2006, yet another grand challenge was held but this time in urban environments \cite{darpa-urban-challenge}. The objective was to navigate through an urban area while complying with traffic regulations and avoiding collisions with obstacles. The winner of this contest was vehicle Boss of team Tartan \cite{urmson2007tartan} developed in a collaborative effort by Carnegie Mellon University and General Motors.

Researchers have also worked on multi-robot setups. Swarm-bots project \cite{Gro-etal:amire2005} was focused on self-organizing swarm of simple robots to achieve complex tasks. Swarm of autonomous underwater vehicles for performing stealthy underwater missions has also been investigated yielding a method called Suave (Swarm Underwater Autonomous Vehicle localization) \cite{liu2014suave}. This method reduces the frequency of going to water surface for obtaining position information from external sources such as satellites. This way, the probability of being compromised for the stealthy underwater mission is reduced. In 2010, there has also been an international contest called MAGIC (Multi Autonomous Ground-Robotic International Challenge) \cite{aplin20101025} devoted to group of autonomous robots working together.

Obviously, multiple survey papers have already been published on the subject of autonomous mobile robots. To highlight the differences between our survey and existing ones, table \ref{compare-to-prev-works-table} has been prepared in which existing surveys have been sorted according to their publication year. As can be seen in table \ref{compare-to-prev-works-table}, some of the topics are missing in previous surveys but are covered in this paper. For example, Niloy et al. \cite{niloy2021critical} have only focused on indoor autonomous navigation. The survey by Pol and Murugan \cite{pol2015review} is also limited to indoor navigation except that human presence in the environment has been considered.

Mohanty and Parhi \cite{mohanty2013controlling} have primarily focused on requirements of global and local path planning and navigation. In Guzel's survey \cite{guzel2013autonomous}, the primary topic is mapless, vision-based navigation but some map-based methods have been reviewed as well. However, SLAM and several other important topics are absent. Injarapu and Gawre \cite{injarapu2017survey} have only considered path planning and legacy methods for obstacle avoidance. Pandey et al. \cite{pandey2017mobile} have divided mobile robot navigation algorithms into three groups deterministic, non-deterministic, and evolutionary algorithms. Similar to some other surveys, Pandey et al. divide navigation into global and local \cite{ni2016bioinspired}. Again, some topics such as SLAM and modern obstacle avoidance methods are missing from \cite{pandey2017mobile}. Victerpaul et al. \cite{victerpaul2017path} have special emphasis on path planning approaches and have also covered related topics such as navigation and legacy obstacle avoidance. Tzafestas \cite{tzafestas2018mobile} has covered most of the topics but lacks the review of modern obstacle avoidance methods, simulation tools, etc. Alatise and Hancke \cite{alatise2020review} have covered multiple topics such as sensor types, autonomous mobile robots challenges, classical obstacle avoidance methods, sensor fusion, etc.; however, other important topics such as SLAM and newer obstacle avoidance methods based on reinforcement learning (RL) or deep learning (DL) have not been covered.

Ibanez et al. \cite{sanchez2021path} have reviewed path planning methods with special focus on path planners for ground vehicles which can be extended to other platforms such as autonomous boats. However, SLAM, modern obstacle avoidance, sensor fusion, etc. are missing from \cite{sanchez2021path}. Panigrahi and Bisoy \cite{panigrahi2021localization} have not investigated path following/planning, simulation tools, etc. Zghair and Al-Araji \cite{zghair2021one} have investigated the evolution of control systems of mobile robots for the past decade. Again, modern obstacle avoidance methods and simulation tools are missing from their survey.

\begin{table}[]
\caption{Comparison between our survey and previous ones.}\label{compare-to-prev-works-table}
\begin{tabular}{|c|c|c|@{\hspace{1pt}}c@{\hspace{1pt}}|@{\hspace{1pt}}c@{\hspace{1pt}}|@{\hspace{1pt}}c@{\hspace{1pt}}|@{\hspace{1pt}}c@{\hspace{1pt}}|@{\hspace{1pt}}c@{\hspace{1pt}}|@{\hspace{1pt}}c@{\hspace{1pt}}|@{\hspace{1pt}}c@{\hspace{1pt}}|@{\hspace{1pt}}c@{\hspace{1pt}}|@{\hspace{1pt}}c@{\hspace{1pt}}|@{\hspace{1pt}}c@{\hspace{1pt}}|@{\hspace{1pt}}c@{\hspace{1pt}}|@{\hspace{1pt}}c@{\hspace{1pt}}|@{\hspace{1pt}}c@{\hspace{1pt}}|@{\hspace{1pt}}c@{\hspace{1pt}}|}
\hline
\multirow{2}{*}{}                & Year                                                              &           & 2013 & 2013 & 2015 & 2017 & 2017 & 2017 & 2018 & 2020 & 2021 & 2021 & 2021 & 2021 & -    \\ \cline{2-16} 
                                 & Reference                                                                    &           & \cite{mohanty2013controlling}  & \cite{guzel2013autonomous}   & \cite{pol2015review}  & \cite{injarapu2017survey}  & \cite{pandey2017mobile}  & \cite{victerpaul2017path}  & \cite{tzafestas2018mobile}  & \cite{alatise2020review}   & \cite{sanchez2021path}  & \cite{panigrahi2021localization}  & \cite{niloy2021critical}  & \cite{zghair2021one}  & ours \\ \hline
\multirow{11}{*}{\rotatebox{90}{Covered topics}} & Sensor types                                                                  &           &      &      &      &      &      &      &      & \checkmark    &      &      & \checkmark    & \checkmark    & \checkmark    \\ \cline{2-16} 
                                 & \begin{tabular}[c]{@{}c@{}}Mobile robot\\platform types\end{tabular}         &           &      & \checkmark    &      &      &      &      &      & \checkmark    &      &      & \checkmark    &      & \checkmark    \\ \cline{2-16} 
                                 & Simulation tools                                                              &           &      &      &      &      &      &      &      &      &      &      &      &      & \checkmark    \\ \cline{2-16} 
                                 & Path planning                                                                 &           & \checkmark    &      & \checkmark    & \checkmark    & \checkmark    & \checkmark    & \checkmark    & \checkmark    & \checkmark    &      & \checkmark    & \checkmark    & \checkmark    \\ \cline{2-16} 
                                 & Path following                                                                &           &      &      &      &      & \checkmark    & \checkmark    & \checkmark    &      &      &      &      & \checkmark    & \checkmark    \\ \cline{2-16} 
                                 & Sensor fusion                                                                 &           &      &      &      &      &      &      & \checkmark    & \checkmark    &      & \checkmark    & \checkmark    & \checkmark    & \checkmark    \\ \cline{2-16} 
                                 & \multirow{2}{*}{\begin{tabular}[c]{@{}c@{}}Obstacle\\ avoidance\end{tabular}} & Legacy    & \checkmark    &      & \checkmark    & \checkmark    & \checkmark    & \checkmark    & \checkmark    & \checkmark    & \checkmark    & \checkmark    & \checkmark    & \checkmark    & \checkmark    \\ \cline{3-16} 
                                 &                                                                               & RL/DL     &      &      &      &      & \checkmark    &      &      &      &      & \checkmark    &      &      & \checkmark    \\ \cline{2-16} 
                                 & \multirow{2}{*}{Navigation}                                                   & Map-based & \checkmark    & \checkmark    & \checkmark    &      & \checkmark    & \checkmark    & \checkmark    & \checkmark    & \checkmark    & \checkmark    & \checkmark    & \checkmark    & \checkmark    \\ \cline{3-16} 
                                 &                                                                               & Mapless   & \checkmark    & \checkmark    & \checkmark    &      & \checkmark    & \checkmark    & \checkmark    & \checkmark    & \checkmark    & \checkmark    & \checkmark    & \checkmark    & \checkmark    \\ \cline{2-16} 
                                 & SLAM                                                                          &           &      &      &      &      &      &      & \checkmark    &      &      & \checkmark    & \checkmark    & \checkmark    & \checkmark    \\ \hline
\end{tabular}
\end{table}

Based on the shortcomings of the previous surveys mentioned above, this survey has been prepared to cover all the topics listed in table \ref{compare-to-prev-works-table}. We have striven to:
\begin{itemize}
\item
give a comprehensive treatment on legacy as well as modern obstacle avoidance methods (based on RL and DL)
\item
review the most notable SLAM methods
\item
introduce well-known robotic simulators and whether they can be linked to robotic operating system (ROS) or not \cite{ros}
\item
types of mobile robot platforms and their characteristics based on their operational environment
\item
Simple and concise review of famous sensor fusion approaches such as Kalman filter \cite{kalman-filter} and its extensions as well as particle filter \cite{gordon1993novel}
\item
review open source SLAM datasets
\end{itemize}
The rest of this survey is structured as follows. Section \ref{sensors-sec} is devoted to different types of sensors used for autonomous navigation. Section \ref{robot-platforms-sec} reviews different types of mobile robot platforms and their characteristics. Common software for robotic simulations is reviewed in section \ref{sim-tools-sec}. Path planning and path following approaches are reviewed in sections \ref{path-planning-sec} and \ref{path-following-sec}, respectively. Sensor fusion methods are reviewed in section \ref{sensor-fusion-sec}. Obstacle avoidance is investigated in section \ref{obstacle-avoidance-sec}. Navigation and SLAM methods are reviewed in sections \ref{nav-methods-sec} and \ref{slam-sec}, respectively. The survey is closed with future works and conclusion in sections \ref{future-work-sec} and \ref{conc-sec}, respectively.
\section{Sensors}\label{sensors-sec}
Acting autonomously is not possible without the ability to sense the surrounding environment \cite{russell2016artificial}. Any autonomous robot must be equipped with a perception module which is combination of hardware (sensors) and software components. This section is devoted to various types of sensors that are commonly used in autonomous mobile robots.
\subsection{Absolution positioning sensors}\label{absolute-pose-sensor-sec}
Mobile robots need to move in their operational environment in order to carry out their objectives at different positions in the environment. For example, a rescue robot needs to navigate toward the wounded victim in order to perform the rescue operation. An important prerequisite to autonomous navigation is the self-localization ability of the robot. Usually, localization is performed with respect to some reference frame in the environment.

\begin{figure}[!t] 
\centering
\subfloat[]{\includegraphics[width=2.6in]{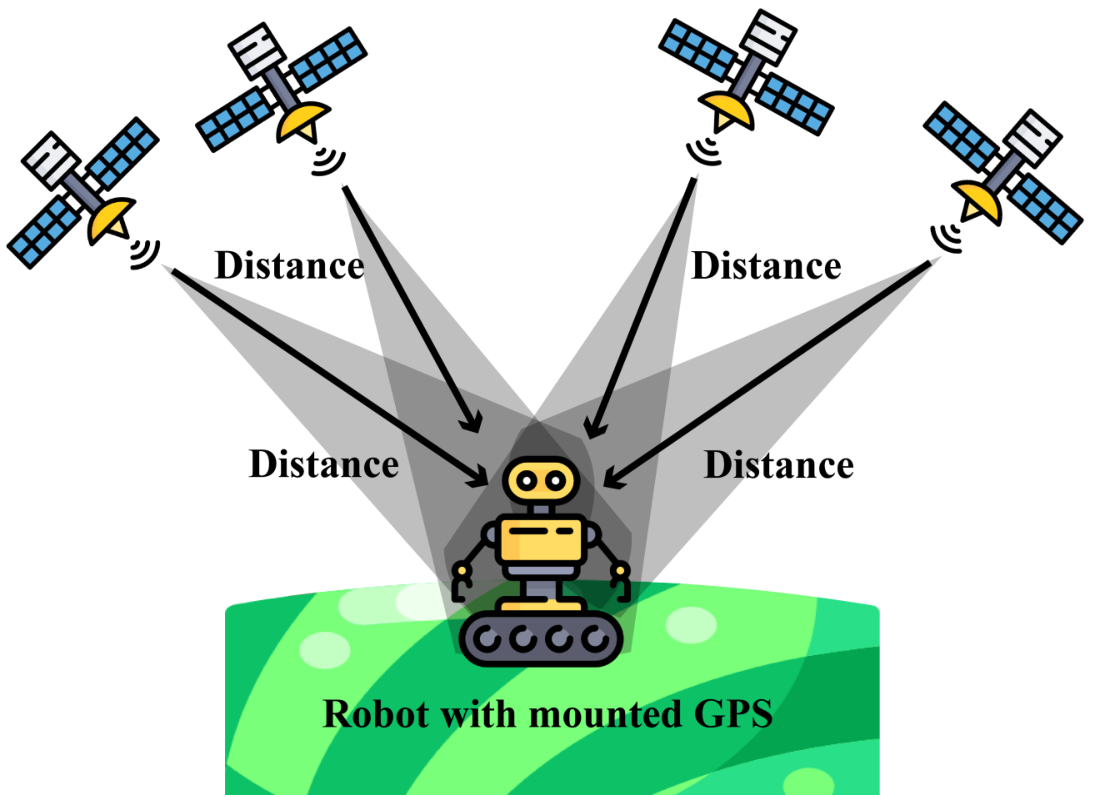}%
\label{gps-localization-fig}}
\subfloat[]{\includegraphics[width=2.6in]{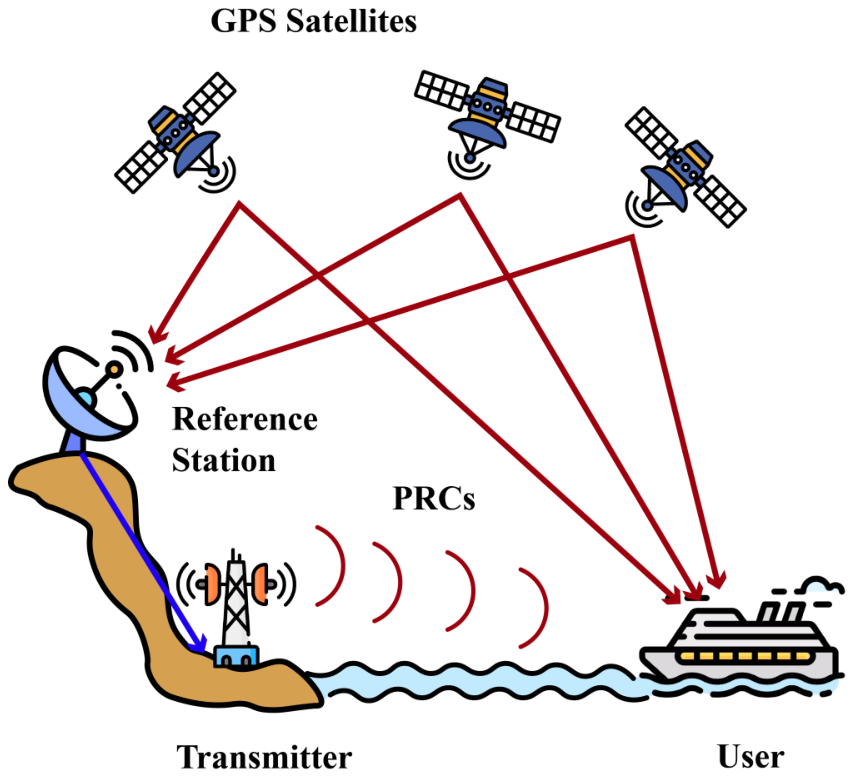}%
\label{dgps-localization-fig}}
\caption{Illustration of (a) robot localization using GPS receiver and (b) DGPS localization and error correction using reference station.}
\label{}			
\end{figure}

Global positioning system (GPS) receiver is one of the common sensors that provides absolute positioning in terms of longitude and latitude as well as time data anywhere on or near the Earth  based on information received from low Earth orbiting satellites. The receiver uses the difference between the reception time of satellite signal and its broadcast time to compute its distance from the satellite. As shown in figure \ref{gps-localization-fig}, having access to information from at least four satellites, the receiver can provide positioning information accurate to a few meters for outdoor navigation. The drawback of GPS is that localization data becomes unreliable if unobstructed line of sight to at least four satellites is unavailable. In urban areas, signal outages may be caused by tall buildings or mountains \cite{gps-wikipedia}. Additionally GPS signals are not available inside indoor environments.

The localization error of GPS can be reduced by using differential GPS (DGPS). Basically, DGPS relies on multiple fixed beacons with known locations on the Earth. These beacons compare their known positions with the ones computed based on received satellite signals and broadcast correction information as shown in figure \ref{dgps-localization-fig}. GPS receivers can reduce their positioning error using corrective signals broadcast by these beacons. The drawback of DGPS is that the quality of correction degrades as we move away from base/reference station \cite{barr2013performance}.
\subsection{Obstacle detection sensors}
In this section, different types of sensors that are commonly used for obstacle detection and avoidance are introduced.
\begin{figure}[!t] 
\centering
\subfloat[]{\includegraphics[width=1.5in]{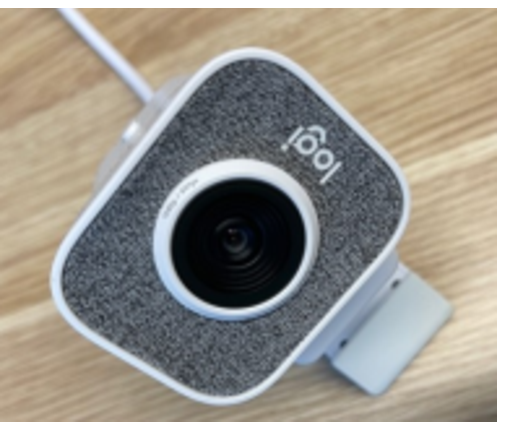}
\label{mono-cam-fig}}
\subfloat[]{\includegraphics[width=1.5in]{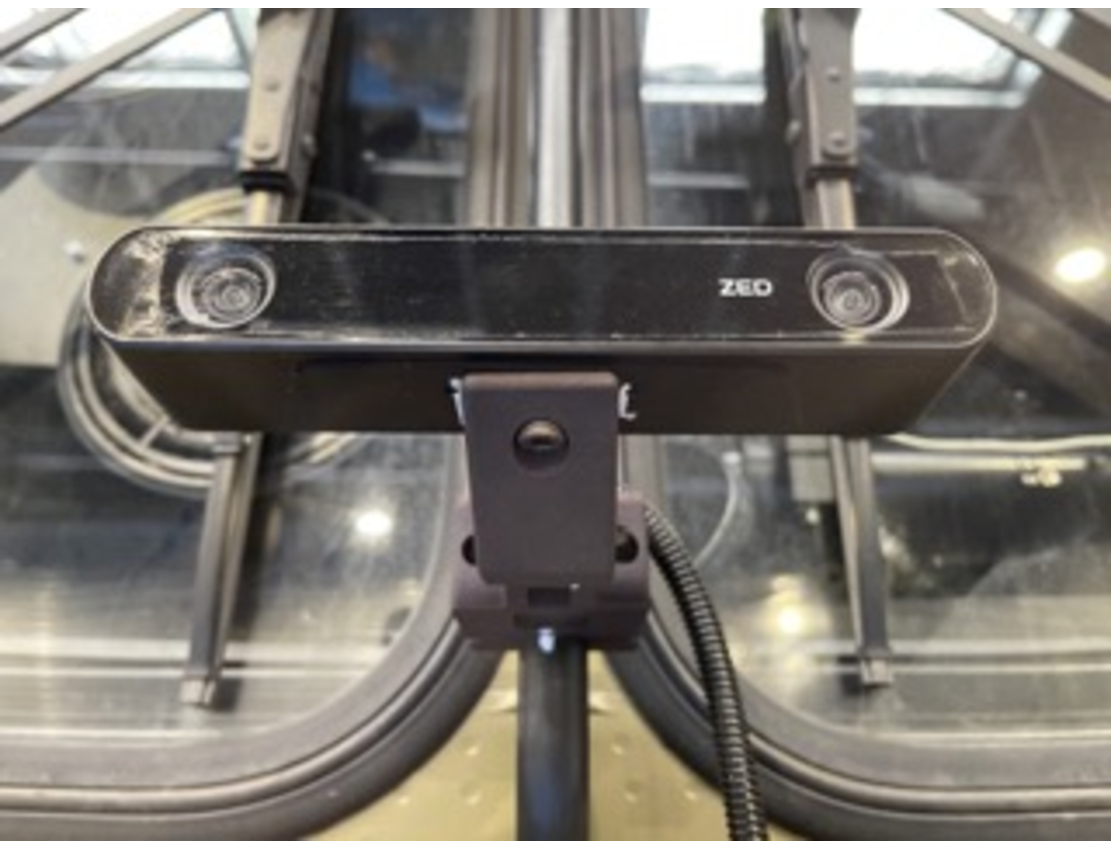}
\label{rgbd,stereo-cam-fig}}
\subfloat[]{\includegraphics[width=1.in]{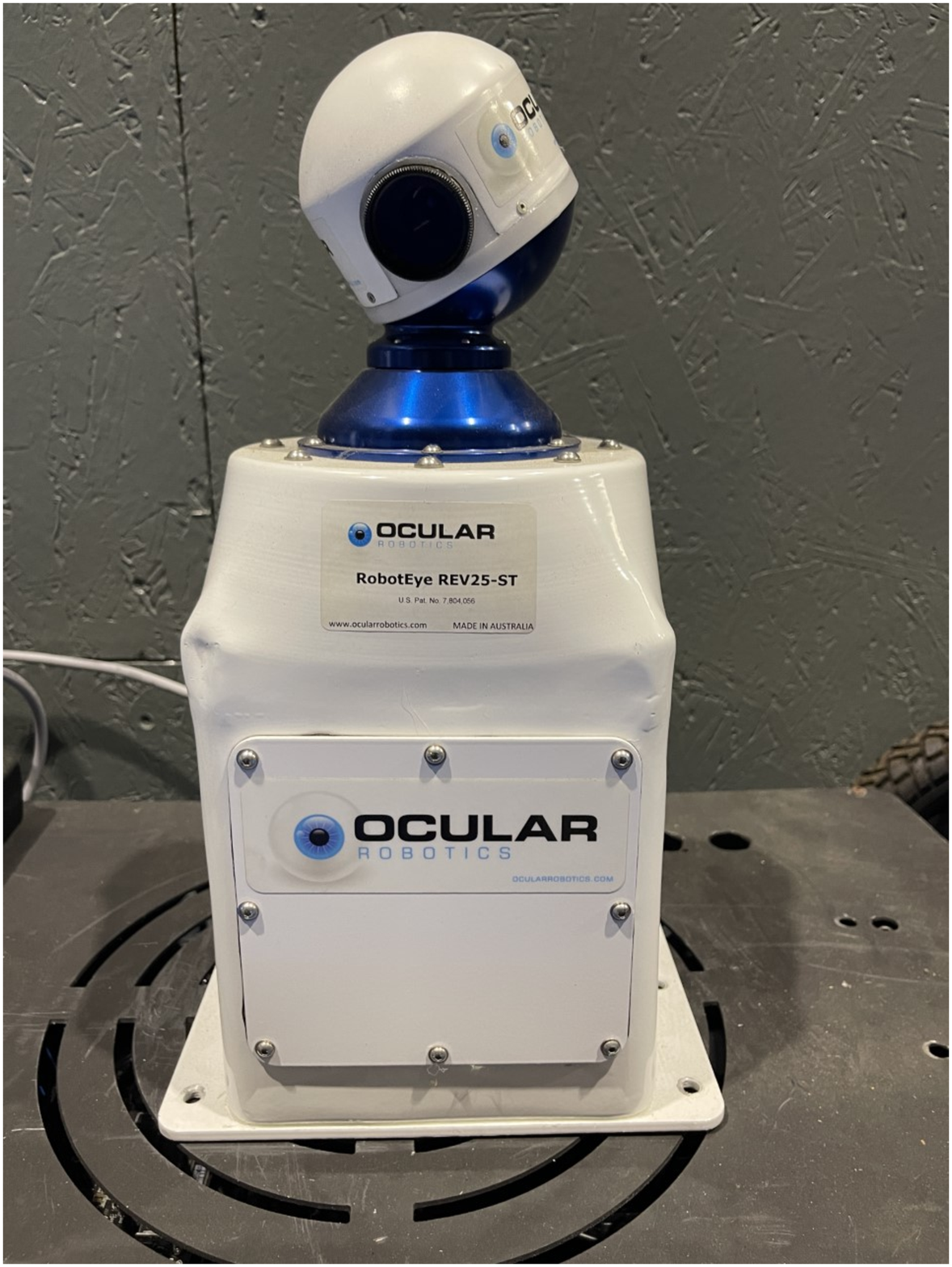}
\label{robot-eye-rev25-st-stabilized-cam-fig}}
\caption{Different types of cameras: (a) mono, (b) stereo/RGBD, and (c) Stabilized vision system}
\label{cam-fig}			
\end{figure}
\subsubsection{Visual sensors}
Claiming that a robotic system has autonomous navigation capability is almost impossible without equipping the robot with means of obstacle detection and avoidance. One can think of variety of sensor types when it comes to obstacle detection. Monocular (figure \ref{mono-cam-fig}) and stereo/RGBD cameras (figure \ref{rgbd,stereo-cam-fig}) provide feature-rich visual data at a reasonable price. The emergence of powerful vision algorithms based on DL and relatively affordable edge devices with reasonable processing power such as Jetson boards \cite{jetson-boards} has made visual sensors an ideal choice for performing various vision tasks including obstacle detection. During the robot movement, the onboard cameras may be subject to vibration making captured images blurry. This issue is addressed by stabilized vision systems \cite{RobotEye-REV25-ST} an example of which is shown in figure \ref{robot-eye-rev25-st-stabilized-cam-fig}. Stabilized vision systems also improve the tracking capability. 
\subsubsection{Range sensors}
\begin{figure}[!t] 
\centering
\includegraphics[width=0.8\textwidth]{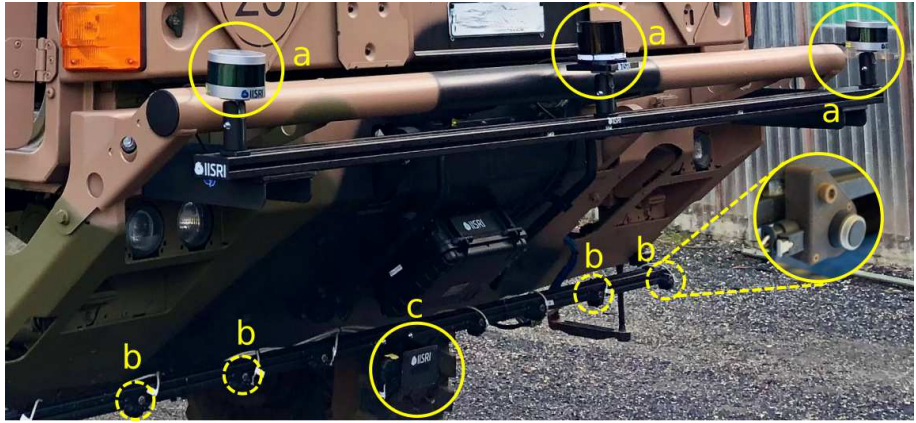}
\caption{Different types of range sensors: (a) LiDAR, (b) Ultrasonic, (c) RADAR.}
\label{iisri-robot-fig}			
\end{figure}

One of the most convenient solutions for obstacle detection is using range sensors which measure time-of-flight. Radar (Radio Detection And Ranging) relies on radio waves to detect distance and angle to objects and/or their velocity. A typical Radar has been depicted in figure \ref{iisri-robot-fig}. LiDAR (light detection and ranging) does similar thing to Radar but instead of using radio waves, it uses pulsed laser light. LiDAR sensors are capable of measuring range data with higher accuracy compared to Radar. The downside of these sensors is their high price. Moreover, laser scanners cannot sense glass obstacles and they cannot be used underwater due to laser disruption by water \cite{riisgaard2003slam}. Typical LiDARs have been shown in figure \ref{iisri-robot-fig}.

Ultrasonic sensors also provide range data but with much less precision compared to LiDAR and RADAR. The total range of ultrasonic sensor is lower (about seven meters) than LiDAR and RADAR. The lower precision of ultrasonic sensor is due to the fact that its radiation pattern is bat-like. Therefore, after receiving the reflection of an emitted signal that hits an obstacle, the position of the sensed obstacle can only be estimated vaguely within the region covered by the emitted wave. This is in contrast to LiDAR and RADAR that provide much more accurate data about the obstacle position. A typical ultrasonic sensor has been shown in figure \ref{iisri-robot-fig}.
\subsection{Relative positioning sensors}
In section \ref{absolute-pose-sensor-sec}, absolute positioning sensors were reviewed. However, there are scenarios that absolution positioning is not possible which is the motivation for using relative positioning sensors such as inertial measurement unit (IMU) and encoder. IMU is composed of three accelerometers and three gyroscopes along $\{x, y, z\}$ axes. When IMU is mounted on the robot, the accelerometers and gyroscopes measure the robot acceleration and rotation rate along the three axes, respectively. Using the well-established mathematics of an inertial navigation system (INS) \cite{titterton2004strapdown}, it is possible to turn IMU readings into useful 3D position and 3D orientation for the robot. Due to successive integration of IMU data, the estimated robot pose is subject to accumulated error which can be reduced by fusing INS data with other sensors such as GPS and camera(s) (see section \ref{MSCKF-sec}).

Another sensor that can be used for relative positioning in ground robots is rotary encoder. Mounting encoder on a robot wheel, it is possible to count the number of rounds the wheel has completed. Knowing the perimeter of the wheel, the amount of robot displacement can be estimated. The drawback of using encoders is that wheel slippage leads to erroneous displacement estimates.
\section{Mobile robot platforms}\label{robot-platforms-sec}
Over the years, different types of mobile robot platforms have been designed and developed. Mobile robots may be ground \cite{bagnell2010learning,urmson2008autonomous, schafer2006ravon,braid2006terramax}, aerial \cite{parrot-uav,meier2012pixhawk,heng2014autonomous} or underwater \cite{codd2017milton} vehicles.
\subsection{Ground vehicles}
\begin{figure}[!tbp]
\centering
\includegraphics[width=0.5\textwidth]{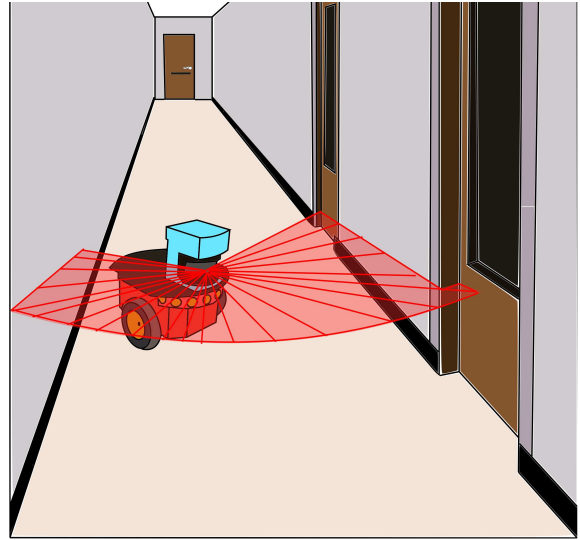}
    \caption{Indoor environment illustration.}
    \label{indoor-env-fig}
\end{figure}

Ground vehicles come in variety of sizes and shapes depending on the objectives they have been designed for. Apart from structural differences, suite of sensors mounted on these robots depends on their operational environment. In indoor environments, GPS is not accessible but the environment is usually feature-rich due to surrounding walls, doors, furniture, etc. Thus, range sensors can be used to carry out scan matching \cite{kohlbrecher2013hector} in order to achieve accurate localization. Using stereo cameras is also an affordable solution to achieve depth data. Ring of ultrasonic sensors around the robot chassis is also a viable solution for obstacle detection \cite{kim2012simple} or localization \cite{moreno2002genetic,moreno1999global} in indoor environments. An illustration of indoor environment has been shown in figure \ref{indoor-env-fig}. Regardless of being indoor or outdoor, INS can be used for localization which relies on IMU data. Indoor environments usually have smooth floors so wheel slippage is not likely and rotary encoders are quite suitable for estimating the robot displacement.

\begin{figure}[!t] 
\centering
\subfloat[]{\includegraphics[width=0.52\textwidth]{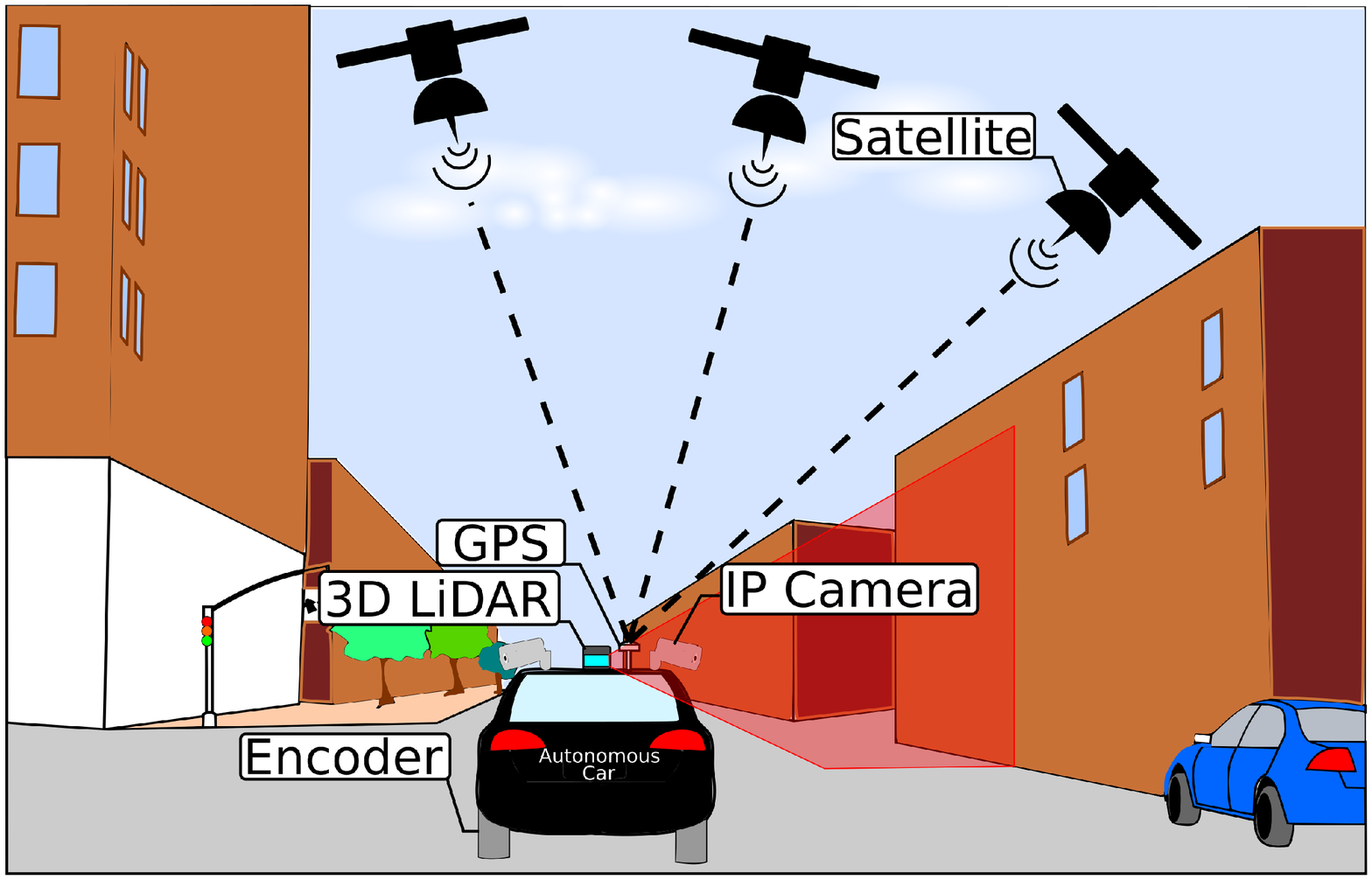}%
\label{outdoor-urban-fig}}
\subfloat[]{\includegraphics[width=0.445\textwidth]{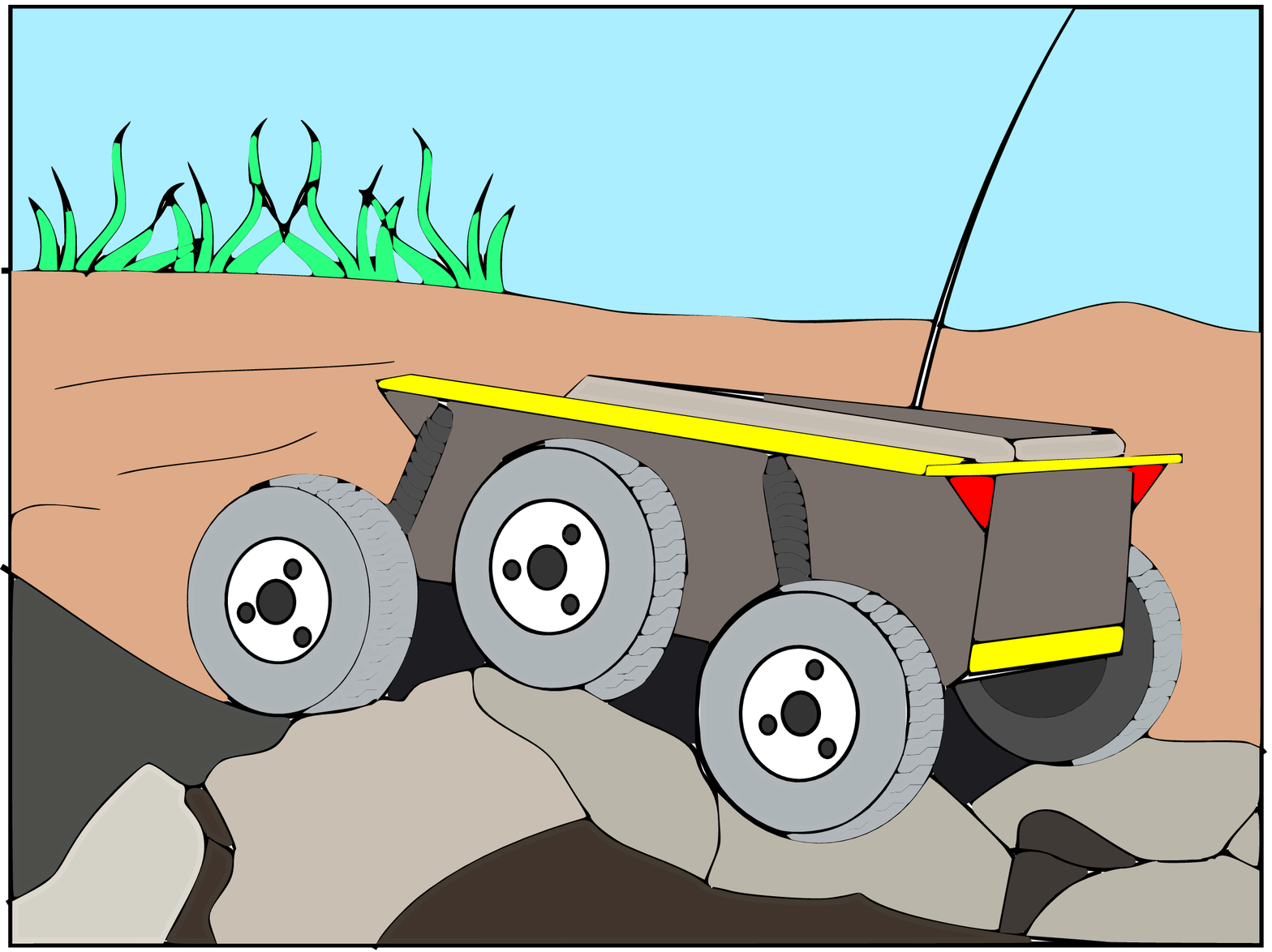}%
\label{outdoor-offroad-fig}}
\caption{Ground robots for outdoor scenarios: (a) structured urban environment, (b) unstructured offroad environment.}
\label{outdoor-envs}			
\end{figure}

For outdoor environments, GPS can aid the absolute positioning provided that clear line-of-sight between the GPS receiver and at least four satellites is available. Such conditions may be violated in urban environments (figure \ref{outdoor-urban-fig}) due to signal blockage by tall buildings. However, urban environments are well-structured in nature. This property can be exploited to improve localization accuracy by performing scan matching on LiDAR data \cite{yang2018mobile}. Even if GPS signal is available, scan matching can still lead to better localization accuracy \cite{levinson2007map}. An alternate approach is visual place recognition (VPR) \cite{DBLP:journals/corr/abs-2103-01994,chen2022self, majdik2015air} which plays a vital role in reducing localization error by recognizing the places that have been visited previously.

In some outdoor applications (e.g. military), the mobile robot encounters offroad environments which are harsh, unstructured, and sometimes impossible to traverse. Offroad environments push the mechanical \cite{cmu-12years} and autonomous capabilities \cite{jones2006autonomous} of mobile robots to their limits. As can be seen in figure \ref{outdoor-offroad-fig}, the offroad robots must be equipped with large wheels, long travel suspension, and preferably all-wheel drive ability to be able to traverse different types of lands such as muddy, rocky, snowy, etc. 
Apart from robot mechanical requirements for offroad environments, several challenges must be addressed by autonomous software modules of the robot to achieve successful navigation:
\begin{enumerate}
\item
Given that offroad environments do not follow any specific structure \cite{thakker2020autonomous}, autonomous navigation cannot make any assumptions about them beforehand.
\item
Some offroad environments like deserts are featureless which hinders localization of the robot due to lack of recognizable landmarks. For example in deserts, even the terrain shape may vary due to movement of sands by blowing winds.
\item
The offroad environments are hazardous due to deep valleys, rough rivers, volcanoes, etc. Therefore, it is crucial to take into account impassable areas during path planning to enforce the robot safety.
\item
In offroad environments, some of the obstacles may be passable like bushes while others are impassable like mountains. The ability to detect passable obstacles is vital to shorten the length of the planned path by moving through the passable obstacles.
\item
Autonomous navigation relies on the robot sensor readings to make decisions. However, the field of view of sensors may be limited due to dense vegetation which hinders efficient localization.
\item
Depending on the offroad environment, the robot motion controller parameters must be tuned differently \cite{bradley2010domain}. For example, the amount of force applied to the robot wheels  in snowy environments is different from dirt lands. 
\end{enumerate}
\subsection{Unmanned aerial vehicles}
While controlling ground vehicles are simpler than unmanned aerial vehicles (UAVs), their movement is limited to traversable ground. Aerial vehicles are not bound to such constraints and they can reach the desired location in less time compared to ground vehicles. However, aerial vehicles have some limitations as well. For example, the flight duration may be short, especially for low-grade UAVs. Additionally, failure in control module may lead to a disastrous crash damaging the UAV. Another drawback is limited payload. UAVs can carry limited load which means the set of hardware (e.g. sensors and processing units) mountable on them must be chosen wisely. Given that IMUs are usually small and lightweight, using them to implement INS for efficient localization of UAVs is quite popular.  As shown in figure \ref{uav-env-fig}, UAVs may be deployed to variety of outdoor environments e.g. urban or unstructured and offroad. To keep the INS error bounded, GPS sensor can be mounted on UAVs \cite{nemra2010robust, 4602149, da2016fuzzy, zhang2018intelligent}. Using UAVs in indoor environments is also possible. The only difference to outdoor scenarios is unavailability of GPS signals. Instead of GPS, INS data can be fused with LiDAR \cite{7049479},  mix of laser data and vision sensors \cite{5983054}, Ultra Wideband (UWB)  \cite{song2021tightly}, mix of UWB and 3D LiDAR \cite{7831874}. Even seamless switch between indoor/outdoor navigation modes for UAVs has been attempted \cite{cebrian2019low}. Moreover, Hall effect sensor data has been used to aid the velocity update of INS \cite{8594610}. Hall effect sensor detects magnetic filed presence and magnitude by exploiting Hall effect. According to this effect, a magnetic field perpendicular to current in an electrical conductor produces voltage difference across the electrical conductor such that it is transverse to the electric current and the magnetic field. 
\subsection{Underwater vehicles}
\begin{figure}[!t] 
\centering
\subfloat[]{\includegraphics[width=0.5\textwidth]{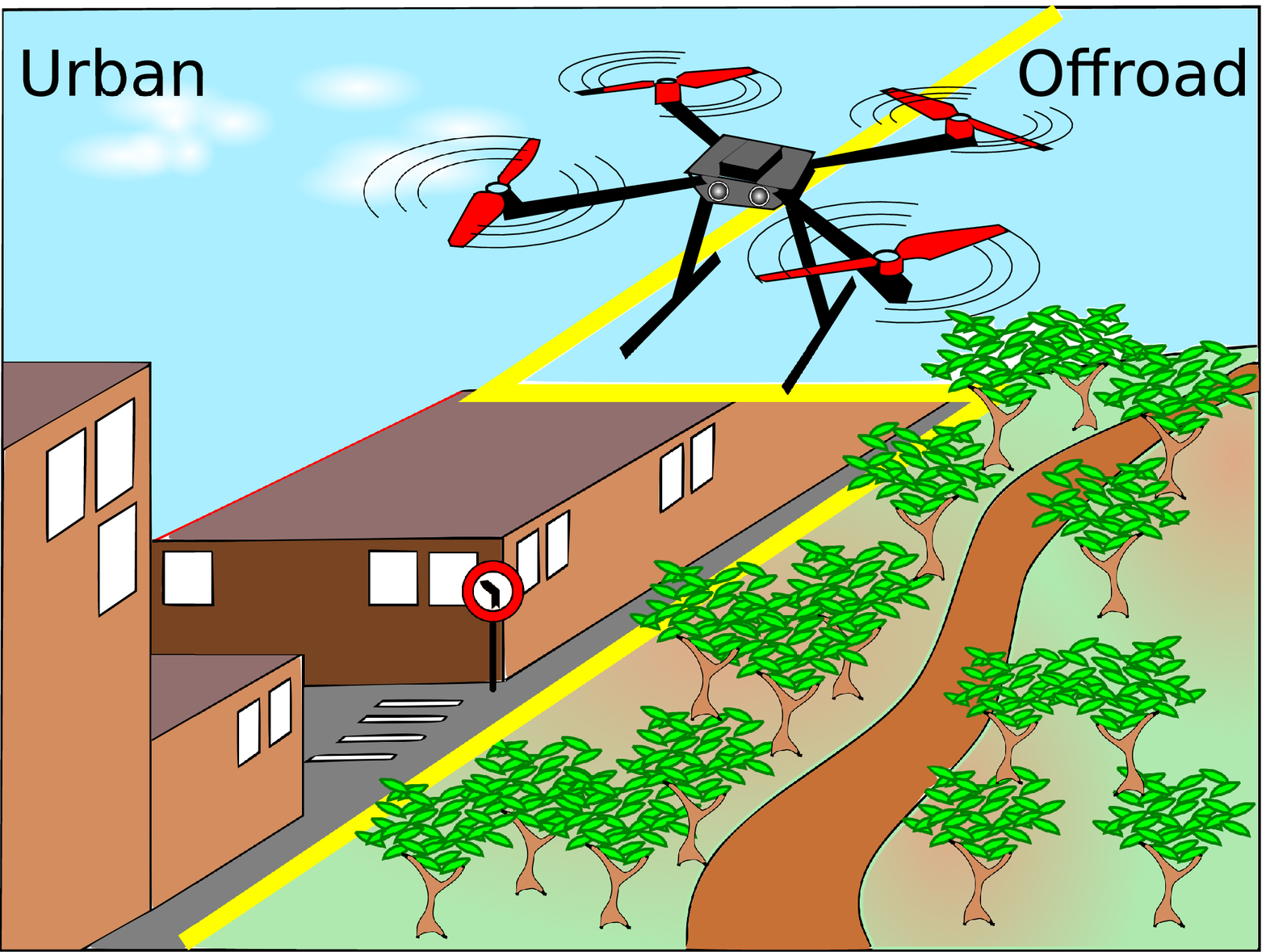}%
\label{uav-env-fig}}
\subfloat[]{\includegraphics[width=0.5\textwidth]{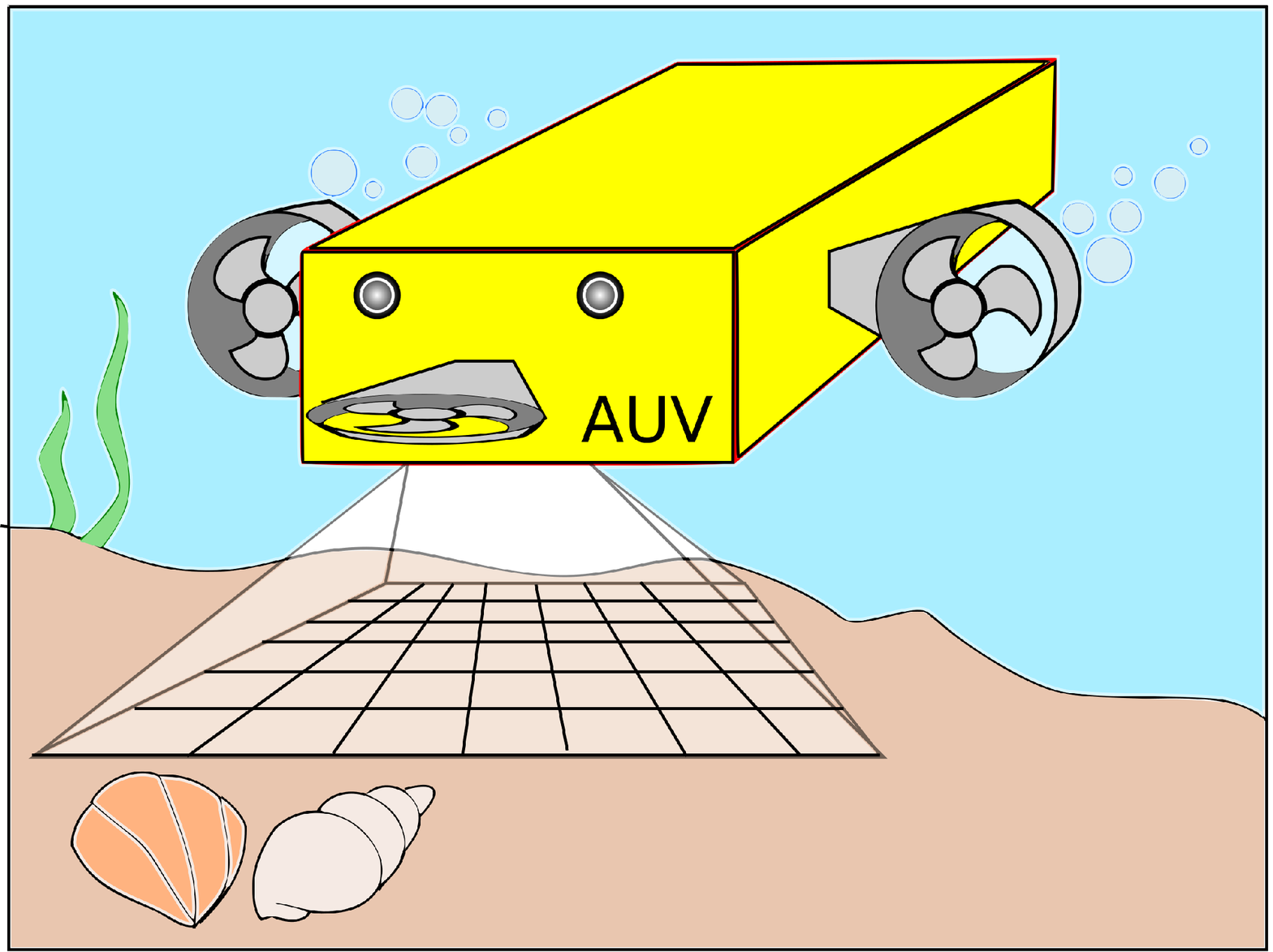}%
\label{auv-env-fig}}
\caption{Outdoor (a) aerial and (b) under-water environments and their respective robots.}
\label{}			
\end{figure}

AUVs can be used in various applications such as retrieving black boxes of airplanes \cite{jian2015target,ridao2015intervention} crashed in oceans/seas, enforcing security of ports and harbors by detection and disposal of explosives and mines \cite{abreu2014minehunting, reed2010detection}, and infrastructure maintenance for oil and gas industries \cite{xiang2010coordinated, zhang2021subsea, calvo2009low, 6608089}. However, underwater autonomous navigation (figure \ref{auv-env-fig}) is challenging due to unavailability of GPS \cite{gonzalez2020autonomous} and limited visibility. Failing to register the current position of the robot using visual features leads to inaccurate localization which in turn hinders autonomous navigation \cite{qin2022real}. Using DR and INS to localize AUVs has a long history \cite{leonard1998autonomous}. However, these methods suffer from accumulated error due to noisy sensor data, ocean currents, and Earth's gravity in underwater environments \cite{gonzalez2020autonomous}. To deal with the accumulated error, Geophysical Navigation (GN) \cite{rice2004geophysical} can be used in which AUV sensor readings are matched with geophysical map of the underwater environment. Unfortunately, GN suffers from some drawbacks such as  requiring geophysical map of the environment before the navigation starts and computational complexity of matching sensor data with the map. As an alternative approach to GN, acoustic ranging systems \cite{fallon2011efficient} may be used but they rely on complex infrastructure and their cost of deployment is high. Researchers have explored new alternatives for AUV localization such as optical technologies \cite{bosch2016close}. Yet again, development of these technologies has been slowed down due to tough conditions of underwater environments. In the presence of proper lighting conditions,  visual processing systems can improve localization accuracy significantly \cite{gonzalez2020autonomous}. For example, Mehdi et al. \cite{mehdi2021autonomous} have fused IMU and depth data extracted from stereo camera system to plot the map of the underwater environment using SLAM implemented in ROS. The map is used for path planning and autonomous navigation.
\subsubsection{AUV suitable sensors}
The first sensor to perform localization using INS underwater is IMU. To deal with INS accumulated error, fusing IMU data with other sensors is crucial. Doppler velocity loggers (DVL)\cite{doppler-velocity-logs} is  a sonar system for measuring motion underwater that can determine the speed and the direction of AUV movement. Regarding absolute positioning, GPS cannot be used underwater since its electromagnetic signals decay quickly. Acoustic signaling is an alternative solution since acoustic signals decay very slowly underwater \cite{he2011autonomous}. The downside of this method is its reliance on baseline stations deployed in the navigation environment. The baseline batteries need frequent charging in order to stay functional and respond to interrogator device mounted on AUV. In addition, baseline deployment and recovery is time-consuming and costly which makes acoustic navigation impractical in large-scale environments \cite{kinsey2006survey}. To address this problem, MIT researchers have developed underwater backscatter localization (UBL) \cite{ghaffarivardavagh2020underwater} which is a battery-free pinpointing system. Instead of emitting acoustic signals which require battery consumption, UBL reflects the modulated signals from its surrounding environment. This way, position information is provided without needing any battery-operated device. UBL takes the role of underwater GPS.

Family of sonar sensors such as forward looking sonars and side scan sonar \cite{881735} is widely used for obstacle detection and localization underwater. Off-the-shelf underwater camera systems \cite{massot2015optical} are also very common for extracting depth data based on stereo vision.

\section{Simulation tools}\label{sim-tools-sec}
To conduct research in the field of robotics, a suitable robotic platform is needed which is a combination of software and hardware components. During research process, various experiments are needed to be carried out. It is quite likely for these experiments to fail due to software and/or hardware module malfunction. Sometimes, these failures may damage the robot. On the other hand, performing experiments on real robots is tedious and time consuming. An alternative approach is to implement and test the robot software in simulation. This way, researchers will not have to worry about accidentally damaging robots and no hardware is needed to test an initial idea. Resetting the simulation scenario is also much easier than resetting the robot status in real world. In table \ref{main-robo-sims-table}, the characteristics of some of main robotic simulators have been provided \cite{robotic-main-simulators-comapre}. 

While testing new ideas in simulation is beneficial, it is not the ultimate goal of robotic projects. Sooner or later, the simulated project must be evaluated on real robots. The drawback of using simulators is that for modeling the physical laws of the real world, some compromises may have been made. Therefore, the project tested in simulation may need some refinements and manual tuning to be ready for deployment on a real robot. Nevertheless, putting the time and effort into developing robotic projects in simulation before deployment on real robots is definitely worth it since it reduces development time and cost significantly. Among the existing simulators, some like Gazebo rely on high quality physics engines such as open dynamic engine (ODE) \cite{ode-physics-endgine} to reduce the gap between simulated and real robots drastically. That is why many well known robotic companies provide their customers with simulation tools of their commercial robots in the form of software packages relying on ROS and Gazebo. ROS is a set of tools and software libraries specifically designed to facilitate robot application development. Using ROS, development time of robotic projects is drastically reduced due to the possibility to use high quality of-the-shelf software packages that implement famous algorithms and interface with various sensors (hardware abstraction). Some examples of well-known robots with descent simulation in ROS and Gazebo include but are not limited to Fetch robotics \cite{fetch-robot-sim}, PR2 robot \cite{pr2-robot-sim}, KUKA iiwa \cite{kuka-iiwa-robot-sim}, and humanoid Nao robot \cite{nao-robot-sim}.
\begin{table}
    \centering
    \caption{Characteristics comparison of main robotic simulators}\label{main-robo-sims-table}
\setlength\tabcolsep{1.5pt} 
\begin{tabular}{|c|c|c|c|cccc|}
\hline
\multirow{2}{*}{Simulator} & \multirow{2}{*}{\begin{tabular}[c]{@{}c@{}}Supported\\ OS\end{tabular}}       & \multirow{2}{*}{\begin{tabular}[c]{@{}c@{}}Physics\\ engine\end{tabular}}                            & \multirow{2}{*}{\begin{tabular}[c]{@{}c@{}}ROS\\ support\end{tabular}} & \multicolumn{4}{c|}{Supported}                                                                                                                                                                                                                                                                                                                                                                                                                                                                                                                                                                                               \\ \cline{5-8} 
                           &                                                                               &                                                                                                      &                                                                        & \multicolumn{1}{c|}{Robots}                                                                   & \multicolumn{1}{c|}{Actuators}                                                                                                                                                                              & \multicolumn{1}{c|}{Tools}                                                                                                               & Sensors                                                                                                                                                             \\ \hline
Gazebo\cite{gazebo-sim}                     & Linux                                                                         & \begin{tabular}[c]{@{}c@{}}ODE, Bullet,\\ Simbody, DART\end{tabular}                                 & \begin{tabular}[c]{@{}c@{}}ROS1,\\ ROS2\end{tabular}                   & \multicolumn{1}{c|}{\begin{tabular}[c]{@{}c@{}}Mobile,\\ humanoid,\\ industrial\end{tabular}} & \multicolumn{1}{c|}{\begin{tabular}[c]{@{}c@{}}Revolute,\\ prismatic,\\ screw and\\ spherical joints\end{tabular}}                                                                                          & \multicolumn{1}{c|}{Grippers}                                                                                                            & \begin{tabular}[c]{@{}c@{}}Camera, distance \&\\ proximity sensors,\\ laser, force sensors\end{tabular}                                                             \\ \hline
RoboDK                     & \begin{tabular}[c]{@{}c@{}}Linux,\\ Mac OS,\\ Windows,\\ Android\end{tabular} & None                                                                                                 & -                                                                      & \multicolumn{1}{c|}{\begin{tabular}[c]{@{}c@{}}Industrial\\ robots\end{tabular}}              & \multicolumn{1}{c|}{\begin{tabular}[c]{@{}c@{}}Possibility to\\create rotated\\and linear\\axis\end{tabular}}                                                                                                & \multicolumn{1}{c|}{\begin{tabular}[c]{@{}c@{}}Grippers, weld\\ tools, spindle,\\ grinding,\\polishing,\\ tool changer, ...\end{tabular}} & \begin{tabular}[c]{@{}c@{}}Lasers, cameras,\\ laser tracker\end{tabular}                                                                                            \\ \hline
Webots\cite{webots-sim}                     & \begin{tabular}[c]{@{}c@{}}Linux,\\ Mac OS,\\ Windows\end{tabular}            & \begin{tabular}[c]{@{}c@{}}Proprietary,\\ ODE-based\end{tabular}                                     & \begin{tabular}[c]{@{}c@{}}ROS1,\\ ROS2\end{tabular}                   & \multicolumn{1}{c|}{\begin{tabular}[c]{@{}c@{}}Mobile,\\ humanoid,\\ industrial\end{tabular}} & \multicolumn{1}{c|}{\begin{tabular}[c]{@{}c@{}}Brake, connector,\\ display, emitter,\\ linear/rotational\\ motor, muscle,\\ pen, propeller,\\speaker, track\\(conveyor belt\\ or tank robots)\end{tabular}} & \multicolumn{1}{c|}{Grippers}                                                                                                            & \begin{tabular}[c]{@{}c@{}}Accelerometer,\\ camera, compass,\\ distance sensor,\\ GPS, gyroscope,\\ lidar, position\\ sensor, receiver,\\ touch sensor\end{tabular} \\ \hline
\begin{tabular}[c]{@{}c@{}}CoppeliaSim\\\cite{coppeliasim-sim}\end{tabular}              & \begin{tabular}[c]{@{}c@{}}Linux,\\ Mac OS,\\ Windows\end{tabular}            & \begin{tabular}[c]{@{}c@{}}ODE, Bullet,\\ Vortex, Newton\end{tabular}                                & \begin{tabular}[c]{@{}c@{}}ROS1,\\ ROS2\end{tabular}                   & \multicolumn{1}{c|}{\begin{tabular}[c]{@{}c@{}}Mobile,\\ humanoid,\\ industrial\end{tabular}} & \multicolumn{1}{c|}{\begin{tabular}[c]{@{}c@{}}Revolute,\\ prismatic \&\\ spherical motors\end{tabular}}                                                                                                    & \multicolumn{1}{c|}{\begin{tabular}[c]{@{}c@{}}Pens, paint gun,\\ welding torch,\\ grippers\end{tabular}}                                & \begin{tabular}[c]{@{}c@{}}Vision, force\\ sensors, proximity\\ sensor,\\accelerometer,\\ gyroscope, lasers,\\ LiDARs\end{tabular}                                   \\ \hline
OpenRave                   & \begin{tabular}[c]{@{}c@{}}Linux,\\ Windows\end{tabular}                      & ODE, Bullet                                                                                          & ROS1                                                                   & \multicolumn{1}{c|}{\begin{tabular}[c]{@{}c@{}}Mobile,\\ humanoid,\\ industrial\end{tabular}} & \multicolumn{1}{c|}{\begin{tabular}[c]{@{}c@{}}Revolute \&\\ linear joints\end{tabular}}                                                                                                                    & \multicolumn{1}{c|}{Grippers}                                                                                                            & \begin{tabular}[c]{@{}c@{}}Cameras,\\ ray-casting laser\end{tabular}                                                                                                \\ \hline
Unity                      & \begin{tabular}[c]{@{}c@{}}Linux,\\ Mac OS,\\ Windows\end{tabular}            & \begin{tabular}[c]{@{}c@{}}NVidia PhysX,\\ Box2D, other\\ open-source\\ physics engines\end{tabular} & \begin{tabular}[c]{@{}c@{}}ROS1,\\ ROS2\end{tabular}                   & \multicolumn{1}{c|}{\begin{tabular}[c]{@{}c@{}}Mobile,\\ humanoid,\\ industrial\end{tabular}} & \multicolumn{1}{c|}{\begin{tabular}[c]{@{}c@{}}Revolute \&\\ prismatic joints\end{tabular}}                                                                                                                 & \multicolumn{1}{c|}{Grippers}                                                                                                            & Cameras                                                                                                                                                             \\ \hline
\end{tabular}
\end{table}
\section{Path planning}\label{path-planning-sec}
After percepting the environment using robot onboard sensors, it is necessary to plan a feasible path in order to reach the desired target starting from the current position. The choice of the path planning algorithm depends on whether the environment map is available or not. In case the map is known, the choice of the path planner depends on the map representation. For occupancy grid maps, D* \cite{stentz1997optimal} and its extensions such as focused D* \cite{stentz1995focussed} and  D* Lite \cite{koenig2005fast} are suitable choices. D* and its extensions are all considered as incremental search algorithms. D* assumes that unseen parts of the environment are free of obstacles. Based on this assumption, D* uses the map to compute the shortest path from the start to the goal. In the event of observing new obstacles, the map is updated with their information and a new shortest path is computed if necessary. This trend repeats until the goal is reached or it is concluded that the goal is indeed unreachable. Considering that observing new obstacles during navigation is quite likely, recomputing the shortest path must be fast which is what D* and its extensions are famous for. It is worth noting that focused D* is a combination of A* \cite{hart1968formal} and the ideas from the original D* algorithm; and D* Lite has been built based on life long planning A* (LPA*)\cite{koenig2004lifelong} and another incremental search method made of combining A* ideas and dynamic SWSF-FP \cite{ramalingam1996incremental}. 
\subsection{DMP-based path planning}
One of the popular approaches for trajectory planning and control signals generation in manipulator robots is dynamic movement primitive (DMP) \cite{schaal2006dynamic,ijspeert2013dynamical}. DMP is a highly adaptable mathematical model capable of representing complex trajectories without the need for manual tuning of its parameters. Roughly speaking, a DMP is a second order nonlinear dynamical system capable of imitating complex trajectories provided that its parameters are set appropriately. DMPs are either discrete or rhythmic which are suitable for generation of aperiodic and periodic trajectories, respectively. Discrete DMPs are suitable for path following since planned paths are not periodic trajectories. The crux of a discrete DMP is a point-attractor system:
\begin{equation}\label{point-attractor-eq}
\ddot{y} = \alpha_y \left[ \beta_y (g-y) - \dot{y} \right],
\end{equation}
where $y$, $\dot{y}$, and $\ddot{y}$ are the system position, velocity, and acceleration, respectively. Moreover, $g$ is the goal point and $\alpha_y >0$ and $\beta_y>0$ are gain coefficients. Careful inspection of equation \ref{point-attractor-eq} reveals that it is just a PD controller whose objective is to reach $g$ given the current state $y$. To specify the shape of the trajectory between $y$ and $g$, equation \ref{point-attractor-eq} is augmented with a special function $f(x)$ known as the forcing function\cite{schaal2006dynamic}:
\begin{equation}
f(x) = \frac{\sum_{i=1}^{N} \psi_i w_i}{\sum_{j=1}^{N} \psi_j} (g-y_0) x,
\end{equation}
where $\psi_i$ is the ith Gaussian basis function, $w_i$ is the ith weight parameter, $y_0$ is the initial position of the system and $x$ is state variable of another dynamical system known as the canonical system. By setting the parameter set $\{w_i, i=1,...,N\}$ of the forcing function, any arbitrary trajectory can be produced using DMP. The canonical system
\begin{equation}\label{canonical-system-eq}
\zeta \dot{x} = - \alpha_x x,
\end{equation}
controls the life time of the DMP. In equation \ref{canonical-system-eq}, $\alpha_x$ is a constant which is set to $-\log(0.01)$ for ensuring 99\% convergence of the DMP after its life time $\zeta$ is passed.

Despite the fact that DMP is primarily designed for motion planning of robotic manipulators, it has also been used for path planning in mobile robots. For example, Jiang et al. \cite{jiang2016mobile} have carried out mobile robot path planning using DMPs. A desired path is first planned and trajectories are extracted from it to be used as training samples for adjustment of DMP parameters. Thanks to the DMP high adaptability, Jiang et al. were able to achieve smooth trajectories for the mobile robot even when start and goal positions were altered.
\section{Path following}\label{path-following-sec}
To execute planned paths, mobile robots must be equipped with the necessary means to follow them. Over the years, various path following methods have been developed. Some of these methods are as simple as pure pursuit while others take a more involved approach by relying on dynamical systems or deep learning. Some of these path following methods are reviewed in this section. 
\subsection{Pure pursuit algorithm}
Pure pursuit \cite{coulter1992implementation} is a path tracking algorithm with respectable history as it has been used in various projects such as the Terragator \cite{champeny1991terregator}, NavLab\cite{10.1007/978-3-642-74585-0_6}, and NavLabs2\cite{coulter1994reconfiguration}. In a nutshell,  pure pursuit algorithm computes the robot angular velocity while assuming constant linear velocity. The robot is moved from its current position to some look-ahead point in front of it. After reaching the specified look-ahead point, it is moved further on the path which is to be followed.

\begin{figure}[!tbp]
\centering
\includegraphics[width=0.25\textwidth]{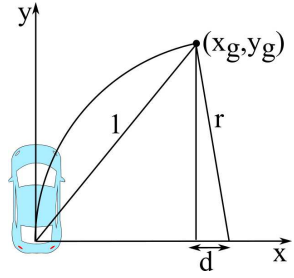}
    \caption{Geometry of the pure pursuit algorithm: $1$ is the chord length (look-ahead distance), $r$ is the radius of the arc joining robot position to the goal point, $(x_g,y_g)$ is the goal point}
    \label{pure-pursuit-geometry-fig}
\end{figure}

The algorithm computes an arc joining the robot current position to the goal position. The goal point is one look-ahead distance from the current position of the robot. In figure \ref{pure-pursuit-geometry-fig}, the robot's local coordinate frame as well as the goal point $(x_g,y_g)$ have been shown. The objective is to compute the curvature $\gamma$ of the arc joining the robot current position to the goal position assuming chord length (look-ahead distance) of $1$. The goal point is enforced to be on the robot path. 
Computation of the curvature value is fairly straightforward. Looking at figure \ref{pure-pursuit-geometry-fig}, one can write the following equations:
\begin{eqnarray}
x^2 + y^2 = 1^2 \label{x+y=1-eq}\\
x + d = r \label{x+d=r-eq}
\end{eqnarray}
where equation \ref{x+y=1-eq} follows from Pythagorean theorem in the left bigger triangle and equation \ref{x+d=r-eq} is due to the fact that sum of x and d is equal to the radius of curvature $r$. Using equations \ref{x+y=1-eq} and \ref{x+d=r-eq}, the curvature $\gamma$ can be computed as:
\begin{align}
&d = r - x \nonumber \\
&(r - x)^2 + y^2 = r^2 \implies
r^2 - 2rx + x^2 + y^2 = r^2 \implies
2rx = l^2 \implies
r = \frac{l^2}{2x} \implies
\gamma = \frac{2x}{l^2}
\end{align}
\subsection{DMP-based path following}
As mentioned in section \ref{path-planning-sec}, DMP is a popular tool that generates smooth trajectories for manipulator robots. However, it has also been used for path following of mobile robots. Sharma el al.\cite{8793911} have implemented a DMP-based path tracking approach for nonholonomic robots which is capable of avoiding an obstacle on the path. Two DMPs (to represent movement along x and y axes) were used and forcing function of each of the DMPs was parameterized using radial basis functions (RBFNs). Thanks to the adaptive nature of DMP, the proposed approach can preserve the shape of the planned trajectory and reach the goal even if the start or goal position is altered. Moreover, the proposed approach is able to suggest both trajectories passing both sides of the obstacle. This is particularly useful when one of the two paths is not traversable for some reason.
\subsection{DL and RL-based path following}
Inspired by interesting learning results achieved  by RL, it has been extended to handle problems with continuous  state and action spaces. This trend has been reinforced by borrowing ideas from DL giving birth to robust deep RL (DRL) methods capable of controlling robots with continuous and high dimensional states and action spaces. For example, Cheng et al. \cite{cheng2022path} applied DRL for path following and obstacle avoidance. The path following is realized by designing appropriate state and reward functions which are central to successful training of any RL agent. To realize obstacle avoidance, the aforementioned reward and state functions are modified to take into account distance and angle to detect obstacles. The RL algorithm used in this work was DDPG \cite{lillicrap2015continuous}. The proposed approach has been evaluated in simulation.
\section{Sensor fusion methods}\label{sensor-fusion-sec}
Effective localization of mobile robots is a challenging task. There are multiple sources of uncertainty that the localization module has to deal with in order to estimate robot location with reasonable accuracy at each instance in time. The simplest form of localization is dead reckoning (DR) \cite{choset2005principles}. In this method, equations of motion of the mobile robot are devised based on robot structural properties such as wheel circumference, onboard sensors e.g. IMU, magnetometer, and rotary encoders. Knowing number of wheel rotations (using rotary encoder) and direction of movement, it is possible to estimate the robot position and heading at each instance in time. However, sensor readings are always noisy so estimated positions are prone to errors. The situation gets worse by stochastic phenomena of the surrounding environment. For example, in case of wheel slippage, the rotary encoder reports wheel movement, however, due to slippage the robot may not have move at all leading to incorrect position estimation. To deal with localization error, it is common practice to fuse readings from multiple sensors to get a better position estimation.  In the remainder of this section, some of the most famous filtering algorithms that are used in mobile robot localization are reviewed.
\subsection{Kalman filter}
\begin{figure}[!t] 
\centering
\subfloat[]{\includegraphics[width=2in]{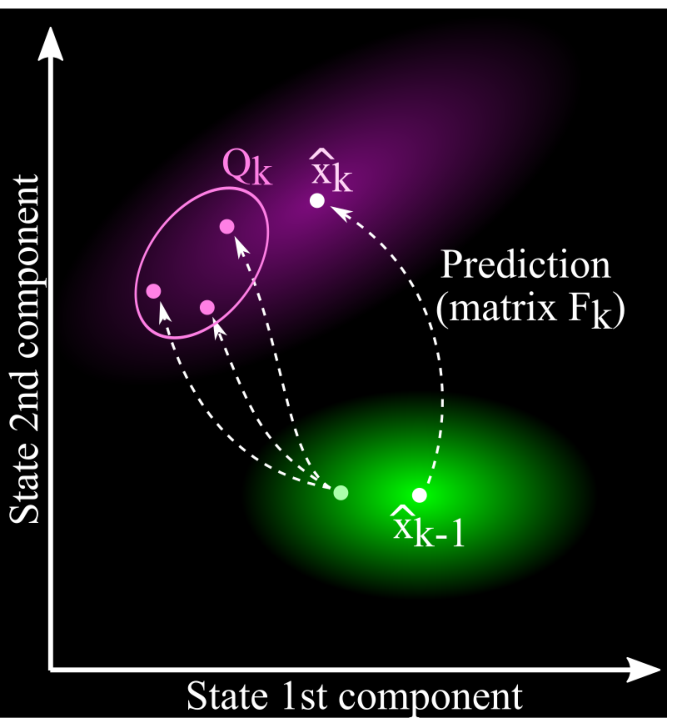}
\label{kf-prediction-phase-fig}}
\subfloat[]{\includegraphics[width=3.97in]{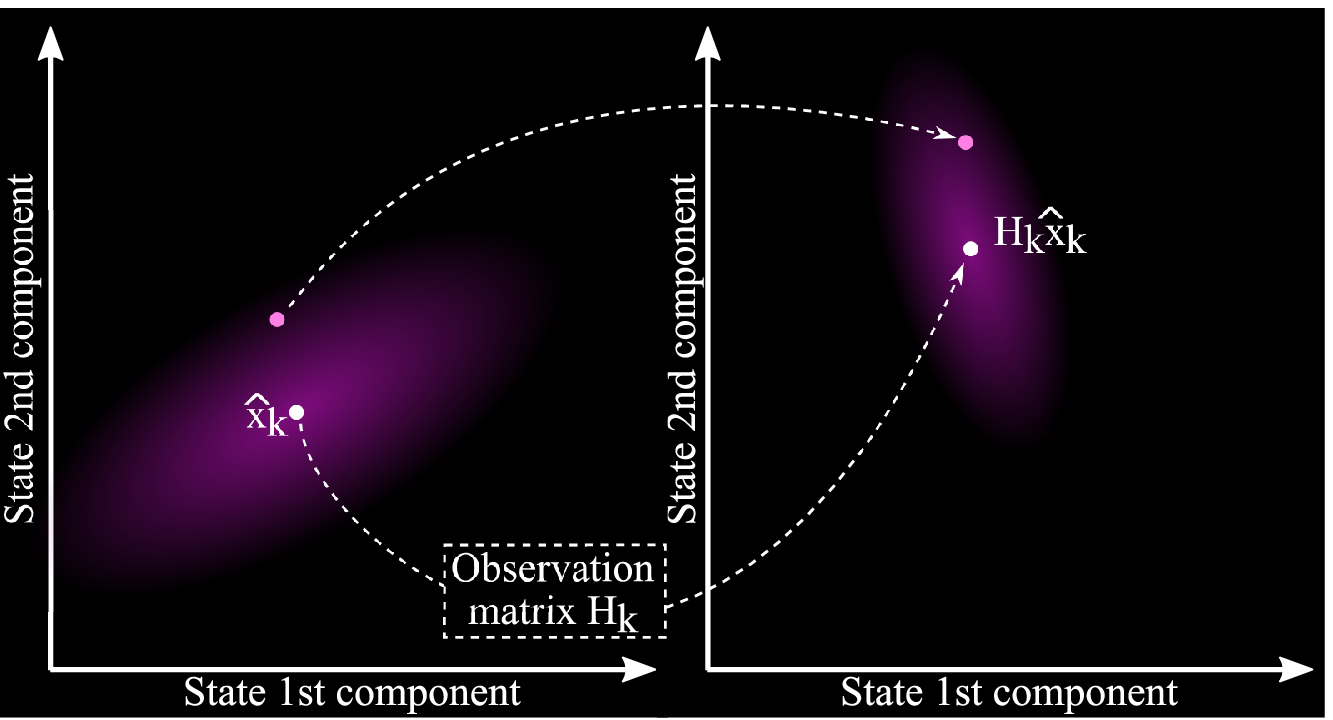}
\label{kf-expected-sensor-readings-fig}}
\\
\subfloat[]{\includegraphics[width=5.2in]{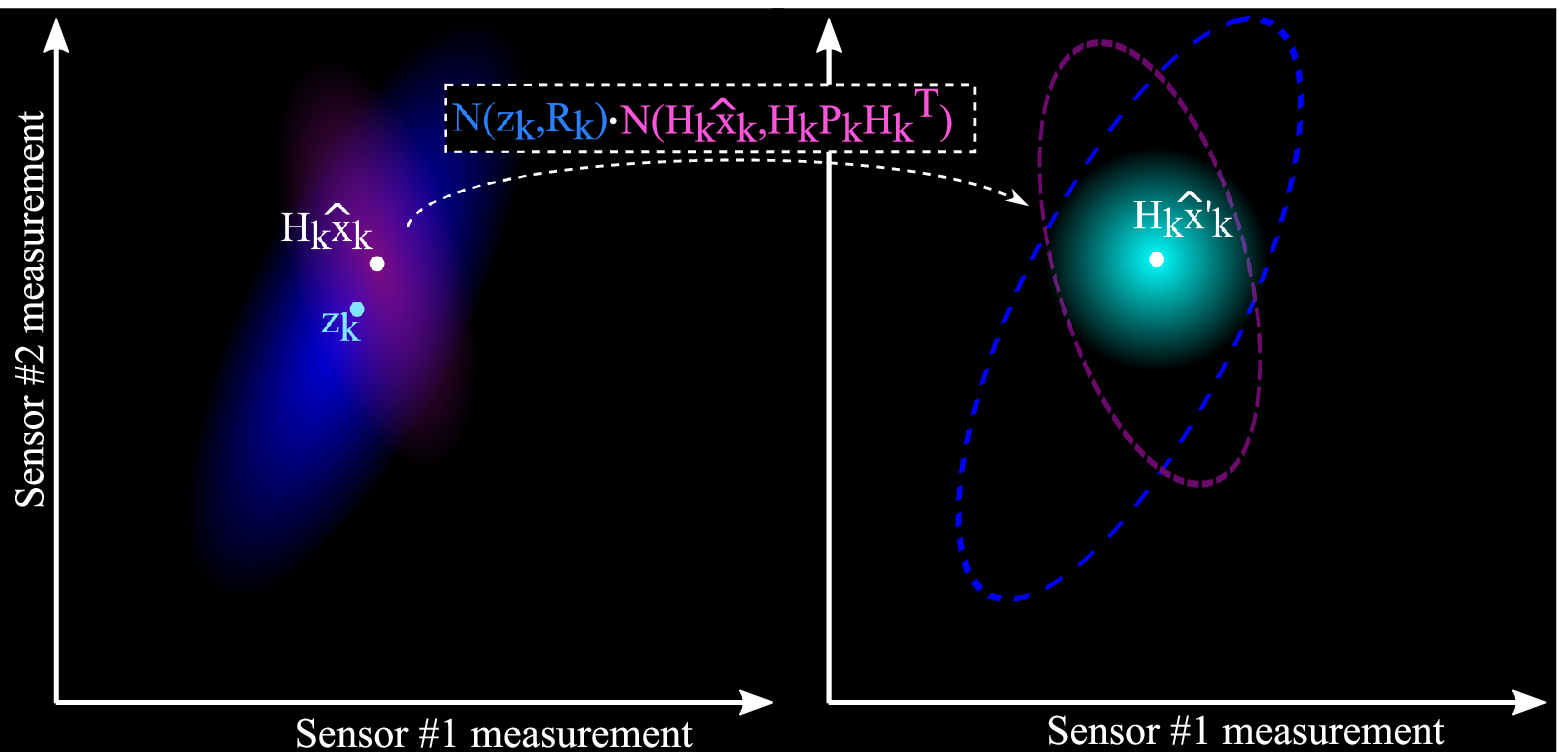}%
\label{kf-correction-phase-fig}}
\caption{Kalman filter phases for 2D state vector: (a) prediction phase, (b) expected observation computation based on predicted state $\hat{x}_k$, (c) correction phase.}
\label{}			
\end{figure}

Kalman filter (KF) is an algorithm for estimation of unknown variables based on a sequence of possibly noisy observations (measurements) collected over time. The estimated values will be more precise compared to the values estimated based on a single observation. KFs do not keep history of past observations. Only the previous state is kept so memory requirements are light. KFs computations are fast enough to be used in real-time applications such as robot localization.
KF models the unknown current state vector $x_{k-1}$ with a Gaussian distribution $\mathcal{N}(\hat{x}_{k-1}, P_{k-1})$ in which $k-1$ is the current time step, $\hat{x}_{k-1}$ and $P_{k-1}$ are the mean and covariance matrix of the distribution. The hat accent of $\hat{x}_{k-1}$ emphasizes that mean is not necessarily equal to the true current state $x_{k-1}$. The motivation for this representation is that the true state $x_{k-1}$ is not known but some states are more likely than others and the elements of the state vector may be correlated. These two properties are captured by the Gaussian distribution covariance matrix.
KF cycles through prediction and correction phases. In the prediction phase, the linear equations governing the robot motion are represented as matrix $F_k$ which can be multiplied with the current state estimate $\hat{x}_{k-1}$ to predict the next state $\hat{x}_k$. However, the effect of external control commands (or forces) such as turning the wheel or stopping must be taken into account as well. In KF terminology, the control vector is denoted as $u_k$. How $u_k$ affects the next state is represented by matrix $B_k$. The final equation for predicting next state using KF reads as:
\begin{equation}\label{kf-prop-eq}
\textcolor{magenta}{\hat{x}_k} = F_k \textcolor{Green}{\hat{x}_{k-1}} + B_k u_k.
\end{equation}
It is also necessary to update the covariance matrix $P_k$. This is done using matrix $F_k$. However, the uncertainty due to external events such as robot wheel slippage or wind force (for aerial robots) must be considered as well. KF models external uncertainty by mapping each point from the current Gaussian $\mathcal{N}(\hat{x}_{k-1}, P_{k-1})$ to multiple new points which will be part of the next Gaussian distribution $\mathcal{N}(\hat{x}_{k}, P_{k})$. The external uncertainty is captured by adding covariance matrix $Q_k$ to the update equation of $P_k$ yielding: 
\begin{equation}
\textcolor{magenta}{P_k} = F_k \textcolor{Green}{P_{k-1}} F_k^T + Q_k,
\end{equation}
which means the new estimate of $P_k$ is computed from the current estimate $P_{k-1}$ and additional environment uncertainty represented by $Q_k$. The prediction process is graphically illustrated in figure \ref{kf-prediction-phase-fig}.
In the correction phase, the expected observations (i.e. expected sensor readings) are computed by multiplying matrix $H_k$ with the new estimates $\{\hat{x}_k, P_k\}$ (figure \ref{kf-expected-sensor-readings-fig}). Actual observations are also obtained from the onboard sensors.

Therefore, the overlap of Gaussian distributions $\mathcal{N}(H_k\hat{x}_k, H_k P_k H_k^T)$ and $\mathcal{N}(z_k, R_k)$ can be computed to obtain the corrected observation distribution with mean $H_k\hat{x}_k'$ as shown in figure \ref{kf-correction-phase-fig}. After some manipulation, the $H_k$ in front of $H_k\hat{x}_k'$ can be dropped to get the final equations for correcting the prediction ($\{\hat{x}_k, P_k\}$):
\begin{eqnarray*}
\textcolor{RoyalBlue}{\hat{x}_k'} = \textcolor{magenta}{\hat{x}_k} + \textcolor{Sepia}{K_k} \left( \textcolor{blue}{z_k} - H_k\textcolor{magenta}{ \hat{x}_k} \right),\\
\textcolor{RoyalBlue}{P'_k} = \textcolor{magenta}{P_k} - \textcolor{Sepia}{K_k} H_k\textcolor{magenta}{P_k}, \\
\textcolor{Sepia}{K_k} = \textcolor{magenta}{P_k} H_k^T \left( H_k \textcolor{magenta}{P_k} H_k^T + \textcolor{blue}{R_k} \right) ^{-1},
\end{eqnarray*}
where $R_k$ is the covariance matrix describing sensors uncertainty and $K_k$ is known as the Kalman gain.
\subsection{Extended Kalman filter}\label{ekf-sec}
KF is a powerful state estimator even in the presence of noisy sensor readings. However, to be able to use KF, the state propagation equations must be linear. This requirement rarely holds in real-world applications. Even in the case of a simple differential drive robot \cite{CS-W4733-NOTES-Differential-Drive-Robots}, the kinematics equations are nonlinear. That is where the extended Kalman filter (EKF) \cite{welch1995introduction} comes in. The main idea behind EKF is to linearize nonlinear state propagation equations using Jacobian (matrix of partial derivatives) so that the propagation equations can be expressed as equation \ref{kf-prop-eq}. To this end, partial derivatives of the next state $\hat{x}_{k}$ are computed with respect to the current state $\hat{x}_{k-1}$ yielding matrix $F_k$ and partial derivatives of the next state $\hat{x}_{k}$ with respect to the control vector $u_k$ are computed yielding matrix $B_k$:
\begin{equation*}
F_k = \begin{bmatrix}
\frac{\partial \hat{x}_{k,1}}{\partial \hat{x}_{k-1,1}} & \hdots & \frac{\partial \hat{x}_{k,1}}{\partial \hat{x}_{k-1,n}} \\ 
\vdots & \ddots & \vdots\\ 
\frac{\partial \hat{x}_{k,n}}{\partial \hat{x}_{k-1,1}} & \hdots & \frac{\partial \hat{x}_{k,n}}{\partial \hat{x}_{k-1,n}}  & 
\end{bmatrix}, \quad 
B_k = \begin{bmatrix}
\frac{\partial \hat{x}_{k,1}}{\partial u_{k,1}} & \hdots & \frac{\partial \hat{x}_{k,1}}{\partial u_{k,m}} \\ 
\vdots & \ddots & \vdots\\ 
\frac{\partial \hat{x}_{k,n}}{\partial u_{k,1}} & \hdots & \frac{\partial \hat{x}_{k,n}}{\partial u_{k,m}}  & 
\end{bmatrix},
\end{equation*}
where state vectors are assumed to be n-dimensional. Following a similar pattern, nonlinear equations for mapping predicted next state $\hat{x}_{k}$ to predicted observations $\hat{z}$ are linearized using partial derivatives of observations (sensor readings) with respect to $\hat{x}_{k}$:
\begin{equation}
H_k = \begin{bmatrix}
\frac{\partial \hat{z}_{k,1}}{\partial \hat{x}_{k,1}} & \hdots & \frac{\partial \hat{z}_{k,1}}{\partial \hat{x}_{k,n}} \\ 
\vdots & \ddots & \vdots\\ 
\frac{\partial \hat{z}_{k,h}}{\partial \hat{x}_{k,1}} & \hdots & \frac{\partial \hat{z}_{k,h}}{\partial \hat{x}_{k,n}}  & 
\end{bmatrix},
\end{equation}
where observation vector is assumed to be h-dimensional.
\subsection{Unscented Kalman Filter}
As mentioned in section \ref{ekf-sec}, EKF tackles the nonlinear  system equations using their first order approximations. In practice, this linearization method may lead to a posterior distribution with corrupted mean and covariance. This issue has been addressed by Unsented Kalman Filter (UKF) \cite{wan2001unscented} via taking a derivative-free approach based on deterministic sampling. Instead of employing a Gaussian, UKF represents the state distribution by carefully choosing a set of sample points around the distribution mean. These points are known as the sigma points $\{ \sigma^{(k-1)}_i, i=1,..., 2n+1 \}$ where $n$ is the dimension of the state vector and $k-1$ refers to the current time step. As pointed out in \cite{wan2001unscented}, the sigma points are selected as:
\begin{align*}
&\sigma^{(k-1)}_0 = \hat{x}_{k-1}, \\
&\sigma^{(k-1)}_i = \hat{x}_{k-1} \textcolor{blue}{+} \left( \sqrt{(n + \lambda) P_{k-1}} \right)_{\textcolor{blue}{i}}, \quad i = 1, ..., n, \\
&\sigma^{(k-1)}_i = \hat{x}_{k-1} \textcolor{blue}{-} \left( \sqrt{(n + \lambda) P_{k-1}} \right)_{\textcolor{blue}{i-n}}, \quad i = n+1, ..., 2n
\end{align*}
where $\lambda$ is a scaling parameter and $\left( \sqrt{(n + \lambda) P_{k-1}} \right)_{i}$ refers to the i-th column of matrix square root of $(n + \lambda) P_{k-1}$. The square root may be computed e.g. by lower triangular Cholesky factorization.

Similar to KF, UKF workflow includes prediction and correction phases followed by a new phase for sigma points selection. Assuming that the next state $\hat{x}_k$ is calculated as nonlinear mapping $f(.)$ of the current state $\hat{x}_{k-1}$, in the prediction phase, the sigma points undergo unscented transform \cite{julier1997new} by being fed to the nonlinear function $f(.)$:
\begin{equation*}
\sigma^{(k)}_i = f(\sigma^{(k-1)}_i), \quad i=1, ..., 2n+1.
\end{equation*}
The weighted average of the new sigma points are then used to estimate the mean and covariance of the predicted state:
\begin{align*}
&\hat{x}_k \simeq \sum_{i=1}^{2n+1} w_i^{(m)} \sigma^{(k)}_i, \\
&P_k \simeq \sum_{i=1}^{2n+1} w_i^{(c)} \left(\sigma^{(k)}_i - \hat{x}_k \right) \left(\sigma^{(k)}_i - \hat{x}_k \right)^T.
\end{align*}
The transformed sigma points are then used to calculate the mean and covariance of the new state Gaussian distribution. The estimated mean and covariance are accurate up to the second order of Taylor series expansion \cite{wan2001unscented}.

\begin{figure}[!t] 
\centering
\subfloat[True mapping]{\includegraphics[width=1.98in]{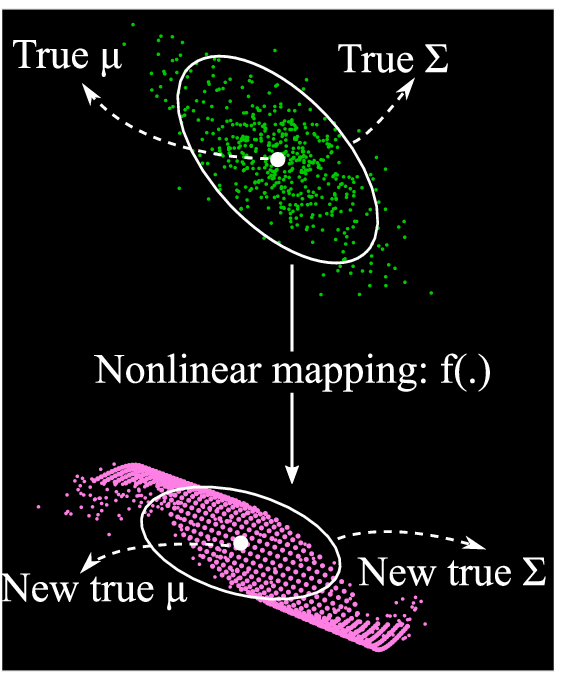}%
\label{ukf-true-mapping-fig}}
\subfloat[EKF prediction]{\includegraphics[width=1.98in]{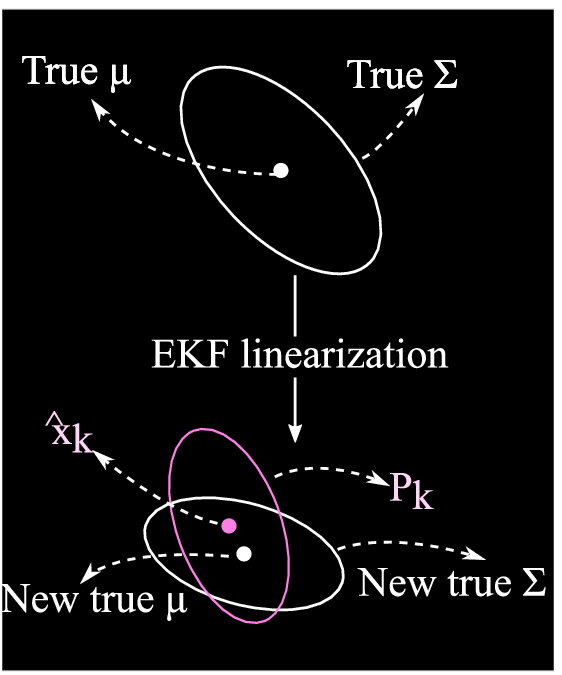}%
\label{ukf-ekf-mapping-fig}}
\subfloat[UKF prediction]{\includegraphics[width=1.98in]{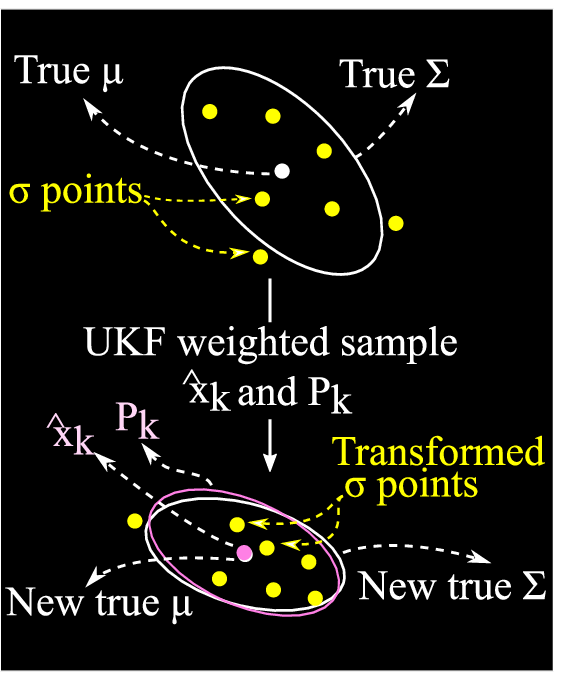}%
\label{ukf-ukf-mapping-fig}}
\caption{Illustrating the state propagation mechanisms of EKF and UKF compared to the true mapping.}
\label{ukf-mapping-illustration-fig}			
\end{figure}

The difference between EKF and UKF state propagation mechanism has been depicted in figure \ref{ukf-mapping-illustration-fig}. Given the current distribution with mean $\mu$ and covariance $\Sigma$, in the absence of noise, applying nonlinear mapping $f(.)$ yields the new mean and covariance (figure \ref{ukf-true-mapping-fig}). The drawback of EKF linearization is apparent in figure \ref{ukf-ekf-mapping-fig}. The error introduced by the first order linearization has led to mismatch between true ($\mathcal{N}(\mu_{New}, \Sigma_{New})$) and predicted ($\mathcal{N}(\hat{x}_k, P_k)$) distributions. The distribution mismatch has been addressed by UKF in figure \ref{ukf-ukf-mapping-fig}. To this end, multiple sigma points sampled from the current distribution have been transformed based on $f(.)$. The transformed points are then used to calculate the new mean and covariance. Such an approach is more flexible and reliable compared to EKF first order linearization.
\subsection{Multi-State Constraint Kalman Filter}\label{MSCKF-sec}
Inertial navigation system (INS) is the backbone of autonomous navigation that can be applied on aerial  \cite{sayre2018visual, geneva2020openvins, morales2021tightly}, ground \cite{lobo1995inertial, cai2019multi, chae2020robust}, and underwater vehicles \cite{hegrenas2008model, hegrenaes2009doppler, bao2020integrated}. Relying primarily on IMU, INS provides very cost-efficient navigation solution. However, INS localization accuracy degrades over time due to accumulative error. INS  employs numerical integration over gyroscope and accelerometer readings to estimate the position and orientation of the platform it is mounted on. Applying numerical integration in successive time steps causes the aforementioned accumulated error. That is why INS is usually paired with auxiliary sensor(s) e.g. GPS, magnetometer, camera, etc. for error correction. The choice of the auxiliary sensors depends on the characteristics of the environment (e.g. indoor/outdoor, in the air/on the ground/underwater).

\begin{figure}[!tbp]
\centering
\includegraphics[width=0.7\textwidth]{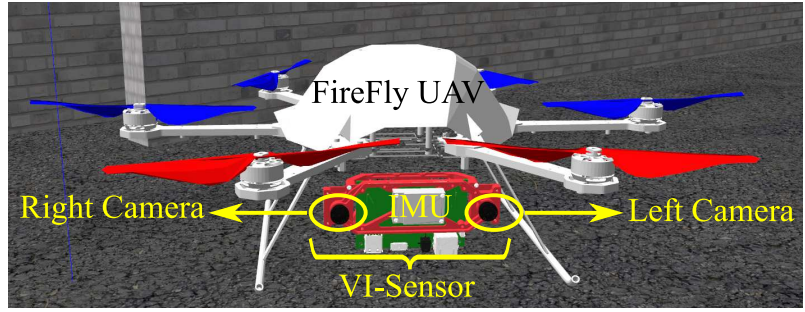}
    \caption{Simulated Astec Firefly UAV equipped with VI-sensor: the image is captured while running Rotors\_Simulator \cite{eth-rotors-sim, Furrer2016} in Gazebo.}
    \label{vi-sensor-fig}
\end{figure}

Visual sensors such as cameras can be used in variety of environments and they are available at reasonable prices. That is the motivation behind Multi-State Constraint Kalman Filter (MSCKF) which is an EKF-based method designed to implement visually-aided inertial navigation in real-time  \cite{mourikis2007multi}. The main difference between MSCKF and ordinary EKF is the fact that MSCKF  uses a novel measurement model capable of capturing geometric constraints  that arise when a static feature appears in multiple cameras with different poses. Thanks to this new measurement model, MSCKF is able to perform localization which is accurate up to linearization error of EKF only using visual sensor(s). Moreover,  MSCKF computational complexity scales linearly with the number of features which is remarkable. One of the notable applications of MSCKF is visual-inertial navigation system OpenVINS implemented by Geneva et al. \cite{geneva2020openvins} which utilizes a VI-sensor (visual-inertial sensor) similar to the one shown in figure \ref{vi-sensor-fig}. The VI-sensor consists of two cameras and one 3-axis IMU which are hardware-synchronized.

The goal of MSCKF is 3D pose estimation of reference frame $\{I\}$ of an IMU strapped to a mobile (robot) platform with respect to some global reference frame $\{G\}$. To this end, the IMU state is written as \cite{mourikis2007multi}:
\begin{equation}\label{imu-state-eq}
X_{IMU} = \left[ {}^{I}_{G}\bar{q}^{T}_{}, b_g^T,   {}^{G}_{}v^{T}_{I}, b_a^T,  {}^{G}_{}p^{T}_{I} \right]^T,
\end{equation} 
where the rotation from global frame $\{G\}$ to IMU frame $\{I\}$ is expressed as a unit quaternion ${}^{I}_{G}\bar{q}^{T}_{}$. Moreover, ${}^{G}_{}v^{T}_{I}$ and ${}^{G}_{}p^{T}_{I}$ represent the IMU position and velocity with respect to $\{G\}$, and the $3 \times 1$ vectors $b_g$ and $b_a$ express the gyroscope and accelerometer bias affecting their measurements. Based on equation \ref{imu-state-eq}, the error state of the IMU reads as \cite{mourikis2007multi}:
\begin{equation}
\tilde{X}_{IMU} = \left[ \delta \theta_I^T, \tilde{b}_g^T, {}^{G}_{}\tilde{v}^{T}_{I}, \tilde{b}_a^T,  {}^{G}_{}\tilde{p}^{T}_{I} \right]^T,
\end{equation}
where the additive error definition (i.e. difference between true value and estimated value) is used to compute error for position, velocity, and biases but the orientation error is described by the error quaternion $\delta \bar{q}$ such that the relation $\bar{q} = \delta \bar{q} \otimes \hat{\bar{q}}$ holds in which $\otimes$ stands for quaternion multiplication.

As the objective of MSCKF is an accurate estimation of IMU pose and orientation, the IMU state (equation \ref{imu-state-eq}) is used as part of the EKF state vector \cite{mourikis2007multi}:
\begin{equation}\label{ekf-state-vec-eq}
\hat{X}_k = \left[ \hat{X}_{IMU_k}^T, {}^{C_1}_{G}\hat{\bar{q}}^{T}_{},  {}^{G}_{}\hat{p}^{T}_{C_1}, ...,  {}^{C_N}_{G}\hat{\bar{q}}^{T}_{},  {}^{G}_{}\hat{p}^{T}_{C_N} \right]^T,
\end{equation}
where subscript $k$ in $\hat{X}_k$ denotes the EKF state at time step $k$.
In addition to the IMU state, equation \ref{ekf-state-vec-eq} contains $N$ pairs of $\{{}^{C_i}_{G}\hat{\bar{q}}^{T}_{},  {}^{G}_{}\hat{p}^{T}_{C_i}\}$ items where ${}^{C_i}_{G}\hat{\bar{q}}^{T}_{}$ is the quaternion expressing the rotation from the i-th camera frame $\{C_i\}$ to the global frame and ${}^{G}_{}\hat{p}^{T}_{C_i}$ denotes the origin of frame $\{C_i\}$ expressed in the global frame. The aforementioned pairs corresponding to $N$ cameras are considered in the EKF state to establish geometric constraints induced by observation of static features from multiple cameras poses. Looking at equation \ref{ekf-state-vec-eq}, the EKF error-state vector reads as \cite{mourikis2007multi}:
\begin{equation}
\tilde{X}_k = \left[ \tilde{X}_{IMU_k}^T, \delta \theta_{C_1}^T,  {}^{G}_{}\tilde{p}^{T}_{C_1}, ..., \delta \theta_{C_N}^T,  {}^{G}_{}\tilde{p}^{T}_{C_N} \right]^T,
\end{equation}
where $\{\delta \theta_{C_i}^T,  {}^{G}_{}\tilde{p}^{T}_{C_i}\}$ expresses the i-th camera attitude and position error.

Recall that EKF consists of prediction and correction phases. In the prediction phase, the IMU state (i.e. position and attitude of frame $\{I\}$) is propagated according to the set of equations that will be described below.

The first equation expresses how quaternion ${}^{I}_{G}\hat{\bar{q}}^{}_{}$ evolves over time \cite{mourikis2007multi}:
\begin{align}\label{imu-quat-propagate-eq}
{}^{I}_{G}\dot{\hat{\bar{q}}}^{}_{} = \frac{1}{2} \Omega(\hat{\omega}) {}^{I}_{G}\hat{\bar{q}}^{}_{}, 
\end{align}
where ${}^{I}_{G}\dot{\hat{\bar{q}}}^{}_{}$ is the derivative of quaternion $\bar{q}$ at time step $t$ and matrix $\Omega(\hat{\omega})$ is computed as \cite{mourikis2007multi}:
\begin{equation}\label{big-omega-mat-eq}
\Omega(\hat{\omega}) = \begin{bmatrix}
- \lfloor \hat{\omega} \times \rfloor & \hat{\omega} \\
-\hat{\omega}^T & 0
\end{bmatrix}.
\end{equation}
In equation \ref{big-omega-mat-eq}, $\hat{\omega} = \omega_m - \hat{b}_g - C({}^{I}_{G}\hat{\bar{q}}^{}_{}) \omega_G$ where $\omega_m$ is the gyroscope sensor reading from which $\hat{b}_g$ and  $C({}^{I}_{G}\hat{\bar{q}}^{}_{}) \omega_G$ are subtracted to cancel the effect of gyroscope bias and angular rate $\omega_G$ caused by planet's rotation, respectively. Note that the term $C({}^{I}_{G}\hat{\bar{q}}^{}_{})$ represents the rotational matrix corresponding to quaternion ${}^{I}_{G}\hat{\bar{q}}^{}_{}$. The other component of matrix $\Omega(\hat{\omega})$ (equation \ref{big-omega-mat-eq}) is $- \lfloor \hat{\omega} \times \rfloor$ which is known as the skew symmetric matrix of angular velocity vector $\hat{\omega} = [\hat{\omega}_x, \hat{\omega}_y, \hat{\omega}_z]^T$ \cite{titterton2004strapdown}.
The second and third equations of IMU state propagation express the evolution of gyroscope and accelerometer biases ($\dot{\hat{b}}_g,\dot{\hat{b}}_a$) over time \cite{mourikis2007multi}:
\begin{equation}
\dot{\hat{b}}_g = \mathbb{E}\left[ n_g \right]=[0,0,0]^T, \: \dot{\hat{b}}_a = \mathbb{E}\left[ n_a \right] = [0,0,0]^T, \label{bg-dot,ba-dot-eq}
\end{equation}
where $n_g$ and $n_a$ are the gyroscope and accelerometer biases, respectively and they are modeled as zero-mean white Gaussian noise processes. That is why expectation of $n_g$ and $n_a$ are zero vectors as shown in equation \ref{bg-dot,ba-dot-eq}. The fourth equation related to IMU state propagation reads as \cite{mourikis2007multi}:
\begin{equation}\label{imu-accl-propagate-eq}
{}^{G}_{}\dot{\hat{v}}^{}_{I} = C({}^{I}_{G}\hat{\bar{q}}^{}_{})^T \hat{a} - 2 \lfloor \omega_G \times \rfloor {}^{G}_{}\hat{v}^{}_{I} - \lfloor \omega_G \times \rfloor^2 \: {}^{G}_{}\hat{p}^{}_{I} + G_g, 
\end{equation}
where ${}^{G}_{}\dot{\hat{v}}^{}_{I}$ is the derivative of IMU frame velocity which expresses the IMU frame acceleration with respect to frame $\{G\}$. Moreover, $\hat{a} = a_m - \hat{b}_a$ where $a_m$ is the accelerometer sensor reading from which the accelerometer bias $\hat{b}_a$ is subtracted. The term $\lfloor \omega_G \times \rfloor$ in equation \ref{imu-accl-propagate-eq} is the skew symmetric matrix corresponding to planet's rotation $\omega_G$ and as mentioned before, ${}^{G}_{}\hat{v}^{}_{I}$ and ${}^{G}_{}\hat{p}^{}_{I}$ represent the IMU frame velocity and position with respect to frame $\{G\}$ and $G_g$ stands for gravitational acceleration expressed in frame $\{I\}$. The last equation for IMU state propagation establishes the relation between the IMU velocity and position \cite{mourikis2007multi}:
\begin{equation}\label{imu-vel-pos-relation-eq}
{}^{G}_{}\dot{\hat{p}}^{}_{I} = {}^{G}_{}\hat{v}^{}_{I}.
\end{equation}
To sum up, the set of equations \ref{imu-quat-propagate-eq}-\ref{imu-accl-propagate-eq} form the basis of the prediction phase of MSCKF. Considering that MSCKF is based on EKF, it is necessary to linearize equations  \ref{imu-quat-propagate-eq}-\ref{imu-accl-propagate-eq}:
\begin{equation}
\dot{\tilde{X}}_{IMU} = F \tilde{X}_{IMU} + G_{n_{IMU}},
\end{equation}
where matrices $F$ and $G_{n_{IMU}}$ are the results of linearizing equations \ref{imu-quat-propagate-eq}-\ref{imu-accl-propagate-eq} (see \cite{mourikis2007multi} for further details) and $n_{IMU} = \left[ n_g^T, n_{wg}^T, n_a^T, n_{wa}^T \right]^T$ is the system noise consisting of gyroscope and accelerometer noise densities ($n_g, n_a$) and random walks ($n_{wg}, n_{wa}$). The covariance matrix $Q_{IMU}$ corresponding to the system noise $n_{IMU}$ is computed during the calibration process offline. Kalibr \cite{eth-kalibr-tool,rehder2016extending, furgale2013unified, furgale2012continuous,maye2013self,oth2013rolling} is one of the famous calibration tools implemented in ROS which can be used to determine IMU noise density and random walk parameters as well as camera intrinsic and extrinsic parameters. Moreover, Kalibr offers spatial (coordinate transform) and temporal (time synchronization) calibration of an IMU with respect to a mono/stereo camera system.

As mentioned before, MSCKF exploits the presence of 3D features in multiple camera poses. To this end, the EKF state vector is augmented with the camera poses from which the 3D features are observed. The camera poses are determined based on the current estimate of the IMU pose. The EKF covariance matrix is updated as well.

To implement the correction phase of the EKF, a new measurement model is used. Consider a single feature $f_j$ which has been observed in $M_j$ camera poses. Then according to the measurement model, each of these $M_j$ observations is expressed as \cite{mourikis2007multi}:
\begin{equation}\label{msckf-measurement-model-eq}
z_i^{(j)} = \frac{1}{{}^{C_i}_{}Z^{}_{j}}
 \begin{bmatrix}
{}^{C_i}_{}X^{}_{j} \\
{}^{C_i}_{}Y^{}_{j}
\end{bmatrix}
+ n_i^{(j)}, \quad i \in S_j,
\end{equation}
where $n_i^{(j)}$ is a $2 \times 1$ vector expressing image noise and ${}^{C_i}_{}p^{}_{f_j}=\left[{}^{C_i}_{}X^{}_{j}, {}^{C_i}_{}Y^{}_{j}, {}^{C_i}_{}Z^{}_{j} \right]$ is the 3D position of the feature $f_j$ in i-th camera pose frame $\{C_i\}$. To determine ${}^{C_i}_{}p^{}_{f_j}$, the 3D position of the feature $f_j$ in global frame (${}^{G}_{}p^{}_{f_j}$) is needed which is unknown. To estimate ${}^{G}_{}p^{}_{f_j}$, lease-squares minimization is employed. The estimated vector ${}^{G}_{}\hat{p}^{}_{f_j}$ is used to compute estimated 3D position of the feature $f_j$ in frame $\{C_i\}$ (i.e. ${}^{C_i}_{}p^{}_{f_j}$) using the transformation from frame $\{G\}$ to frame $\{C_i\}$. Using ${}^{C_i}_{}p^{}_{f_j}$ and equation \ref{msckf-measurement-model-eq}, the measurement model output (i.e. $\hat{z}_i^{(j)}$) can be computed. The difference between $\hat{z}_i^{(j)}$ and the measurement obtained from the environment ($z_i^{(j)}$) forms the measurement residual $r_i^{(j)}$ which is linearized to comply with EKF requirements. Gathering all residuals $\{r_i^{(j)}, i=1, ..., M_j\}$ corresponding to feature $f_j$ forms an equation enforcing constraint between all the camera poses from which feature $f_j$ has been observed. The constraint equation can then be used in the update (correction) phase of the EKF.
%
\subsection{Particle filter}
Particle filter (PF) is one of the popular sample-based approaches for optimal state estimation in nonlinear and non-Gaussian scenarios. Contrary to the EKF,  PF does not use local linearization which leads to more flexibility and representation power. However, more flexibility of PF comes at the expense of higher computational complexity. The goal of PF is estimation of state vector $x_k$ at each time step $k$. PF requires process and measurement models. The process model $f(\hat{x}_{k-1}, u_k, v_{k-1})$ can be nonlinear and represents the mapping from current state $\hat{x}_{k-1}$ to the next state $\hat{x}_{k}$. Apart from the state vector $\hat{x}_{k-1}$, the function $f(.)$ receives deterministic control input $u_k$ and process noise $v_{k-1}$ which is independently and identically distributed across different time steps and captures the process model uncertainties. The measurement model $z_k = h_k(\hat{x}_k,u_k,\zeta_k)$ establishes the relation between measurements $z_k$ and state $\hat{x}_{k}$. The measurement noise is represented by $\zeta_k$ which is independently and identically distributed. In case of having multiple sensors, each sensor will have its own measurement model.

As pointed out in \cite{gustafsson2002particle}, PF expresses the measurement update of  the next state as a probability density function (PDF) conditioned on the past observations $z_{1:k-1}$:
\begin{equation}\label{pf-eq1-eq}
p(\hat{x}_k | z_{1:k-1}) = \int p(\hat{x}_{k}|\hat{x}_{k-1}) \textcolor{blue}{p(\hat{x}_{k-1} | z_{1:k-1})} d\hat{x}_{k-1},
\end{equation}
in which the probability of being at state $\hat{x}_k$ depends on sequence of applied control signals $u_{1:k} = \{u_i, i = 1, ..., k \}$ (not included in equation (\ref{pf-eq1-eq}) for brevity) and sequence of obtained measurements $z_{1:k-1} = \{ z_i, i = 1, ..., k-1 \}$ in the past $k-1$ time steps. The probability $\textcolor{blue}{p(\hat{x}_{k-1} | z_{1:k-1})}$ of equation (\ref{pf-eq1-eq}) can be represented as:
\begin{equation}\label{pf-p(x|z)-eq}
\textcolor{blue}{p(\hat{x}_{k-1} | z_{1:k-1})} = \frac{p(z_{k - 1} | \hat{x}_{k - 1}) p(\hat{x}_{k-1} | z_{-1:k-2})}{p(z_{k - 1} | z_{-1:k-2})},
\end{equation}
where $z_{-1:k-2}$ refers to the set of last $k-1$ observations starting from time step $k-2$.
Analytical solution of the integral in equation (\ref{pf-eq1-eq}) is only possible under heavy restrictive assumptions that rarely hold in practice \cite{elfring2021particle}. Therefore, instead of computing this integral, mean and covariance of the next state posterior (after measurement update) are approximated using equation (\ref{pf-p(x|z)-eq}) and weighted sums over $N$ sampled points $\{x_k^i, i=1, ..., N\}$:
\begin{align}\label{sample-based-mean-pf-eq}
\hat{x}_{k} = \sum_{i = 1}^{N} w_k^i x_k^i.
\end{align}
In equation (\ref{sample-based-mean-pf-eq}), the set of points $\{x_k^i, i=1, ..., N\}$ are known as particles. During the initialization step, the particles are sampled randomly from the initial distribution on the system state. At each time step, in the prediction phase, the particles are propagated according to the process model $f(\hat{x}_{k-1}, u_k, v_{k-1})$ and in the correction phase (i.e. measurement update), they are weighted according to the measurement model and the obtained measurements \cite{gustafsson2002particle}:
\begin{equation}\label{pf-w-update-eq}
w_{k}^{(i)} = w_{k-1}^{(i)} p(z_k | x_k^{(i)}) =  w_{k-1}^{(i)} p_{\zeta_k}\left( z_k - h(x_{k}^{(i)}) \right),
\end{equation}
where in equation (\ref{pf-w-update-eq}), $p_{\zeta_k}(.)$ is the PDF of measurement noise.

\begin{figure}[!t] 
\centering
\subfloat[]{\includegraphics[width=1.7in]{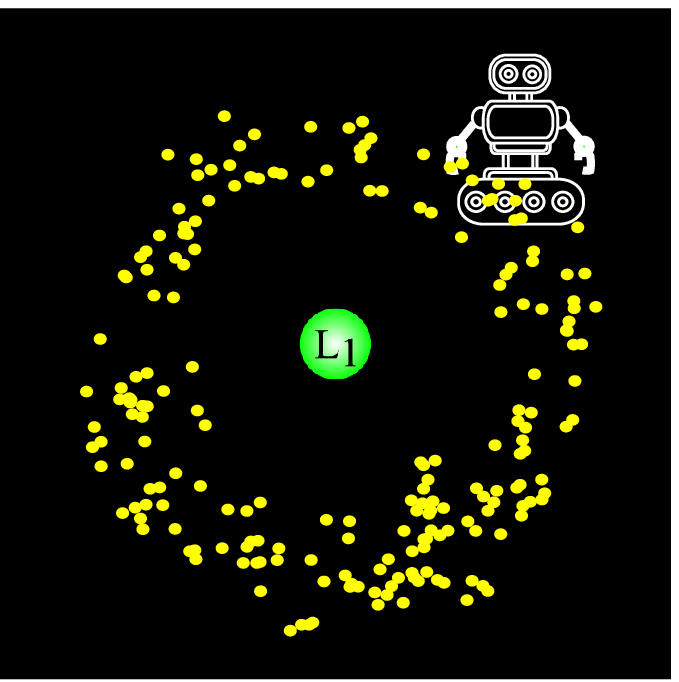}%
\label{pf-1-landmark-fig}}
\hspace{1pt}  
\subfloat[]{\includegraphics[width=1.7in]{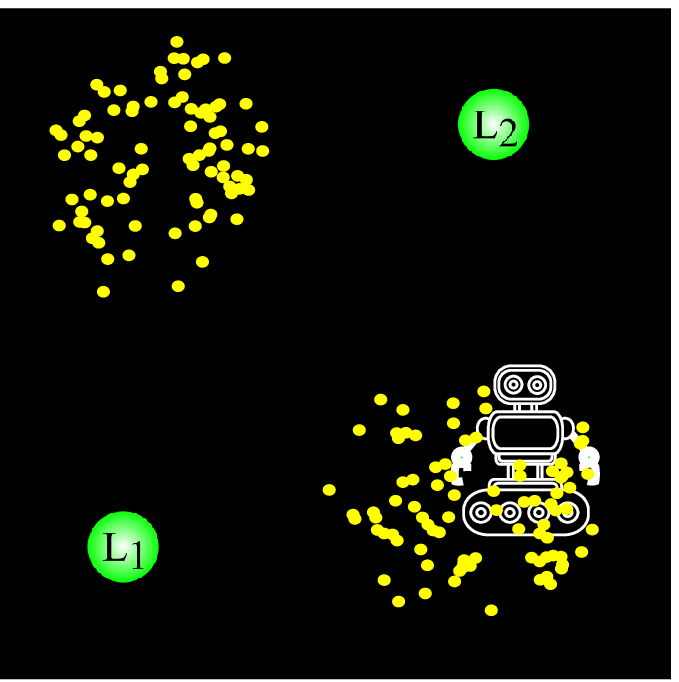}%
\label{pf-2-landmarks-fig}}
\caption{Illustration of particles distribution with (a) one landmark $L_1$, (b) two landmarks $\{L_1,L_2\}$.}
\label{pf-landmarks-fig}			
\end{figure}

Like any other method, PF has its own advantages and disadvantages. Following a sample-based approach, PF is capable of representing any state distribution with arbitrary shape. To make this point clear, consider the localization scenarios of a mobile robot as shown in figure \ref{pf-landmarks-fig}. In figure \ref{pf-1-landmark-fig}, the robot can only sense its distance to landmark $L_1$. Without having access to the landmark angle with respect to the robot, after PF update step, the particles will be spread around the landmark forming a circular shape. Adding a second landmark $L_2$ to the scenario restricts the possible region for robot position leading to a bimodal distribution of the particles (figure \ref{pf-2-landmarks-fig}).

Using PF, process and measurement models are no longer required to be linear which has made PF very popular in the robotics community. The disadvantage of using samples-based representation is that the target distribution will not be captured with reasonable accuracy if the number of particles is not sufficient. On the other hand, as the number of particles is increased, so will the computational complexity. PF performance may degrade due to loss of diversity of particle samples. Methods like resample-move algorithm \cite{gilks2001following}, regularization \cite{musso2001improving}, Rao-Blackwellisation \cite{liu1998sequential}, and multiple Monte-Carlo \cite{liu1998sequential} can be utilized to deal with particles loss of diversity issue. PF has seen multiple variants such as sampling importance resampling (SIR) \cite{arulampalam2002tutorial}, auxiliary sampling importance resampling (ASIR) \cite{pitt1999filtering},  and regularized particle filter (RPF) \cite{doucet2001sequential}.
\section{Obstacle avoidance}\label{obstacle-avoidance-sec}
As mobile robots navigate through their operational environment, they have to deal with variety of obstacles in order to reach their destination. Therefore, the ability to avoid obstacles during navigation is a crucial module in autonomous systems. The location of obstacles may be known as part of the environment map but oftentimes the map is not known beforehand. Thus, it is highly desirable for autonomous mobile robots to be able to detect and avoid obstacles in real-time during navigation. Several obstacle avoidance methods exist in the literature which are reviewed in this section.
\subsection{Legacy methods}
The history of obstacle avoidance methods is almost as old as the emergence of mobile robots which is not surprising because without obstacle avoidance, mobility is not of much use. In this subsection, the classic methods of obstacle avoidance are reviewed.
\subsubsection{Family of Bug algorithms}
\begin{figure}[!t] 
\centering
\subfloat[]{\includegraphics[width=2.5in]{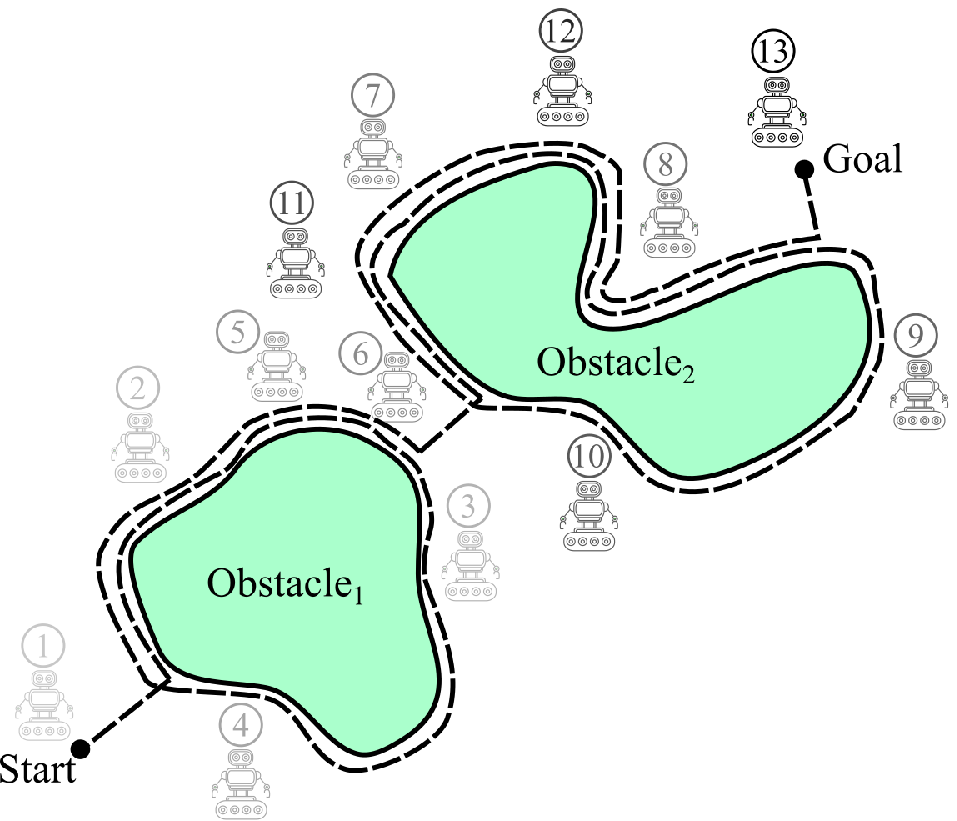}%
\label{bug1-fig}}
\hfil
\subfloat[]{\includegraphics[width=2.5in]{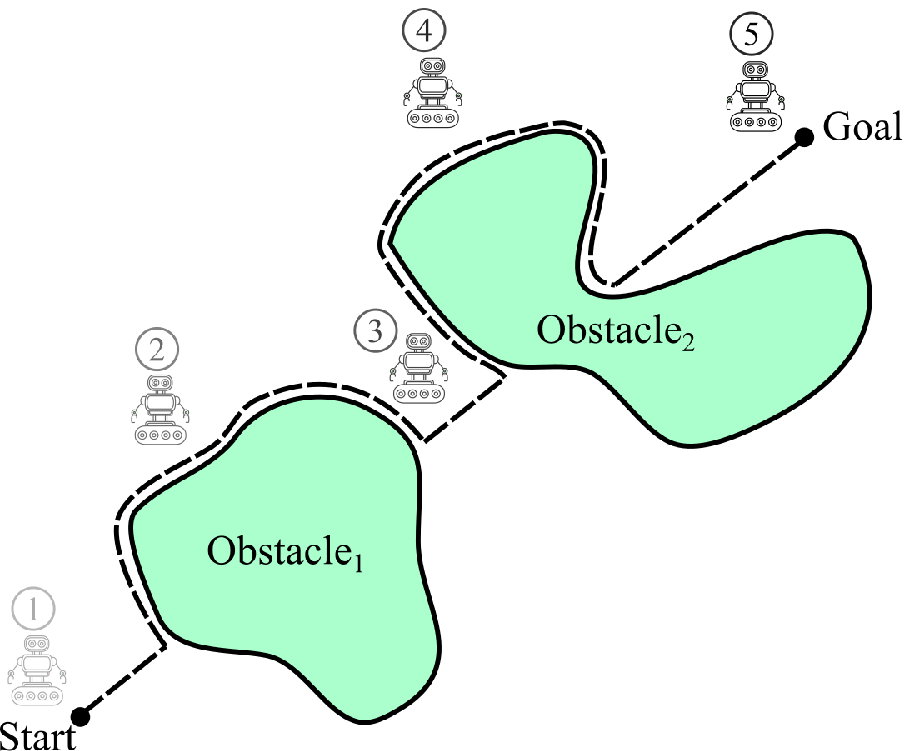}%
\label{bug2-fig}}
\\
\subfloat[]{\includegraphics[width=3in]{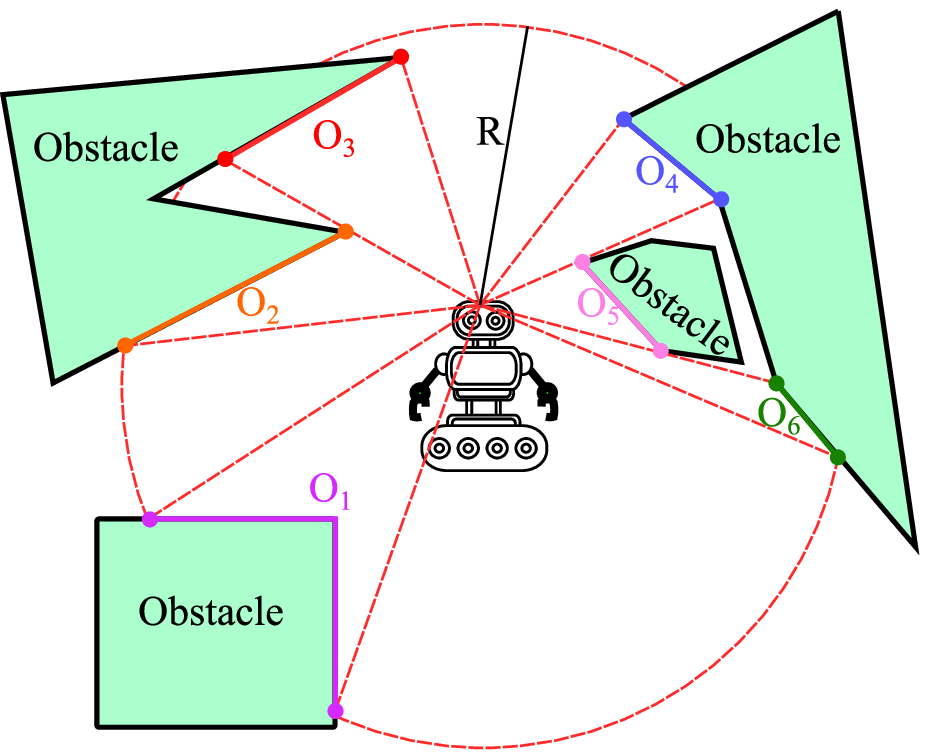}%
\label{tangentbug-fig}}
\hfil
\subfloat[]{\includegraphics[width=2.25in]{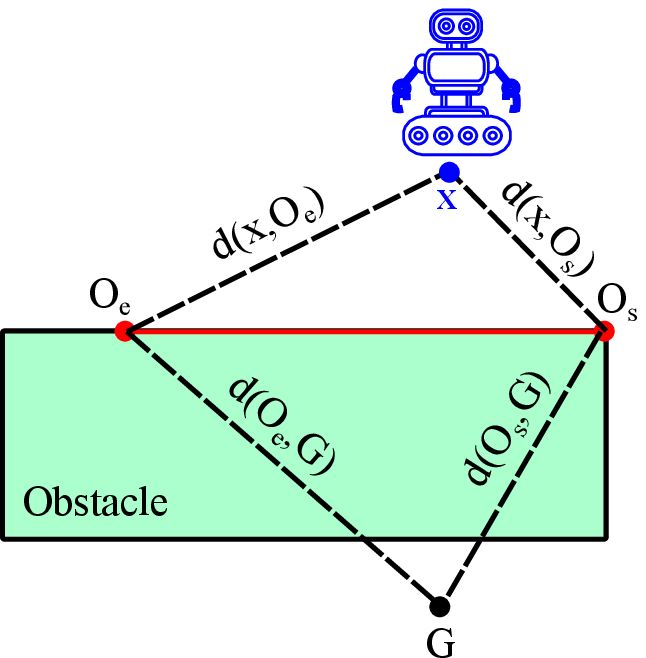}%
\label{tangentbug-heuristic-dist-fig}}
\caption{Illustration of Bug algorithms: (a) Bug1 behavior, (b) Bug2 behavior, (c) local tangent graph used by TangentBug, (d) heuristic distance computation by TangentBug.}
\label{bug-methods-fig}			
\end{figure}

One of the simplest sets of obstacle avoidance methods is the family of Bug algorithms. As the name implies, these algorithms mimic the behavior of bugs upon encountering an obstacle on their way. The family of Bug algorithms began with the proposal of Bug1 and Bug2 \cite{lumelsky1986dynamic}. The obstacle avoidance mechanisms of Bug1 and Bug2 have been depicted in figure \ref{bug1-fig} and \ref{bug2-fig}. According to figure \ref{bug1-fig}, whenever Bug1 bumps into an obstacle, it completes a cycle around it and in the second cycle, departs from the obstacle and continues its path toward the goal until the goal is reached or a new obstacle is detected. As shown in figure \ref{bug2-fig}, Bug2 assumes a hypothetical line from the start point to the goal point and circumvents any obstacle that intersects this line. Clearly, Bug2 is more efficient in terms of traveled distance compared to Bug1. Bug2+ \cite{antich2009bug2+} is an improved version of Bug2. In Bug2+ the condition for switching from obstacle-boundary-following to motion-to-goal has been revised in order to reduce the length of the path traveled from the start to the goal.

Another algorithm from the Bug family is called CBUG \cite{gabriely2008cbug} which considers an ellipse with area equal to $A_0$ such that the start and the goal points form the ellipse's focus points. Within this ellipse, Bug1 is executed in an attempt to reach the goal. If the goal is not reachable in the current ellipse, its area is increasd to $2^i A_0$ where $i$ is the iteration count. Within the new ellipse Bug1 is executed again. This trend is repeated until the goal is reached or it is concluded that the goal is not reachable. As shown in \cite{gabriely2008cbug}, CBUG is an optimal online planner with constant memory requirement.

Bug2+ was originally designed to work with unknown environments but it has been combined with A* to exploit the map of the environment in order to reduce the length of the path to be traveled. The resulting method is an anytime planner called ABUG \cite{antich2009bug}. The advantage of anytime planners is that they plan a sub-optimal path quickly and revise the path during the navigation. The term anytime planner means that the planning process can be interrupted at any time and a valid path is available. The more time is given to the planner the better the planned path will be. In the case of ABUG, whenever an obstacle is encountered, Bug2+ can circumvent it either from the its left or right side. Given the map of the environment, all of the obstacles on the path is known beforehand. That is where A* comes in. For each obstacle, a binary node is created corresponding to going around the left or right side of the obstacle. Set of the obstacle form a binary search tree. Running A* on the tree gives us the optimal path from the start to the goal. 

Perhaps one of the most important Bug algorithms is TangetBug \cite{kamon1998tangentbug} based on which other methods such as WedgeBug \cite{laubach1999autonomous} and CautiousBug \cite{magid2004cautiousbug} have been proposed. WedgeBug uses stereo vision for obstacle detection and its energy consumption, computational complexity, and memory usage has been minimized to meet the limited resources of mars rovers. The name WedgeBug is due to the fact that this algorithm is capable of navigation using the limited ($30^{\circ}$ to $45^{\circ}$) wedge-shaped field-of-view of stereo vision. This algorithm inspects a portion of the environment and if necessary, rotates its stereo vision system to inspect regions in the vicinity of the already observed region. This way, the limited field-of-view is dealt with.  WedgeBug has been further modified to give birth to RoverBug \cite{laubach2000roverbug} used in Rock7 mars rover \cite{laubach1999autonomous}. 

TangentBug utilizes range sensors such as laser scanners to detect obstacles before actually bumping into them. This way, the traveled path toward the goal is reduced significantly. Each detected obstacle is presented by its two endpoints. For example, in figure \ref{tangentbug-fig}, the robot is surrounded by four obstacles that have been represented by the set of six endpoints $\{O_i, i=1,...,6\}$. TangentBug has two primary behavior modes namely motion-to-goal (M2G) and wall-following (WF). Initially, the behavior is set to M2G. In this mode, the robot takes the direct route toward goal $G$. In case an obstacle interferes with the direct route, the robot computes the heuristic distance for the two endpoints of the obstacle as depicted in figure \ref{tangentbug-heuristic-dist-fig}. The heuristic distance $H(.,.)$ is computed as
\begin{equation}
H(x,G) = d(x,O_*) + d(O_*, G)
\end{equation}
where $x$ is the robot current position, $O_*$ is one of the endpoints of the detected obstacle and $d(x,O_*)$ is the Euclidean distance between robot position $x$ and endpoint $O_*$. After computing $H(x,G)$ for $O_s$ and $O_e$, the endpoint with minimum distance is chosen for bypassing the obstacle. M2G behavior is continued until the goal is reached or $d(x,G)$ starts to increase in which case the behavior is switched to WF. During WF, the robot follows the obstacle boundary until one of the conditions below is satisfied:
\begin{itemize}
\item
The goal is reached $\rightarrow$ the algorithm halts with success.
\item
The robot completes a full cycle around the obstacle $\rightarrow$ the algorithm halts with failure since the goal is not reachable.
\item
The equation 
\begin{equation}\label{dreach-dfollowed-eq}
d_{Reach} < d_{Followed}
\end{equation}
holds $\rightarrow$ the algorithm switches back to M2G behavior. $d_{Followed}$ is the distance between the goal $G$ and the point on the blocking obstacle boundary which is closest to $G$. Additionally, $d_{Reach}=d(v_{leave}, G)$ where $v_{leave}$ is a node on the local tangent graph such that equation \ref{dreach-dfollowed-eq} holds. 
\end{itemize}

Another method that has been proposed with practical limitations in mind is I-Bug \cite{taylor2009bug}. This method has been designed to deal with inaccurate localization and noisy sensor readings. It is assumed that the goal point emits a signal which is detectable by a special sensor onboard the robot. As the robot gets closer to the goal, the signal strength increases and as it moves away from the goal, the signal strength drops. The signal strength measurement sensor is the only accurate sensor onboard the robot. In addition to this sensor, the robot is also equipped with a sensor that determines whether the robot is facing the goal or not. I-Bug is able to reach the goal although the robot does not know its position and obstacle positions in the environment. The family of Bug algorithm has even been extended for handling moving obstacles. The resulting algorith is called distance histogram Bug (DH-Bug) \cite{zhu2012new}.
\subsubsection{Curvature velocity method}
Bug algorithms are simple and easy to implement but they do not take into account the kinematics and/or dynamic constraints of the robot. On the contrary, curvature velocity method (CVM) \cite{simmons1996curvature} considers the robot kinematics constraints and even dynamic constraints to some extent.

To perform obstacle avoidance, CVM transforms the physical constraints imposed by environment blocking obstacles as well as acceleration and speed limitations of the robot to the velocity space. Assuming that the robot only moves along circle arcs with curvature $c=\omega/v$, the velocity space consists of rotational ($\omega$) and translational velocity ($v$). CVM maps blocking obstacles from Cartesian grid to velocity space. The mapping is done based on the distance between the robot and the obstacles. After mapping, the range of $v$ and $\omega$ that must be filtered out due to the blocking obstacles can be determined. 

\begin{figure}[!tbp]
\centering
\includegraphics[width=0.35\textwidth]{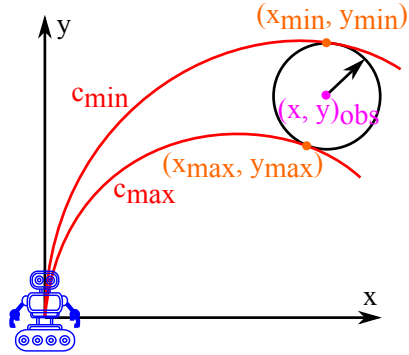}
    \caption{CVM computing valid curvatures for bypassing the blocking of the obstacle.}
    \label{cmv-fig}
\end{figure}

Naturally, obstacle avoidance methods must be real-time. To satisfy this requirement, CVM models the obstacles with a circular shape (figure \ref{cmv-fig}). Such modeling may lead to poor representation for the shape of some of the obstacles. This is the main drawback of CVM.

\subsubsection{Potential field}\label{potential-field-sec}
Potential field \cite{khatib1986real} was originally designed for control and obstacle avoidance of manipulator arms. However, due to its simplicity and effectiveness, potential field has been adopted by mobile robots community as well. Potential field models the robot as a point that is moved in the environment under influence of various attractive and repulsive potential fields. The goal is the source of attraction while the obstacles act as repulsive forces. In this setup, the total force $F(q)$ applied to the robot is the sum of repulsive ($F_{rep}$) and attractive ($F_{att}$) forces \cite{khatib1986real}:
\begin{equation}
F(q) = F_{att}(q) + F_{rep}(q).
\end{equation} 
The attractive and repulsive forces are computed by taking the gradient from the attractive and repulsive potential functions. Therefore, it is necessary to define these functions such that they are differentiable:
\begin{align}
U_{att}(q) &= \frac{1}{2} k_{att} \rho_{goal}^2 (q), \label{pf_u_att-eq} \\
U_{rep}(q) &= 
\begin{cases}
\frac{1}{2} k_{rep} \left( \frac{1}{\rho(q)} - \frac{1}{\rho_0}  \right)^2, \: if \: \rho(q) < \rho_0\\
0, \: if \rho(q) \geq \rho_0\\
\end{cases}\label{pf_u_rep-eq}
\end{align}
In equation \ref{pf_u_att-eq}, the attractive potential field is defined where $k_{att}$ is the scaling factor and $\rho_{goal}(q)$ is the Euclidean distance between point $q$ and goal $q_{goal}$. Looking at equation \ref{pf_u_att-eq}, it is clear that as the robot gets closer to the goal, the strength of the attractive field is reduced. This way when the robot reaches the goal, it is stopped.

In equation \ref{pf_u_rep-eq}, the repulsive potential field is defined where $k_{rep}$ is the scaling factor and $\rho(q)$ is the minimum distance between the robot and the obstacle the repulsive field of which is to be computed for the robot. Additionally, in equation \ref{pf_u_rep-eq}, $\rho_0$ is a threshold value. If the robot distance to the obstacle is less than $\rho_0$, it is affected by the repulsive field of the obstacle and  otherwise, the robot is not affected by the field. As $\rho(q)$ is in the denominator of equation \ref{pf_u_rep-eq}, the more the robot gets close to an obstacle, the stronger the repulsive field of the obstacle gets.
\subsubsection{Dynamic window approach}
Similar to CVM, dynamic window approach (DWA) \cite{fox1997dynamic} takes into account robot kinematics. To this end, DWA searches in velocity space which consists of all possible pairs of values of linear and angular velocities $(v, \omega)$. Another similarity between DWA and CVM is the assumption that the robot movement is only limited to circular arcs expressed as $(v, \omega)$. DWA forms a dynamic window in the velocity space such that the current linear and angular velocities of the robot will be at the center of that window. The dimensions of the window are determined based on the acceleration capabilities of the robot and the control cycle time. The infeasible velocity tuples $(v, \omega)$ within the dynamic window are pruned. The selection criterion of the tuples is that by using them, the robot will be able to stop before colliding with obstacles. The remaining tuples after the pruning are called admissible tuples. An objective function is applied on admissible tuples to determine the final movement direction for avoiding the obstacles and reaching the goal. The objective function reads as \cite{fox1997dynamic}:
\begin{equation}
O = a . heading(v, \omega) + b . velocity(v, \omega) + c . dist(v,\omega)
\end{equation}
where heading measures the progress made toward the goal, velocity encourages fast forward motion, and dist is the distance between the robot and the closes obstacle during navigation.
\subsubsection{Vector field histogram}
\begin{figure}[!t] 
\centering
\subfloat[]{\includegraphics[width=2in]{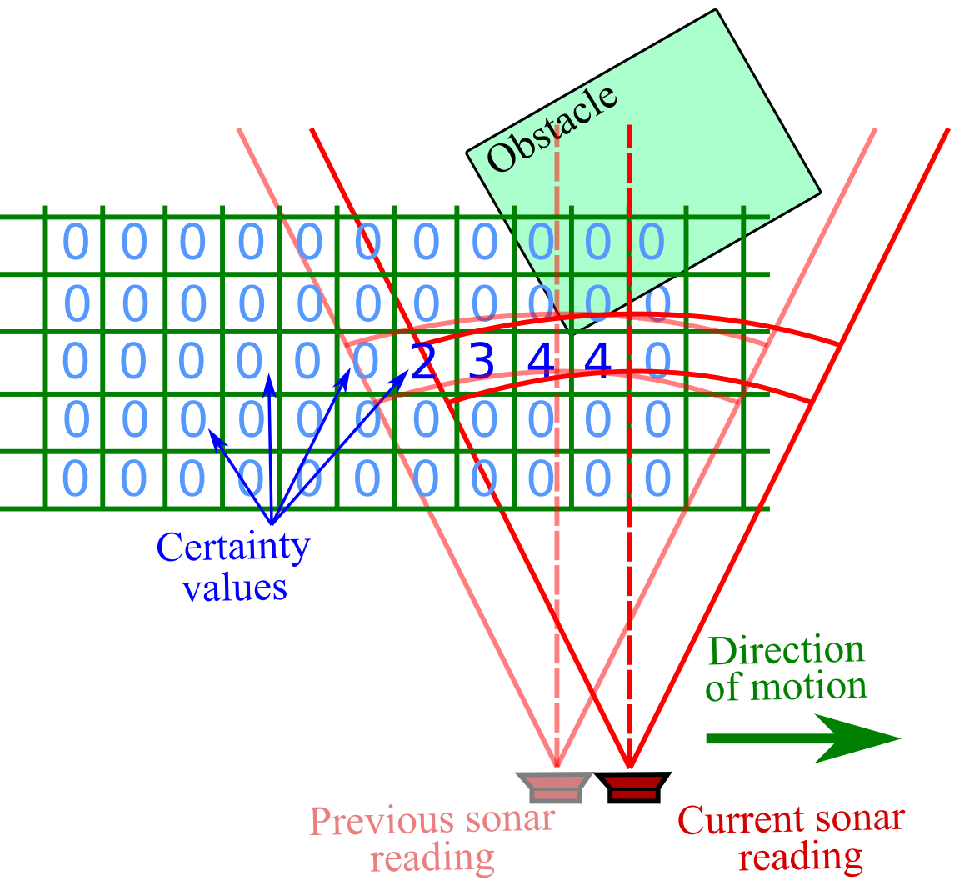}%
\label{vfh-sonar-reading-fig}}
\hfil
\subfloat[]{\includegraphics[width=3.2in]{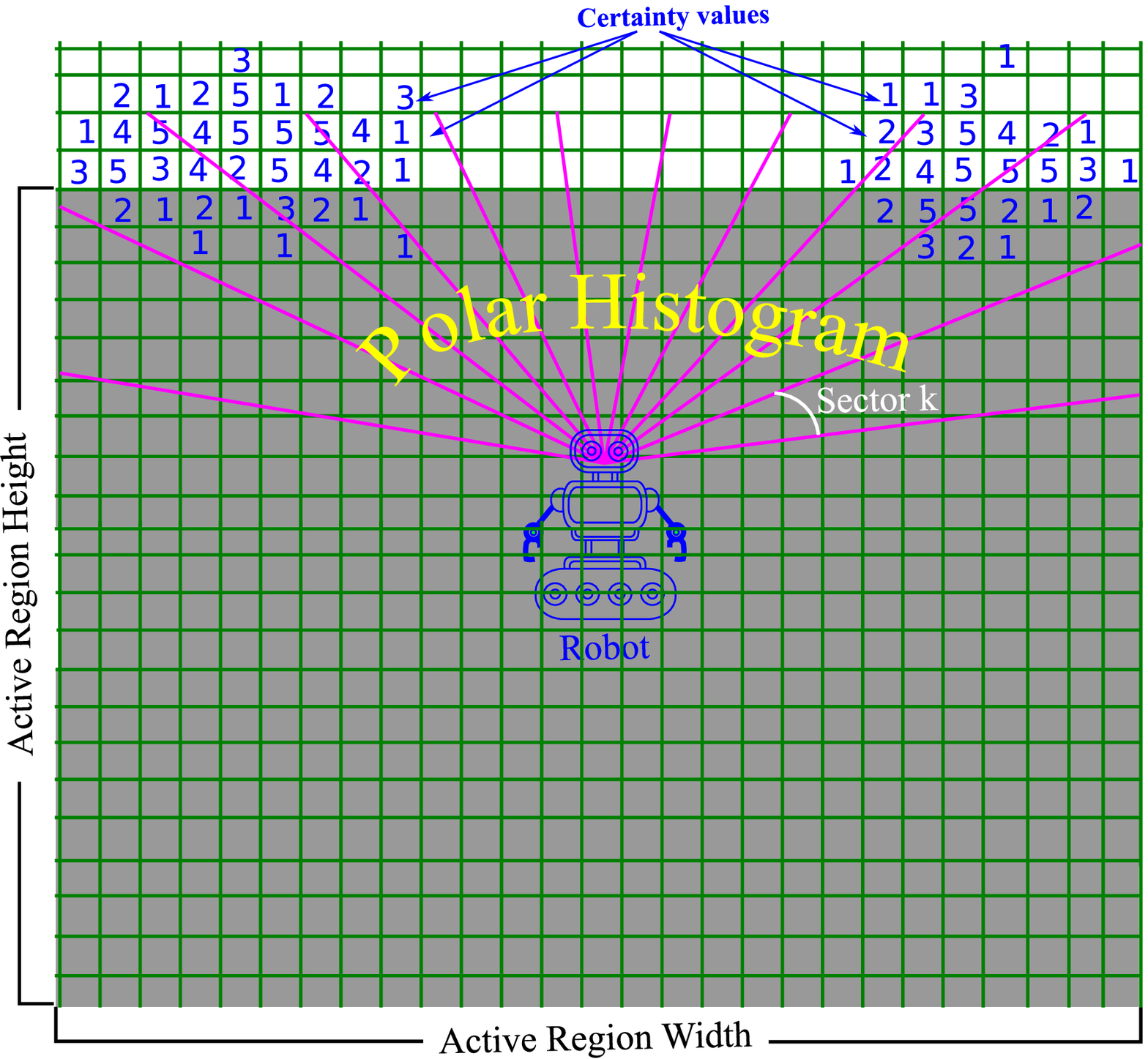}%
\label{vfh-polar-histogram-fig}}
\caption{(a) VFF certainty grid update process using successive sonar readings, (b) illustration of how certainty grid is mapped to polar histogram by VFH method.}
\label{}			
\end{figure}

Virtual force field (VFF) \cite{borenstein1989real} has been proposed to meet the obstacle avoidance requirements of relatively fast moving robots. VFF is capable of working with unknown environments having different configurations of obstacles using a ring of cheap sonar sensors around the robot chassis. VFF models the sensed obstacles as a 2D Cartesian histogram called certainty grid. The value in each cell of the grid represents the certainty (likelihood) that the cell is occupied with an obstacle. The grid is initialized with all zeros indicating that there are no cells occupied with obstacles.
 Each time a sonar reading is available, the certainty value of the appropriate histogram cell is incremented. The appropriate cell is determined based on the orientation of the sensor acoustic axis and measured range. During navigation, robot sonar readings are accumulated in the appropriate grid cells which leads to a reasonable estimation of the obstacle locations in the environment even though the sonar readings are not accurate. The process of accumulating sonar readings in the grid has been illustrated in figure \ref{vfh-sonar-reading-fig}. As can be seen, the certainty values of the cells close to the obstacle have been increased due to successive sonar readings.
 
After populating the certainty grid with reasonable values, a rectangular sub-region of the grid (called active region) is considered with the robot at its center. The obstacles within the active region are considered for obstacle avoidance using the potential field (section \ref{potential-field-sec}) method.

The main drawback of VFF is the inability to bypass certain obstacles configuration. For example, while moving down a hallway, the repulsive force exerted on the robot due to the left and right walls paralyzes VFF obstacle avoidance. The reason is the sudden reduction of the certainty grid to direction and magnitude of the repulsive force vector $F_r$ in a single step \cite{borenstein1991vector}. This issue has been addressed by vector field histogram (VFH)\cite{borenstein1991vector} method using a two-step reduction procedure. After populating the certainty grid, it is reduced to a 1D polar histogram. To this end, the active region (the gray region in figure \ref{vfh-polar-histogram-fig}) is partitioned into multiple sectors with angle $\alpha$. To compute the obstacle density for each sector $k$, for each certainty grid cell $\left(i,j \right)$, an obstacle vector with magnitude $m_{i,j}$ and direction $\beta_{i,j}$ is computed \cite{borenstein1991vector}:
\begin{align}
\beta_{i,j} = \tan^{-1} \left( \frac{y_i - y_r}{x_i - x_r} \right), \\
m_{i,j} = \left( c_{i,j}^* \right)^2 \left( a - bd_{i,j}  \right),
\end{align}
where $\left(x_i, y_i \right)$ is the coordinates of cell $\left( i,j \right)$ and $c_{i,j}$ is its certainty value, $\left(x_r, y_r \right)$ is the robot current position, $a$ and $b$ are positive constants, and $d_{i,j}$ is the distance between cell $\left(i, j \right)$ and the robot position. Based on the direction component $\beta_{i,j}$, the sector $k$ within which cell $\left(i,j \right)$ resides is decided \cite{borenstein1991vector}:
\begin{equation*}
k = \lfloor  \frac{\beta_{i,j}}{\alpha} \rfloor.
\end{equation*}
After determining the appropriate sector for all of the certainty cells, the obstacle density for each sector $k$ is computed \cite{borenstein1991vector}:
\begin{equation}\label{vfh-h-k-eq}
h_k = \sum_{i,j} m_{i,j}.
\end{equation}
Using equation \ref{vfh-h-k-eq}, the polar histogram is formed with horizontal axis showing $[0, 2\pi]$ degrees and vertical axis showing the obstacle density values for sectors at different angles. To determine the safe portions of the active region for robot movement, a threshold is applied on the polar histogram vertical axis. Sectors with obstacle density below the threshold are considered safe. VFH has been further extended to VFH+\cite{ulrich1998vfh+} to improve robot safety during navigation.

\subsection{Obstacle avoidance using RL and DL}
Over the years, computers have evolved dramatically reducing the computational burden for mobile robot platforms. This has paved the road toward using more powerful methods like DL and reinforcement learning which have higher processing power requirements. In this subsection, the state-of-the-art obstacle avoidance methods are reviewed.

To be able to avoid environmental obstacles, the distance between the robot and the obstacle is required. LiDAR sensors provide high precision range data but they are costly. Much cheaper alternatives are stereo cameras that provide depth data via disparity map computed between the two cameras. However, acquiring correct range data depends on appropriate calibration of stereo cameras and the estimated range is limited to a few meters (which is much less than LiDAR). To avoid the calibration process, one can use monocular cameras at the cost of losing depth data. However, recent advancements in DL have made it possible to estimate depth using only a single camera. To this end, Wenzel et al. \cite{wenzel2021vision} utilized generative adversarial networks (GAN) \cite{goodfellow2014generative} to convert RGB images to their corresponding depth images. This approach not only makes depth estimation from a single camera possible but also bridges the gap between simulation and real-world applications. The depth estimation is made possible thanks to the powerful feature extraction of DL methods. Bridging the gap between simulation and real-world is realized due to the fact that GAN does not care about the source of RGB images and it can estimate the depth maps for simulated as well as real images. The authors trained deep reinforcement learning agents such as deep Q networks (DQN) \cite{mnih2015human} and proximal policy optimization (PPO) \cite{schulman2017proximal} to generate appropriate control commands for input depth images. History of utilizing RL for obstacle avoidance precedes the birth of the infamous DQN approach. As an example, instead of relying on a lookup table, Huang et al. \cite{huang2005reinforcement} used a NN to estimate the Q function of a Q-learning agent.

In addition to RL, DL-based obstacle avoidance modules can be trained in supervised manner as well. Liu et al. \cite{liu2017cnn} implemented indoor obstacle avoidance by feeding raw images to a DL model based on AlexNet\cite{krizhevsky2017imagenet} and using the model output as steering commands. Based on the input image, the robot is steered left, right and commanded to go straight. The training of the DL model has been done using data collected during robot motion controlled by a human operator. To facilitate the implementation, ROS has been used.
\section{Navigation methods}\label{nav-methods-sec}
Autonomous navigation methods are either map-based or mapless. In map-based approaches, the environment map is known before the navigation starts but it is not the case for mapless methods. Obviously having the environment map facilitates efficient path planning. However, in many scenarios, the environment map is not known beforehand which has led to the development of mapless navigation and simultaneous localization and mapping (SLAM) reviewed in section \ref{slam-sec}.
\subsection{Map-based navigation}
\begin{figure}[!t] 
\centering
\subfloat[]{\includegraphics[width=0.3\textwidth]{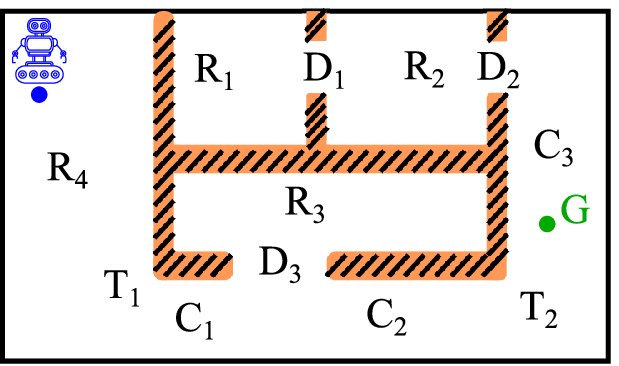}%
\label{env-c,r,d-fig}}
\hfill
\subfloat[]{\includegraphics[width=0.3\textwidth]{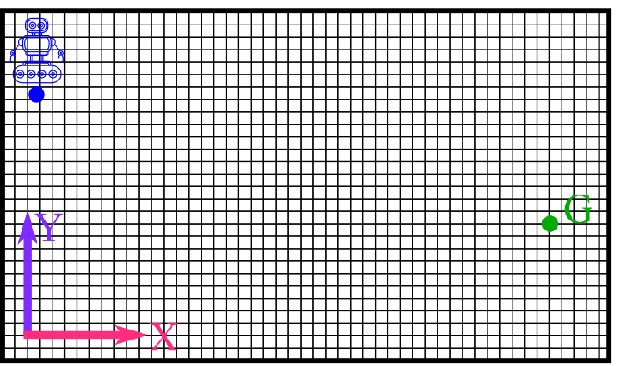}%
\label{metric-map-fig}}
\hfill
\subfloat[]{\includegraphics[width=0.3\textwidth]{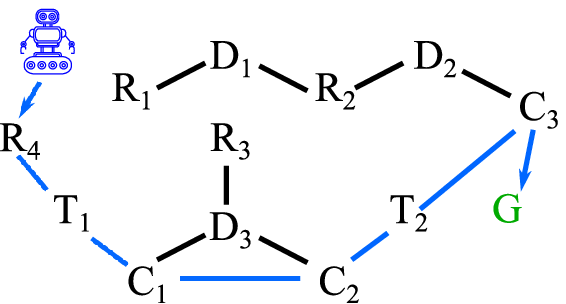}%
\label{topological-map-fig}}
\caption{Types of map representation: (a) actual environment, (b) metric map, (c) topological map.}
\label{}			
\end{figure}

The quality of map-based navigation is directly affected by how the environment map is represented and how accurate it is. Map representations are either metric or topological. In metric maps, the location of detected obstacles, landmarks, robot, etc. are expressed with respect to a certain reference frame  \cite{meyer2003map}. A typical metric map corresponding to environment in figure \ref{env-c,r,d-fig} has been illustrated in figure \ref{metric-map-fig} in the form of a 2D grid. As can be seen, in the grid map, the detected obstacles, robot position, etc. are represented with respect to some specific $XY$ reference frame. Each cell of this grid represents the probability of being occupied with an obstacle. Grid cells with darker black are more likely to be occupied by obstacles. Probabilistic occupancy grids can be implemented using multiple approaches. Bayesian occupancy grid (BOG) \cite{moravec1985high} relies on Baye's rule to estimate the probability of being occupied for each cell of the grid map. BOG has been used in some of the mapping approaches \cite{elfes1989using, hornung2013efficient}. Other methods to implement occupancy grid are based on reflection map \cite{thrun2002probabilistic} and expectation maximization \cite{thrun2001learning}.

Naturally, when range sensors are available, expressing the environment map as a 2D or 3D grid is useful \cite{triebel2006multi, hahnel2003efficient, grisetti2007improved}. However, the memory requirement for storing map of large scale environments may be prohibitive\cite{fisher2021colmap} since the memory size needed to store the map increases as range data are obtained and processed \cite{fisher2021colmap}. Several attempts have been made to keep the memory requirement of grid maps manageable \cite{hornung2013efficient, khan2014rmap, einhorn2010building}.

Apart from range sensors, camera sensors provide rich visual features as well. These features can be used to implement landmark maps \cite{smith1990estimating, montemerlo2002fastslam} which are effective for scenarios that locally distinguishable features are present in the environment. As an example, consider the environment in figure \ref{env-c,r,d-fig} again and assume that there are distinguishable features in different regions of this environment. Using these features, it is possible to generate a topological map similar to the one in figure \ref{topological-map-fig}. Each node of the topological map represents a particular region in the environment based on the aforementioned distinguishable visual features. The regions that are accessible from one another are connected by edges in the topological map. For example $R_4$ (room 4) is accessible via $T_1$ (turn 1) which is shown in the topological map (\ref{topological-map-fig}) with an edge connecting $R_4$ to $T_1$. Treating the topological map as an undirected graph, off-the-shelf graph path planning algorithms can be used to find the shortest path between robot position and the goal position (denoted by $G$ in figure \ref{topological-map-fig}). In real-world applications, it is likely that the environment map is only partially known. To carry out autonomous navigation under such circumstances, Sgorbissa and Renato proposed a hybrid navigation approach to exploit a possibly incomplete map of the environemnt and local perceptions during robot motion in order to achieve safe and collision-free navigation \cite{sgorbissa2012planning}.
\subsection{Mapless}
Autonomous mapless navigation is the act of controlling a mobile robot toward the desired destination without having access to the environment map. The problem of mapless navigation has been tackled several times. As pointed out in  \cite{guzel2013autonomous}, mapless vision-based techniques for indoor robot navigation can be categorized as:
\begin{itemize}
\item
methods based on optical flow: in a nutshell, the motion of visual features in an image sequence is defined as optical flow. One of the mapless navigation methods based on optical flow is robee \cite{bernardino1998visual}. In this method, vision system of a bee is mimicked using stereo cameras to navigate through the environment without colliding with obstacles. Optical flow has also been used to implement corridor following behavior \cite{duchon1998ecological} as well as a feature-based navigation system \cite{temizer2001optical}.
\item
appearance-based matching: in this approach, certain templates and images of the environment are memorized. The memorized data are then associated with controlling commands to guide the robot toward desired destination \cite{desouza2002vision}. The main challenge of appearance-based methods is finding appropriate methods to express environment in terms of prominent features as well as defining matching criteria for recognizing certain regions of the environment \cite{bonin2008visual}. Visual memory \cite{gaussier1997visual} is one of the appearance-based methods in which a sequence of images are captured during robot motion. The captured images are associated with appropriate motions required for reaching a specific destination. The images, their associated motions and a template matching algorithm can be used to construct the robot route.
Another appearance-based approach is multidimensional histograms that are extracted based on statistical analysis (related to color, texture, and edge density) of captured images \cite{schiele1996object,1241767}. The recognition of places is done via matching histograms of stored places and histograms computed for the current position of the robot.
\item
object recognition: appearance-based approaches are primarily focused on memorizing their surrounding environment. However, symbolic navigation can be used instead \cite{kim1998recognition, kim1999symbolic}. In this approach, the detected obstacle are stored in a 2D grid map called the S-map. Using this map, path planning is done and robot is given symbolic commands such as "go to the main exit" or "go to the corner in front of you" \cite{guzel2013autonomous}
\item
feature-based navigation methods: in these methods, the motion of the robot is determined by extracting features from image sequence and tracking the relative displacement of those features during robot movement. An example of feature-based navigation methods is funnel lane \cite{chen2009qualitative} in which a robot is tasked to follow a previously traversed path by matching image feed from its mono camera with the previously recorded image sequence called visual path. Funnel lane method is fully qualitative and does not need environment map of any kind. This method has been extended to sloped funnel lanes \cite{kassir2020qualitative} to address issues such as ambiguity between translation and rotation in certain scenarios and achieve better maneuverability.
\end{itemize}

RL has also been utilized to implement map-less autonomous navigation in highly dynamic environments where quick and adaptive action selection is required. As an example, Fan et al. \cite{fan2018crowdmove} optimized a mapless navigation policy capable of robot control in highly dynamic environments such as pedestrian crowds. The optimization of the policy as a local planner was done using PPO and the policy was expressed as a CNN with four hidden layers.
\section{SLAM}\label{slam-sec}
Simultaneous Localization And Mapping (SLAM) can be thought as a concept rather than a specific algorithm \cite{riisgaard2003slam}. The term was coined after the original work of Hugh Durrant-Whyte and John J. Leonard \cite{leonard1991mobile} whose research was founded based on earlier work by Smith, Self and Cheeseman \cite{smith1990estimating}. In a nutshell, SLAM objective is to localize a mobile robot in an unknown environment and at the same time estimate the environment map in terms of observed environment landmarks. Roughly speaking, the main components of SLAM are landmark extraction, data association, state estimation, and state and landmark update.

\begin{figure}[!t] 
\centering
\subfloat[]{\includegraphics[width=1.3in]{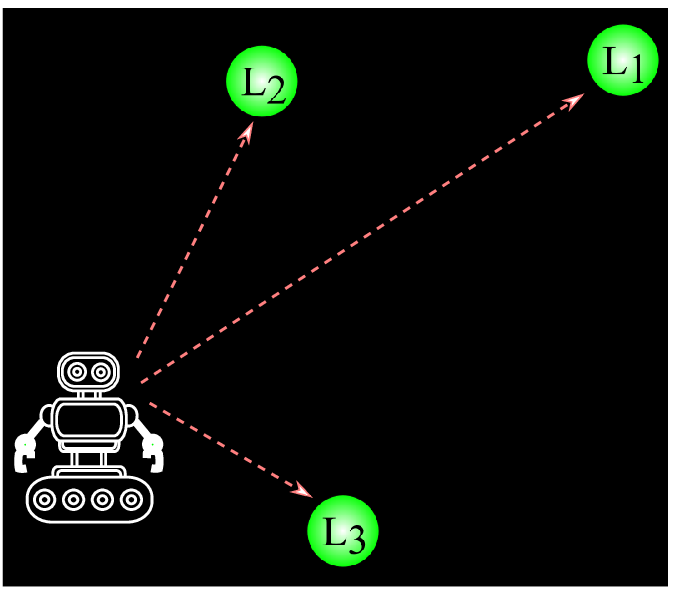}%
\label{slam-1st-measuremnt-fig}}
\subfloat[]{\includegraphics[width=1.3in]{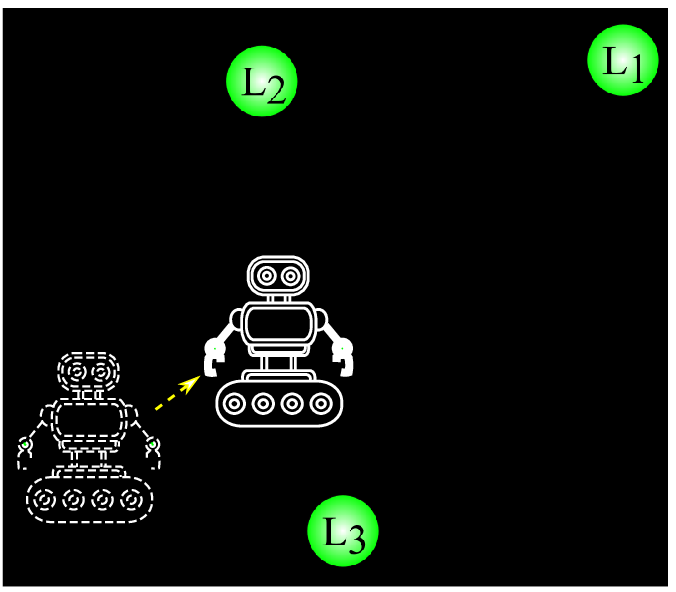}%
\label{slam-odometry-update-fig}}
\subfloat[]{\includegraphics[width=1.3in]{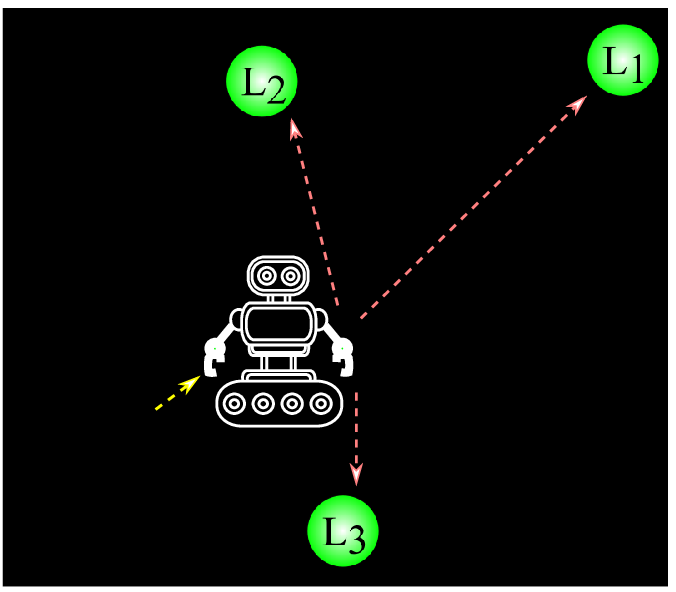}%
\label{slam-2nd-measurement-fig}}
\subfloat[]{\includegraphics[width=1.3in]{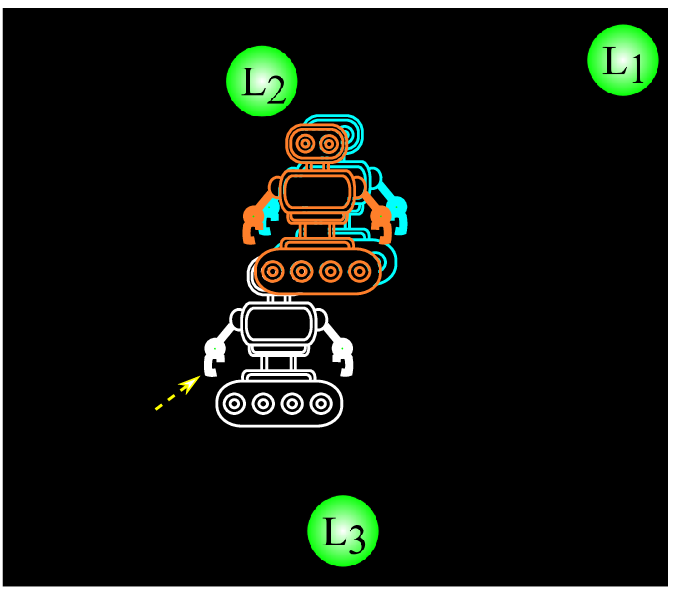}%
\label{slam-updated-vs-true-pose-fig}}
\caption{Illustration of mobile robot localization using SLAM and detected landmarks: (a) distance measurement to three landmarks $\{L_1, L_2, L_3\}$, (b) robot pose prediction after movement (dotted robot: old pose, solid robot: new predicted pose), (c) sensing landmarks again, (d) pose correction based on measurements in part c (white robot: predicted pose before correction, orange robot: pose after correction, cyan robot: true pose).}
\label{slam-localization-fig}			
\end{figure}

The high level steps of robot pose prediction and correction during SLAM have been depicted in figure \ref{slam-localization-fig}. As shown in figure \ref{slam-1st-measuremnt-fig}, the robot measures its distance to three landmarks ($L_1,L_2, L_3$) and then moves to a new location. Using robot odometry model, the new location is predicted which has been shown as a robot figure with solid lines in figure \ref{slam-odometry-update-fig}. Once again, the distance to the three landmarks is checked (figure \ref{slam-2nd-measurement-fig}). Based on the obtained measurements, the robot pose is corrected leading to orange robot in figure \ref{slam-updated-vs-true-pose-fig}. As can be seen, the corrected pose is much closer to the ground truth pose (cyan robot in figure \ref{slam-updated-vs-true-pose-fig}). Sensors like IMU provide valuable information about the robot motion and play vital role in the prediction step of the SLAM \cite{karam2019integrating}. On the other hand, measured range data or visual features obtained from cameras are used in the update step of the SLAM in order to reduce the prediction error \cite{mohamed2019survey}.

SLAM solutions can be categorized into two broad classes of filtering (online SLAM) and smoothing (full SLAM) methods \cite{alsadik2021simultaneous} which are briefly introduced below.

Smoothing approaches rely on least-square optimization to estimate the robot full trajectory using all of the sensor measurements collected so far. Therefore, the memory consumption is high however, it is still possible to implement full SLAM in real-time \cite{strasdat2010scale}. Estimatating the robot full trajectory $\hat{x}_{1:K} = \{\hat{x}_k, k=1, ..., K\}$ (with length $K$) given the sequence of control data $u_{1:K} = \{u_k, k = 1, ..., K\}$ obtained from odometry sensors and observations $z_{1:K} = \{z_k, k =1, ..., K\}$ obtained from the environment is what separates full SLAM from online SLAM \cite{korkmaz2016comparison}. In full SLAM, the following posterior probability must be estimated \cite{grisetti2010tutorial}:
\begin{equation}\label{graph-slam-posterior-eq}
p(\hat{x}_{1:K} | u_{1:K}, z_{1:K}, x_0)
\end{equation}
where $x_0$ is the initial state of the robot at the start of the navigation. To estimate the posterior probability in equation (\ref{graph-slam-posterior-eq}), one has to deal with intractable high dimensional state spaces unless certain conditions such as static world and Markov property are assumed. Taking these two assumptions into account, Graph SLAM was proposed by Lu and Millios \cite{lu1997globally} which uses a graph-based representation to capture the underlying structure of the SLAM problem. Each graph node consists of a robot pose
The graph edges represent spatial constraints between nodes which are based on the collected observations $z_k$ or odometry data $u_k$. 

\begin{figure}[!t] 
\centering
\subfloat[]{\includegraphics[width=0.5\textwidth]{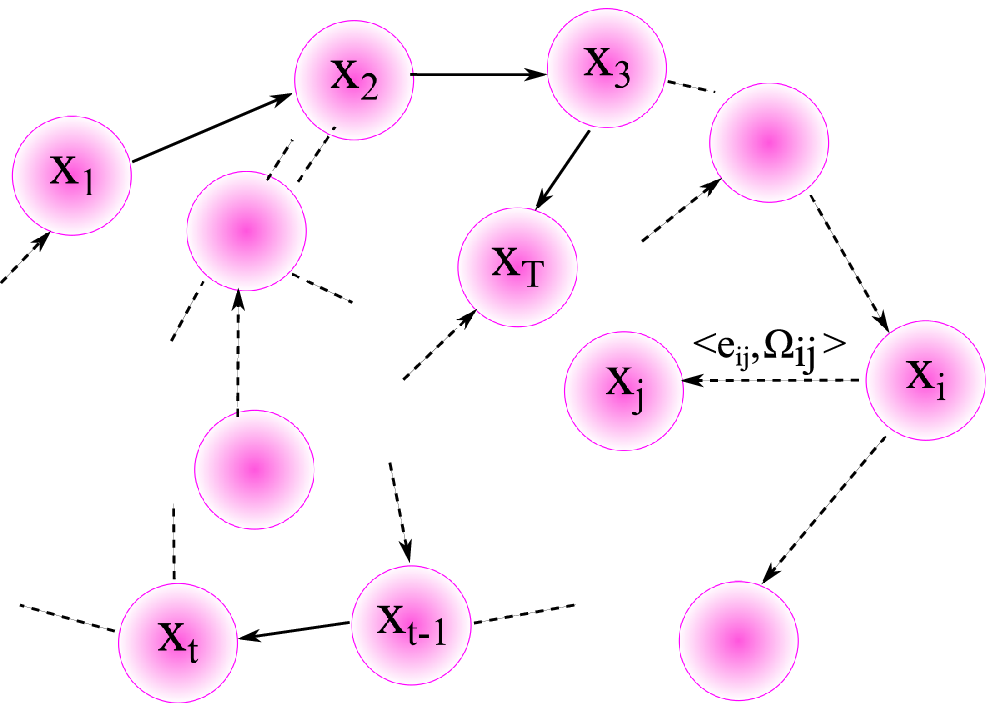}%
\label{graph-slam-pose-graph-representation-fig}}
\hfil
\subfloat[]{\includegraphics[width=0.3\textwidth]{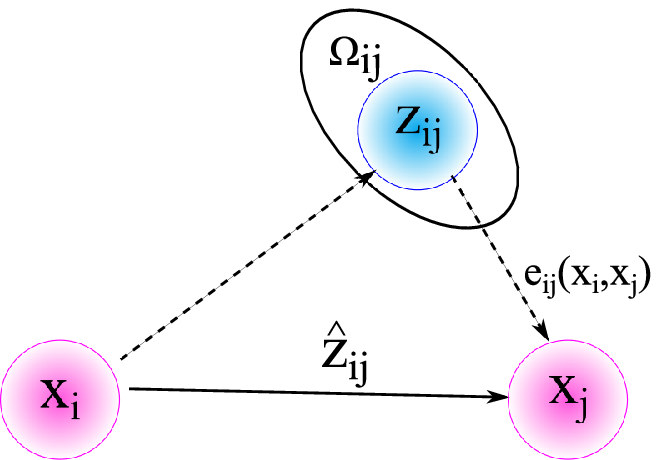}%
\label{graph-slam-edge-fig}}
\caption{(a) a typical pose-graph representation for Graph SLAM: nodes represent robot pose and edges express spatial constraint between nodes due to obtained measurements, (b) nodes $x_i$ and $x_j$ connection due to mutual observation $z_{ij}$.}
\label{graph-slam-pose-graph-fig}			
\end{figure}

Graph SLAM decomposes the problem into two parts: (a) graph construction (known as front-end) based on raw measurements and (b) finding the most likely configuration of the robot poses (known as back-end) such that the set of constraints imposed by the graph edges is satisfied \cite{grisetti2010tutorial}. A typical pose-graph has been depicted in figure \ref{graph-slam-pose-graph-representation-fig} in which $\Omega_{ij}$ is the information matrix and $e_{ij}=z_{ij} - \hat{z}(x_i, x_j)$ (the error function) is the difference between actual observation $z_{ij}$ gathered by the robot and the predicted observation $\hat{z}_{ij}(x_i, x_j)$ which represents $x_j$ in the coordinate frame of $x_i$. For more clarity, the connection between the two poses $x_i$ and $x_j$ due to their mutual observation $z_{ij}$ has been depicted in figure \ref{graph-slam-edge-fig}. As can be seen, the uncertainty of observing $z_{ij}$ is captured by $\Omega_{ij}$.

In SLAM, 
data association is the matching of measurements obtained from the environment with the existing map which is vital for successful SLAM implementation \cite{zhou2016review}. Data association is usually handled by the SLAM front-end. To achieve a consistent estimate of the prior over robot trajectory $p(\hat{x}_{1:K} | z_{1:K}, u_{1:K})$, interleaving of the front-end and the back-end is crucial. The data association problem has been tackled using different approaches based on joint compatibility branch and bound \cite{neira2001data}, spectral clustering \cite{teller2006fast}, and backtracking \cite{hahnel2005towards}.

The second category of SLAM methods is online approaches which have lower memory consumption compared to full SLAM methods. This stems from the fact that online methods only rely on the previous pose estimate and sensor readings. However, lower memory consumption comes at the cost of decreased accuracy compared to the full SLAM approaches.

Apart from being online or full, SLAM methods can be also categorized based on the sensor(s) they use for collecting observations from the environment. Methods that use range sensors as their source of observation and methods that use camera sensors are called LiDAR SLAM and visual SLAM, respectively. 
Sections \ref{lidar-slam-sec} and \ref{visual-slam-sec} are devoted to reviewing some of well-known methods belonging to LiDAR and visual SLAM and they are summarized in table \ref{slam-summary-table}.

\subsection{LiDAR SLAM}\label{lidar-slam-sec}
 Indoor environments are heavily structured and having obstacles within 60 meters from the robot is quite likely. Such conditions are ideal for using laser scanners to sense and map the environment. SLAM approaches based on LiDAR sensors are known as LiDAR SLAM. Despite the ability to provide high precision data, LiDAR sensors are usually accompanied by hefty price tags making LiDAR SLAM an expensive solution.

The SLAM literature consists of various LiDAR-based SLAM approaches such as Hector slam \cite{kohlbrecher2013hector}, keyframe-based 3D LiDAR SLAM \cite{guo2022lidar}, Google Cartographer \cite{hess2016real}, GMapping \cite{grisetti2007improved}, and TinySLAM \cite{steux2010tinyslam}. Traditionally, LiDAR sensors used rotary mechanisms to scan the surrounding environments. However, recent developments have led to low power and low cost LiDAR sensors based on Micro-Electro-Mechanical Systems MEMS technology. MEMS LiDARs are available at much more affordable prices which makes them ideal choices for the robotic researchers. However, lower cost of MEMS LiDAR comes at the price of shorter range (about nine meters) and smaller field of view (FOV) which makes LiDAR SLAM challenging. Fusing MEMS LiDAR data with other data sources such as IMU can improve the SLAM performance \cite{vonk2022graph}. It has been shown that utilizing fast point cloud registration can further improve the localization and mapping quality \cite{wang2022high}. While using LiDAR sensors provide range data with outstanding accuracy, their relatively costly price tag is the motivation for using cheaper alternatives such as ultrasonic sensors \cite{cotugno2013extended, d2013polynomial, jung2009simultaneous, yang2012efficient}.
\subsection{Visual SLAM}\label{visual-slam-sec}
Another type of SLAM methods is visual SLAM which is primarily focused on mono/stereo camera(s). Visual SLAM must not be confused with visual odometry (VO) \cite{yousif2015overview}. Visual SLAM is all about preparing the global map of the explored environment whereas VO is used to estimate the robot pose locally.

One natural approach to implement visual SLAM is extracting features from camera images and use them to determine the scene geometry and camera pose. Feature-based methods are either based on filtering like the EKF-based visual-inertial odometry \cite{li2013high} or based on keyframes \cite{4538852}. At first glance, feature-based approaches seem very intuitive. However, these methods are limited to the information that are captured by the type of the features extracted. As pointed out in \cite{engel2014lsd}, feature-based methods will have difficulty with man-made environments. Some edge-based \cite{klein2008improving, eade2009edge} and region-based      \cite{concha2014using} have been proposed to tackle the aforementioned limitation. Among feature-based methods, Parallel Tracking And Mapping (PTAM) \cite{klein2007parallel, klein2008improving, klein2009parallel} computational complexity is low enough to be runnable on a mobile phone using its camera as the source of visual data. PTAM uses FAST-10 corner detector \cite{rosten2006machine} for feature extraction. ORB-SLAM \cite{mur2015orb} is based on ORB features \cite{rublee2011orb} capable of handling small/large and indoor/outdoor environments. The use of ORB features has made ORB-SLAM runnable in real-time without requiring a graphical processing unit (GPU). ORB-SLAM has received two important extensions resulting in second \cite{mur2017orb} and third \cite{campos2021orb} versions of this method. The second version provides support for stereo and RGBD cameras. The first version was limited to pin-hole camera model but the third version (based on first and second versions and ORB-SLAM-VI \cite{mur2017visual}) works with any camera type assuming that the camera module is provided. OV$^2$SLAM \cite{ferrera2021ov} also offers a feature-based approach which is accurate, robust, and real-time outperforming ORB-SLAM.

It is also possible to take a direct (feature-less) approach in which all of the information in the image intensities are used. Using the direct approach, loss of prominent information due to limited representation power of features is avoided altogether. One of the direct methods is Large-Scale Direct Monocular SLAM (LSD-SLAM) \cite{engel2014lsd} capable of building large-scale and consistent environment 3D maps in real-time. Direct Sparse Mapping (DSM) \cite{zubizarreta2020direct} is based on photometric bundle adjustment (PBA) that introduces the idea of map reuse in direct methods. DSM showed the importance of mid-term data association.

While majority of SLAM methods are either visual or LiDAR-based, Multi-cue Direct SLAM (MD-SLAM) \cite{di2022md} is the only open source approach that supports LiDAR and RGBD sensors in a unified manner.

\begin{table}[]
\caption{Summary of some of well known SLAM methods.}\label{slam-summary-table}
\begin{tabular}{|c|ccccc|cc|c|}
\hline
\multirow{3}{*}{Method}                                                     & \multicolumn{5}{c|}{Sensor}                                                                                                                                & \multicolumn{2}{c|}{SLAM approach}                                   & \multirow{3}{*}{\begin{tabular}[c]{@{}c@{}}Loop closure\\ detection\end{tabular}} \\ \cline{2-8}
                                                                            & \multicolumn{1}{c|}{\multirow{2}{*}{Radar}} & \multicolumn{1}{c|}{\multirow{2}{*}{LiDAR}} & \multicolumn{3}{c|}{Visual}                                    & \multicolumn{1}{c|}{\multirow{2}{*}{Online}} & \multirow{2}{*}{Full} &                                                                                   \\ \cline{4-6}
                                                                            & \multicolumn{1}{c|}{}                       & \multicolumn{1}{c|}{}                       & \multicolumn{1}{c|}{Mono} & \multicolumn{1}{c|}{Stereo} & RGBD & \multicolumn{1}{c|}{}                        &                       &                                                                                   \\ \hline
                                                                            
Hector SLAM \cite{kohlbrecher2013hector}                                                                 & \multicolumn{1}{c|}{}                       & \multicolumn{1}{c|}{\checkmark}                      & \multicolumn{1}{c|}{}     & \multicolumn{1}{c|}{}       &      & \multicolumn{1}{c|}{\checkmark}                       &                       &                                                                                   \\ \hline

LSD-SLAM \cite{engel2014lsd, engel2015large}                                                                    & \multicolumn{1}{c|}{}                       & \multicolumn{1}{c|}{}                       & \multicolumn{1}{c|}{\checkmark}    & \multicolumn{1}{c|}{}       &      & \multicolumn{1}{c|}{}                        & \checkmark                     & \checkmark                                                                                 \\ \hline

ORB-SLAM \cite{mur2015orb}                                                                    & \multicolumn{1}{c|}{}                       & \multicolumn{1}{c|}{}                       & \multicolumn{1}{c|}{\checkmark}    & \multicolumn{1}{c|}{}      &     & \multicolumn{1}{c|}{}                        & \checkmark                     & \checkmark                                                                                 \\ \hline

ORB-SLAM2 \cite{mur2017orb}                                                                    & \multicolumn{1}{c|}{}                       & \multicolumn{1}{c|}{}                       & \multicolumn{1}{c|}{\checkmark}    & \multicolumn{1}{c|}{\checkmark}      & \checkmark    & \multicolumn{1}{c|}{}                        & \checkmark                     & \checkmark                                                                                 \\ \hline

ORB-SLAM3 \cite{campos2021orb}                                                                    & \multicolumn{1}{c|}{}                       & \multicolumn{1}{c|}{}                       & \multicolumn{1}{c|}{\checkmark}    & \multicolumn{1}{c|}{\checkmark}      & \checkmark    & \multicolumn{1}{c|}{}                        & \checkmark                     & \checkmark                                                                                 \\ \hline

ORB-SLAM-VI \cite{mur2017visual}                                                                    & \multicolumn{1}{c|}{}                       & \multicolumn{1}{c|}{}                       & \multicolumn{1}{c|}{\checkmark}    & \multicolumn{1}{c|}{\checkmark}      & \checkmark    & \multicolumn{1}{c|}{}                        & \checkmark                     & \checkmark                                                                                 \\ \hline

Mono-SLAM \cite{davison2003real, davison2004real, molton2004locally, davison2007monoslam}                                                                    & \multicolumn{1}{c|}{}                       & \multicolumn{1}{c|}{}                       & \multicolumn{1}{c|}{\checkmark}    & \multicolumn{1}{c|}{}      &     & \multicolumn{1}{c|}{\checkmark}                        &                      &                                                                                  \\ \hline

PTAM \cite{klein2007parallel, klein2008improving, klein2009parallel}                                                                    & \multicolumn{1}{c|}{}                       & \multicolumn{1}{c|}{}                       & \multicolumn{1}{c|}{\checkmark}    & \multicolumn{1}{c|}{}      &     & \multicolumn{1}{c|}{}                        &        \checkmark              &                                                                                  \\ \hline

DSM \cite{zubizarreta2020direct}                                                                    & \multicolumn{1}{c|}{}                       & \multicolumn{1}{c|}{}                       & \multicolumn{1}{c|}{\checkmark}    & \multicolumn{1}{c|}{}      &     & \multicolumn{1}{c|}{}                        &        \checkmark              &                                                                                  \\ \hline

OV$^2$SLAM \cite{ferrera2021ov}                                                                    & \multicolumn{1}{c|}{}                       & \multicolumn{1}{c|}{}                       & \multicolumn{1}{c|}{\checkmark}    & \multicolumn{1}{c|}
{\checkmark}      &      & \multicolumn{1}{c|}{}                        & \checkmark                     & \checkmark                                                                                 \\ \hline

RT-SLAM \cite{roussillon2011rt}                                                                     & \multicolumn{1}{c|}{}                       & \multicolumn{1}{c|}{}                       & \multicolumn{1}{c|}{\checkmark}    & \multicolumn{1}{c|}{\checkmark}      &      & \multicolumn{1}{c|}{\checkmark}                       &                       & \checkmark                                                                                 \\ \hline

\begin{tabular}[c]{@{}c@{}}Google Cartographer \cite{hess2016real}\end{tabular} & \multicolumn{1}{c|}{}                       & \multicolumn{1}{c|}{\checkmark}                      & \multicolumn{1}{c|}{}     & \multicolumn{1}{c|}{}       &      & \multicolumn{1}{c|}{}                        & \checkmark                     & \checkmark                                                                                 \\ \hline

\begin{tabular}[c]{@{}c@{}}GMapping \cite{grisetti2007improved}\end{tabular}     & \multicolumn{1}{c|}{}                       & \multicolumn{1}{c|}{\checkmark}                      & \multicolumn{1}{c|}{}     & \multicolumn{1}{c|}{}       &      & \multicolumn{1}{c|}{\checkmark}                       &                       & \checkmark                                                                                 \\ \hline

TinySLAM \cite{steux2010tinyslam}                                                                   & \multicolumn{1}{c|}{}                       & \multicolumn{1}{c|}{\checkmark}                      & \multicolumn{1}{c|}{}     & \multicolumn{1}{c|}{}       &      & \multicolumn{1}{c|}{\checkmark}                       &                       & \checkmark                                                                                 \\ \hline

\begin{tabular}[c]{@{}c@{}}RadarSLAM \cite{hong2022radarslam}\end{tabular}       & \multicolumn{1}{c|}{\checkmark}                      & \multicolumn{1}{c|}{}                       & \multicolumn{1}{c|}{}     & \multicolumn{1}{c|}{}       &      & \multicolumn{1}{c|}{}                        & \checkmark                     & \checkmark                                                                                 \\ \hline
\begin{tabular}[c]{@{}c@{}}MD-SLAM \cite{di2022md}\end{tabular}       & \multicolumn{1}{c|}{}                      & \multicolumn{1}{c|}{\checkmark}                       & \multicolumn{1}{c|}{}     & \multicolumn{1}{c|}{}       &    \checkmark  & \multicolumn{1}{c|}{}                        &                     \checkmark &                                    \checkmark                                             \\ \hline

\end{tabular}
\end{table}
\subsection{Loop closure techniques}
An efficient SLAM module must be capable of recognizing previously visited places in order to reduce the localization and mapping error that has been accumulated during multiple time steps. Such capability is called loop closure detection. Traditionally, feature descriptors such as SIFT \cite{LoweDavidG} and SURF \cite{Herbert} were used to loop closure detection. However, these methods are computationally inefficient for real-time applications. Researchers have developed loop closure mechanisms \cite{stumm2016robust, cummins2008fab, garcia2018ibow} based on bag-of-words (BoWs) approach to remedy the high computational load of descriptors such as SIFT and SURF. Bag-of-words refers to a dictionary of visual features extracted from the environment and clustered to form the so-called words. Despite being fast, BoWs rely on fixed visual features that must be known beforehand. It comes as no surprise that BoWs will not perform well in unexplored environments. To make BoWs adaptable to new regions (not explored before), it may be created in an online step \cite{garcia2018ibow}. However, the memory requirements of the online approach are high. Following the significant achievements of DL in various domains, it has been adopted by the robotics community \cite{naseer2015robust, zhang2017loop, shin2018loop, qin2018loop}. A notable example of reinforcing SLAM with DL is DXSLAM \cite{li2020dxslam,dxslam-repo} which uses CNN to extract salient features for boosting SLAM performance.
\subsection{Public SLAM datasets}
Conducting research on SLAM methods requires access to multiple sensors and mobile robot platforms. Apart from the cost of preparing the aforementioned hardware, it may be difficult for individual researchers to put together a SLAM-capable hardware setup. An alternative approach is development and evaluation of SLAM methods on public datasets obtained from real robots. Yin and Berger \cite{yin2017use} have reviewed 27 publicly available datasets that have been collected on public roads. Some of these datasets are suited for evaluation of SLAM approaches. The summary of these datasets is available in table \ref{slam-dataset-table}. As can be seen, existing datasets cover variety of environments being indoor, and outdoor, having different weather and lighting conditions, etc. Recently, semantic SLAM has gained popularity which is the inspiration for preparing synthetic datasets suitable for semantic SLAM. SSS-dataset (ShanghaiTech Semantic SLAM) \cite{cao2020representations} is an example of such datasets which provides accurate ground truth for the scene composition and individual object shapes and poses.
\begin{table}[]
\caption{Summary of public SLAM datasets.}\label{slam-dataset-table}
\begin{tabular}{|c|c|c|c|c|c|c|c|}
\hline
Dataset                                                                                   & \rotatebox{90}{Indoor} & \rotatebox{90}{Urban} & \rotatebox{90}{Suburb} & \rotatebox{90}{Rural} & Attributes                                                                                                                                                                                                                                                                    & \rotatebox{90}{\begin{tabular}[c]{@{}c@{}}Loop\\ closure\end{tabular}} & Sensor data                                                                                                                                                                                           \\ \hline
\begin{tabular}[c]{@{}c@{}}AMUSE\\ \cite{amuse2013gt}\end{tabular}                                                                                     &        & \checkmark     &        &       & \begin{tabular}[c]{@{}c@{}}Weather: \{snowy, water \&\\ snow on lens\}\end{tabular}                                                                                                                                                                                            & \checkmark                                                      & \begin{tabular}[c]{@{}c@{}}Omni-directional multi-camera,\\ height sensor, IMU, velocity\\ sensor, GPS\end{tabular}                                                                                   \\ \hline
\begin{tabular}[c]{@{}c@{}}CCSAD\\\cite{guzman2015towards}\end{tabular}                                                                                     &        & \checkmark     &        &       & \begin{tabular}[c]{@{}c@{}}Recorded in developing countries,\\ variable lighting, tunnel at night,\\ narrow roads\end{tabular}                                                                                                                                                &                                                        & Stereo camera, IMU, GPS                                                                                                                                                                               \\ \hline
\begin{tabular}[c]{@{}c@{}}Cheddar\\Gorge \cite{simpson2011cheddar}\end{tabular}                                   &        &       &        & \checkmark     & \begin{tabular}[c]{@{}c@{}}Weather: \{dry, sunny,\\clear, cold\}\end{tabular}                                                                                                                                                                                                                                          & \checkmark                                                      & \begin{tabular}[c]{@{}c@{}}Stereo/mono camera,\\ mono IR camera,\\ Velodyne 64 LiDAR,\\ high-end GPS/IMU,\\low-end IMU,\\wheel encoders, laser\\tracker for sensor\\pose measurement\end{tabular} \\ \hline
\begin{tabular}[c]{@{}c@{}}CMU Visual\\Localization\\Dataset \cite{badino2011visual}\end{tabular}               &        & \checkmark     & \checkmark      &       & \begin{tabular}[c]{@{}c@{}}Versatile weather \&\\ light conditions\end{tabular}                                                                                                                                                                                               & \checkmark                                                      & \begin{tabular}[c]{@{}c@{}}2 Mono cameras,\\ 4 Sick LiDARs,\\ GPS, IMU\end{tabular}                                                                                                                   \\ \hline
ETH3D \cite{schops2019bad}                                                                                     & \checkmark      & \checkmark     &        &       & \begin{tabular}[c]{@{}c@{}}Suitable for evaluating visual inertial\\ mono, stereo, and RGB-D-SLAM \cite{eth3d_dataset_},\\accompanying source code \cite{eth3d_dataset_open_source_code},\\ synchronized global shutter\\cameras, modeling rolling shutter\\effect not required\end{tabular} & \checkmark                                                      & \begin{tabular}[c]{@{}c@{}}Stereo/mono/RGBD\\ cameras, IMU,\\ Faro Focus X 330\\ laser scanner\end{tabular}                                                                                           \\ \hline
Ford \cite{gpandey-2011a}                                                                                      &        & \checkmark     &        &       & \begin{tabular}[c]{@{}c@{}}Weather condition: sunny,\\time-registered data,\\accompanied with utility\\source codes\end{tabular}                                                                                                                                           & \checkmark                                                      & \begin{tabular}[c]{@{}c@{}}GPS, IMU, Velodyne 3D\\ LiDAR, 2 push-broom\\ forward looking Riegl\\ LiDARs, omnidirectional\\ camera\end{tabular}                                                       \\ \hline
\begin{tabular}[c]{@{}c@{}}2D\\Cartographer\\Backpack\\Deutsches\\Museum \cite{2DCartDtchMuse-dataset}\end{tabular} & \checkmark      &       &        &       & \begin{tabular}[c]{@{}c@{}}ROS .bag format, intended\\ for 2D SLAM\end{tabular}                                                                                                                                                                                               & \checkmark                                                      & \begin{tabular}[c]{@{}c@{}}IMU, two 2D LiDAR\\ (horizontal and vertical)\end{tabular}                                                                                                                  \\ \hline
KITTI \cite{geiger2013vision}                                                                                     &        & \checkmark     &        & \checkmark     & \begin{tabular}[c]{@{}c@{}}grayscale/color images,\\ ground truth provided based\\ on GPS and laser scanner\end{tabular}                                                                                                                                                      & \checkmark                                                      & \begin{tabular}[c]{@{}c@{}}2 Mono grayscale\\ cameras, 2 mono\\ color cameras, Velodyne 64\\ LiDAR, GPS, IMU\end{tabular}                                                                           \\ \hline
Malaga \cite{blanco2014malaga}                                                                                    &        & \checkmark     & \checkmark      &       & \begin{tabular}[c]{@{}c@{}}Direct sun, accompanied with C++\\sample codes for parsing\\raw log files\end{tabular}                                                                                                                                                           & \checkmark                                                      & \begin{tabular}[c]{@{}c@{}}Stereo camera,\\ 3 Hokuyo UTM-30LX\\ LiDAR, 2 Sick\\ LiDARs, GPS, IMU\end{tabular}                                                                                       \\ \hline
Oxford \cite{maddern20171}                                                                                    &        & \checkmark     &        &       & \begin{tabular}[c]{@{}c@{}}Different lighting/weather\\conditions, accompanied with\\Matlab, Python tools\end{tabular}                                                                                                                                                       & \checkmark                                                      & \begin{tabular}[c]{@{}c@{}}Stereo camera,\\ 3 monocular color\\ cameras, 2 Sick 2D\\ LiDARs, Sick 3D\\ LiDAR, GPS, INS\end{tabular}                                                                 \\ \hline
\end{tabular}
\end{table}
\section{Future work}\label{future-work-sec}
In this section, the future research direction for autonomous mobile robots is presented.

\subsection{Mobile robots deployment in real-world}
One of the ultimate goals of pursuing research on autonomous mobile robots is their deployment in manufacturing, hospitals, warehousing, etc. Implementing a multi-robot setup is quite natural to complete tasks faster\cite{nahavandi2022autonomous}. Deciding on the operations that must be carried out in centralized or decentralized manner is an important future direction \cite{fragapane2021planning}. It is also necessary to support multi-robot setup with efficient decision making about the set of tools that must be mounted on the robots to carry out specific tasks \cite{fragapane2021planning}.

Autonomous navigation has also been used to develop autonomous ships with different levels of autonomy. As future works, it is important to establish cooperative behavior between maritime autonomous surface ships (MASS) which have autonomy levels of D3 and D4. D3 autonomy is remote control of ships without having seafarers onboard while D4 refers to fully autonomous ships \cite{international2018working}. It is also critical to analyze safety in human-machine interaction setup \cite{kim2022safety}.
\subsection{SLAM}
Although there is multitude of works in SLAM literature, due to complex and challenging nature of SLAM, there are still important research gaps for future work. For efficient SLAM implementation, it is crucial to keep balance between exploration (visiting new places) and exploitation (revisiting known places for uncertainty reduction) \cite{thrun1995exploration} which is addressed by active SLAM. Future works \cite{placed2022survey} for active SLAM consist of active spatial perception, proposing evaluation metrics to facilitate comparison of methods, and performing benchmarks to move toward real-world applications.

SLAM methods commonly assume that the world is static. However, real world is full of moving objects which contribute to failure of feature recognition causing tracking failure in SLAM \cite{deng2018feature}. As future work, DL and spectral methods \cite{xu2021rethinking, xu2019improved} can be used to achieve more robust feature recognition and tracking \cite{macario2022comprehensive}. It may be possible to detect and exclude dynamic objects from map construction process of SLAM. Currently, dynamic SLAM methods such as \cite{sun2017improving,sun2018motion} exist to deal with dynamicity of the environment. Using semantic information to devise semantic SLAM is an ongoing field of research for addressing dynamic SLAM \cite{macario2022comprehensive}. Additionally, using semantic SLAM to take the autonomy level of robots to the next level is challenging future task \cite{chen2022overview}.

While DL has shown some merits for loop closure detection, due to its limitations, the true loop detection is still unaddressed \cite{arshad2021role}. The limitation is due to sensitivity of vision-based loop closure detection methods to different illumination conditions. LiDAR-based loop does not suffer from such limitations however its performance is degraded due to weather conditions such as rain. Coming up with hybrid methods, exploiting the strengths and covering the weaknesses of both LiDAR-based and DL-based methods seems to be a necessary future direction \cite{arshad2021role}. 

To deploy visual SLAM methods in real world applications, they must be executable on embedded platforms such as smartphones which have limited processing power. Therefore, one of the important future works for SLAM is to keep the balance between real-time execution and accuracy \cite{chen2022overview}.
\subsection{Autonomous exploration in unseen environment}
Being able to navigate and explore unseen environments autonomously is highly desirable. To this end, Shah et al. exploited RL to propose a method called RECON (Rapid Exploration Controllers for Outcome-driven Navigation) \cite{shah2021rapid} which was trained on visual dataset \cite{recon-dataset} collected during 18 months from diverse real-world environments including different seasons and lighting conditions. Two primary components of RECON are uncertainty-aware and context conditioned goal representation as well as topological map representation. Nodes of the map represent egocentric observations and the edges are distances between nodes.

Extending RECON, Shah and Levine \cite{shah2022viking} focused on long-range autonomous navigation through unseen outdoor environments without requiring accurate map and/or sensor data. The advantage of their method called ViKiNG (Vision-Based Kilometer-Scale Navigation with Geographic Hints) is twofold. First, ViKiNG manages to reach the given destination using inaccurate data (GPS coordinates, satellite maps and schematic roadmaps ) as a heuristic during planning. Second, this method does not build geometric maps and only relies on topological representation of the environment. This way, the issue of building and storing environment dense map for long distances is not encountered on robots with lightweight processing/storage platforms. The authors have trained ViKiNG using trajectories that are only 80 meters long but have managed to reach goals up to three kilometers away. As pointed out by Shah and Levine, one of the important future works is devising models similar to ViKiNG that are capable of accepting inputs such as textual instructions and paper maps.
\section{Conclusion}\label{conc-sec}
Autonomous navigation has come a long way but the journey still continues. Different sensor combinations and mathematical approaches have been tried leading to variety of approaches; each having advantages and disadvantages of their own. In this paper, we strove to present a comprehensive review on different aspects of autonomous mobile robots including localization methods, obstacle avoidance, sensor types, SLAM, etc. We also tried to highlight the role of DL in reinforcing obstacle avoidance and SLAM approaches. The common simulation tools and their properties were reviewed. Additionally, publicly available datasets (containing real sensor data) for implementing SLAM without having an actual robot were introduced. Finally, the possible directions for future research were pointed out.
\begin{acks}
The authors have no conflict of interest to declare.
\end{acks}
\bibliographystyle{ACM-Reference-Format}
\bibliography{Main}


\begin{thebibliography}{315}


\ifx \showCODEN    \undefined \def \showCODEN     #1{\unskip}     \fi
\ifx \showDOI      \undefined \def \showDOI       #1{#1}\fi
\ifx \showISBNx    \undefined \def \showISBNx     #1{\unskip}     \fi
\ifx \showISBNxiii \undefined \def \showISBNxiii  #1{\unskip}     \fi
\ifx \showISSN     \undefined \def \showISSN      #1{\unskip}     \fi
\ifx \showLCCN     \undefined \def \showLCCN      #1{\unskip}     \fi
\ifx \shownote     \undefined \def \shownote      #1{#1}          \fi
\ifx \showarticletitle \undefined \def \showarticletitle #1{#1}   \fi
\ifx \showURL      \undefined \def \showURL       {\relax}        \fi
\providecommand\bibfield[2]{#2}
\providecommand\bibinfo[2]{#2}
\providecommand\natexlab[1]{#1}
\providecommand\showeprint[2][]{arXiv:#2}

\bibitem[\protect\citeauthoryear{??}{ros}{[n. d.]}]%
        {ros}
 \bibinfo{year}{[n. d.]}\natexlab{}.
\newblock \bibinfo{howpublished}{url = {https://www.ros.org/}}.
  (\bibinfo{year}{[n. d.]}).
\newblock
\newblock
\shownote{accessed February 27, 2022.}


\bibitem[\protect\citeauthoryear{??}{gps}{[n. d.]}]%
        {gps-wikipedia}
 \bibinfo{year}{[n. d.]}\natexlab{}.
\newblock \bibinfo{howpublished}{url =
  {https://en.wikipedia.org/wiki/Global\_Positioning\_System}}.
  (\bibinfo{year}{[n. d.]}).
\newblock
\newblock
\shownote{accessed February 27, 2022.}


\bibitem[\protect\citeauthoryear{??}{jet}{[n. d.]}]%
        {jetson-boards}
 \bibinfo{year}{[n. d.]}\natexlab{}.
\newblock \bibinfo{howpublished}{url =
  {https://developer.nvidia.com/buy-jetson}}.   (\bibinfo{year}{[n. d.]}).
\newblock
\newblock
\shownote{accessed February 27, 2022.}


\bibitem[\protect\citeauthoryear{??}{rob}{[n. d.]}]%
        {robotic-main-simulators-comapre}
 \bibinfo{year}{[n. d.]}\natexlab{}.
\newblock \bibinfo{howpublished}{url =
  {https://learn.e.ros4.pro/en/robotic\_simulators/comparison/}}.
  (\bibinfo{year}{[n. d.]}).
\newblock
\newblock
\shownote{accessed February 27, 2022.}


\bibitem[\protect\citeauthoryear{??}{ode}{[n. d.]}]%
        {ode-physics-endgine}
 \bibinfo{year}{[n. d.]}\natexlab{}.
\newblock \bibinfo{howpublished}{url = {https://www.ode.org/}}.
  (\bibinfo{year}{[n. d.]}).
\newblock
\newblock
\shownote{accessed February 27, 2022.}


\bibitem[\protect\citeauthoryear{??}{fet}{[n. d.]}]%
        {fetch-robot-sim}
 \bibinfo{year}{[n. d.]}\natexlab{}.
\newblock \bibinfo{howpublished}{url =
  {https://docs.fetchrobotics.com/gazebo.html}}.   (\bibinfo{year}{[n. d.]}).
\newblock
\newblock
\shownote{accessed February 27, 2022.}


\bibitem[\protect\citeauthoryear{??}{pr2}{[n. d.]}]%
        {pr2-robot-sim}
 \bibinfo{year}{[n. d.]}\natexlab{}.
\newblock \bibinfo{howpublished}{url =
  {http://wiki.ros.org/Robots/PR2\#Hardware\_Drivers\_and\_Simulation}}.
  (\bibinfo{year}{[n. d.]}).
\newblock
\newblock
\shownote{accessed February 27, 2022.}


\bibitem[\protect\citeauthoryear{??}{kuk}{[n. d.]a}]%
        {kuka-iiwa-robot-sim}
 \bibinfo{year}{[n. d.]}\natexlab{a}.
\newblock \bibinfo{howpublished}{url =
  {https://github.com/IFL-CAMP/iiwa\_stack}}.   (\bibinfo{year}{[n. d.]}).
\newblock
\newblock
\shownote{accessed February 27, 2022.}


\bibitem[\protect\citeauthoryear{??}{nao}{[n. d.]}]%
        {nao-robot-sim}
 \bibinfo{year}{[n. d.]}\natexlab{}.
\newblock \bibinfo{howpublished}{url = {http://wiki.ros.org/nao}}.
  (\bibinfo{year}{[n. d.]}).
\newblock
\newblock
\shownote{accessed February 27, 2022.}


\bibitem[\protect\citeauthoryear{??}{gaz}{[n. d.]}]%
        {gazebo-sim}
 \bibinfo{year}{[n. d.]}\natexlab{}.
\newblock \bibinfo{howpublished}{url = {https://gazebosim.org/home}}.
  (\bibinfo{year}{[n. d.]}).
\newblock
\newblock
\shownote{accessed February 27, 2022.}


\bibitem[\protect\citeauthoryear{??}{web}{[n. d.]}]%
        {webots-sim}
 \bibinfo{year}{[n. d.]}\natexlab{}.
\newblock \bibinfo{howpublished}{url = {https://cyberbotics.com/}}.
  (\bibinfo{year}{[n. d.]}).
\newblock
\newblock
\shownote{accessed February 27, 2022.}


\bibitem[\protect\citeauthoryear{??}{cop}{[n. d.]}]%
        {coppeliasim-sim}
 \bibinfo{year}{[n. d.]}\natexlab{}.
\newblock \bibinfo{howpublished}{url = {https://www.coppeliarobotics.com/}}.
  (\bibinfo{year}{[n. d.]}).
\newblock
\newblock
\shownote{accessed February 27, 2022.}


\bibitem[\protect\citeauthoryear{??}{dxs}{[n. d.]}]%
        {dxslam-repo}
 \bibinfo{year}{[n. d.]}\natexlab{}.
\newblock \bibinfo{howpublished}{url =
  {https://github.com/ivipsourcecode/dxslam}}.   (\bibinfo{year}{[n. d.]}).
\newblock
\newblock
\shownote{accessed February 27, 2022.}


\bibitem[\protect\citeauthoryear{??}{eth}{[n. d.]a}]%
        {eth3d_dataset_}
 \bibinfo{year}{[n. d.]}\natexlab{a}.
\newblock \bibinfo{howpublished}{url = {https://www.eth3d.net/slam\_overview}}.
    (\bibinfo{year}{[n. d.]}).
\newblock
\newblock
\shownote{accessed February 27, 2022.}


\bibitem[\protect\citeauthoryear{??}{eth}{[n. d.]b}]%
        {eth3d_dataset_open_source_code}
 \bibinfo{year}{[n. d.]}\natexlab{b}.
\newblock \bibinfo{howpublished}{url = {https://github.com/ETH3D}}.
  (\bibinfo{year}{[n. d.]}).
\newblock
\newblock
\shownote{accessed February 27, 2022.}


\bibitem[\protect\citeauthoryear{??}{rec}{[n. d.]}]%
        {recon-dataset}
 \bibinfo{year}{[n. d.]}\natexlab{}.
\newblock \bibinfo{howpublished}{url = {sites.google.com/view/recon-robot}}.
  (\bibinfo{year}{[n. d.]}).
\newblock
\newblock
\shownote{accessed February 27, 2022.}


\bibitem[\protect\citeauthoryear{??}{2DC}{[n. d.]}]%
        {2DCartDtchMuse-dataset}
 \bibinfo{year}{[n. d.]}\natexlab{}.
\newblock \bibinfo{title}{2D Cartographer Backpack - Deutsches Museum}.
\newblock \bibinfo{howpublished}{url =
  {https://google-cartographer-ros.readthedocs.io/en/latest/data.html}}.
  (\bibinfo{year}{[n. d.]}).
\newblock
\newblock
\shownote{accessed February 27, 2022.}


\bibitem[\protect\citeauthoryear{??}{dar}{[n. d.]a}]%
        {darpa-grand2004}
 \bibinfo{year}{[n. d.]}\natexlab{a}.
\newblock \bibinfo{title}{DARPA Grand Challenge (2004)}.
\newblock \bibinfo{howpublished}{url =
  {https://en.wikipedia.org/wiki/DARPA\_Grand\_Challenge\_(2004)}}.
  (\bibinfo{year}{[n. d.]}).
\newblock
\newblock
\shownote{accessed February 27, 2022.}


\bibitem[\protect\citeauthoryear{??}{dar}{[n. d.]b}]%
        {darpa-urban-challenge}
 \bibinfo{year}{[n. d.]}\natexlab{b}.
\newblock \bibinfo{title}{DARPA Grand Challenge (2007)}.
\newblock \bibinfo{howpublished}{url =
  {https://en.wikipedia.org/wiki/DARPA\_Grand\_Challenge\_(2007)}}.
  (\bibinfo{year}{[n. d.]}).
\newblock
\newblock
\shownote{accessed February 27, 2022.}


\bibitem[\protect\citeauthoryear{??}{dop}{[n. d.]}]%
        {doppler-velocity-logs}
 \bibinfo{year}{[n. d.]}\natexlab{}.
\newblock \bibinfo{title}{DVLs - Doppler Velocity Logs}.
\newblock \bibinfo{howpublished}{url =
  {https://geo-matching.com/dvls-doppler-velocity-logs}}.   (\bibinfo{year}{[n.
  d.]}).
\newblock
\newblock
\shownote{accessed February 27, 2022.}


\bibitem[\protect\citeauthoryear{??}{kuk}{[n. d.]b}]%
        {kuka-industial-robots}
 \bibinfo{year}{[n. d.]}\natexlab{b}.
\newblock \bibinfo{title}{Industrial robots from KUKA}.
\newblock \bibinfo{howpublished}{url =
  {https://www.kuka.com/en-de/products/robot-systems/industrial-robots}}.
  (\bibinfo{year}{[n. d.]}).
\newblock
\newblock
\shownote{accessed February 27, 2022.}


\bibitem[\protect\citeauthoryear{??}{eth}{[n. d.]c}]%
        {eth-kalibr-tool}
 \bibinfo{year}{[n. d.]}\natexlab{c}.
\newblock \bibinfo{title}{Kalibr}.
\newblock \bibinfo{howpublished}{url = {https://github.com/ethz-asl/kalibr}}.
  (\bibinfo{year}{[n. d.]}).
\newblock
\newblock
\shownote{accessed February 27, 2022.}


\bibitem[\protect\citeauthoryear{??}{Lun}{[n. d.]}]%
        {Lunokhod1}
 \bibinfo{year}{[n. d.]}\natexlab{}.
\newblock \bibinfo{title}{Luna - Exploring the Moon}.
\newblock \bibinfo{howpublished}{url =
  {http://www.orbitalfocus.uk/Diaries/Luna/Luna17.php}}.   (\bibinfo{year}{[n.
  d.]}).
\newblock
\newblock
\shownote{accessed February 27, 2022.}


\bibitem[\protect\citeauthoryear{??}{par}{[n. d.]}]%
        {parrot-uav}
 \bibinfo{year}{[n. d.]}\natexlab{}.
\newblock \bibinfo{title}{Parrot drones}.
\newblock \bibinfo{howpublished}{url = {https://www.parrot.com/us/drones}}.
  (\bibinfo{year}{[n. d.]}).
\newblock
\newblock
\shownote{accessed February 27, 2022.}


\bibitem[\protect\citeauthoryear{??}{Dic}{[n. d.]}]%
        {Dickmanns-car}
 \bibinfo{year}{[n. d.]}\natexlab{}.
\newblock \bibinfo{title}{Prof. Schmidhuber's highlights of robot car history}.
\newblock \bibinfo{howpublished}{url =
  {https://people.idsia.ch/~juergen/robotcars.html}}.   (\bibinfo{year}{[n.
  d.]}).
\newblock
\newblock
\shownote{accessed February 27, 2022.}


\bibitem[\protect\citeauthoryear{??}{Rob}{[n. d.]}]%
        {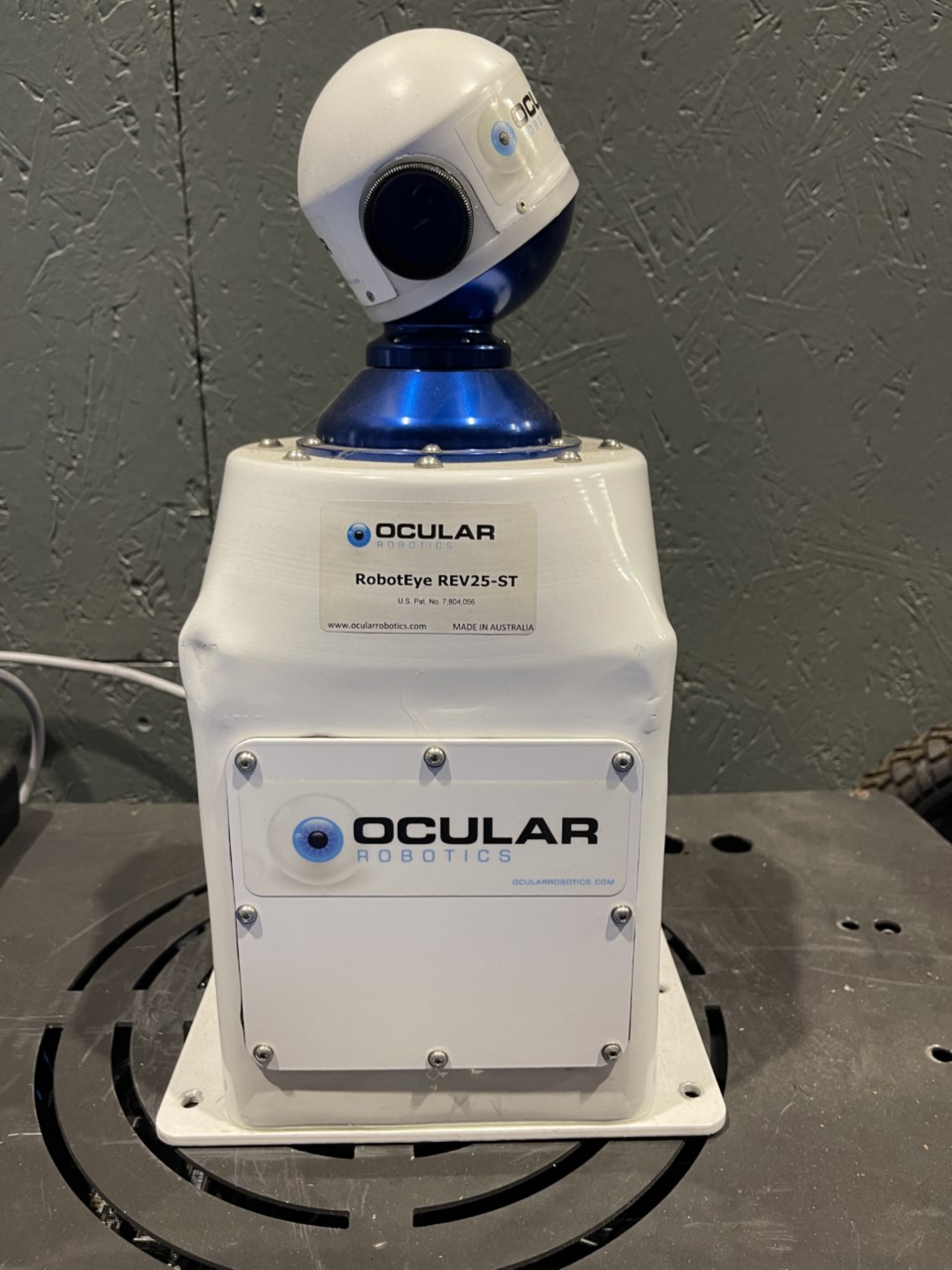}
 \bibinfo{year}{[n. d.]}\natexlab{}.
\newblock \bibinfo{title}{RobotEye REV25-ST}.
\newblock \bibinfo{howpublished}{url =
  {https://www.ocularrobotics.com/products/camera-pointing-devices/rev25-st/}}.
    (\bibinfo{year}{[n. d.]}).
\newblock
\newblock
\shownote{accessed July 14, 2022.}


\bibitem[\protect\citeauthoryear{??}{eth}{[n. d.]d}]%
        {eth-rotors-sim}
 \bibinfo{year}{[n. d.]}\natexlab{d}.
\newblock \bibinfo{title}{Rotors\_simulator}.
\newblock \bibinfo{howpublished}{url =
  {https://github.com/ethz-asl/rotors\_simulator}}.   (\bibinfo{year}{[n.
  d.]}).
\newblock
\newblock
\shownote{accessed February 27, 2022.}


\bibitem[\protect\citeauthoryear{Abdelkrim, Aouf, Tsourdos, and
  White}{Abdelkrim et~al\mbox{.}}{2008}]%
        {4602149}
\bibfield{author}{\bibinfo{person}{Nemra Abdelkrim}, \bibinfo{person}{Nabil
  Aouf}, \bibinfo{person}{Antonios Tsourdos}, {and} \bibinfo{person}{Brian
  White}.} \bibinfo{year}{2008}\natexlab{}.
\newblock \showarticletitle{Robust nonlinear filtering for INS/GPS UAV
  localization}. In \bibinfo{booktitle}{\emph{2008 16th Mediterranean
  Conference on Control and Automation}}. \bibinfo{pages}{695--702}.
\newblock
\urldef\tempurl%
\url{https://doi.org/10.1109/MED.2008.4602149}
\showDOI{\tempurl}


\bibitem[\protect\citeauthoryear{Abreu and Matos}{Abreu and Matos}{2014}]%
        {abreu2014minehunting}
\bibfield{author}{\bibinfo{person}{Nuno Abreu} {and}
  \bibinfo{person}{An{\'\i}bal Matos}.} \bibinfo{year}{2014}\natexlab{}.
\newblock \showarticletitle{Minehunting mission planning for autonomous
  underwater systems using evolutionary algorithms}.
\newblock \bibinfo{journal}{\emph{Unmanned Systems}} \bibinfo{volume}{2},
  \bibinfo{number}{04} (\bibinfo{year}{2014}), \bibinfo{pages}{323--349}.
\newblock


\bibitem[\protect\citeauthoryear{Ackerman}{Ackerman}{2017}]%
        {cmu-12years}
\bibfield{author}{\bibinfo{person}{Evan Ackerman}.}
  \bibinfo{year}{2017}\natexlab{}.
\newblock \bibinfo{title}{Carnegie Mellon Solves 12-Year-Old DARPA Grand
  Challenge Mystery}.
\newblock \bibinfo{howpublished}{url =
  {https://spectrum.ieee.org/cmu-solves-12-year-old-darpa-grand-challenge-mystery}}.
    (\bibinfo{year}{2017}).
\newblock
\newblock
\shownote{accessed February 27, 2022.}


\bibitem[\protect\citeauthoryear{Adams}{Adams}{1961}]%
        {adams1961remote}
\bibfield{author}{\bibinfo{person}{James~Lowell Adams}.}
  \bibinfo{year}{1961}\natexlab{}.
\newblock \bibinfo{booktitle}{\emph{Remote control with long transmission
  delays}}.
\newblock \bibinfo{publisher}{Stanford University}.
\newblock


\bibitem[\protect\citeauthoryear{Alatise and Hancke}{Alatise and
  Hancke}{2020}]%
        {alatise2020review}
\bibfield{author}{\bibinfo{person}{Mary~B Alatise} {and}
  \bibinfo{person}{Gerhard~P Hancke}.} \bibinfo{year}{2020}\natexlab{}.
\newblock \showarticletitle{A review on challenges of autonomous mobile robot
  and sensor fusion methods}.
\newblock \bibinfo{journal}{\emph{IEEE Access}}  \bibinfo{volume}{8}
  (\bibinfo{year}{2020}), \bibinfo{pages}{39830--39846}.
\newblock


\bibitem[\protect\citeauthoryear{Alsadik and Karam}{Alsadik and Karam}{2021}]%
        {alsadik2021simultaneous}
\bibfield{author}{\bibinfo{person}{Bashar Alsadik} {and} \bibinfo{person}{Samer
  Karam}.} \bibinfo{year}{2021}\natexlab{}.
\newblock \showarticletitle{The simultaneous localization and mapping (SLAM)-An
  overview}.
\newblock \bibinfo{journal}{\emph{Surveying and Geospatial Engineering
  Journal}} \bibinfo{volume}{2}, \bibinfo{number}{01} (\bibinfo{year}{2021}),
  \bibinfo{pages}{01--12}.
\newblock


\bibitem[\protect\citeauthoryear{Antich, Ortiz, and M{\'\i}nguez}{Antich
  et~al\mbox{.}}{2009a}]%
        {antich2009bug}
\bibfield{author}{\bibinfo{person}{Javier Antich}, \bibinfo{person}{Alberto
  Ortiz}, {and} \bibinfo{person}{Javier M{\'\i}nguez}.}
  \bibinfo{year}{2009}\natexlab{a}.
\newblock \showarticletitle{A bug-inspired algorithm for efficient anytime path
  planning}. In \bibinfo{booktitle}{\emph{2009 IEEE/RSJ International
  Conference on Intelligent Robots and Systems}}. IEEE,
  \bibinfo{pages}{5407--5413}.
\newblock


\bibitem[\protect\citeauthoryear{Antich, Ortiz, and Minguez}{Antich
  et~al\mbox{.}}{2009b}]%
        {antich2009bug2+}
\bibfield{author}{\bibinfo{person}{Javier Antich}, \bibinfo{person}{Alberto
  Ortiz}, {and} \bibinfo{person}{J Minguez}.} \bibinfo{year}{2009}\natexlab{b}.
\newblock \showarticletitle{Bug2+: Details and formal proofs}.
\newblock \bibinfo{journal}{\emph{Technical Report A-1-2009, University of the
  Balearic Islands}} (\bibinfo{year}{2009}).
\newblock


\bibitem[\protect\citeauthoryear{Aplin, Cazzolato, Grainger, and Madden}{Aplin
  et~al\mbox{.}}{2010}]%
        {aplin20101025}
\bibfield{author}{\bibinfo{person}{Richard Aplin}, \bibinfo{person}{Ben
  Cazzolato}, \bibinfo{person}{Steven Grainger}, {and} \bibinfo{person}{Chris
  Madden}.} \bibinfo{year}{2010}\natexlab{}.
\newblock \bibinfo{booktitle}{\emph{1025: MAGIC 2010 Multi Autonomous Ground
  International Challenge. Volume I}}.
\newblock \bibinfo{type}{{T}echnical {R}eport}. \bibinfo{institution}{ADELAIDE
  UNIV (AUSTRALIA)}.
\newblock


\bibitem[\protect\citeauthoryear{Arshad and Kim}{Arshad and Kim}{2021}]%
        {arshad2021role}
\bibfield{author}{\bibinfo{person}{Saba Arshad} {and} \bibinfo{person}{Gon-Woo
  Kim}.} \bibinfo{year}{2021}\natexlab{}.
\newblock \showarticletitle{Role of deep learning in loop closure detection for
  visual and lidar slam: A survey}.
\newblock \bibinfo{journal}{\emph{Sensors}} \bibinfo{volume}{21},
  \bibinfo{number}{4} (\bibinfo{year}{2021}), \bibinfo{pages}{1243}.
\newblock


\bibitem[\protect\citeauthoryear{Arulampalam, Maskell, Gordon, and
  Clapp}{Arulampalam et~al\mbox{.}}{2002}]%
        {arulampalam2002tutorial}
\bibfield{author}{\bibinfo{person}{M~Sanjeev Arulampalam},
  \bibinfo{person}{Simon Maskell}, \bibinfo{person}{Neil Gordon}, {and}
  \bibinfo{person}{Tim Clapp}.} \bibinfo{year}{2002}\natexlab{}.
\newblock \showarticletitle{A tutorial on particle filters for online
  nonlinear/non-Gaussian Bayesian tracking}.
\newblock \bibinfo{journal}{\emph{IEEE Transactions on signal processing}}
  \bibinfo{volume}{50}, \bibinfo{number}{2} (\bibinfo{year}{2002}),
  \bibinfo{pages}{174--188}.
\newblock


\bibitem[\protect\citeauthoryear{Bader and Rahimifard}{Bader and
  Rahimifard}{2020}]%
        {bader2020methodology}
\bibfield{author}{\bibinfo{person}{Farah Bader} {and} \bibinfo{person}{Shahin
  Rahimifard}.} \bibinfo{year}{2020}\natexlab{}.
\newblock \showarticletitle{A methodology for the selection of industrial
  robots in food handling}.
\newblock \bibinfo{journal}{\emph{Innovative Food Science \& Emerging
  Technologies}}  \bibinfo{volume}{64} (\bibinfo{year}{2020}),
  \bibinfo{pages}{102379}.
\newblock


\bibitem[\protect\citeauthoryear{Badino, Huber, and Kanade}{Badino
  et~al\mbox{.}}{2011}]%
        {badino2011visual}
\bibfield{author}{\bibinfo{person}{Hern{\'a}n Badino}, \bibinfo{person}{Daniel
  Huber}, {and} \bibinfo{person}{Takeo Kanade}.}
  \bibinfo{year}{2011}\natexlab{}.
\newblock \showarticletitle{Visual topometric localization}. In
  \bibinfo{booktitle}{\emph{2011 IEEE Intelligent vehicles symposium (IV)}}.
  IEEE, \bibinfo{pages}{794--799}.
\newblock


\bibitem[\protect\citeauthoryear{Bagnell, Bradley, Silver, Sofman, and
  Stentz}{Bagnell et~al\mbox{.}}{2010}]%
        {bagnell2010learning}
\bibfield{author}{\bibinfo{person}{James~Andrew Bagnell},
  \bibinfo{person}{David Bradley}, \bibinfo{person}{David Silver},
  \bibinfo{person}{Boris Sofman}, {and} \bibinfo{person}{Anthony Stentz}.}
  \bibinfo{year}{2010}\natexlab{}.
\newblock \showarticletitle{Learning for autonomous navigation}.
\newblock \bibinfo{journal}{\emph{IEEE Robotics \& Automation Magazine}}
  \bibinfo{volume}{17}, \bibinfo{number}{2} (\bibinfo{year}{2010}),
  \bibinfo{pages}{74--84}.
\newblock


\bibitem[\protect\citeauthoryear{Bajracharya, Maimone, and Helmick}{Bajracharya
  et~al\mbox{.}}{2008}]%
        {bajracharya2008autonomy}
\bibfield{author}{\bibinfo{person}{Max Bajracharya}, \bibinfo{person}{Mark~W
  Maimone}, {and} \bibinfo{person}{Daniel Helmick}.}
  \bibinfo{year}{2008}\natexlab{}.
\newblock \showarticletitle{Autonomy for mars rovers: Past, present, and
  future}.
\newblock \bibinfo{journal}{\emph{Computer}} \bibinfo{volume}{41},
  \bibinfo{number}{12} (\bibinfo{year}{2008}), \bibinfo{pages}{44--50}.
\newblock


\bibitem[\protect\citeauthoryear{Bao, Li, Qiao, and Rauschenbach}{Bao
  et~al\mbox{.}}{2020}]%
        {bao2020integrated}
\bibfield{author}{\bibinfo{person}{Jianhua Bao}, \bibinfo{person}{Daoliang Li},
  \bibinfo{person}{Xi Qiao}, {and} \bibinfo{person}{Thomas Rauschenbach}.}
  \bibinfo{year}{2020}\natexlab{}.
\newblock \showarticletitle{Integrated navigation for autonomous underwater
  vehicles in aquaculture: A review}.
\newblock \bibinfo{journal}{\emph{Information processing in agriculture}}
  \bibinfo{volume}{7}, \bibinfo{number}{1} (\bibinfo{year}{2020}),
  \bibinfo{pages}{139--151}.
\newblock


\bibitem[\protect\citeauthoryear{Barducci, Pittiglio, Norton, Obstein, and
  Valdastri}{Barducci et~al\mbox{.}}{2019}]%
        {barducci2019adaptive}
\bibfield{author}{\bibinfo{person}{Lavinia Barducci}, \bibinfo{person}{Giovanni
  Pittiglio}, \bibinfo{person}{Joseph~C Norton}, \bibinfo{person}{Keith~L
  Obstein}, {and} \bibinfo{person}{Pietro Valdastri}.}
  \bibinfo{year}{2019}\natexlab{}.
\newblock \showarticletitle{Adaptive dynamic control for magnetically actuated
  medical robots}.
\newblock \bibinfo{journal}{\emph{IEEE robotics and automation letters}}
  \bibinfo{volume}{4}, \bibinfo{number}{4} (\bibinfo{year}{2019}),
  \bibinfo{pages}{3633--3640}.
\newblock


\bibitem[\protect\citeauthoryear{Bares and Wettergreen}{Bares and
  Wettergreen}{1999}]%
        {bares1999dante}
\bibfield{author}{\bibinfo{person}{John~E Bares} {and} \bibinfo{person}{David~S
  Wettergreen}.} \bibinfo{year}{1999}\natexlab{}.
\newblock \showarticletitle{Dante II: Technical description, results, and
  lessons learned}.
\newblock \bibinfo{journal}{\emph{The International Journal of robotics
  research}} \bibinfo{volume}{18}, \bibinfo{number}{7} (\bibinfo{year}{1999}),
  \bibinfo{pages}{621--649}.
\newblock


\bibitem[\protect\citeauthoryear{Barr, Swaszek, Hartnett, and Johnson}{Barr
  et~al\mbox{.}}{2013}]%
        {barr2013performance}
\bibfield{author}{\bibinfo{person}{Simon~P Barr}, \bibinfo{person}{Peter~F
  Swaszek}, \bibinfo{person}{Richard~J Hartnett}, {and}
  \bibinfo{person}{Gregory~W Johnson}.} \bibinfo{year}{2013}\natexlab{}.
\newblock \showarticletitle{Performance of Multi-Beacon DGPS}. In
  \bibinfo{booktitle}{\emph{Proceedings of the International Technical Meeting
  of The Institute of Navigation}}. ION, \bibinfo{pages}{359--373}.
\newblock


\bibitem[\protect\citeauthoryear{Bay, Tuytelaars, and Van~Gool}{Bay
  et~al\mbox{.}}{2006}]%
        {Herbert}
\bibfield{author}{\bibinfo{person}{Herbert Bay}, \bibinfo{person}{Tinne
  Tuytelaars}, {and} \bibinfo{person}{Luc Van~Gool}.}
  \bibinfo{year}{2006}\natexlab{}.
\newblock \showarticletitle{SURF: Speeded Up Robust Features}. In
  \bibinfo{booktitle}{\emph{Computer Vision -- ECCV 2006}},
  \bibfield{editor}{\bibinfo{person}{Alevs Leonardis}, \bibinfo{person}{Horst
  Bischof}, {and} \bibinfo{person}{Axel Pinz}} (Eds.).
  \bibinfo{publisher}{Springer Berlin Heidelberg}, \bibinfo{address}{Berlin,
  Heidelberg}, \bibinfo{pages}{404--417}.
\newblock


\bibitem[\protect\citeauthoryear{Bernardino and Santos-Victor}{Bernardino and
  Santos-Victor}{1998}]%
        {bernardino1998visual}
\bibfield{author}{\bibinfo{person}{Alexandre Bernardino} {and}
  \bibinfo{person}{Jos{\'e} Santos-Victor}.} \bibinfo{year}{1998}\natexlab{}.
\newblock \showarticletitle{Visual behaviours for binocular tracking}.
\newblock \bibinfo{journal}{\emph{Robotics and Autonomous Systems}}
  \bibinfo{volume}{25}, \bibinfo{number}{3-4} (\bibinfo{year}{1998}),
  \bibinfo{pages}{137--146}.
\newblock


\bibitem[\protect\citeauthoryear{Blanco-Claraco, Moreno-Duenas, and
  Gonzalez-Jimenez}{Blanco-Claraco et~al\mbox{.}}{2014}]%
        {blanco2014malaga}
\bibfield{author}{\bibinfo{person}{Jose-Luis Blanco-Claraco},
  \bibinfo{person}{Francisco-Angel Moreno-Duenas}, {and}
  \bibinfo{person}{Javier Gonzalez-Jimenez}.} \bibinfo{year}{2014}\natexlab{}.
\newblock \showarticletitle{The Malaga urban dataset: High-rate stereo and
  LiDAR in a realistic urban scenario}.
\newblock \bibinfo{journal}{\emph{The International Journal of Robotics
  Research}} \bibinfo{volume}{33}, \bibinfo{number}{2} (\bibinfo{year}{2014}),
  \bibinfo{pages}{207--214}.
\newblock


\bibitem[\protect\citeauthoryear{Bonin-Font, Ortiz, and Oliver}{Bonin-Font
  et~al\mbox{.}}{2008}]%
        {bonin2008visual}
\bibfield{author}{\bibinfo{person}{Francisco Bonin-Font},
  \bibinfo{person}{Alberto Ortiz}, {and} \bibinfo{person}{Gabriel Oliver}.}
  \bibinfo{year}{2008}\natexlab{}.
\newblock \showarticletitle{Visual navigation for mobile robots: A survey}.
\newblock \bibinfo{journal}{\emph{Journal of intelligent and robotic systems}}
  \bibinfo{volume}{53}, \bibinfo{number}{3} (\bibinfo{year}{2008}),
  \bibinfo{pages}{263--296}.
\newblock


\bibitem[\protect\citeauthoryear{Borenstein and Koren}{Borenstein and
  Koren}{1989}]%
        {borenstein1989real}
\bibfield{author}{\bibinfo{person}{Johann Borenstein} {and}
  \bibinfo{person}{Yoram Koren}.} \bibinfo{year}{1989}\natexlab{}.
\newblock \showarticletitle{Real-time obstacle avoidance for fast mobile
  robots}.
\newblock \bibinfo{journal}{\emph{IEEE Transactions on systems, Man, and
  Cybernetics}} \bibinfo{volume}{19}, \bibinfo{number}{5}
  (\bibinfo{year}{1989}), \bibinfo{pages}{1179--1187}.
\newblock


\bibitem[\protect\citeauthoryear{Borenstein, Koren, et~al\mbox{.}}{Borenstein
  et~al\mbox{.}}{1991}]%
        {borenstein1991vector}
\bibfield{author}{\bibinfo{person}{Johann Borenstein}, \bibinfo{person}{Yoram
  Koren}, {et~al\mbox{.}}} \bibinfo{year}{1991}\natexlab{}.
\newblock \showarticletitle{The vector field histogram-fast obstacle avoidance
  for mobile robots}.
\newblock \bibinfo{journal}{\emph{IEEE transactions on robotics and
  automation}} \bibinfo{volume}{7}, \bibinfo{number}{3} (\bibinfo{year}{1991}),
  \bibinfo{pages}{278--288}.
\newblock


\bibitem[\protect\citeauthoryear{Bosch, Gracias, Ridao, Isteni{\v{c}}, and
  Ribas}{Bosch et~al\mbox{.}}{2016}]%
        {bosch2016close}
\bibfield{author}{\bibinfo{person}{Josep Bosch}, \bibinfo{person}{Nuno
  Gracias}, \bibinfo{person}{Pere Ridao}, \bibinfo{person}{Klemen
  Isteni{\v{c}}}, {and} \bibinfo{person}{David Ribas}.}
  \bibinfo{year}{2016}\natexlab{}.
\newblock \showarticletitle{Close-range tracking of underwater vehicles using
  light beacons}.
\newblock \bibinfo{journal}{\emph{Sensors}} \bibinfo{volume}{16},
  \bibinfo{number}{4} (\bibinfo{year}{2016}), \bibinfo{pages}{429}.
\newblock


\bibitem[\protect\citeauthoryear{Bradley and Bagnell}{Bradley and
  Bagnell}{2010}]%
        {bradley2010domain}
\bibfield{author}{\bibinfo{person}{David~M Bradley} {and}
  \bibinfo{person}{J~Andrew Bagnell}.} \bibinfo{year}{2010}\natexlab{}.
\newblock \bibinfo{booktitle}{\emph{Domain adaptation for mobile robot
  navigation}}.
\newblock \bibinfo{type}{{T}echnical {R}eport}. \bibinfo{institution}{Number
  CMU-RI-TR}.
\newblock


\bibitem[\protect\citeauthoryear{Braid, Broggi, and Schmiedel}{Braid
  et~al\mbox{.}}{2006}]%
        {braid2006terramax}
\bibfield{author}{\bibinfo{person}{Deborah Braid}, \bibinfo{person}{Alberto
  Broggi}, {and} \bibinfo{person}{Gary Schmiedel}.}
  \bibinfo{year}{2006}\natexlab{}.
\newblock \showarticletitle{The TerraMax autonomous vehicle}.
\newblock \bibinfo{journal}{\emph{Journal of Field Robotics}}
  \bibinfo{volume}{23}, \bibinfo{number}{9} (\bibinfo{year}{2006}),
  \bibinfo{pages}{693--708}.
\newblock


\bibitem[\protect\citeauthoryear{Cai, Hsu, Chai, Ding, and Wang}{Cai
  et~al\mbox{.}}{2019}]%
        {cai2019multi}
\bibfield{author}{\bibinfo{person}{Xiaobo Cai}, \bibinfo{person}{Houtse Hsu},
  \bibinfo{person}{Hua Chai}, \bibinfo{person}{Leixiang Ding}, {and}
  \bibinfo{person}{Yong Wang}.} \bibinfo{year}{2019}\natexlab{}.
\newblock \showarticletitle{Multi-antenna GNSS and INS integrated position and
  attitude determination without base station for land vehicles}.
\newblock \bibinfo{journal}{\emph{The Journal of Navigation}}
  \bibinfo{volume}{72}, \bibinfo{number}{2} (\bibinfo{year}{2019}),
  \bibinfo{pages}{342--358}.
\newblock


\bibitem[\protect\citeauthoryear{Calvo, Sousa, Bibiloni, Curti, Acosta, and
  Rozenfeld}{Calvo et~al\mbox{.}}{2009}]%
        {calvo2009low}
\bibfield{author}{\bibinfo{person}{O Calvo}, \bibinfo{person}{A Sousa},
  \bibinfo{person}{J Bibiloni}, \bibinfo{person}{H Curti}, \bibinfo{person}{G
  Acosta}, {and} \bibinfo{person}{A Rozenfeld}.}
  \bibinfo{year}{2009}\natexlab{}.
\newblock \showarticletitle{Low-cost autonomous underwater vehicle for
  underwater acoustic inspections}.
\newblock \bibinfo{journal}{\emph{Journal of Maritime Research}}
  \bibinfo{volume}{6}, \bibinfo{number}{2} (\bibinfo{year}{2009}),
  \bibinfo{pages}{37--52}.
\newblock


\bibitem[\protect\citeauthoryear{Campos, Elvira, Rodr{\'\i}guez, Montiel, and
  Tard{\'o}s}{Campos et~al\mbox{.}}{2021}]%
        {campos2021orb}
\bibfield{author}{\bibinfo{person}{Carlos Campos}, \bibinfo{person}{Richard
  Elvira}, \bibinfo{person}{Juan J~G{\'o}mez Rodr{\'\i}guez},
  \bibinfo{person}{Jos{\'e}~MM Montiel}, {and} \bibinfo{person}{Juan~D
  Tard{\'o}s}.} \bibinfo{year}{2021}\natexlab{}.
\newblock \showarticletitle{Orb-slam3: An accurate open-source library for
  visual, visual--inertial, and multimap slam}.
\newblock \bibinfo{journal}{\emph{IEEE Transactions on Robotics}}
  \bibinfo{volume}{37}, \bibinfo{number}{6} (\bibinfo{year}{2021}),
  \bibinfo{pages}{1874--1890}.
\newblock


\bibitem[\protect\citeauthoryear{Cao, Hu, and Kneip}{Cao et~al\mbox{.}}{2020}]%
        {cao2020representations}
\bibfield{author}{\bibinfo{person}{Yuchen Cao}, \bibinfo{person}{Lan Hu}, {and}
  \bibinfo{person}{Laurent Kneip}.} \bibinfo{year}{2020}\natexlab{}.
\newblock \showarticletitle{Representations and Benchmarking of Modern Visual
  SLAM Systems}.
\newblock \bibinfo{journal}{\emph{Sensors}} \bibinfo{volume}{20},
  \bibinfo{number}{9} (\bibinfo{year}{2020}), \bibinfo{pages}{2572}.
\newblock


\bibitem[\protect\citeauthoryear{Cebrian, Bell{\'e}s, Martin, Salas,
  Fern{\'a}ndez, Arribas, Vil{\`a}-Valls, and Navarro}{Cebrian
  et~al\mbox{.}}{2019}]%
        {cebrian2019low}
\bibfield{author}{\bibinfo{person}{{\'A}lvaro Cebrian}, \bibinfo{person}{Andrea
  Bell{\'e}s}, \bibinfo{person}{Carla Martin}, \bibinfo{person}{Aitor Salas},
  \bibinfo{person}{Javier Fern{\'a}ndez}, \bibinfo{person}{Javier Arribas},
  \bibinfo{person}{Jordi Vil{\`a}-Valls}, {and} \bibinfo{person}{Monica
  Navarro}.} \bibinfo{year}{2019}\natexlab{}.
\newblock \showarticletitle{Low-cost hybrid GNSS/UWB/INS integration for
  seamless indoor/outdoor UAV navigation}. In
  \bibinfo{booktitle}{\emph{Proceedings of the 32nd International Technical
  Meeting of the Satellite Division of The Institute of Navigation (ION GNSS+
  2019)}}. \bibinfo{pages}{2680--2691}.
\newblock


\bibitem[\protect\citeauthoryear{Chae, Choi, and Song}{Chae
  et~al\mbox{.}}{2020}]%
        {chae2020robust}
\bibfield{author}{\bibinfo{person}{Hee-Won Chae}, \bibinfo{person}{Ji-Hoon
  Choi}, {and} \bibinfo{person}{Jae-Bok Song}.}
  \bibinfo{year}{2020}\natexlab{}.
\newblock \showarticletitle{Robust and autonomous stereo visual-inertial
  navigation for non-holonomic mobile robots}.
\newblock \bibinfo{journal}{\emph{IEEE Transactions on Vehicular Technology}}
  \bibinfo{volume}{69}, \bibinfo{number}{9} (\bibinfo{year}{2020}),
  \bibinfo{pages}{9613--9623}.
\newblock


\bibitem[\protect\citeauthoryear{Champeny-Bares, Coppersmith, and
  Dowling}{Champeny-Bares et~al\mbox{.}}{1991}]%
        {champeny1991terregator}
\bibfield{author}{\bibinfo{person}{Lee Champeny-Bares}, \bibinfo{person}{Syd
  Coppersmith}, {and} \bibinfo{person}{Kevin Dowling}.}
  \bibinfo{year}{1991}\natexlab{}.
\newblock \bibinfo{booktitle}{\emph{The terregator mobile robot}}.
\newblock \bibinfo{type}{{T}echnical {R}eport}.
  \bibinfo{institution}{CARNEGIE-MELLON UNIV PITTSBURGH PA ROBOTICS INST}.
\newblock


\bibitem[\protect\citeauthoryear{Chen, Liu, Xu, Li, Ding, Wang, and Feng}{Chen
  et~al\mbox{.}}{2022a}]%
        {chen2022self}
\bibfield{author}{\bibinfo{person}{Chao Chen}, \bibinfo{person}{Xinhao Liu},
  \bibinfo{person}{Xuchu Xu}, \bibinfo{person}{Yiming Li}, \bibinfo{person}{Li
  Ding}, \bibinfo{person}{Ruoyu Wang}, {and} \bibinfo{person}{Chen Feng}.}
  \bibinfo{year}{2022}\natexlab{a}.
\newblock \showarticletitle{Self-Supervised Visual Place Recognition by Mining
  Temporal and Feature Neighborhoods}.
\newblock \bibinfo{journal}{\emph{arXiv preprint arXiv:2208.09315}}
  (\bibinfo{year}{2022}).
\newblock


\bibitem[\protect\citeauthoryear{Chen, Shang, Ji, Zhou, Wang, Xu, Li, and
  Hu}{Chen et~al\mbox{.}}{2022b}]%
        {chen2022overview}
\bibfield{author}{\bibinfo{person}{Weifeng Chen}, \bibinfo{person}{Guangtao
  Shang}, \bibinfo{person}{Aihong Ji}, \bibinfo{person}{Chengjun Zhou},
  \bibinfo{person}{Xiyang Wang}, \bibinfo{person}{Chonghui Xu},
  \bibinfo{person}{Zhenxiong Li}, {and} \bibinfo{person}{Kai Hu}.}
  \bibinfo{year}{2022}\natexlab{b}.
\newblock \showarticletitle{An Overview on Visual SLAM: From Tradition to
  Semantic}.
\newblock \bibinfo{journal}{\emph{Remote Sensing}} \bibinfo{volume}{14},
  \bibinfo{number}{13} (\bibinfo{year}{2022}), \bibinfo{pages}{3010}.
\newblock


\bibitem[\protect\citeauthoryear{Chen and Birchfield}{Chen and
  Birchfield}{2009}]%
        {chen2009qualitative}
\bibfield{author}{\bibinfo{person}{Zhichao Chen} {and}
  \bibinfo{person}{Stanley~T Birchfield}.} \bibinfo{year}{2009}\natexlab{}.
\newblock \showarticletitle{Qualitative vision-based path following}.
\newblock \bibinfo{journal}{\emph{IEEE Transactions on Robotics}}
  \bibinfo{volume}{25}, \bibinfo{number}{3} (\bibinfo{year}{2009}),
  \bibinfo{pages}{749--754}.
\newblock


\bibitem[\protect\citeauthoryear{Cheng, Zhang, Cheng, Xia, and Zhang}{Cheng
  et~al\mbox{.}}{2022}]%
        {cheng2022path}
\bibfield{author}{\bibinfo{person}{Xiuquan Cheng}, \bibinfo{person}{Shaobo
  Zhang}, \bibinfo{person}{Sizhu Cheng}, \bibinfo{person}{Qinxiang Xia}, {and}
  \bibinfo{person}{Junhao Zhang}.} \bibinfo{year}{2022}\natexlab{}.
\newblock \showarticletitle{Path-Following and Obstacle Avoidance Control of
  Nonholonomic Wheeled Mobile Robot Based on Deep Reinforcement Learning}.
\newblock \bibinfo{journal}{\emph{Applied Sciences}} \bibinfo{volume}{12},
  \bibinfo{number}{14} (\bibinfo{year}{2022}), \bibinfo{pages}{6874}.
\newblock


\bibitem[\protect\citeauthoryear{Choset, Lynch, Hutchinson, Kantor, and
  Burgard}{Choset et~al\mbox{.}}{2005}]%
        {choset2005principles}
\bibfield{author}{\bibinfo{person}{Howie Choset}, \bibinfo{person}{Kevin~M
  Lynch}, \bibinfo{person}{Seth Hutchinson}, \bibinfo{person}{George~A Kantor},
  {and} \bibinfo{person}{Wolfram Burgard}.} \bibinfo{year}{2005}\natexlab{}.
\newblock \bibinfo{booktitle}{\emph{Principles of robot motion: theory,
  algorithms, and implementations}}.
\newblock \bibinfo{publisher}{MIT press}.
\newblock


\bibitem[\protect\citeauthoryear{Codd-Downey, Jenkin, and Allison}{Codd-Downey
  et~al\mbox{.}}{2017}]%
        {codd2017milton}
\bibfield{author}{\bibinfo{person}{Robert Codd-Downey},
  \bibinfo{person}{Michael Jenkin}, {and} \bibinfo{person}{Katherine Allison}.}
  \bibinfo{year}{2017}\natexlab{}.
\newblock \showarticletitle{Milton: An open hardware underwater autonomous
  vehicle}. In \bibinfo{booktitle}{\emph{2017 IEEE International Conference on
  Information and Automation (ICIA)}}. IEEE, \bibinfo{pages}{30--34}.
\newblock


\bibitem[\protect\citeauthoryear{compiled~from Dudek and Jenkin}{compiled~from
  Dudek and Jenkin}{[n. d.]}]%
        {CS-W4733-NOTES-Differential-Drive-Robots}
\bibfield{author}{\bibinfo{person}{compiled~from Dudek} {and}
  \bibinfo{person}{Computational Principles of Mobile~Robotics Jenkin}.}
  \bibinfo{year}{[n. d.]}\natexlab{}.
\newblock \bibinfo{title}{CS W4733 NOTES - Differential Drive Robots}.
\newblock \bibinfo{howpublished}{url =
  {https://www.cs.columbia.edu/~allen/F19/NOTES/icckinematics.pdf}}.
  (\bibinfo{year}{[n. d.]}).
\newblock
\newblock
\shownote{accessed February 27, 2022.}


\bibitem[\protect\citeauthoryear{Concha and Civera}{Concha and Civera}{2014}]%
        {concha2014using}
\bibfield{author}{\bibinfo{person}{Alejo Concha} {and} \bibinfo{person}{Javier
  Civera}.} \bibinfo{year}{2014}\natexlab{}.
\newblock \showarticletitle{Using superpixels in monocular SLAM}. In
  \bibinfo{booktitle}{\emph{2014 IEEE international conference on robotics and
  automation (ICRA)}}. IEEE, \bibinfo{pages}{365--372}.
\newblock


\bibitem[\protect\citeauthoryear{Cotugno, D'Alfonso, Lucia, Muraca, and
  Pugliese}{Cotugno et~al\mbox{.}}{2013}]%
        {cotugno2013extended}
\bibfield{author}{\bibinfo{person}{Giuseppe Cotugno}, \bibinfo{person}{Luigi
  D'Alfonso}, \bibinfo{person}{Walter Lucia}, \bibinfo{person}{Pietro Muraca},
  {and} \bibinfo{person}{Paolo Pugliese}.} \bibinfo{year}{2013}\natexlab{}.
\newblock \showarticletitle{Extended and Unscented Kalman Filters for mobile
  robot localization and environment reconstruction}. In
  \bibinfo{booktitle}{\emph{21st Mediterranean Conference on Control and
  Automation}}. IEEE, \bibinfo{pages}{19--26}.
\newblock


\bibitem[\protect\citeauthoryear{Coulter}{Coulter}{1992}]%
        {coulter1992implementation}
\bibfield{author}{\bibinfo{person}{R~Craig Coulter}.}
  \bibinfo{year}{1992}\natexlab{}.
\newblock \bibinfo{booktitle}{\emph{Implementation of the pure pursuit path
  tracking algorithm}}.
\newblock \bibinfo{type}{{T}echnical {R}eport}.
  \bibinfo{institution}{Carnegie-Mellon UNIV Pittsburgh PA Robotics INST}.
\newblock


\bibitem[\protect\citeauthoryear{Coulter and Mueller}{Coulter and
  Mueller}{1994}]%
        {coulter1994reconfiguration}
\bibfield{author}{\bibinfo{person}{R~Craig Coulter} {and}
  \bibinfo{person}{George~G Mueller}.} \bibinfo{year}{1994}\natexlab{}.
\newblock \bibinfo{booktitle}{\emph{A Reconfiguration Study for the NavLab II
  Mobile Robot}}.
\newblock \bibinfo{type}{{T}echnical {R}eport}.
  \bibinfo{institution}{CARNEGIE-MELLON UNIV PITTSBURGH PA ROBOTICS INST}.
\newblock


\bibitem[\protect\citeauthoryear{Cummins and Newman}{Cummins and
  Newman}{2008}]%
        {cummins2008fab}
\bibfield{author}{\bibinfo{person}{Mark Cummins} {and} \bibinfo{person}{Paul
  Newman}.} \bibinfo{year}{2008}\natexlab{}.
\newblock \showarticletitle{FAB-MAP: Probabilistic localization and mapping in
  the space of appearance}.
\newblock \bibinfo{journal}{\emph{The International Journal of Robotics
  Research}} \bibinfo{volume}{27}, \bibinfo{number}{6} (\bibinfo{year}{2008}),
  \bibinfo{pages}{647--665}.
\newblock


\bibitem[\protect\citeauthoryear{da~Silva and da~Cruz}{da~Silva and
  da~Cruz}{2016}]%
        {da2016fuzzy}
\bibfield{author}{\bibinfo{person}{Andr{\'e}~Lu{\'\i}s da Silva} {and}
  \bibinfo{person}{Jos{\'e}~Jaime da Cruz}.} \bibinfo{year}{2016}\natexlab{}.
\newblock \showarticletitle{Fuzzy adaptive extended Kalman filter for UAV
  INS/GPS data fusion}.
\newblock \bibinfo{journal}{\emph{Journal of the Brazilian Society of
  Mechanical Sciences and Engineering}} \bibinfo{volume}{38},
  \bibinfo{number}{6} (\bibinfo{year}{2016}), \bibinfo{pages}{1671--1688}.
\newblock


\bibitem[\protect\citeauthoryear{D'Alfonso, Grano, Muraca, and
  Pugliese}{D'Alfonso et~al\mbox{.}}{2013}]%
        {d2013polynomial}
\bibfield{author}{\bibinfo{person}{Luigi D'Alfonso}, \bibinfo{person}{Antonio
  Grano}, \bibinfo{person}{Pietro Muraca}, {and} \bibinfo{person}{Paolo
  Pugliese}.} \bibinfo{year}{2013}\natexlab{}.
\newblock \showarticletitle{A polynomial based SLAM algorithm for mobile robots
  using ultrasonic sensors-Experimental results}. In
  \bibinfo{booktitle}{\emph{2013 16th International Conference on Advanced
  Robotics (ICAR)}}. IEEE, \bibinfo{pages}{1--6}.
\newblock


\bibitem[\protect\citeauthoryear{Davison}{Davison}{2003}]%
        {davison2003real}
\bibfield{author}{\bibinfo{person}{Andrew~J Davison}.}
  \bibinfo{year}{2003}\natexlab{}.
\newblock \showarticletitle{Real-time simultaneous localisation and mapping
  with a single camera}. In \bibinfo{booktitle}{\emph{Computer Vision, IEEE
  International Conference on}}, Vol.~\bibinfo{volume}{3}. IEEE Computer
  Society, \bibinfo{pages}{1403--1403}.
\newblock


\bibitem[\protect\citeauthoryear{Davison, Cid, and Kita}{Davison
  et~al\mbox{.}}{2004}]%
        {davison2004real}
\bibfield{author}{\bibinfo{person}{Andrew~J Davison},
  \bibinfo{person}{Yolanda~Gonzalez Cid}, {and} \bibinfo{person}{Nobuyuki
  Kita}.} \bibinfo{year}{2004}\natexlab{}.
\newblock \showarticletitle{Real-time 3D SLAM with wide-angle vision}.
\newblock \bibinfo{journal}{\emph{IFAC Proceedings Volumes}}
  \bibinfo{volume}{37}, \bibinfo{number}{8} (\bibinfo{year}{2004}),
  \bibinfo{pages}{868--873}.
\newblock


\bibitem[\protect\citeauthoryear{Davison, Reid, Molton, and Stasse}{Davison
  et~al\mbox{.}}{2007}]%
        {davison2007monoslam}
\bibfield{author}{\bibinfo{person}{Andrew~J Davison}, \bibinfo{person}{Ian~D
  Reid}, \bibinfo{person}{Nicholas~D Molton}, {and} \bibinfo{person}{Olivier
  Stasse}.} \bibinfo{year}{2007}\natexlab{}.
\newblock \showarticletitle{MonoSLAM: Real-time single camera SLAM}.
\newblock \bibinfo{journal}{\emph{IEEE transactions on pattern analysis and
  machine intelligence}} \bibinfo{volume}{29}, \bibinfo{number}{6}
  (\bibinfo{year}{2007}), \bibinfo{pages}{1052--1067}.
\newblock


\bibitem[\protect\citeauthoryear{Deng, Zhang, Sintov, Huang, and Bretl}{Deng
  et~al\mbox{.}}{2018}]%
        {deng2018feature}
\bibfield{author}{\bibinfo{person}{Xinke Deng}, \bibinfo{person}{Zixu Zhang},
  \bibinfo{person}{Avishai Sintov}, \bibinfo{person}{Jing Huang}, {and}
  \bibinfo{person}{Timothy Bretl}.} \bibinfo{year}{2018}\natexlab{}.
\newblock \showarticletitle{Feature-constrained active visual SLAM for mobile
  robot navigation}. In \bibinfo{booktitle}{\emph{2018 IEEE international
  conference on robotics and automation (ICRA)}}. IEEE,
  \bibinfo{pages}{7233--7238}.
\newblock


\bibitem[\protect\citeauthoryear{DeSouza and Kak}{DeSouza and Kak}{2002}]%
        {desouza2002vision}
\bibfield{author}{\bibinfo{person}{Guilherme~N DeSouza} {and}
  \bibinfo{person}{Avinash~C Kak}.} \bibinfo{year}{2002}\natexlab{}.
\newblock \showarticletitle{Vision for mobile robot navigation: A survey}.
\newblock \bibinfo{journal}{\emph{IEEE transactions on pattern analysis and
  machine intelligence}} \bibinfo{volume}{24}, \bibinfo{number}{2}
  (\bibinfo{year}{2002}), \bibinfo{pages}{237--267}.
\newblock


\bibitem[\protect\citeauthoryear{Di~Giammarino, Brizi, Guadagnino, Stachniss,
  and Grisetti}{Di~Giammarino et~al\mbox{.}}{2022}]%
        {di2022md}
\bibfield{author}{\bibinfo{person}{Luca Di~Giammarino},
  \bibinfo{person}{Leonardo Brizi}, \bibinfo{person}{Tiziano Guadagnino},
  \bibinfo{person}{Cyrill Stachniss}, {and} \bibinfo{person}{Giorgio
  Grisetti}.} \bibinfo{year}{2022}\natexlab{}.
\newblock \showarticletitle{MD-SLAM: Multi-cue Direct SLAM}.
\newblock \bibinfo{journal}{\emph{arXiv preprint arXiv:2203.13237}}
  (\bibinfo{year}{2022}).
\newblock


\bibitem[\protect\citeauthoryear{Di~Lallo, Murphy, Krieger, Zhu, Taylor, and
  Su}{Di~Lallo et~al\mbox{.}}{2021}]%
        {di2021medical}
\bibfield{author}{\bibinfo{person}{Antonio Di~Lallo}, \bibinfo{person}{Robin
  Murphy}, \bibinfo{person}{Axel Krieger}, \bibinfo{person}{Junxi Zhu},
  \bibinfo{person}{Russell~H Taylor}, {and} \bibinfo{person}{Hao Su}.}
  \bibinfo{year}{2021}\natexlab{}.
\newblock \showarticletitle{Medical robots for infectious diseases: lessons and
  challenges from the COVID-19 pandemic}.
\newblock \bibinfo{journal}{\emph{IEEE Robotics \& Automation Magazine}}
  \bibinfo{volume}{28}, \bibinfo{number}{1} (\bibinfo{year}{2021}),
  \bibinfo{pages}{18--27}.
\newblock


\bibitem[\protect\citeauthoryear{Dickmanns}{Dickmanns}{2007}]%
        {dickmanns2007dynamic}
\bibfield{author}{\bibinfo{person}{Ernst~Dieter Dickmanns}.}
  \bibinfo{year}{2007}\natexlab{}.
\newblock \bibinfo{booktitle}{\emph{Dynamic vision for perception and control
  of motion}}.
\newblock \bibinfo{publisher}{Springer Science \& Business Media}.
\newblock


\bibitem[\protect\citeauthoryear{Doucet, De~Freitas, Gordon,
  et~al\mbox{.}}{Doucet et~al\mbox{.}}{2001}]%
        {doucet2001sequential}
\bibfield{author}{\bibinfo{person}{Arnaud Doucet}, \bibinfo{person}{Nando
  De~Freitas}, \bibinfo{person}{Neil~James Gordon}, {et~al\mbox{.}}}
  \bibinfo{year}{2001}\natexlab{}.
\newblock \bibinfo{booktitle}{\emph{Sequential Monte Carlo methods in
  practice}}. Vol.~\bibinfo{volume}{1}.
\newblock \bibinfo{publisher}{Springer}.
\newblock


\bibitem[\protect\citeauthoryear{Duchon, Kaelbling, and Warren}{Duchon
  et~al\mbox{.}}{1998}]%
        {duchon1998ecological}
\bibfield{author}{\bibinfo{person}{Andrew~P Duchon},
  \bibinfo{person}{Leslie~Pack Kaelbling}, {and} \bibinfo{person}{William~H
  Warren}.} \bibinfo{year}{1998}\natexlab{}.
\newblock \showarticletitle{Ecological robotics}.
\newblock \bibinfo{journal}{\emph{Adaptive Behavior}} \bibinfo{volume}{6},
  \bibinfo{number}{3-4} (\bibinfo{year}{1998}), \bibinfo{pages}{473--507}.
\newblock


\bibitem[\protect\citeauthoryear{Eade and Drummond}{Eade and Drummond}{2009}]%
        {eade2009edge}
\bibfield{author}{\bibinfo{person}{Ethan Eade} {and} \bibinfo{person}{Tom
  Drummond}.} \bibinfo{year}{2009}\natexlab{}.
\newblock \showarticletitle{Edge landmarks in monocular SLAM}.
\newblock \bibinfo{journal}{\emph{Image and Vision Computing}}
  \bibinfo{volume}{27}, \bibinfo{number}{5} (\bibinfo{year}{2009}),
  \bibinfo{pages}{588--596}.
\newblock


\bibitem[\protect\citeauthoryear{Earnest}{Earnest}{2012}]%
        {stanford-cart}
\bibfield{author}{\bibinfo{person}{Les Earnest}.}
  \bibinfo{year}{2012}\natexlab{}.
\newblock \bibinfo{title}{Stanford Cart}.
\newblock \bibinfo{howpublished}{url =
  {https://web.stanford.edu/~learnest/sail/oldcart.html}}.
  (\bibinfo{year}{2012}).
\newblock
\newblock
\shownote{accessed February 27, 2022.}


\bibitem[\protect\citeauthoryear{Edan}{Edan}{1995}]%
        {edan1995design}
\bibfield{author}{\bibinfo{person}{Yael Edan}.}
  \bibinfo{year}{1995}\natexlab{}.
\newblock \showarticletitle{Design of an autonomous agricultural robot}.
\newblock \bibinfo{journal}{\emph{Applied Intelligence}} \bibinfo{volume}{5},
  \bibinfo{number}{1} (\bibinfo{year}{1995}), \bibinfo{pages}{41--50}.
\newblock


\bibitem[\protect\citeauthoryear{Einhorn, Schr{\"o}ter, and Gross}{Einhorn
  et~al\mbox{.}}{2010}]%
        {einhorn2010building}
\bibfield{author}{\bibinfo{person}{Erik Einhorn}, \bibinfo{person}{Christof
  Schr{\"o}ter}, {and} \bibinfo{person}{Horst-Michael Gross}.}
  \bibinfo{year}{2010}\natexlab{}.
\newblock \showarticletitle{Building 2d and 3d adaptiveresolution occupancy
  maps using nd-trees}.
\newblock \bibinfo{journal}{\emph{Proceeding 55th Int. Scientific Colloquiium,
  Ilmenau, Germany. Verlag ISLE}} (\bibinfo{year}{2010}),
  \bibinfo{pages}{306--311}.
\newblock


\bibitem[\protect\citeauthoryear{Elfes}{Elfes}{1989}]%
        {elfes1989using}
\bibfield{author}{\bibinfo{person}{Alberto Elfes}.}
  \bibinfo{year}{1989}\natexlab{}.
\newblock \showarticletitle{Using occupancy grids for mobile robot perception
  and navigation}.
\newblock \bibinfo{journal}{\emph{Computer}} \bibinfo{volume}{22},
  \bibinfo{number}{6} (\bibinfo{year}{1989}), \bibinfo{pages}{46--57}.
\newblock


\bibitem[\protect\citeauthoryear{Elfring, Torta, and van~de Molengraft}{Elfring
  et~al\mbox{.}}{2021}]%
        {elfring2021particle}
\bibfield{author}{\bibinfo{person}{Jos Elfring}, \bibinfo{person}{Elena Torta},
  {and} \bibinfo{person}{Ren{\'e} van~de Molengraft}.}
  \bibinfo{year}{2021}\natexlab{}.
\newblock \showarticletitle{Particle filters: A hands-on tutorial}.
\newblock \bibinfo{journal}{\emph{Sensors}} \bibinfo{volume}{21},
  \bibinfo{number}{2} (\bibinfo{year}{2021}), \bibinfo{pages}{438}.
\newblock


\bibitem[\protect\citeauthoryear{Engel, Sch{\"o}ps, and Cremers}{Engel
  et~al\mbox{.}}{2014}]%
        {engel2014lsd}
\bibfield{author}{\bibinfo{person}{Jakob Engel}, \bibinfo{person}{Thomas
  Sch{\"o}ps}, {and} \bibinfo{person}{Daniel Cremers}.}
  \bibinfo{year}{2014}\natexlab{}.
\newblock \showarticletitle{LSD-SLAM: Large-scale direct monocular SLAM}. In
  \bibinfo{booktitle}{\emph{European conference on computer vision}}. Springer,
  \bibinfo{pages}{834--849}.
\newblock


\bibitem[\protect\citeauthoryear{Engel, St{\"u}ckler, and Cremers}{Engel
  et~al\mbox{.}}{2015}]%
        {engel2015large}
\bibfield{author}{\bibinfo{person}{Jakob Engel}, \bibinfo{person}{J{\"o}rg
  St{\"u}ckler}, {and} \bibinfo{person}{Daniel Cremers}.}
  \bibinfo{year}{2015}\natexlab{}.
\newblock \showarticletitle{Large-scale direct SLAM with stereo cameras}. In
  \bibinfo{booktitle}{\emph{2015 IEEE/RSJ international conference on
  intelligent robots and systems (IROS)}}. IEEE, \bibinfo{pages}{1935--1942}.
\newblock


\bibitem[\protect\citeauthoryear{Fallon, Kaess, Johannsson, and Leonard}{Fallon
  et~al\mbox{.}}{2011}]%
        {fallon2011efficient}
\bibfield{author}{\bibinfo{person}{Maurice~F Fallon}, \bibinfo{person}{Michael
  Kaess}, \bibinfo{person}{Hordur Johannsson}, {and} \bibinfo{person}{John~J
  Leonard}.} \bibinfo{year}{2011}\natexlab{}.
\newblock \showarticletitle{Efficient AUV navigation fusing acoustic ranging
  and side-scan sonar}. In \bibinfo{booktitle}{\emph{2011 IEEE International
  Conference on Robotics and Automation}}. IEEE, \bibinfo{pages}{2398--2405}.
\newblock


\bibitem[\protect\citeauthoryear{Fan, Cheng, Pan, Manocha, and Yang}{Fan
  et~al\mbox{.}}{2018}]%
        {fan2018crowdmove}
\bibfield{author}{\bibinfo{person}{Tingxiang Fan}, \bibinfo{person}{Xinjing
  Cheng}, \bibinfo{person}{Jia Pan}, \bibinfo{person}{Dinesh Manocha}, {and}
  \bibinfo{person}{Ruigang Yang}.} \bibinfo{year}{2018}\natexlab{}.
\newblock \showarticletitle{Crowdmove: Autonomous mapless navigation in crowded
  scenarios}.
\newblock \bibinfo{journal}{\emph{arXiv preprint arXiv:1807.07870}}
  (\bibinfo{year}{2018}).
\newblock


\bibitem[\protect\citeauthoryear{Ferrera, Eudes, Moras, Sanfourche, and
  Le~Besnerais}{Ferrera et~al\mbox{.}}{2021}]%
        {ferrera2021ov}
\bibfield{author}{\bibinfo{person}{Maxime Ferrera}, \bibinfo{person}{Alexandre
  Eudes}, \bibinfo{person}{Julien Moras}, \bibinfo{person}{Martial Sanfourche},
  {and} \bibinfo{person}{Guy Le~Besnerais}.} \bibinfo{year}{2021}\natexlab{}.
\newblock \showarticletitle{OV2SLAM: A Fully Online and Versatile Visual SLAM
  for Real-Time Applications}.
\newblock \bibinfo{journal}{\emph{IEEE Robotics and Automation Letters}}
  \bibinfo{volume}{6}, \bibinfo{number}{2} (\bibinfo{year}{2021}),
  \bibinfo{pages}{1399--1406}.
\newblock


\bibitem[\protect\citeauthoryear{Fisher, Cannizzaro, Cochrane, Nagahawatte, and
  Palmer}{Fisher et~al\mbox{.}}{2021}]%
        {fisher2021colmap}
\bibfield{author}{\bibinfo{person}{Alex Fisher}, \bibinfo{person}{Ricardo
  Cannizzaro}, \bibinfo{person}{Madeleine Cochrane}, \bibinfo{person}{Chatura
  Nagahawatte}, {and} \bibinfo{person}{Jennifer~L Palmer}.}
  \bibinfo{year}{2021}\natexlab{}.
\newblock \showarticletitle{ColMap: A memory-efficient occupancy grid mapping
  framework}.
\newblock \bibinfo{journal}{\emph{Robotics and Autonomous Systems}}
  \bibinfo{volume}{142} (\bibinfo{year}{2021}), \bibinfo{pages}{103755}.
\newblock


\bibitem[\protect\citeauthoryear{Foglia and Reina}{Foglia and Reina}{2006}]%
        {foglia2006agricultural}
\bibfield{author}{\bibinfo{person}{Mario~M Foglia} {and}
  \bibinfo{person}{Giulio Reina}.} \bibinfo{year}{2006}\natexlab{}.
\newblock \showarticletitle{Agricultural robot for radicchio harvesting}.
\newblock \bibinfo{journal}{\emph{Journal of Field Robotics}}
  \bibinfo{volume}{23}, \bibinfo{number}{6-7} (\bibinfo{year}{2006}),
  \bibinfo{pages}{363--377}.
\newblock


\bibitem[\protect\citeauthoryear{Fox, Burgard, and Thrun}{Fox
  et~al\mbox{.}}{1997}]%
        {fox1997dynamic}
\bibfield{author}{\bibinfo{person}{Dieter Fox}, \bibinfo{person}{Wolfram
  Burgard}, {and} \bibinfo{person}{Sebastian Thrun}.}
  \bibinfo{year}{1997}\natexlab{}.
\newblock \showarticletitle{The dynamic window approach to collision
  avoidance}.
\newblock \bibinfo{journal}{\emph{IEEE Robotics \& Automation Magazine}}
  \bibinfo{volume}{4}, \bibinfo{number}{1} (\bibinfo{year}{1997}),
  \bibinfo{pages}{23--33}.
\newblock


\bibitem[\protect\citeauthoryear{Fragapane, De~Koster, Sgarbossa, and
  Strandhagen}{Fragapane et~al\mbox{.}}{2021}]%
        {fragapane2021planning}
\bibfield{author}{\bibinfo{person}{Giuseppe Fragapane}, \bibinfo{person}{Rene
  De~Koster}, \bibinfo{person}{Fabio Sgarbossa}, {and} \bibinfo{person}{Jan~Ola
  Strandhagen}.} \bibinfo{year}{2021}\natexlab{}.
\newblock \showarticletitle{Planning and control of autonomous mobile robots
  for intralogistics: Literature review and research agenda}.
\newblock \bibinfo{journal}{\emph{European Journal of Operational Research}}
  \bibinfo{volume}{294}, \bibinfo{number}{2} (\bibinfo{year}{2021}),
  \bibinfo{pages}{405--426}.
\newblock


\bibitem[\protect\citeauthoryear{Furgale, Barfoot, and Sibley}{Furgale
  et~al\mbox{.}}{2012}]%
        {furgale2012continuous}
\bibfield{author}{\bibinfo{person}{Paul Furgale}, \bibinfo{person}{Timothy~D
  Barfoot}, {and} \bibinfo{person}{Gabe Sibley}.}
  \bibinfo{year}{2012}\natexlab{}.
\newblock \showarticletitle{Continuous-time batch estimation using temporal
  basis functions}. In \bibinfo{booktitle}{\emph{2012 IEEE International
  Conference on Robotics and Automation}}. IEEE, \bibinfo{pages}{2088--2095}.
\newblock


\bibitem[\protect\citeauthoryear{Furgale, Rehder, and Siegwart}{Furgale
  et~al\mbox{.}}{2013}]%
        {furgale2013unified}
\bibfield{author}{\bibinfo{person}{Paul Furgale}, \bibinfo{person}{Joern
  Rehder}, {and} \bibinfo{person}{Roland Siegwart}.}
  \bibinfo{year}{2013}\natexlab{}.
\newblock \showarticletitle{Unified temporal and spatial calibration for
  multi-sensor systems}. In \bibinfo{booktitle}{\emph{2013 IEEE/RSJ
  International Conference on Intelligent Robots and Systems}}. IEEE,
  \bibinfo{pages}{1280--1286}.
\newblock


\bibitem[\protect\citeauthoryear{Furrer, Burri, Achtelik, and Siegwart}{Furrer
  et~al\mbox{.}}{2016}]%
        {Furrer2016}
\bibfield{author}{\bibinfo{person}{Fadri Furrer}, \bibinfo{person}{Michael
  Burri}, \bibinfo{person}{Markus Achtelik}, {and} \bibinfo{person}{Roland
  Siegwart}.} \bibinfo{year}{2016}\natexlab{}.
\newblock \bibinfo{booktitle}{\emph{Robot Operating System (ROS): The Complete
  Reference (Volume 1)}}.
\newblock \bibinfo{publisher}{Springer International Publishing},
  \bibinfo{address}{Cham}, Chapter RotorS---A Modular Gazebo MAV Simulator
  Framework, \bibinfo{pages}{595--625}.
\newblock
\showISBNx{978-3-319-26054-9}
\urldef\tempurl%
\url{https://doi.org/10.1007/978-3-319-26054-9_23}
\showDOI{\tempurl}


\bibitem[\protect\citeauthoryear{Gabriely and Rimon}{Gabriely and
  Rimon}{2008}]%
        {gabriely2008cbug}
\bibfield{author}{\bibinfo{person}{Yoav Gabriely} {and} \bibinfo{person}{Elon
  Rimon}.} \bibinfo{year}{2008}\natexlab{}.
\newblock \showarticletitle{Cbug: A quadratically competitive mobile robot
  navigation algorithm}.
\newblock \bibinfo{journal}{\emph{IEEE Transactions on Robotics}}
  \bibinfo{volume}{24}, \bibinfo{number}{6} (\bibinfo{year}{2008}),
  \bibinfo{pages}{1451--1457}.
\newblock


\bibitem[\protect\citeauthoryear{Garcia-Fidalgo and Ortiz}{Garcia-Fidalgo and
  Ortiz}{2018}]%
        {garcia2018ibow}
\bibfield{author}{\bibinfo{person}{Emilio Garcia-Fidalgo} {and}
  \bibinfo{person}{Alberto Ortiz}.} \bibinfo{year}{2018}\natexlab{}.
\newblock \showarticletitle{ibow-lcd: An appearance-based loop-closure
  detection approach using incremental bags of binary words}.
\newblock \bibinfo{journal}{\emph{IEEE Robotics and Automation Letters}}
  \bibinfo{volume}{3}, \bibinfo{number}{4} (\bibinfo{year}{2018}),
  \bibinfo{pages}{3051--3057}.
\newblock


\bibitem[\protect\citeauthoryear{Gaussier, Joulain, Zrehen, Banquet, and
  Revel}{Gaussier et~al\mbox{.}}{1997}]%
        {gaussier1997visual}
\bibfield{author}{\bibinfo{person}{Philippe Gaussier},
  \bibinfo{person}{C{\'e}dric Joulain}, \bibinfo{person}{St{\'e}phane Zrehen},
  \bibinfo{person}{Jean-Paul Banquet}, {and} \bibinfo{person}{Arnaud Revel}.}
  \bibinfo{year}{1997}\natexlab{}.
\newblock \showarticletitle{Visual navigation in an open environment without
  map}. In \bibinfo{booktitle}{\emph{Proceedings of the 1997 IEEE/RSJ
  International Conference on Intelligent Robot and Systems. Innovative
  Robotics for Real-World Applications. IROS'97}}, Vol.~\bibinfo{volume}{2}.
  IEEE, \bibinfo{pages}{545--550}.
\newblock


\bibitem[\protect\citeauthoryear{Geiger, Lenz, Stiller, and Urtasun}{Geiger
  et~al\mbox{.}}{2013}]%
        {geiger2013vision}
\bibfield{author}{\bibinfo{person}{Andreas Geiger}, \bibinfo{person}{Philip
  Lenz}, \bibinfo{person}{Christoph Stiller}, {and} \bibinfo{person}{Raquel
  Urtasun}.} \bibinfo{year}{2013}\natexlab{}.
\newblock \showarticletitle{Vision meets robotics: The kitti dataset}.
\newblock \bibinfo{journal}{\emph{The International Journal of Robotics
  Research}} \bibinfo{volume}{32}, \bibinfo{number}{11} (\bibinfo{year}{2013}),
  \bibinfo{pages}{1231--1237}.
\newblock


\bibitem[\protect\citeauthoryear{Geneva, Eckenhoff, Lee, Yang, and
  Huang}{Geneva et~al\mbox{.}}{2020}]%
        {geneva2020openvins}
\bibfield{author}{\bibinfo{person}{Patrick Geneva}, \bibinfo{person}{Kevin
  Eckenhoff}, \bibinfo{person}{Woosik Lee}, \bibinfo{person}{Yulin Yang}, {and}
  \bibinfo{person}{Guoquan Huang}.} \bibinfo{year}{2020}\natexlab{}.
\newblock \showarticletitle{Openvins: A research platform for visual-inertial
  estimation}. In \bibinfo{booktitle}{\emph{2020 IEEE International Conference
  on Robotics and Automation (ICRA)}}. IEEE, \bibinfo{pages}{4666--4672}.
\newblock


\bibitem[\protect\citeauthoryear{Ghaffarivardavagh, Afzal, Rodriguez, and
  Adib}{Ghaffarivardavagh et~al\mbox{.}}{2020}]%
        {ghaffarivardavagh2020underwater}
\bibfield{author}{\bibinfo{person}{Reza Ghaffarivardavagh},
  \bibinfo{person}{Sayed~Saad Afzal}, \bibinfo{person}{Osvy Rodriguez}, {and}
  \bibinfo{person}{Fadel Adib}.} \bibinfo{year}{2020}\natexlab{}.
\newblock \showarticletitle{Underwater backscatter localization: Toward a
  battery-free underwater GPS}. In \bibinfo{booktitle}{\emph{Proceedings of the
  19th ACM Workshop on Hot Topics in Networks}}. \bibinfo{pages}{125--131}.
\newblock


\bibitem[\protect\citeauthoryear{Gilks and Berzuini}{Gilks and
  Berzuini}{2001}]%
        {gilks2001following}
\bibfield{author}{\bibinfo{person}{Walter~R Gilks} {and} \bibinfo{person}{Carlo
  Berzuini}.} \bibinfo{year}{2001}\natexlab{}.
\newblock \showarticletitle{Following a moving target—Monte Carlo inference
  for dynamic Bayesian models}.
\newblock \bibinfo{journal}{\emph{Journal of the Royal Statistical Society:
  Series B (Statistical Methodology)}} \bibinfo{volume}{63},
  \bibinfo{number}{1} (\bibinfo{year}{2001}), \bibinfo{pages}{127--146}.
\newblock


\bibitem[\protect\citeauthoryear{Gonz{\'a}lez-Garc{\'\i}a, G{\'o}mez-Espinosa,
  Cuan-Urquizo, Garc{\'\i}a-Valdovinos, Salgado-Jim{\'e}nez, and
  Escobedo~Cabello}{Gonz{\'a}lez-Garc{\'\i}a et~al\mbox{.}}{2020}]%
        {gonzalez2020autonomous}
\bibfield{author}{\bibinfo{person}{Josu{\'e} Gonz{\'a}lez-Garc{\'\i}a},
  \bibinfo{person}{Alfonso G{\'o}mez-Espinosa}, \bibinfo{person}{Enrique
  Cuan-Urquizo}, \bibinfo{person}{Luis~Govinda Garc{\'\i}a-Valdovinos},
  \bibinfo{person}{Tom{\'a}s Salgado-Jim{\'e}nez}, {and}
  \bibinfo{person}{Jesus~Arturo Escobedo~Cabello}.}
  \bibinfo{year}{2020}\natexlab{}.
\newblock \showarticletitle{Autonomous underwater vehicles: Localization,
  navigation, and communication for collaborative missions}.
\newblock \bibinfo{journal}{\emph{Applied sciences}} \bibinfo{volume}{10},
  \bibinfo{number}{4} (\bibinfo{year}{2020}), \bibinfo{pages}{1256}.
\newblock


\bibitem[\protect\citeauthoryear{Goodfellow, Pouget-Abadie, Mirza, Xu,
  Warde-Farley, Ozair, Courville, and Bengio}{Goodfellow et~al\mbox{.}}{2014}]%
        {goodfellow2014generative}
\bibfield{author}{\bibinfo{person}{Ian Goodfellow}, \bibinfo{person}{Jean
  Pouget-Abadie}, \bibinfo{person}{Mehdi Mirza}, \bibinfo{person}{Bing Xu},
  \bibinfo{person}{David Warde-Farley}, \bibinfo{person}{Sherjil Ozair},
  \bibinfo{person}{Aaron Courville}, {and} \bibinfo{person}{Yoshua Bengio}.}
  \bibinfo{year}{2014}\natexlab{}.
\newblock \showarticletitle{Generative adversarial nets}.
\newblock \bibinfo{journal}{\emph{Advances in neural information processing
  systems}}  \bibinfo{volume}{27} (\bibinfo{year}{2014}).
\newblock


\bibitem[\protect\citeauthoryear{Gordon, Salmond, and Smith}{Gordon
  et~al\mbox{.}}{1993}]%
        {gordon1993novel}
\bibfield{author}{\bibinfo{person}{Neil~J Gordon}, \bibinfo{person}{David~J
  Salmond}, {and} \bibinfo{person}{Adrian~FM Smith}.}
  \bibinfo{year}{1993}\natexlab{}.
\newblock \showarticletitle{Novel approach to nonlinear/non-Gaussian Bayesian
  state estimation}. In \bibinfo{booktitle}{\emph{IEE Proceedings F-radar and
  signal processing}}, Vol.~\bibinfo{volume}{140}. IET,
  \bibinfo{pages}{107--113}.
\newblock


\bibitem[\protect\citeauthoryear{Grisetti, K{\"u}mmerle, Stachniss, and
  Burgard}{Grisetti et~al\mbox{.}}{2010}]%
        {grisetti2010tutorial}
\bibfield{author}{\bibinfo{person}{Giorgio Grisetti}, \bibinfo{person}{Rainer
  K{\"u}mmerle}, \bibinfo{person}{Cyrill Stachniss}, {and}
  \bibinfo{person}{Wolfram Burgard}.} \bibinfo{year}{2010}\natexlab{}.
\newblock \showarticletitle{A tutorial on graph-based SLAM}.
\newblock \bibinfo{journal}{\emph{IEEE Intelligent Transportation Systems
  Magazine}} \bibinfo{volume}{2}, \bibinfo{number}{4} (\bibinfo{year}{2010}),
  \bibinfo{pages}{31--43}.
\newblock


\bibitem[\protect\citeauthoryear{Grisetti, Stachniss, and Burgard}{Grisetti
  et~al\mbox{.}}{2007}]%
        {grisetti2007improved}
\bibfield{author}{\bibinfo{person}{Giorgio Grisetti}, \bibinfo{person}{Cyrill
  Stachniss}, {and} \bibinfo{person}{Wolfram Burgard}.}
  \bibinfo{year}{2007}\natexlab{}.
\newblock \showarticletitle{Improved techniques for grid mapping with
  rao-blackwellized particle filters}.
\newblock \bibinfo{journal}{\emph{IEEE transactions on Robotics}}
  \bibinfo{volume}{23}, \bibinfo{number}{1} (\bibinfo{year}{2007}),
  \bibinfo{pages}{34--46}.
\newblock


\bibitem[\protect\citeauthoryear{Gro{\ss}, Bonani, Mondada, and
  Dorigo}{Gro{\ss} et~al\mbox{.}}{2006}]%
        {Gro-etal:amire2005}
\bibfield{author}{\bibinfo{person}{R. Gro{\ss}}, \bibinfo{person}{M. Bonani},
  \bibinfo{person}{F. Mondada}, {and} \bibinfo{person}{M. Dorigo}.}
  \bibinfo{year}{2006}\natexlab{}.
\newblock \showarticletitle{Autonomous Self-assembly in a Swarm-bot}. In
  \bibinfo{booktitle}{\emph{Proc. of the 3rd Int. Symp. on Autonomous
  Minirobots for Research and Edutainment (AMiRE 2005)}},
  \bibfield{editor}{\bibinfo{person}{K.~Murase}, \bibinfo{person}{K.~Sekiyama},
  \bibinfo{person}{N.~Kubota}, \bibinfo{person}{T.~Naniwa}, {and}
  \bibinfo{person}{J.~Sitte}} (Eds.). \bibinfo{publisher}{Springer, Berlin,
  Germany}, \bibinfo{pages}{314--322}.
\newblock


\bibitem[\protect\citeauthoryear{Guo, Rong, Wang, and Wu}{Guo
  et~al\mbox{.}}{2022}]%
        {guo2022lidar}
\bibfield{author}{\bibinfo{person}{Shiyi Guo}, \bibinfo{person}{Zheng Rong},
  \bibinfo{person}{Shuo Wang}, {and} \bibinfo{person}{Yihong Wu}.}
  \bibinfo{year}{2022}\natexlab{}.
\newblock \showarticletitle{A LiDAR SLAM with PCA-based Feature Extraction and
  Two-stage Matching}.
\newblock \bibinfo{journal}{\emph{IEEE Transactions on Instrumentation and
  Measurement}} (\bibinfo{year}{2022}).
\newblock


\bibitem[\protect\citeauthoryear{Gustafsson, Gunnarsson, Bergman, Forssell,
  Jansson, Karlsson, and Nordlund}{Gustafsson et~al\mbox{.}}{2002}]%
        {gustafsson2002particle}
\bibfield{author}{\bibinfo{person}{Fredrik Gustafsson},
  \bibinfo{person}{Fredrik Gunnarsson}, \bibinfo{person}{Niclas Bergman},
  \bibinfo{person}{Urban Forssell}, \bibinfo{person}{Jonas Jansson},
  \bibinfo{person}{Rickard Karlsson}, {and} \bibinfo{person}{P-J Nordlund}.}
  \bibinfo{year}{2002}\natexlab{}.
\newblock \showarticletitle{Particle filters for positioning, navigation, and
  tracking}.
\newblock \bibinfo{journal}{\emph{IEEE Transactions on signal processing}}
  \bibinfo{volume}{50}, \bibinfo{number}{2} (\bibinfo{year}{2002}),
  \bibinfo{pages}{425--437}.
\newblock


\bibitem[\protect\citeauthoryear{G{\"u}zel}{G{\"u}zel}{2013}]%
        {guzel2013autonomous}
\bibfield{author}{\bibinfo{person}{Mehmet~Serdar G{\"u}zel}.}
  \bibinfo{year}{2013}\natexlab{}.
\newblock \showarticletitle{Autonomous vehicle navigation using vision and
  mapless strategies: a survey}.
\newblock \bibinfo{journal}{\emph{Advances in Mechanical Engineering}}
  \bibinfo{volume}{5} (\bibinfo{year}{2013}), \bibinfo{pages}{234747}.
\newblock


\bibitem[\protect\citeauthoryear{Guzm{\'a}n, Hayet, and Klette}{Guzm{\'a}n
  et~al\mbox{.}}{2015}]%
        {guzman2015towards}
\bibfield{author}{\bibinfo{person}{Roberto Guzm{\'a}n},
  \bibinfo{person}{Jean-Bernard Hayet}, {and} \bibinfo{person}{Reinhard
  Klette}.} \bibinfo{year}{2015}\natexlab{}.
\newblock \showarticletitle{Towards ubiquitous autonomous driving: The CCSAD
  dataset}. In \bibinfo{booktitle}{\emph{International Conference on Computer
  Analysis of Images and Patterns}}. Springer, \bibinfo{pages}{582--593}.
\newblock


\bibitem[\protect\citeauthoryear{Ha, Yen, and Balaguer}{Ha
  et~al\mbox{.}}{2019}]%
        {ha2019robotic}
\bibfield{author}{\bibinfo{person}{QP Ha}, \bibinfo{person}{L Yen}, {and}
  \bibinfo{person}{C Balaguer}.} \bibinfo{year}{2019}\natexlab{}.
\newblock \showarticletitle{Robotic autonomous systems for earthmoving in
  military applications}.
\newblock \bibinfo{journal}{\emph{Automation in Construction}}
  \bibinfo{volume}{107} (\bibinfo{year}{2019}), \bibinfo{pages}{102934}.
\newblock


\bibitem[\protect\citeauthoryear{Hahnel, Burgard, Fox, and Thrun}{Hahnel
  et~al\mbox{.}}{2003}]%
        {hahnel2003efficient}
\bibfield{author}{\bibinfo{person}{Dirk Hahnel}, \bibinfo{person}{Wolfram
  Burgard}, \bibinfo{person}{Dieter Fox}, {and} \bibinfo{person}{Sebastian
  Thrun}.} \bibinfo{year}{2003}\natexlab{}.
\newblock \showarticletitle{An efficient FastSLAM algorithm for generating maps
  of large-scale cyclic environments from raw laser range measurements}. In
  \bibinfo{booktitle}{\emph{Proceedings 2003 IEEE/RSJ International Conference
  on Intelligent Robots and Systems (IROS 2003)(Cat. No. 03CH37453)}},
  Vol.~\bibinfo{volume}{1}. IEEE, \bibinfo{pages}{206--211}.
\newblock


\bibitem[\protect\citeauthoryear{H{\"a}hnel, Thrun, Wegbreit, and
  Burgard}{H{\"a}hnel et~al\mbox{.}}{2005}]%
        {hahnel2005towards}
\bibfield{author}{\bibinfo{person}{Dirk H{\"a}hnel}, \bibinfo{person}{Sebastian
  Thrun}, \bibinfo{person}{Ben Wegbreit}, {and} \bibinfo{person}{Wolfram
  Burgard}.} \bibinfo{year}{2005}\natexlab{}.
\newblock \showarticletitle{Towards lazy data association in SLAM}. In
  \bibinfo{booktitle}{\emph{Robotics Research. The Eleventh International
  Symposium}}. Springer, \bibinfo{pages}{421--431}.
\newblock


\bibitem[\protect\citeauthoryear{Hart, Nilsson, and Raphael}{Hart
  et~al\mbox{.}}{1968}]%
        {hart1968formal}
\bibfield{author}{\bibinfo{person}{Peter~E Hart}, \bibinfo{person}{Nils~J
  Nilsson}, {and} \bibinfo{person}{Bertram Raphael}.}
  \bibinfo{year}{1968}\natexlab{}.
\newblock \showarticletitle{A formal basis for the heuristic determination of
  minimum cost paths}.
\newblock \bibinfo{journal}{\emph{IEEE transactions on Systems Science and
  Cybernetics}} \bibinfo{volume}{4}, \bibinfo{number}{2}
  (\bibinfo{year}{1968}), \bibinfo{pages}{100--107}.
\newblock


\bibitem[\protect\citeauthoryear{He, Zhang, Li, Zhang, Liang, and Yan}{He
  et~al\mbox{.}}{2011}]%
        {he2011autonomous}
\bibfield{author}{\bibinfo{person}{Bo He}, \bibinfo{person}{Hongjin Zhang},
  \bibinfo{person}{Chao Li}, \bibinfo{person}{Shujing Zhang},
  \bibinfo{person}{Yan Liang}, {and} \bibinfo{person}{Tianhong Yan}.}
  \bibinfo{year}{2011}\natexlab{}.
\newblock \showarticletitle{Autonomous navigation for autonomous underwater
  vehicles based on information filters and active sensing}.
\newblock \bibinfo{journal}{\emph{Sensors}} \bibinfo{volume}{11},
  \bibinfo{number}{11} (\bibinfo{year}{2011}), \bibinfo{pages}{10958--10980}.
\newblock


\bibitem[\protect\citeauthoryear{Hegrenaes and Berglund}{Hegrenaes and
  Berglund}{2009}]%
        {hegrenaes2009doppler}
\bibfield{author}{\bibinfo{person}{Oyvind Hegrenaes} {and}
  \bibinfo{person}{Einar Berglund}.} \bibinfo{year}{2009}\natexlab{}.
\newblock \showarticletitle{Doppler water-track aided inertial navigation for
  autonomous underwater vehicle}. In \bibinfo{booktitle}{\emph{OCEANS
  2009-EUROPE}}. IEEE, \bibinfo{pages}{1--10}.
\newblock


\bibitem[\protect\citeauthoryear{Hegrenas, Berglund, and Hallingstad}{Hegrenas
  et~al\mbox{.}}{2008}]%
        {hegrenas2008model}
\bibfield{author}{\bibinfo{person}{Oyvind Hegrenas}, \bibinfo{person}{Einar
  Berglund}, {and} \bibinfo{person}{Oddvar Hallingstad}.}
  \bibinfo{year}{2008}\natexlab{}.
\newblock \showarticletitle{Model-aided inertial navigation for underwater
  vehicles}. In \bibinfo{booktitle}{\emph{2008 IEEE International Conference on
  Robotics and Automation}}. IEEE, \bibinfo{pages}{1069--1076}.
\newblock


\bibitem[\protect\citeauthoryear{Heng, Honegger, Lee, Meier, Tanskanen,
  Fraundorfer, and Pollefeys}{Heng et~al\mbox{.}}{2014}]%
        {heng2014autonomous}
\bibfield{author}{\bibinfo{person}{Lionel Heng}, \bibinfo{person}{Dominik
  Honegger}, \bibinfo{person}{Gim~Hee Lee}, \bibinfo{person}{Lorenz Meier},
  \bibinfo{person}{Petri Tanskanen}, \bibinfo{person}{Friedrich Fraundorfer},
  {and} \bibinfo{person}{Marc Pollefeys}.} \bibinfo{year}{2014}\natexlab{}.
\newblock \showarticletitle{Autonomous visual mapping and exploration with a
  micro aerial vehicle}.
\newblock \bibinfo{journal}{\emph{Journal of Field Robotics}}
  \bibinfo{volume}{31}, \bibinfo{number}{4} (\bibinfo{year}{2014}),
  \bibinfo{pages}{654--675}.
\newblock


\bibitem[\protect\citeauthoryear{Hess, Henry, Leovy, Ryan, and Tillman}{Hess
  et~al\mbox{.}}{1977}]%
        {hess1977meteorological}
\bibfield{author}{\bibinfo{person}{SL Hess}, \bibinfo{person}{RM Henry},
  \bibinfo{person}{Conway~B Leovy}, \bibinfo{person}{JA Ryan}, {and}
  \bibinfo{person}{James~E Tillman}.} \bibinfo{year}{1977}\natexlab{}.
\newblock \showarticletitle{Meteorological results from the surface of Mars:
  Viking 1 and 2}.
\newblock \bibinfo{journal}{\emph{Journal of Geophysical Research}}
  \bibinfo{volume}{82}, \bibinfo{number}{28} (\bibinfo{year}{1977}),
  \bibinfo{pages}{4559--4574}.
\newblock


\bibitem[\protect\citeauthoryear{Hess, Kohler, Rapp, and Andor}{Hess
  et~al\mbox{.}}{2016}]%
        {hess2016real}
\bibfield{author}{\bibinfo{person}{Wolfgang Hess}, \bibinfo{person}{Damon
  Kohler}, \bibinfo{person}{Holger Rapp}, {and} \bibinfo{person}{Daniel
  Andor}.} \bibinfo{year}{2016}\natexlab{}.
\newblock \showarticletitle{Real-time loop closure in 2D LIDAR SLAM}. In
  \bibinfo{booktitle}{\emph{2016 IEEE international conference on robotics and
  automation (ICRA)}}. IEEE, \bibinfo{pages}{1271--1278}.
\newblock


\bibitem[\protect\citeauthoryear{Hong, Petillot, Wallace, and Wang}{Hong
  et~al\mbox{.}}{2022}]%
        {hong2022radarslam}
\bibfield{author}{\bibinfo{person}{Ziyang Hong}, \bibinfo{person}{Yvan
  Petillot}, \bibinfo{person}{Andrew Wallace}, {and} \bibinfo{person}{Sen
  Wang}.} \bibinfo{year}{2022}\natexlab{}.
\newblock \showarticletitle{RadarSLAM: A robust simultaneous localization and
  mapping system for all weather conditions}.
\newblock \bibinfo{journal}{\emph{The International Journal of Robotics
  Research}} (\bibinfo{year}{2022}), \bibinfo{pages}{02783649221080483}.
\newblock


\bibitem[\protect\citeauthoryear{Hornung, Wurm, Bennewitz, Stachniss, and
  Burgard}{Hornung et~al\mbox{.}}{2013}]%
        {hornung2013efficient}
\bibfield{author}{\bibinfo{person}{A Hornung}, \bibinfo{person}{KM Wurm},
  \bibinfo{person}{M Bennewitz}, \bibinfo{person}{C Stachniss}, {and}
  \bibinfo{person}{W Burgard}.} \bibinfo{year}{2013}\natexlab{}.
\newblock \showarticletitle{An efficient probabilistic 3d mapping framework
  based on octrees armin hornung}.
\newblock \bibinfo{journal}{\emph{Autonomous Robots Journal. Springer}}
  (\bibinfo{year}{2013}).
\newblock


\bibitem[\protect\citeauthoryear{Huang, Cao, and Guo}{Huang
  et~al\mbox{.}}{2005}]%
        {huang2005reinforcement}
\bibfield{author}{\bibinfo{person}{Bing-Qiang Huang}, \bibinfo{person}{Guang-Yi
  Cao}, {and} \bibinfo{person}{Min Guo}.} \bibinfo{year}{2005}\natexlab{}.
\newblock \showarticletitle{Reinforcement learning neural network to the
  problem of autonomous mobile robot obstacle avoidance}. In
  \bibinfo{booktitle}{\emph{2005 International conference on machine learning
  and cybernetics}}, Vol.~\bibinfo{volume}{1}. IEEE, \bibinfo{pages}{85--89}.
\newblock


\bibitem[\protect\citeauthoryear{Huynh, Assadi, Rivi{\`e}re-Lorph{\`e}vre,
  Verlinden, and Ahmadi}{Huynh et~al\mbox{.}}{2020}]%
        {huynh2020modelling}
\bibfield{author}{\bibinfo{person}{Hoai~Nam Huynh}, \bibinfo{person}{Hamed
  Assadi}, \bibinfo{person}{Edouard Rivi{\`e}re-Lorph{\`e}vre},
  \bibinfo{person}{Olivier Verlinden}, {and} \bibinfo{person}{Keivan Ahmadi}.}
  \bibinfo{year}{2020}\natexlab{}.
\newblock \showarticletitle{Modelling the dynamics of industrial robots for
  milling operations}.
\newblock \bibinfo{journal}{\emph{Robotics and Computer-Integrated
  Manufacturing}}  \bibinfo{volume}{61} (\bibinfo{year}{2020}),
  \bibinfo{pages}{101852}.
\newblock


\bibitem[\protect\citeauthoryear{Ijspeert, Nakanishi, Hoffmann, Pastor, and
  Schaal}{Ijspeert et~al\mbox{.}}{2013}]%
        {ijspeert2013dynamical}
\bibfield{author}{\bibinfo{person}{Auke~Jan Ijspeert}, \bibinfo{person}{Jun
  Nakanishi}, \bibinfo{person}{Heiko Hoffmann}, \bibinfo{person}{Peter Pastor},
  {and} \bibinfo{person}{Stefan Schaal}.} \bibinfo{year}{2013}\natexlab{}.
\newblock \showarticletitle{Dynamical movement primitives: learning attractor
  models for motor behaviors}.
\newblock \bibinfo{journal}{\emph{Neural computation}} \bibinfo{volume}{25},
  \bibinfo{number}{2} (\bibinfo{year}{2013}), \bibinfo{pages}{328--373}.
\newblock


\bibitem[\protect\citeauthoryear{Injarapu and Gawre}{Injarapu and
  Gawre}{2017}]%
        {injarapu2017survey}
\bibfield{author}{\bibinfo{person}{Anantha Sai Hari Haran~V Injarapu} {and}
  \bibinfo{person}{Suresh~Kumar Gawre}.} \bibinfo{year}{2017}\natexlab{}.
\newblock \showarticletitle{A survey of autonomous mobile robot path planning
  approaches}. In \bibinfo{booktitle}{\emph{2017 International conference on
  recent innovations in signal processing and embedded systems (RISE)}}. IEEE,
  \bibinfo{pages}{624--628}.
\newblock


\bibitem[\protect\citeauthoryear{Jacobi and Karimanzira}{Jacobi and
  Karimanzira}{2013}]%
        {6608089}
\bibfield{author}{\bibinfo{person}{Marco Jacobi} {and} \bibinfo{person}{Divas
  Karimanzira}.} \bibinfo{year}{2013}\natexlab{}.
\newblock \showarticletitle{Underwater pipeline and cable inspection using
  autonomous underwater vehicles}. In \bibinfo{booktitle}{\emph{2013 MTS/IEEE
  OCEANS - Bergen}}. \bibinfo{pages}{1--6}.
\newblock
\urldef\tempurl%
\url{https://doi.org/10.1109/OCEANS-Bergen.2013.6608089}
\showDOI{\tempurl}


\bibitem[\protect\citeauthoryear{Jian, Xiaoyuan, Xiaoping, and Hang}{Jian
  et~al\mbox{.}}{2015}]%
        {jian2015target}
\bibfield{author}{\bibinfo{person}{Xu Jian}, \bibinfo{person}{Chen Xiaoyuan},
  \bibinfo{person}{Song Xiaoping}, {and} \bibinfo{person}{Li Hang}.}
  \bibinfo{year}{2015}\natexlab{}.
\newblock \showarticletitle{Target recognition and location based on binocular
  vision system of UUV}. In \bibinfo{booktitle}{\emph{2015 34th Chinese Control
  Conference (CCC)}}. IEEE, \bibinfo{pages}{3959--3963}.
\newblock


\bibitem[\protect\citeauthoryear{Jiang, Chen, Zheng, Wu, and Cheng}{Jiang
  et~al\mbox{.}}{2016}]%
        {jiang2016mobile}
\bibfield{author}{\bibinfo{person}{Minghao Jiang}, \bibinfo{person}{Yang Chen},
  \bibinfo{person}{Wenlei Zheng}, \bibinfo{person}{Huaiyu Wu}, {and}
  \bibinfo{person}{Lei Cheng}.} \bibinfo{year}{2016}\natexlab{}.
\newblock \showarticletitle{Mobile robot path planning based on dynamic
  movement primitives}. In \bibinfo{booktitle}{\emph{2016 IEEE International
  Conference on Information and Automation (ICIA)}}. IEEE,
  \bibinfo{pages}{980--985}.
\newblock


\bibitem[\protect\citeauthoryear{Jones, Fulkerson, Frazzoli, Kumar, Walters,
  Radford, and Mason}{Jones et~al\mbox{.}}{2006}]%
        {jones2006autonomous}
\bibfield{author}{\bibinfo{person}{Eagle Jones}, \bibinfo{person}{Brian
  Fulkerson}, \bibinfo{person}{Emilio Frazzoli}, \bibinfo{person}{Deepak
  Kumar}, \bibinfo{person}{Robert Walters}, \bibinfo{person}{Jim Radford},
  {and} \bibinfo{person}{Richard Mason}.} \bibinfo{year}{2006}\natexlab{}.
\newblock \showarticletitle{Autonomous off-road driving in the DARPA Grand
  Challenge}. In \bibinfo{booktitle}{\emph{Proceedings of IEEE/ION PLANS
  2006}}. \bibinfo{pages}{366--371}.
\newblock


\bibitem[\protect\citeauthoryear{Julier and Uhlmann}{Julier and
  Uhlmann}{1997}]%
        {julier1997new}
\bibfield{author}{\bibinfo{person}{Simon~J Julier} {and}
  \bibinfo{person}{Jeffrey~K Uhlmann}.} \bibinfo{year}{1997}\natexlab{}.
\newblock \showarticletitle{New extension of the Kalman filter to nonlinear
  systems}. In \bibinfo{booktitle}{\emph{Signal processing, sensor fusion, and
  target recognition VI}}, Vol.~\bibinfo{volume}{3068}. International Society
  for Optics and Photonics, \bibinfo{pages}{182--193}.
\newblock


\bibitem[\protect\citeauthoryear{Jung, Kim, and Kim}{Jung
  et~al\mbox{.}}{2009}]%
        {jung2009simultaneous}
\bibfield{author}{\bibinfo{person}{Sungyoung Jung}, \bibinfo{person}{Jungmin
  Kim}, {and} \bibinfo{person}{Sungshin Kim}.} \bibinfo{year}{2009}\natexlab{}.
\newblock \showarticletitle{Simultaneous localization and mapping of a
  wheel-based autonomous vehicle with ultrasonic sensors}.
\newblock \bibinfo{journal}{\emph{Artificial Life and Robotics}}
  \bibinfo{volume}{14}, \bibinfo{number}{2} (\bibinfo{year}{2009}),
  \bibinfo{pages}{186--190}.
\newblock


\bibitem[\protect\citeauthoryear{Kalman}{Kalman}{1960}]%
        {kalman-filter}
\bibfield{author}{\bibinfo{person}{Rudolph~Emil Kalman}.}
  \bibinfo{year}{1960}\natexlab{}.
\newblock \showarticletitle{A New Approach to Linear Filtering and Prediction
  Problems}.
\newblock \bibinfo{journal}{\emph{Transactions of the ASME--Journal of Basic
  Engineering}} \bibinfo{volume}{82}, \bibinfo{number}{Series D}
  (\bibinfo{year}{1960}), \bibinfo{pages}{35--45}.
\newblock


\bibitem[\protect\citeauthoryear{Kamon, Rimon, and Rivlin}{Kamon
  et~al\mbox{.}}{1998}]%
        {kamon1998tangentbug}
\bibfield{author}{\bibinfo{person}{Ishay Kamon}, \bibinfo{person}{Elon Rimon},
  {and} \bibinfo{person}{Ehud Rivlin}.} \bibinfo{year}{1998}\natexlab{}.
\newblock \showarticletitle{Tangentbug: A range-sensor-based navigation
  algorithm}.
\newblock \bibinfo{journal}{\emph{The International Journal of Robotics
  Research}} \bibinfo{volume}{17}, \bibinfo{number}{9} (\bibinfo{year}{1998}),
  \bibinfo{pages}{934--953}.
\newblock


\bibitem[\protect\citeauthoryear{Kara~Mohamed, Patra, and Lanzon}{Kara~Mohamed
  et~al\mbox{.}}{2011}]%
        {5983054}
\bibfield{author}{\bibinfo{person}{Mohamed Kara~Mohamed},
  \bibinfo{person}{Sourav Patra}, {and} \bibinfo{person}{Alexander Lanzon}.}
  \bibinfo{year}{2011}\natexlab{}.
\newblock \showarticletitle{Designing simple indoor navigation system for
  UAVs}. In \bibinfo{booktitle}{\emph{2011 19th Mediterranean Conference on
  Control \& Automation (MED)}}. \bibinfo{pages}{1223--1228}.
\newblock
\urldef\tempurl%
\url{https://doi.org/10.1109/MED.2011.5983054}
\showDOI{\tempurl}


\bibitem[\protect\citeauthoryear{Karam, Lehtola, and Vosselman}{Karam
  et~al\mbox{.}}{2019}]%
        {karam2019integrating}
\bibfield{author}{\bibinfo{person}{S Karam}, \bibinfo{person}{V Lehtola}, {and}
  \bibinfo{person}{G Vosselman}.} \bibinfo{year}{2019}\natexlab{}.
\newblock \showarticletitle{INTEGRATING A LOW-COST MEMS IMU INTO A LASER-BASED
  SLAM FOR INDOOR MOBILE MAPPING.}
\newblock \bibinfo{journal}{\emph{International Archives of the Photogrammetry,
  Remote Sensing \& Spatial Information Sciences}} (\bibinfo{year}{2019}).
\newblock


\bibitem[\protect\citeauthoryear{Kassel}{Kassel}{1971}]%
        {kassel1971lunokhod}
\bibfield{author}{\bibinfo{person}{Simon Kassel}.}
  \bibinfo{year}{1971}\natexlab{}.
\newblock \bibinfo{booktitle}{\emph{Lunokhod-1 Soviet lunar surface vehicle}}.
\newblock \bibinfo{type}{{T}echnical {R}eport}. \bibinfo{institution}{RAND CORP
  SANTA MONICA CA}.
\newblock


\bibitem[\protect\citeauthoryear{Kassir, Palhang, and Ahmadzadeh}{Kassir
  et~al\mbox{.}}{2020}]%
        {kassir2020qualitative}
\bibfield{author}{\bibinfo{person}{Mohamad~Mahdi Kassir},
  \bibinfo{person}{Maziar Palhang}, {and} \bibinfo{person}{Mohammad~Reza
  Ahmadzadeh}.} \bibinfo{year}{2020}\natexlab{}.
\newblock \showarticletitle{Qualitative vision-based navigation based on sloped
  funnel lane concept}.
\newblock \bibinfo{journal}{\emph{Intelligent Service Robotics}}
  \bibinfo{volume}{13}, \bibinfo{number}{2} (\bibinfo{year}{2020}),
  \bibinfo{pages}{235--250}.
\newblock


\bibitem[\protect\citeauthoryear{Keirsey, Mitchell, Payton, Tseng, and
  Wong}{Keirsey et~al\mbox{.}}{1987}]%
        {keirsey1987autonomous}
\bibfield{author}{\bibinfo{person}{DM Keirsey}, \bibinfo{person}{JS Mitchell},
  \bibinfo{person}{DW Payton}, \bibinfo{person}{DY Tseng}, {and}
  \bibinfo{person}{VS Wong}.} \bibinfo{year}{1987}\natexlab{}.
\newblock \bibinfo{booktitle}{\emph{Autonomous Land Vehicle (ALV) Planning and
  Navigation System.}}
\newblock \bibinfo{type}{{T}echnical {R}eport}. \bibinfo{institution}{HUGHES
  RESEARCH LABS MALIBU CA}.
\newblock


\bibitem[\protect\citeauthoryear{Key}{Key}{2000}]%
        {881735}
\bibfield{author}{\bibinfo{person}{W.H. Key}.} \bibinfo{year}{2000}\natexlab{}.
\newblock \showarticletitle{Side scan sonar technology}. In
  \bibinfo{booktitle}{\emph{OCEANS 2000 MTS/IEEE Conference and Exhibition.
  Conference Proceedings (Cat. No.00CH37158)}}, Vol.~\bibinfo{volume}{2}.
  \bibinfo{pages}{1029--1033 vol.2}.
\newblock
\urldef\tempurl%
\url{https://doi.org/10.1109/OCEANS.2000.881735}
\showDOI{\tempurl}


\bibitem[\protect\citeauthoryear{Khan, Dometios, Verginis, Tzafestas, Wollherr,
  and Buss}{Khan et~al\mbox{.}}{2014}]%
        {khan2014rmap}
\bibfield{author}{\bibinfo{person}{Sheraz Khan}, \bibinfo{person}{Athanasios
  Dometios}, \bibinfo{person}{Chris Verginis}, \bibinfo{person}{Costas
  Tzafestas}, \bibinfo{person}{Dirk Wollherr}, {and} \bibinfo{person}{Martin
  Buss}.} \bibinfo{year}{2014}\natexlab{}.
\newblock \showarticletitle{RMAP: a rectangular cuboid approximation framework
  for 3D environment mapping}.
\newblock \bibinfo{journal}{\emph{Autonomous Robots}} \bibinfo{volume}{37},
  \bibinfo{number}{3} (\bibinfo{year}{2014}), \bibinfo{pages}{261--277}.
\newblock


\bibitem[\protect\citeauthoryear{Khatib}{Khatib}{1986}]%
        {khatib1986real}
\bibfield{author}{\bibinfo{person}{Oussama Khatib}.}
  \bibinfo{year}{1986}\natexlab{}.
\newblock \showarticletitle{Real-time obstacle avoidance for manipulators and
  mobile robots}.
\newblock In \bibinfo{booktitle}{\emph{Autonomous robot vehicles}}.
  \bibinfo{publisher}{Springer}, \bibinfo{pages}{396--404}.
\newblock


\bibitem[\protect\citeauthoryear{Kim and Nevatia}{Kim and Nevatia}{1998}]%
        {kim1998recognition}
\bibfield{author}{\bibinfo{person}{Dongsung Kim} {and}
  \bibinfo{person}{Ramakant Nevatia}.} \bibinfo{year}{1998}\natexlab{}.
\newblock \showarticletitle{Recognition and localization of generic objects for
  indoor navigation using functionality}.
\newblock \bibinfo{journal}{\emph{Image and Vision Computing}}
  \bibinfo{volume}{16}, \bibinfo{number}{11} (\bibinfo{year}{1998}),
  \bibinfo{pages}{729--743}.
\newblock


\bibitem[\protect\citeauthoryear{Kim and Nevatia}{Kim and Nevatia}{1999}]%
        {kim1999symbolic}
\bibfield{author}{\bibinfo{person}{Dongsung Kim} {and}
  \bibinfo{person}{Ramakant Nevatia}.} \bibinfo{year}{1999}\natexlab{}.
\newblock \showarticletitle{Symbolic navigation with a generic map}.
\newblock \bibinfo{journal}{\emph{Autonomous Robots}} \bibinfo{volume}{6},
  \bibinfo{number}{1} (\bibinfo{year}{1999}), \bibinfo{pages}{69--88}.
\newblock


\bibitem[\protect\citeauthoryear{Kim and Kim}{Kim and Kim}{2012}]%
        {kim2012simple}
\bibfield{author}{\bibinfo{person}{Sungbok Kim} {and} \bibinfo{person}{Hyunbin
  Kim}.} \bibinfo{year}{2012}\natexlab{}.
\newblock \showarticletitle{Simple and complex obstacle detection using an
  overlapped ultrasonic sensor ring}. In \bibinfo{booktitle}{\emph{2012 12th
  International Conference on Control, Automation and Systems}}. IEEE,
  \bibinfo{pages}{2148--2152}.
\newblock


\bibitem[\protect\citeauthoryear{Kim, Perera, Sollid, Batalden, and Sydnes}{Kim
  et~al\mbox{.}}{2022}]%
        {kim2022safety}
\bibfield{author}{\bibinfo{person}{Tae-eun Kim},
  \bibinfo{person}{Lokukaluge~Prasad Perera}, \bibinfo{person}{Magne-Petter
  Sollid}, \bibinfo{person}{Bj{\o}rn-Morten Batalden}, {and}
  \bibinfo{person}{Are~Kristoffer Sydnes}.} \bibinfo{year}{2022}\natexlab{}.
\newblock \showarticletitle{Safety challenges related to autonomous ships in
  mixed navigational environments}.
\newblock \bibinfo{journal}{\emph{WMU Journal of Maritime Affairs}}
  (\bibinfo{year}{2022}), \bibinfo{pages}{1--18}.
\newblock


\bibitem[\protect\citeauthoryear{Kinsey, Eustice, and Whitcomb}{Kinsey
  et~al\mbox{.}}{2006}]%
        {kinsey2006survey}
\bibfield{author}{\bibinfo{person}{James~C Kinsey}, \bibinfo{person}{Ryan~M
  Eustice}, {and} \bibinfo{person}{Louis~L Whitcomb}.}
  \bibinfo{year}{2006}\natexlab{}.
\newblock \showarticletitle{A survey of underwater vehicle navigation: Recent
  advances and new challenges}. In \bibinfo{booktitle}{\emph{IFAC conference of
  manoeuvering and control of marine craft}}, Vol.~\bibinfo{volume}{88}.
  Lisbon, \bibinfo{pages}{1--12}.
\newblock


\bibitem[\protect\citeauthoryear{Klein and Murray}{Klein and Murray}{2007a}]%
        {4538852}
\bibfield{author}{\bibinfo{person}{Georg Klein} {and} \bibinfo{person}{David
  Murray}.} \bibinfo{year}{2007}\natexlab{a}.
\newblock \showarticletitle{Parallel Tracking and Mapping for Small AR
  Workspaces}. In \bibinfo{booktitle}{\emph{2007 6th IEEE and ACM International
  Symposium on Mixed and Augmented Reality}}. \bibinfo{pages}{225--234}.
\newblock
\urldef\tempurl%
\url{https://doi.org/10.1109/ISMAR.2007.4538852}
\showDOI{\tempurl}


\bibitem[\protect\citeauthoryear{Klein and Murray}{Klein and Murray}{2007b}]%
        {klein2007parallel}
\bibfield{author}{\bibinfo{person}{Georg Klein} {and} \bibinfo{person}{David
  Murray}.} \bibinfo{year}{2007}\natexlab{b}.
\newblock \showarticletitle{Parallel tracking and mapping for small AR
  workspaces}. In \bibinfo{booktitle}{\emph{2007 6th IEEE and ACM international
  symposium on mixed and augmented reality}}. IEEE, \bibinfo{pages}{225--234}.
\newblock


\bibitem[\protect\citeauthoryear{Klein and Murray}{Klein and Murray}{2008}]%
        {klein2008improving}
\bibfield{author}{\bibinfo{person}{Georg Klein} {and} \bibinfo{person}{David
  Murray}.} \bibinfo{year}{2008}\natexlab{}.
\newblock \showarticletitle{Improving the agility of keyframe-based SLAM}. In
  \bibinfo{booktitle}{\emph{European conference on computer vision}}. Springer,
  \bibinfo{pages}{802--815}.
\newblock


\bibitem[\protect\citeauthoryear{Klein and Murray}{Klein and Murray}{2009}]%
        {klein2009parallel}
\bibfield{author}{\bibinfo{person}{Georg Klein} {and} \bibinfo{person}{David
  Murray}.} \bibinfo{year}{2009}\natexlab{}.
\newblock \showarticletitle{Parallel tracking and mapping on a camera phone}.
  In \bibinfo{booktitle}{\emph{2009 8th IEEE International Symposium on Mixed
  and Augmented Reality}}. IEEE, \bibinfo{pages}{83--86}.
\newblock


\bibitem[\protect\citeauthoryear{Koenig and Likhachev}{Koenig and
  Likhachev}{2005}]%
        {koenig2005fast}
\bibfield{author}{\bibinfo{person}{Sven Koenig} {and} \bibinfo{person}{Maxim
  Likhachev}.} \bibinfo{year}{2005}\natexlab{}.
\newblock \showarticletitle{Fast replanning for navigation in unknown terrain}.
\newblock \bibinfo{journal}{\emph{IEEE Transactions on Robotics}}
  \bibinfo{volume}{21}, \bibinfo{number}{3} (\bibinfo{year}{2005}),
  \bibinfo{pages}{354--363}.
\newblock


\bibitem[\protect\citeauthoryear{Koenig, Likhachev, and Furcy}{Koenig
  et~al\mbox{.}}{2004}]%
        {koenig2004lifelong}
\bibfield{author}{\bibinfo{person}{Sven Koenig}, \bibinfo{person}{Maxim
  Likhachev}, {and} \bibinfo{person}{David Furcy}.}
  \bibinfo{year}{2004}\natexlab{}.
\newblock \showarticletitle{Lifelong planning A∗}.
\newblock \bibinfo{journal}{\emph{Artificial Intelligence}}
  \bibinfo{volume}{155}, \bibinfo{number}{1-2} (\bibinfo{year}{2004}),
  \bibinfo{pages}{93--146}.
\newblock


\bibitem[\protect\citeauthoryear{Kohlbrecher, Meyer, Graber, Petersen,
  Klingauf, and Stryk}{Kohlbrecher et~al\mbox{.}}{2013}]%
        {kohlbrecher2013hector}
\bibfield{author}{\bibinfo{person}{Stefan Kohlbrecher},
  \bibinfo{person}{Johannes Meyer}, \bibinfo{person}{Thorsten Graber},
  \bibinfo{person}{Karen Petersen}, \bibinfo{person}{Uwe Klingauf}, {and}
  \bibinfo{person}{Oskar~von Stryk}.} \bibinfo{year}{2013}\natexlab{}.
\newblock \showarticletitle{Hector open source modules for autonomous mapping
  and navigation with rescue robots}. In \bibinfo{booktitle}{\emph{Robot Soccer
  World Cup}}. Springer, \bibinfo{pages}{624--631}.
\newblock


\bibitem[\protect\citeauthoryear{Korkmaz, Y{\i}lmaz, and Durdu}{Korkmaz
  et~al\mbox{.}}{2016}]%
        {korkmaz2016comparison}
\bibfield{author}{\bibinfo{person}{Mehmet Korkmaz}, \bibinfo{person}{Nihat
  Y{\i}lmaz}, {and} \bibinfo{person}{Akif Durdu}.}
  \bibinfo{year}{2016}\natexlab{}.
\newblock \showarticletitle{Comparison of the SLAM algorithms: Hangar
  experiments}. In \bibinfo{booktitle}{\emph{MATEC Web of Conferences}},
  Vol.~\bibinfo{volume}{42}. EDP Sciences, \bibinfo{pages}{03009}.
\newblock


\bibitem[\protect\citeauthoryear{Koschorrek, Piccini, Oberg, Felsberg, Nielsen,
  and Mester}{Koschorrek et~al\mbox{.}}{2013}]%
        {amuse2013gt}
\bibfield{author}{\bibinfo{person}{Philipp Koschorrek},
  \bibinfo{person}{Tommaso Piccini}, \bibinfo{person}{Per Oberg},
  \bibinfo{person}{Michael Felsberg}, \bibinfo{person}{Lars Nielsen}, {and}
  \bibinfo{person}{Rudolf Mester}.} \bibinfo{year}{2013}\natexlab{}.
\newblock \showarticletitle{A multi-sensor traffic scene dataset with
  omnidirectional video}. In \bibinfo{booktitle}{\emph{Ground Truth - What is a
  good dataset? CVPR Workshop 2013}}.
\newblock


\bibitem[\protect\citeauthoryear{Krizhevsky, Sutskever, and Hinton}{Krizhevsky
  et~al\mbox{.}}{2017}]%
        {krizhevsky2017imagenet}
\bibfield{author}{\bibinfo{person}{Alex Krizhevsky}, \bibinfo{person}{Ilya
  Sutskever}, {and} \bibinfo{person}{Geoffrey~E Hinton}.}
  \bibinfo{year}{2017}\natexlab{}.
\newblock \showarticletitle{Imagenet classification with deep convolutional
  neural networks}.
\newblock \bibinfo{journal}{\emph{Commun. ACM}} \bibinfo{volume}{60},
  \bibinfo{number}{6} (\bibinfo{year}{2017}), \bibinfo{pages}{84--90}.
\newblock


\bibitem[\protect\citeauthoryear{Krotkov, Bares, Katragadda, Simmons, and
  Whittaker}{Krotkov et~al\mbox{.}}{1994}]%
        {krotkov1994lunar}
\bibfield{author}{\bibinfo{person}{Eric Krotkov}, \bibinfo{person}{John Bares},
  \bibinfo{person}{Lalitesh Katragadda}, \bibinfo{person}{Reid Simmons}, {and}
  \bibinfo{person}{Red Whittaker}.} \bibinfo{year}{1994}\natexlab{}.
\newblock \showarticletitle{Lunar rover technology demonstrations with Dante
  and Ratler}. In \bibinfo{booktitle}{\emph{JPL, Third International Symposium
  on Artificial Intelligence, Robotics, and Automation for Space 1994}}.
\newblock


\bibitem[\protect\citeauthoryear{Laubach and Burdick}{Laubach and
  Burdick}{2000}]%
        {laubach2000roverbug}
\bibfield{author}{\bibinfo{person}{Sharon Laubach} {and} \bibinfo{person}{Joel
  Burdick}.} \bibinfo{year}{2000}\natexlab{}.
\newblock \showarticletitle{RoverBug: Long range navigation for mars rovers}.
\newblock In \bibinfo{booktitle}{\emph{Experimental Robotics VI}}.
  \bibinfo{publisher}{Springer}, \bibinfo{pages}{339--348}.
\newblock


\bibitem[\protect\citeauthoryear{Laubach and Burdick}{Laubach and
  Burdick}{1999}]%
        {laubach1999autonomous}
\bibfield{author}{\bibinfo{person}{Sharon~L Laubach} {and}
  \bibinfo{person}{Joel~W Burdick}.} \bibinfo{year}{1999}\natexlab{}.
\newblock \showarticletitle{An autonomous sensor-based path-planner for
  planetary microrovers}. In \bibinfo{booktitle}{\emph{Proceedings 1999 IEEE
  International Conference on Robotics and Automation (Cat. No. 99CH36288C)}},
  Vol.~\bibinfo{volume}{1}. IEEE, \bibinfo{pages}{347--354}.
\newblock


\bibitem[\protect\citeauthoryear{Leonard, Bennett, Smith, Jacob, and
  Feder}{Leonard et~al\mbox{.}}{1998}]%
        {leonard1998autonomous}
\bibfield{author}{\bibinfo{person}{John~J Leonard}, \bibinfo{person}{Andrew~A
  Bennett}, \bibinfo{person}{Christopher~M Smith}, \bibinfo{person}{Hans
  Jacob}, {and} \bibinfo{person}{S Feder}.} \bibinfo{year}{1998}\natexlab{}.
\newblock \showarticletitle{Autonomous underwater vehicle navigation}. In
  \bibinfo{booktitle}{\emph{MIT Marine Robotics Laboratory Technical
  Memorandum}}. Citeseer.
\newblock


\bibitem[\protect\citeauthoryear{Leonard and Durrant-Whyte}{Leonard and
  Durrant-Whyte}{1991}]%
        {leonard1991mobile}
\bibfield{author}{\bibinfo{person}{John~J Leonard} {and}
  \bibinfo{person}{Hugh~F Durrant-Whyte}.} \bibinfo{year}{1991}\natexlab{}.
\newblock \showarticletitle{Mobile robot localization by tracking geometric
  beacons}.
\newblock \bibinfo{journal}{\emph{IEEE Transactions on robotics and
  Automation}} \bibinfo{volume}{7}, \bibinfo{number}{3} (\bibinfo{year}{1991}),
  \bibinfo{pages}{376--382}.
\newblock


\bibitem[\protect\citeauthoryear{Levine, Wagener, and Abbeel}{Levine
  et~al\mbox{.}}{2015}]%
        {LevineWA15}
\bibfield{author}{\bibinfo{person}{Sergey Levine}, \bibinfo{person}{Nolan
  Wagener}, {and} \bibinfo{person}{Pieter Abbeel}.}
  \bibinfo{year}{2015}\natexlab{}.
\newblock \showarticletitle{Learning Contact-Rich Manipulation Skills with
  Guided Policy Search}.
\newblock \bibinfo{journal}{\emph{CoRR}}  \bibinfo{volume}{abs/1501.05611}
  (\bibinfo{year}{2015}).
\newblock
\showeprint[arXiv]{1501.05611}
\urldef\tempurl%
\url{http://arxiv.org/abs/1501.05611}
\showURL{%
\tempurl}


\bibitem[\protect\citeauthoryear{Levinson, Montemerlo, and Thrun}{Levinson
  et~al\mbox{.}}{2007}]%
        {levinson2007map}
\bibfield{author}{\bibinfo{person}{Jesse Levinson}, \bibinfo{person}{Michael
  Montemerlo}, {and} \bibinfo{person}{Sebastian Thrun}.}
  \bibinfo{year}{2007}\natexlab{}.
\newblock \showarticletitle{Map-based precision vehicle localization in urban
  environments.}. In \bibinfo{booktitle}{\emph{Robotics: science and systems}},
  Vol.~\bibinfo{volume}{4}. Citeseer, \bibinfo{pages}{1}.
\newblock


\bibitem[\protect\citeauthoryear{Li, Shi, Long, Liu, Yang, Wang, Wei, and
  Qiao}{Li et~al\mbox{.}}{2020}]%
        {li2020dxslam}
\bibfield{author}{\bibinfo{person}{Dongjiang Li}, \bibinfo{person}{Xuesong
  Shi}, \bibinfo{person}{Qiwei Long}, \bibinfo{person}{Shenghui Liu},
  \bibinfo{person}{Wei Yang}, \bibinfo{person}{Fangshi Wang},
  \bibinfo{person}{Qi Wei}, {and} \bibinfo{person}{Fei Qiao}.}
  \bibinfo{year}{2020}\natexlab{}.
\newblock \showarticletitle{DXSLAM: A robust and efficient visual SLAM system
  with deep features}. In \bibinfo{booktitle}{\emph{2020 IEEE/RSJ International
  Conference on Intelligent Robots and Systems (IROS)}}. IEEE,
  \bibinfo{pages}{4958--4965}.
\newblock


\bibitem[\protect\citeauthoryear{Li, Barjuei, Ciuti, Hao, Zhang, Menciassi,
  Huang, and Dario}{Li et~al\mbox{.}}{2018}]%
        {li2018magnetically}
\bibfield{author}{\bibinfo{person}{Jing Li}, \bibinfo{person}{Erfan~Shojaei
  Barjuei}, \bibinfo{person}{Gastone Ciuti}, \bibinfo{person}{Yang Hao},
  \bibinfo{person}{Peisen Zhang}, \bibinfo{person}{Arianna Menciassi},
  \bibinfo{person}{Qiang Huang}, {and} \bibinfo{person}{Paolo Dario}.}
  \bibinfo{year}{2018}\natexlab{}.
\newblock \showarticletitle{Magnetically-driven medical robots: An analytical
  magnetic model for endoscopic capsules design}.
\newblock \bibinfo{journal}{\emph{Journal of Magnetism and Magnetic Materials}}
   \bibinfo{volume}{452} (\bibinfo{year}{2018}), \bibinfo{pages}{278--287}.
\newblock


\bibitem[\protect\citeauthoryear{Li, Wang, Huang, Liang, Wu, and Liao}{Li
  et~al\mbox{.}}{2016}]%
        {7831874}
\bibfield{author}{\bibinfo{person}{Kang Li}, \bibinfo{person}{Can Wang},
  \bibinfo{person}{Sheng Huang}, \bibinfo{person}{Guoyuan Liang},
  \bibinfo{person}{Xinyu Wu}, {and} \bibinfo{person}{Yubin Liao}.}
  \bibinfo{year}{2016}\natexlab{}.
\newblock \showarticletitle{Self-positioning for UAV indoor navigation based on
  3D laser scanner, UWB and INS}. In \bibinfo{booktitle}{\emph{2016 IEEE
  International Conference on Information and Automation (ICIA)}}.
  \bibinfo{pages}{498--503}.
\newblock
\urldef\tempurl%
\url{https://doi.org/10.1109/ICInfA.2016.7831874}
\showDOI{\tempurl}


\bibitem[\protect\citeauthoryear{Li and Mourikis}{Li and Mourikis}{2013}]%
        {li2013high}
\bibfield{author}{\bibinfo{person}{Mingyang Li} {and}
  \bibinfo{person}{Anastasios~I Mourikis}.} \bibinfo{year}{2013}\natexlab{}.
\newblock \showarticletitle{High-precision, consistent EKF-based
  visual-inertial odometry}.
\newblock \bibinfo{journal}{\emph{The International Journal of Robotics
  Research}} \bibinfo{volume}{32}, \bibinfo{number}{6} (\bibinfo{year}{2013}),
  \bibinfo{pages}{690--711}.
\newblock


\bibitem[\protect\citeauthoryear{Li, Liu, Zhang, and Hang}{Li
  et~al\mbox{.}}{2014}]%
        {7049479}
\bibfield{author}{\bibinfo{person}{Rongbing Li}, \bibinfo{person}{Jianye Liu},
  \bibinfo{person}{Ling Zhang}, {and} \bibinfo{person}{Yijun Hang}.}
  \bibinfo{year}{2014}\natexlab{}.
\newblock \showarticletitle{LIDAR/MEMS IMU integrated navigation (SLAM) method
  for a small UAV in indoor environments}. In \bibinfo{booktitle}{\emph{2014
  DGON Inertial Sensors and Systems (ISS)}}. \bibinfo{pages}{1--15}.
\newblock
\urldef\tempurl%
\url{https://doi.org/10.1109/InertialSensors.2014.7049479}
\showDOI{\tempurl}


\bibitem[\protect\citeauthoryear{Libin, Qinghua, Guanjun, Yan, Liyong, Feng,
  and Fang}{Libin et~al\mbox{.}}{2008}]%
        {libin2008overview}
\bibfield{author}{\bibinfo{person}{Zhang Libin}, \bibinfo{person}{Yang
  Qinghua}, \bibinfo{person}{Bao Guanjun}, \bibinfo{person}{Wang Yan},
  \bibinfo{person}{Qi Liyong}, \bibinfo{person}{Gao Feng}, {and}
  \bibinfo{person}{Xu Fang}.} \bibinfo{year}{2008}\natexlab{}.
\newblock \showarticletitle{Overview of research on agricultural robot in
  China}.
\newblock \bibinfo{journal}{\emph{International Journal of Agricultural and
  Biological Engineering}} \bibinfo{volume}{1}, \bibinfo{number}{1}
  (\bibinfo{year}{2008}), \bibinfo{pages}{12--21}.
\newblock


\bibitem[\protect\citeauthoryear{Lillicrap, Hunt, Pritzel, Heess, Erez, Tassa,
  Silver, and Wierstra}{Lillicrap et~al\mbox{.}}{2015}]%
        {lillicrap2015continuous}
\bibfield{author}{\bibinfo{person}{Timothy~P Lillicrap},
  \bibinfo{person}{Jonathan~J Hunt}, \bibinfo{person}{Alexander Pritzel},
  \bibinfo{person}{Nicolas Heess}, \bibinfo{person}{Tom Erez},
  \bibinfo{person}{Yuval Tassa}, \bibinfo{person}{David Silver}, {and}
  \bibinfo{person}{Daan Wierstra}.} \bibinfo{year}{2015}\natexlab{}.
\newblock \showarticletitle{Continuous control with deep reinforcement
  learning}.
\newblock \bibinfo{journal}{\emph{arXiv preprint arXiv:1509.02971}}
  (\bibinfo{year}{2015}).
\newblock


\bibitem[\protect\citeauthoryear{Liu, Zheng, Wang, Zhao, Fu, and Li}{Liu
  et~al\mbox{.}}{2017}]%
        {liu2017cnn}
\bibfield{author}{\bibinfo{person}{Canglong Liu}, \bibinfo{person}{Bin Zheng},
  \bibinfo{person}{Chunyang Wang}, \bibinfo{person}{Yongting Zhao},
  \bibinfo{person}{Shun Fu}, {and} \bibinfo{person}{Haochen Li}.}
  \bibinfo{year}{2017}\natexlab{}.
\newblock \showarticletitle{CNN-based vision model for obstacle avoidance of
  mobile robot}. In \bibinfo{booktitle}{\emph{MATEC Web of Conferences}},
  Vol.~\bibinfo{volume}{139}. EDP Sciences, \bibinfo{pages}{00007}.
\newblock


\bibitem[\protect\citeauthoryear{Liu, Wang, Peng, Cui, and Fiondella}{Liu
  et~al\mbox{.}}{2014}]%
        {liu2014suave}
\bibfield{author}{\bibinfo{person}{Jun Liu}, \bibinfo{person}{Zhaohui Wang},
  \bibinfo{person}{Zheng Peng}, \bibinfo{person}{Jun-Hong Cui}, {and}
  \bibinfo{person}{Lance Fiondella}.} \bibinfo{year}{2014}\natexlab{}.
\newblock \showarticletitle{Suave: Swarm underwater autonomous vehicle
  localization}. In \bibinfo{booktitle}{\emph{IEEE INFOCOM 2014-IEEE Conference
  on Computer Communications}}. IEEE, \bibinfo{pages}{64--72}.
\newblock


\bibitem[\protect\citeauthoryear{Liu and Chen}{Liu and Chen}{1998}]%
        {liu1998sequential}
\bibfield{author}{\bibinfo{person}{Jun~S Liu} {and} \bibinfo{person}{Rong
  Chen}.} \bibinfo{year}{1998}\natexlab{}.
\newblock \showarticletitle{Sequential Monte Carlo methods for dynamic
  systems}.
\newblock \bibinfo{journal}{\emph{Journal of the American statistical
  association}} \bibinfo{volume}{93}, \bibinfo{number}{443}
  (\bibinfo{year}{1998}), \bibinfo{pages}{1032--1044}.
\newblock


\bibitem[\protect\citeauthoryear{Lobo, Lucas, Dias, and De~Almeida}{Lobo
  et~al\mbox{.}}{1995}]%
        {lobo1995inertial}
\bibfield{author}{\bibinfo{person}{Jorge Lobo}, \bibinfo{person}{Paulo Lucas},
  \bibinfo{person}{Jorge Dias}, {and} \bibinfo{person}{A~Traca De~Almeida}.}
  \bibinfo{year}{1995}\natexlab{}.
\newblock \showarticletitle{Inertial navigation system for mobile land
  vehicles}. In \bibinfo{booktitle}{\emph{1995 Proceedings of the IEEE
  International Symposium on Industrial Electronics}},
  Vol.~\bibinfo{volume}{2}. IEEE, \bibinfo{pages}{843--848}.
\newblock


\bibitem[\protect\citeauthoryear{Lowe}{Lowe}{2004}]%
        {LoweDavidG}
\bibfield{author}{\bibinfo{person}{David~G. Lowe}.}
  \bibinfo{year}{2004}\natexlab{}.
\newblock \showarticletitle{Distinctive Image Features from Scale-Invariant
  Keypoints}.
\newblock \bibinfo{journal}{\emph{Int. J. Comput. Vision}}
  \bibinfo{volume}{60}, \bibinfo{number}{2} (\bibinfo{date}{nov}
  \bibinfo{year}{2004}), \bibinfo{pages}{91–110}.
\newblock
\showISSN{0920-5691}
\urldef\tempurl%
\url{https://doi.org/10.1023/B:VISI.0000029664.99615.94}
\showDOI{\tempurl}


\bibitem[\protect\citeauthoryear{Lu and Milios}{Lu and Milios}{1997}]%
        {lu1997globally}
\bibfield{author}{\bibinfo{person}{Feng Lu} {and} \bibinfo{person}{Evangelos
  Milios}.} \bibinfo{year}{1997}\natexlab{}.
\newblock \showarticletitle{Globally consistent range scan alignment for
  environment mapping}.
\newblock \bibinfo{journal}{\emph{Autonomous robots}} \bibinfo{volume}{4},
  \bibinfo{number}{4} (\bibinfo{year}{1997}), \bibinfo{pages}{333--349}.
\newblock


\bibitem[\protect\citeauthoryear{Lumelsky and Stepanov}{Lumelsky and
  Stepanov}{1986}]%
        {lumelsky1986dynamic}
\bibfield{author}{\bibinfo{person}{Vladimir Lumelsky} {and}
  \bibinfo{person}{Alexander Stepanov}.} \bibinfo{year}{1986}\natexlab{}.
\newblock \showarticletitle{Dynamic path planning for a mobile automaton with
  limited information on the environment}.
\newblock \bibinfo{journal}{\emph{IEEE transactions on Automatic control}}
  \bibinfo{volume}{31}, \bibinfo{number}{11} (\bibinfo{year}{1986}),
  \bibinfo{pages}{1058--1063}.
\newblock


\bibitem[\protect\citeauthoryear{Macario~Barros, Michel, Moline, Corre, and
  Carrel}{Macario~Barros et~al\mbox{.}}{2022}]%
        {macario2022comprehensive}
\bibfield{author}{\bibinfo{person}{Andr{\'e}a Macario~Barros},
  \bibinfo{person}{Maugan Michel}, \bibinfo{person}{Yoann Moline},
  \bibinfo{person}{Gwenol{\'e} Corre}, {and} \bibinfo{person}{Fr{\'e}d{\'e}rick
  Carrel}.} \bibinfo{year}{2022}\natexlab{}.
\newblock \showarticletitle{A comprehensive survey of visual slam algorithms}.
\newblock \bibinfo{journal}{\emph{Robotics}} \bibinfo{volume}{11},
  \bibinfo{number}{1} (\bibinfo{year}{2022}), \bibinfo{pages}{24}.
\newblock


\bibitem[\protect\citeauthoryear{Maddern, Pascoe, Linegar, and Newman}{Maddern
  et~al\mbox{.}}{2017}]%
        {maddern20171}
\bibfield{author}{\bibinfo{person}{Will Maddern}, \bibinfo{person}{Geoffrey
  Pascoe}, \bibinfo{person}{Chris Linegar}, {and} \bibinfo{person}{Paul
  Newman}.} \bibinfo{year}{2017}\natexlab{}.
\newblock \showarticletitle{1 year, 1000 km: The Oxford RobotCar dataset}.
\newblock \bibinfo{journal}{\emph{The International Journal of Robotics
  Research}} \bibinfo{volume}{36}, \bibinfo{number}{1} (\bibinfo{year}{2017}),
  \bibinfo{pages}{3--15}.
\newblock


\bibitem[\protect\citeauthoryear{Magid and Rivlin}{Magid and Rivlin}{2004}]%
        {magid2004cautiousbug}
\bibfield{author}{\bibinfo{person}{Evgeni Magid} {and} \bibinfo{person}{Ehud
  Rivlin}.} \bibinfo{year}{2004}\natexlab{}.
\newblock \showarticletitle{CAUTIOUSBUG: A competitive algorithm for
  sensory-based robot navigation}. In \bibinfo{booktitle}{\emph{2004 IEEE/RSJ
  International Conference on Intelligent Robots and Systems (IROS)(IEEE Cat.
  No. 04CH37566)}}, Vol.~\bibinfo{volume}{3}. IEEE,
  \bibinfo{pages}{2757--2762}.
\newblock


\bibitem[\protect\citeauthoryear{Maimone, Johnson, Cheng, Willson, and
  Matthies}{Maimone et~al\mbox{.}}{2006}]%
        {maimone2006autonomous}
\bibfield{author}{\bibinfo{person}{Mark Maimone}, \bibinfo{person}{Andrew
  Johnson}, \bibinfo{person}{Yang Cheng}, \bibinfo{person}{Reg Willson}, {and}
  \bibinfo{person}{Larry Matthies}.} \bibinfo{year}{2006}\natexlab{}.
\newblock \showarticletitle{Autonomous navigation results from the Mars
  Exploration Rover (MER) mission}.
\newblock In \bibinfo{booktitle}{\emph{Experimental robotics IX}}.
  \bibinfo{publisher}{Springer}, \bibinfo{pages}{3--13}.
\newblock


\bibitem[\protect\citeauthoryear{Majdik, Verda, Albers-Schoenberg, and
  Scaramuzza}{Majdik et~al\mbox{.}}{2015}]%
        {majdik2015air}
\bibfield{author}{\bibinfo{person}{Andr{\'a}s~L Majdik},
  \bibinfo{person}{Damiano Verda}, \bibinfo{person}{Yves Albers-Schoenberg},
  {and} \bibinfo{person}{Davide Scaramuzza}.} \bibinfo{year}{2015}\natexlab{}.
\newblock \showarticletitle{Air-ground matching: Appearance-based GPS-denied
  urban localization of micro aerial vehicles}.
\newblock \bibinfo{journal}{\emph{Journal of Field Robotics}}
  \bibinfo{volume}{32}, \bibinfo{number}{7} (\bibinfo{year}{2015}),
  \bibinfo{pages}{1015--1039}.
\newblock


\bibitem[\protect\citeauthoryear{Malavazi, Guyonneau, Fasquel, Lagrange, and
  Mercier}{Malavazi et~al\mbox{.}}{2018}]%
        {malavazi2018lidar}
\bibfield{author}{\bibinfo{person}{Flavio~BP Malavazi}, \bibinfo{person}{Remy
  Guyonneau}, \bibinfo{person}{Jean-Baptiste Fasquel},
  \bibinfo{person}{Sebastien Lagrange}, {and} \bibinfo{person}{Franck
  Mercier}.} \bibinfo{year}{2018}\natexlab{}.
\newblock \showarticletitle{LiDAR-only based navigation algorithm for an
  autonomous agricultural robot}.
\newblock \bibinfo{journal}{\emph{Computers and electronics in agriculture}}
  \bibinfo{volume}{154} (\bibinfo{year}{2018}), \bibinfo{pages}{71--79}.
\newblock


\bibitem[\protect\citeauthoryear{Massot-Campos and Oliver-Codina}{Massot-Campos
  and Oliver-Codina}{2015}]%
        {massot2015optical}
\bibfield{author}{\bibinfo{person}{Miquel Massot-Campos} {and}
  \bibinfo{person}{Gabriel Oliver-Codina}.} \bibinfo{year}{2015}\natexlab{}.
\newblock \showarticletitle{Optical sensors and methods for underwater 3D
  reconstruction}.
\newblock \bibinfo{journal}{\emph{Sensors}} \bibinfo{volume}{15},
  \bibinfo{number}{12} (\bibinfo{year}{2015}), \bibinfo{pages}{31525--31557}.
\newblock


\bibitem[\protect\citeauthoryear{Maurette}{Maurette}{2003}]%
        {maurette2003mars}
\bibfield{author}{\bibinfo{person}{Michel Maurette}.}
  \bibinfo{year}{2003}\natexlab{}.
\newblock \showarticletitle{Mars rover autonomous navigation}.
\newblock \bibinfo{journal}{\emph{Autonomous Robots}} \bibinfo{volume}{14},
  \bibinfo{number}{2} (\bibinfo{year}{2003}), \bibinfo{pages}{199--208}.
\newblock


\bibitem[\protect\citeauthoryear{Maye, Furgale, and Siegwart}{Maye
  et~al\mbox{.}}{2013}]%
        {maye2013self}
\bibfield{author}{\bibinfo{person}{J{\'e}r{\^o}me Maye}, \bibinfo{person}{Paul
  Furgale}, {and} \bibinfo{person}{Roland Siegwart}.}
  \bibinfo{year}{2013}\natexlab{}.
\newblock \showarticletitle{Self-supervised calibration for robotic systems}.
  In \bibinfo{booktitle}{\emph{2013 IEEE Intelligent Vehicles Symposium (IV)}}.
  IEEE, \bibinfo{pages}{473--480}.
\newblock


\bibitem[\protect\citeauthoryear{Mehdi, Mazhar, and Maurelli}{Mehdi
  et~al\mbox{.}}{2021}]%
        {mehdi2021autonomous}
\bibfield{author}{\bibinfo{person}{Syed~Atif Mehdi}, \bibinfo{person}{Suleman
  Mazhar}, {and} \bibinfo{person}{Francesco Maurelli}.}
  \bibinfo{year}{2021}\natexlab{}.
\newblock \showarticletitle{Autonomous navigation of low cost underwater
  vehicle for shallow freshwater applications}. In
  \bibinfo{booktitle}{\emph{OCEANS 2021: San Diego--Porto}}. IEEE,
  \bibinfo{pages}{1--4}.
\newblock


\bibitem[\protect\citeauthoryear{Meier, Tanskanen, Heng, Lee, Fraundorfer, and
  Pollefeys}{Meier et~al\mbox{.}}{2012}]%
        {meier2012pixhawk}
\bibfield{author}{\bibinfo{person}{Lorenz Meier}, \bibinfo{person}{Petri
  Tanskanen}, \bibinfo{person}{Lionel Heng}, \bibinfo{person}{Gim~Hee Lee},
  \bibinfo{person}{Friedrich Fraundorfer}, {and} \bibinfo{person}{Marc
  Pollefeys}.} \bibinfo{year}{2012}\natexlab{}.
\newblock \showarticletitle{PIXHAWK: A micro aerial vehicle design for
  autonomous flight using onboard computer vision}.
\newblock \bibinfo{journal}{\emph{Autonomous Robots}} \bibinfo{volume}{33},
  \bibinfo{number}{1} (\bibinfo{year}{2012}), \bibinfo{pages}{21--39}.
\newblock


\bibitem[\protect\citeauthoryear{Meyer and Filliat}{Meyer and Filliat}{2003}]%
        {meyer2003map}
\bibfield{author}{\bibinfo{person}{Jean-Arcady Meyer} {and}
  \bibinfo{person}{David Filliat}.} \bibinfo{year}{2003}\natexlab{}.
\newblock \showarticletitle{Map-based navigation in mobile robots:: Ii. a
  review of map-learning and path-planning strategies}.
\newblock \bibinfo{journal}{\emph{Cognitive Systems Research}}
  \bibinfo{volume}{4}, \bibinfo{number}{4} (\bibinfo{year}{2003}),
  \bibinfo{pages}{283--317}.
\newblock


\bibitem[\protect\citeauthoryear{Mishkin, Morrison, Nguyen, Stone, Cooper, and
  Wilcox}{Mishkin et~al\mbox{.}}{1998}]%
        {mishkin1998experiences}
\bibfield{author}{\bibinfo{person}{Andrew~H Mishkin}, \bibinfo{person}{Jack~C
  Morrison}, \bibinfo{person}{Tam~T Nguyen}, \bibinfo{person}{Henry~W Stone},
  \bibinfo{person}{Brian~K Cooper}, {and} \bibinfo{person}{Brian~H Wilcox}.}
  \bibinfo{year}{1998}\natexlab{}.
\newblock \showarticletitle{Experiences with operations and autonomy of the
  mars pathfinder microrover}. In \bibinfo{booktitle}{\emph{1998 IEEE aerospace
  conference proceedings (Cat. No. 98TH8339)}}, Vol.~\bibinfo{volume}{2}. IEEE,
  \bibinfo{pages}{337--351}.
\newblock


\bibitem[\protect\citeauthoryear{Mnih, Kavukcuoglu, Silver, Rusu, Veness,
  Bellemare, Graves, Riedmiller, Fidjeland, Ostrovski, et~al\mbox{.}}{Mnih
  et~al\mbox{.}}{2015}]%
        {mnih2015human}
\bibfield{author}{\bibinfo{person}{Volodymyr Mnih}, \bibinfo{person}{Koray
  Kavukcuoglu}, \bibinfo{person}{David Silver}, \bibinfo{person}{Andrei~A
  Rusu}, \bibinfo{person}{Joel Veness}, \bibinfo{person}{Marc~G Bellemare},
  \bibinfo{person}{Alex Graves}, \bibinfo{person}{Martin Riedmiller},
  \bibinfo{person}{Andreas~K Fidjeland}, \bibinfo{person}{Georg Ostrovski},
  {et~al\mbox{.}}} \bibinfo{year}{2015}\natexlab{}.
\newblock \showarticletitle{Human-level control through deep reinforcement
  learning}.
\newblock \bibinfo{journal}{\emph{nature}} \bibinfo{volume}{518},
  \bibinfo{number}{7540} (\bibinfo{year}{2015}), \bibinfo{pages}{529--533}.
\newblock


\bibitem[\protect\citeauthoryear{Mohamed, Haghbayan, Westerlund, Heikkonen,
  Tenhunen, and Plosila}{Mohamed et~al\mbox{.}}{2019}]%
        {mohamed2019survey}
\bibfield{author}{\bibinfo{person}{Sherif~AS Mohamed},
  \bibinfo{person}{Mohammad-Hashem Haghbayan}, \bibinfo{person}{Tomi
  Westerlund}, \bibinfo{person}{Jukka Heikkonen}, \bibinfo{person}{Hannu
  Tenhunen}, {and} \bibinfo{person}{Juha Plosila}.}
  \bibinfo{year}{2019}\natexlab{}.
\newblock \showarticletitle{A survey on odometry for autonomous navigation
  systems}.
\newblock \bibinfo{journal}{\emph{IEEE Access}}  \bibinfo{volume}{7}
  (\bibinfo{year}{2019}), \bibinfo{pages}{97466--97486}.
\newblock


\bibitem[\protect\citeauthoryear{Mohanty and Parhi}{Mohanty and Parhi}{2013}]%
        {mohanty2013controlling}
\bibfield{author}{\bibinfo{person}{Prases~K Mohanty} {and}
  \bibinfo{person}{Dayal~R Parhi}.} \bibinfo{year}{2013}\natexlab{}.
\newblock \showarticletitle{Controlling the motion of an autonomous mobile
  robot using various techniques: a review}.
\newblock \bibinfo{journal}{\emph{Journal of Advance Mechanical Engineering}}
  \bibinfo{volume}{1}, \bibinfo{number}{1} (\bibinfo{year}{2013}),
  \bibinfo{pages}{24--39}.
\newblock


\bibitem[\protect\citeauthoryear{Molton, Davison, and Reid}{Molton
  et~al\mbox{.}}{2004}]%
        {molton2004locally}
\bibfield{author}{\bibinfo{person}{Nicholas Molton}, \bibinfo{person}{Andrew~J
  Davison}, {and} \bibinfo{person}{Ian Reid}.} \bibinfo{year}{2004}\natexlab{}.
\newblock \showarticletitle{Locally planar patch features for real-time
  structure from motion.}. In \bibinfo{booktitle}{\emph{Bmvc}}.
  \bibinfo{pages}{1--10}.
\newblock


\bibitem[\protect\citeauthoryear{Monta, Kondo, and Shibano}{Monta
  et~al\mbox{.}}{1995}]%
        {525635}
\bibfield{author}{\bibinfo{person}{M. Monta}, \bibinfo{person}{N. Kondo}, {and}
  \bibinfo{person}{Y. Shibano}.} \bibinfo{year}{1995}\natexlab{}.
\newblock \showarticletitle{Agricultural robot in grape production system}. In
  \bibinfo{booktitle}{\emph{Proceedings of 1995 IEEE International Conference
  on Robotics and Automation}}, Vol.~\bibinfo{volume}{3}.
  \bibinfo{pages}{2504--2509 vol.3}.
\newblock
\urldef\tempurl%
\url{https://doi.org/10.1109/ROBOT.1995.525635}
\showDOI{\tempurl}


\bibitem[\protect\citeauthoryear{Montemerlo, Thrun, Koller, Wegbreit,
  et~al\mbox{.}}{Montemerlo et~al\mbox{.}}{2002}]%
        {montemerlo2002fastslam}
\bibfield{author}{\bibinfo{person}{Michael Montemerlo},
  \bibinfo{person}{Sebastian Thrun}, \bibinfo{person}{Daphne Koller},
  \bibinfo{person}{Ben Wegbreit}, {et~al\mbox{.}}}
  \bibinfo{year}{2002}\natexlab{}.
\newblock \showarticletitle{FastSLAM: A factored solution to the simultaneous
  localization and mapping problem}.
\newblock \bibinfo{journal}{\emph{Aaai/iaai}}  \bibinfo{volume}{593598}
  (\bibinfo{year}{2002}).
\newblock


\bibitem[\protect\citeauthoryear{Morales and Kassas}{Morales and
  Kassas}{2021}]%
        {morales2021tightly}
\bibfield{author}{\bibinfo{person}{Joshua~J Morales} {and}
  \bibinfo{person}{Zaher~M Kassas}.} \bibinfo{year}{2021}\natexlab{}.
\newblock \showarticletitle{Tightly coupled inertial navigation system with
  signals of opportunity aiding}.
\newblock \bibinfo{journal}{\emph{IEEE Trans. Aerospace Electron. Systems}}
  \bibinfo{volume}{57}, \bibinfo{number}{3} (\bibinfo{year}{2021}),
  \bibinfo{pages}{1930--1948}.
\newblock


\bibitem[\protect\citeauthoryear{Moravec and Elfes}{Moravec and Elfes}{1985}]%
        {moravec1985high}
\bibfield{author}{\bibinfo{person}{Hans Moravec} {and} \bibinfo{person}{Alberto
  Elfes}.} \bibinfo{year}{1985}\natexlab{}.
\newblock \showarticletitle{High resolution maps from wide angle sonar}. In
  \bibinfo{booktitle}{\emph{Proceedings. 1985 IEEE international conference on
  robotics and automation}}, Vol.~\bibinfo{volume}{2}. IEEE,
  \bibinfo{pages}{116--121}.
\newblock


\bibitem[\protect\citeauthoryear{Moreno, Armingol, De~La~Escalera, and
  Salichs}{Moreno et~al\mbox{.}}{1999}]%
        {moreno1999global}
\bibfield{author}{\bibinfo{person}{L Moreno}, \bibinfo{person}{JM Armingol},
  \bibinfo{person}{A De~La~Escalera}, {and} \bibinfo{person}{MA Salichs}.}
  \bibinfo{year}{1999}\natexlab{}.
\newblock \showarticletitle{Global integration of ultrasonic sensors
  information in mobile robot localization}. In
  \bibinfo{booktitle}{\emph{Proceedings of the Ninth International Conference
  on Advanced Robotics}}. Tokyo, Japan, \bibinfo{pages}{283--288}.
\newblock


\bibitem[\protect\citeauthoryear{Moreno, Armingol, Garrido, De~La~Escalera, and
  Salichs}{Moreno et~al\mbox{.}}{2002}]%
        {moreno2002genetic}
\bibfield{author}{\bibinfo{person}{Luis Moreno}, \bibinfo{person}{Jose~M
  Armingol}, \bibinfo{person}{Santiago Garrido}, \bibinfo{person}{Arturo
  De~La~Escalera}, {and} \bibinfo{person}{Miguel~A Salichs}.}
  \bibinfo{year}{2002}\natexlab{}.
\newblock \showarticletitle{A genetic algorithm for mobile robot localization
  using ultrasonic sensors}.
\newblock \bibinfo{journal}{\emph{Journal of Intelligent and Robotic Systems}}
  \bibinfo{volume}{34}, \bibinfo{number}{2} (\bibinfo{year}{2002}),
  \bibinfo{pages}{135--154}.
\newblock


\bibitem[\protect\citeauthoryear{Mourikis, Roumeliotis, et~al\mbox{.}}{Mourikis
  et~al\mbox{.}}{2007}]%
        {mourikis2007multi}
\bibfield{author}{\bibinfo{person}{Anastasios~I Mourikis},
  \bibinfo{person}{Stergios~I Roumeliotis}, {et~al\mbox{.}}}
  \bibinfo{year}{2007}\natexlab{}.
\newblock \showarticletitle{A Multi-State Constraint Kalman Filter for
  Vision-aided Inertial Navigation.}. In \bibinfo{booktitle}{\emph{ICRA}},
  Vol.~\bibinfo{volume}{2}. \bibinfo{pages}{6}.
\newblock


\bibitem[\protect\citeauthoryear{Mur-Artal, Montiel, and Tardos}{Mur-Artal
  et~al\mbox{.}}{2015}]%
        {mur2015orb}
\bibfield{author}{\bibinfo{person}{Raul Mur-Artal}, \bibinfo{person}{Jose
  Maria~Martinez Montiel}, {and} \bibinfo{person}{Juan~D Tardos}.}
  \bibinfo{year}{2015}\natexlab{}.
\newblock \showarticletitle{ORB-SLAM: a versatile and accurate monocular SLAM
  system}.
\newblock \bibinfo{journal}{\emph{IEEE transactions on robotics}}
  \bibinfo{volume}{31}, \bibinfo{number}{5} (\bibinfo{year}{2015}),
  \bibinfo{pages}{1147--1163}.
\newblock


\bibitem[\protect\citeauthoryear{Mur-Artal and Tard{\'o}s}{Mur-Artal and
  Tard{\'o}s}{2017a}]%
        {mur2017orb}
\bibfield{author}{\bibinfo{person}{Raul Mur-Artal} {and}
  \bibinfo{person}{Juan~D Tard{\'o}s}.} \bibinfo{year}{2017}\natexlab{a}.
\newblock \showarticletitle{Orb-slam2: An open-source slam system for
  monocular, stereo, and rgb-d cameras}.
\newblock \bibinfo{journal}{\emph{IEEE transactions on robotics}}
  \bibinfo{volume}{33}, \bibinfo{number}{5} (\bibinfo{year}{2017}),
  \bibinfo{pages}{1255--1262}.
\newblock


\bibitem[\protect\citeauthoryear{Mur-Artal and Tard{\'o}s}{Mur-Artal and
  Tard{\'o}s}{2017b}]%
        {mur2017visual}
\bibfield{author}{\bibinfo{person}{Ra{\'u}l Mur-Artal} {and}
  \bibinfo{person}{Juan~D Tard{\'o}s}.} \bibinfo{year}{2017}\natexlab{b}.
\newblock \showarticletitle{Visual-inertial monocular SLAM with map reuse}.
\newblock \bibinfo{journal}{\emph{IEEE Robotics and Automation Letters}}
  \bibinfo{volume}{2}, \bibinfo{number}{2} (\bibinfo{year}{2017}),
  \bibinfo{pages}{796--803}.
\newblock


\bibitem[\protect\citeauthoryear{Musso, Oudjane, and Gland}{Musso
  et~al\mbox{.}}{2001}]%
        {musso2001improving}
\bibfield{author}{\bibinfo{person}{Christian Musso}, \bibinfo{person}{Nadia
  Oudjane}, {and} \bibinfo{person}{Francois~Le Gland}.}
  \bibinfo{year}{2001}\natexlab{}.
\newblock \showarticletitle{Improving regularised particle filters}.
\newblock In \bibinfo{booktitle}{\emph{Sequential Monte Carlo methods in
  practice}}. \bibinfo{publisher}{Springer}, \bibinfo{pages}{247--271}.
\newblock


\bibitem[\protect\citeauthoryear{Nahavandi, Mohamed, Hossain, Nahavandi,
  Salaken, Rokonuzzaman, Ayoub, and Smith}{Nahavandi et~al\mbox{.}}{2022}]%
        {nahavandi2022autonomous}
\bibfield{author}{\bibinfo{person}{Saeid Nahavandi}, \bibinfo{person}{Shady
  Mohamed}, \bibinfo{person}{Ibrahim Hossain}, \bibinfo{person}{Darius
  Nahavandi}, \bibinfo{person}{Syed~Moshfeq Salaken}, \bibinfo{person}{Mohammad
  Rokonuzzaman}, \bibinfo{person}{Rachael Ayoub}, {and} \bibinfo{person}{Robin
  Smith}.} \bibinfo{year}{2022}\natexlab{}.
\newblock \showarticletitle{Autonomous Convoying: A Survey on Current Research
  and Development}.
\newblock \bibinfo{journal}{\emph{IEEE Access}} (\bibinfo{year}{2022}).
\newblock


\bibitem[\protect\citeauthoryear{Naseer, Ruhnke, Stachniss, Spinello, and
  Burgard}{Naseer et~al\mbox{.}}{2015}]%
        {naseer2015robust}
\bibfield{author}{\bibinfo{person}{Tayyab Naseer}, \bibinfo{person}{Michael
  Ruhnke}, \bibinfo{person}{Cyrill Stachniss}, \bibinfo{person}{Luciano
  Spinello}, {and} \bibinfo{person}{Wolfram Burgard}.}
  \bibinfo{year}{2015}\natexlab{}.
\newblock \showarticletitle{Robust visual SLAM across seasons}. In
  \bibinfo{booktitle}{\emph{2015 IEEE/RSJ International Conference on
  Intelligent Robots and Systems (IROS)}}. IEEE, \bibinfo{pages}{2529--2535}.
\newblock


\bibitem[\protect\citeauthoryear{Neira and Tardos}{Neira and Tardos}{2001}]%
        {neira2001data}
\bibfield{author}{\bibinfo{person}{Jose Neira} {and} \bibinfo{person}{Juan~D
  Tardos}.} \bibinfo{year}{2001}\natexlab{}.
\newblock \showarticletitle{Data association in stochastic mapping using the
  joint compatibility test}.
\newblock \bibinfo{journal}{\emph{IEEE Transactions on robotics and
  automation}} \bibinfo{volume}{17}, \bibinfo{number}{6}
  (\bibinfo{year}{2001}), \bibinfo{pages}{890--897}.
\newblock


\bibitem[\protect\citeauthoryear{Nemra and Aouf}{Nemra and Aouf}{2010}]%
        {nemra2010robust}
\bibfield{author}{\bibinfo{person}{Abdelkrim Nemra} {and}
  \bibinfo{person}{Nabil Aouf}.} \bibinfo{year}{2010}\natexlab{}.
\newblock \showarticletitle{Robust INS/GPS sensor fusion for UAV localization
  using SDRE nonlinear filtering}.
\newblock \bibinfo{journal}{\emph{IEEE Sensors Journal}} \bibinfo{volume}{10},
  \bibinfo{number}{4} (\bibinfo{year}{2010}), \bibinfo{pages}{789--798}.
\newblock


\bibitem[\protect\citeauthoryear{Nesnas, Fesq, and Volpe}{Nesnas
  et~al\mbox{.}}{2021}]%
        {nesnas2021autonomy}
\bibfield{author}{\bibinfo{person}{Issa~AD Nesnas}, \bibinfo{person}{Lorraine~M
  Fesq}, {and} \bibinfo{person}{Richard~A Volpe}.}
  \bibinfo{year}{2021}\natexlab{}.
\newblock \showarticletitle{Autonomy for Space Robots: Past, Present, and
  Future}.
\newblock \bibinfo{journal}{\emph{Current Robotics Reports}}
  \bibinfo{volume}{2}, \bibinfo{number}{3} (\bibinfo{year}{2021}),
  \bibinfo{pages}{251--263}.
\newblock


\bibitem[\protect\citeauthoryear{Ni, Wu, Fan, and Yang}{Ni
  et~al\mbox{.}}{2016}]%
        {ni2016bioinspired}
\bibfield{author}{\bibinfo{person}{Jianjun Ni}, \bibinfo{person}{Liuying Wu},
  \bibinfo{person}{Xinnan Fan}, {and} \bibinfo{person}{Simon~X Yang}.}
  \bibinfo{year}{2016}\natexlab{}.
\newblock \showarticletitle{Bioinspired intelligent algorithm and its
  applications for mobile robot control: a survey}.
\newblock \bibinfo{journal}{\emph{Computational intelligence and neuroscience}}
   \bibinfo{volume}{2016} (\bibinfo{year}{2016}).
\newblock


\bibitem[\protect\citeauthoryear{Niloy, Shama, Chakrabortty, Ryan, Badal,
  Tasneem, Ahamed, Moyeen, Das, Ali, et~al\mbox{.}}{Niloy
  et~al\mbox{.}}{2021}]%
        {niloy2021critical}
\bibfield{author}{\bibinfo{person}{Md~AK Niloy}, \bibinfo{person}{Anika Shama},
  \bibinfo{person}{Ripon~K Chakrabortty}, \bibinfo{person}{Michael~J Ryan},
  \bibinfo{person}{Faisal~R Badal}, \bibinfo{person}{Zinat Tasneem},
  \bibinfo{person}{Md~H Ahamed}, \bibinfo{person}{Sumaya~I Moyeen},
  \bibinfo{person}{Sajal~K Das}, \bibinfo{person}{Md~F Ali}, {et~al\mbox{.}}}
  \bibinfo{year}{2021}\natexlab{}.
\newblock \showarticletitle{Critical design and control issues of indoor
  autonomous mobile robots: A review}.
\newblock \bibinfo{journal}{\emph{IEEE Access}}  \bibinfo{volume}{9}
  (\bibinfo{year}{2021}), \bibinfo{pages}{35338--35370}.
\newblock


\bibitem[\protect\citeauthoryear{Nilsson et~al\mbox{.}}{Nilsson
  et~al\mbox{.}}{1984}]%
        {nilsson1984shakey}
\bibfield{author}{\bibinfo{person}{Nils~J Nilsson} {et~al\mbox{.}}}
  \bibinfo{year}{1984}\natexlab{}.
\newblock \showarticletitle{Shakey the robot}.
\newblock  (\bibinfo{year}{1984}).
\newblock


\bibitem[\protect\citeauthoryear{Nocks}{Nocks}{2007}]%
        {nocks2007robot}
\bibfield{author}{\bibinfo{person}{Lisa Nocks}.}
  \bibinfo{year}{2007}\natexlab{}.
\newblock \bibinfo{booktitle}{\emph{The robot: the life story of a
  technology}}.
\newblock \bibinfo{publisher}{Greenwood Publishing Group}.
\newblock


\bibitem[\protect\citeauthoryear{Oagana}{Oagana}{2013}]%
        {benz-autonomous-driving}
\bibfield{author}{\bibinfo{person}{Alex Oagana}.}
  \bibinfo{year}{2013}\natexlab{}.
\newblock \bibinfo{title}{A Short History of Mercedes-Benz Autonomous Driving
  Technology}.
\newblock \bibinfo{howpublished}{url =
  {https://www.autoevolution.com/news/a-short-history-of-mercedes-benz-autonomous-driving-technology-68148.html}}.
    (\bibinfo{year}{2013}).
\newblock
\newblock
\shownote{accessed February 27, 2022.}


\bibitem[\protect\citeauthoryear{Organization}{Organization}{2018}]%
        {international2018working}
\bibfield{author}{\bibinfo{person}{International~Maritime Organization}.}
  \bibinfo{year}{2018}\natexlab{}.
\newblock \bibinfo{title}{Working Group Report in 100th Session of IMO Maritime
  Safety Committee for the Regulatory Scoping Exercise for the Use of Maritime
  Autonomous Surface Ships (MASS)}.
\newblock   (\bibinfo{year}{2018}).
\newblock


\bibitem[\protect\citeauthoryear{Oth, Furgale, Kneip, and Siegwart}{Oth
  et~al\mbox{.}}{2013}]%
        {oth2013rolling}
\bibfield{author}{\bibinfo{person}{Luc Oth}, \bibinfo{person}{Paul Furgale},
  \bibinfo{person}{Laurent Kneip}, {and} \bibinfo{person}{Roland Siegwart}.}
  \bibinfo{year}{2013}\natexlab{}.
\newblock \showarticletitle{Rolling shutter camera calibration}. In
  \bibinfo{booktitle}{\emph{Proceedings of the IEEE Conference on Computer
  Vision and Pattern Recognition}}. \bibinfo{pages}{1360--1367}.
\newblock


\bibitem[\protect\citeauthoryear{Pandey, Pandey, and Parhi}{Pandey
  et~al\mbox{.}}{2017}]%
        {pandey2017mobile}
\bibfield{author}{\bibinfo{person}{Anish Pandey}, \bibinfo{person}{Shalini
  Pandey}, {and} \bibinfo{person}{DR Parhi}.} \bibinfo{year}{2017}\natexlab{}.
\newblock \showarticletitle{Mobile robot navigation and obstacle avoidance
  techniques: A review}.
\newblock \bibinfo{journal}{\emph{Int Rob Auto J}} \bibinfo{volume}{2},
  \bibinfo{number}{3} (\bibinfo{year}{2017}), \bibinfo{pages}{00022}.
\newblock


\bibitem[\protect\citeauthoryear{Pandey, McBride, and Eustice}{Pandey
  et~al\mbox{.}}{2011}]%
        {gpandey-2011a}
\bibfield{author}{\bibinfo{person}{Gaurav Pandey}, \bibinfo{person}{James~R.
  McBride}, {and} \bibinfo{person}{Ryan~M. Eustice}.}
  \bibinfo{year}{2011}\natexlab{}.
\newblock \showarticletitle{Ford campus vision and lidar data set}.
\newblock \bibinfo{journal}{\emph{International Journal of Robotics Research}}
  \bibinfo{volume}{30}, \bibinfo{number}{13} (\bibinfo{year}{2011}),
  \bibinfo{pages}{1543--1552}.
\newblock


\bibitem[\protect\citeauthoryear{Panigrahi and Bisoy}{Panigrahi and
  Bisoy}{2021}]%
        {panigrahi2021localization}
\bibfield{author}{\bibinfo{person}{Prabin~Kumar Panigrahi} {and}
  \bibinfo{person}{Sukant~Kishoro Bisoy}.} \bibinfo{year}{2021}\natexlab{}.
\newblock \showarticletitle{Localization strategies for autonomous mobile
  robots: A review}.
\newblock \bibinfo{journal}{\emph{Journal of King Saud University-Computer and
  Information Sciences}} (\bibinfo{year}{2021}).
\newblock


\bibitem[\protect\citeauthoryear{Patil, Ansari, Tendulkar, Bhatlekar, Pawar,
  and Aswale}{Patil et~al\mbox{.}}{2020}]%
        {patil2020survey}
\bibfield{author}{\bibinfo{person}{Deepak Patil}, \bibinfo{person}{Munsaf
  Ansari}, \bibinfo{person}{Dilisha Tendulkar}, \bibinfo{person}{Ritesh
  Bhatlekar}, \bibinfo{person}{Vijaykumar~Naik Pawar}, {and}
  \bibinfo{person}{Shailendra Aswale}.} \bibinfo{year}{2020}\natexlab{}.
\newblock \showarticletitle{A survey on autonomous military service robot}. In
  \bibinfo{booktitle}{\emph{2020 International Conference on Emerging Trends in
  Information Technology and Engineering (ic-ETITE)}}. IEEE,
  \bibinfo{pages}{1--7}.
\newblock


\bibitem[\protect\citeauthoryear{Pitt and Shephard}{Pitt and Shephard}{1999}]%
        {pitt1999filtering}
\bibfield{author}{\bibinfo{person}{Michael~K Pitt} {and} \bibinfo{person}{Neil
  Shephard}.} \bibinfo{year}{1999}\natexlab{}.
\newblock \showarticletitle{Filtering via simulation: Auxiliary particle
  filters}.
\newblock \bibinfo{journal}{\emph{Journal of the American statistical
  association}} \bibinfo{volume}{94}, \bibinfo{number}{446}
  (\bibinfo{year}{1999}), \bibinfo{pages}{590--599}.
\newblock


\bibitem[\protect\citeauthoryear{Placed, Strader, Carrillo, Atanasov, Indelman,
  Carlone, and Castellanos}{Placed et~al\mbox{.}}{2022}]%
        {placed2022survey}
\bibfield{author}{\bibinfo{person}{Julio~A Placed}, \bibinfo{person}{Jared
  Strader}, \bibinfo{person}{Henry Carrillo}, \bibinfo{person}{Nikolay
  Atanasov}, \bibinfo{person}{Vadim Indelman}, \bibinfo{person}{Luca Carlone},
  {and} \bibinfo{person}{Jos{\'e}~A Castellanos}.}
  \bibinfo{year}{2022}\natexlab{}.
\newblock \showarticletitle{A Survey on Active Simultaneous Localization and
  Mapping: State of the Art and New Frontiers}.
\newblock \bibinfo{journal}{\emph{arXiv preprint arXiv:2207.00254}}
  (\bibinfo{year}{2022}).
\newblock


\bibitem[\protect\citeauthoryear{Pol and Murugan}{Pol and Murugan}{2015}]%
        {pol2015review}
\bibfield{author}{\bibinfo{person}{Rahul~Shivaji Pol} {and} \bibinfo{person}{M
  Murugan}.} \bibinfo{year}{2015}\natexlab{}.
\newblock \showarticletitle{A review on indoor human aware autonomous mobile
  robot navigation through a dynamic environment survey of different path
  planning algorithm and methods}. In \bibinfo{booktitle}{\emph{2015
  International conference on industrial instrumentation and control (ICIC)}}.
  IEEE, \bibinfo{pages}{1339--1344}.
\newblock


\bibitem[\protect\citeauthoryear{Pomerleau}{Pomerleau}{1994}]%
        {pomerleau1994defense}
\bibfield{author}{\bibinfo{person}{DA Pomerleau}.}
  \bibinfo{year}{1994}\natexlab{}.
\newblock \showarticletitle{Defense and civilian applications of the alvinn
  robot driving system}. In \bibinfo{booktitle}{\emph{Proceedings of 1994
  Government Microcircuit Applications Conference}}. \bibinfo{pages}{358--362}.
\newblock


\bibitem[\protect\citeauthoryear{Qin, Huang, Cao, and Zhang}{Qin
  et~al\mbox{.}}{2018}]%
        {qin2018loop}
\bibfield{author}{\bibinfo{person}{Hao Qin}, \bibinfo{person}{May Huang},
  \bibinfo{person}{Jian Cao}, {and} \bibinfo{person}{Xing Zhang}.}
  \bibinfo{year}{2018}\natexlab{}.
\newblock \showarticletitle{Loop closure detection in SLAM by combining visual
  CNN features and submaps}. In \bibinfo{booktitle}{\emph{2018 4th
  International Conference on Control, Automation and Robotics (ICCAR)}}. IEEE,
  \bibinfo{pages}{426--430}.
\newblock


\bibitem[\protect\citeauthoryear{Qin, Yang, Li, Zhong, and Zhang}{Qin
  et~al\mbox{.}}{2022}]%
        {qin2022real}
\bibfield{author}{\bibinfo{person}{Jiangying Qin}, \bibinfo{person}{Ke Yang},
  \bibinfo{person}{Ming Li}, \bibinfo{person}{Jiageng Zhong}, {and}
  \bibinfo{person}{Hanqi Zhang}.} \bibinfo{year}{2022}\natexlab{}.
\newblock \showarticletitle{Real-Time Positioning and Tracking for Vision-Based
  Unmanned Underwater Vehicles}.
\newblock \bibinfo{journal}{\emph{The International Archives of Photogrammetry,
  Remote Sensing and Spatial Information Sciences}}  \bibinfo{volume}{46}
  (\bibinfo{year}{2022}), \bibinfo{pages}{163--168}.
\newblock


\bibitem[\protect\citeauthoryear{Ramalingam and Reps}{Ramalingam and
  Reps}{1996}]%
        {ramalingam1996incremental}
\bibfield{author}{\bibinfo{person}{Ganesan Ramalingam} {and}
  \bibinfo{person}{Thomas Reps}.} \bibinfo{year}{1996}\natexlab{}.
\newblock \showarticletitle{An incremental algorithm for a generalization of
  the shortest-path problem}.
\newblock \bibinfo{journal}{\emph{Journal of Algorithms}} \bibinfo{volume}{21},
  \bibinfo{number}{2} (\bibinfo{year}{1996}), \bibinfo{pages}{267--305}.
\newblock


\bibitem[\protect\citeauthoryear{Reed, Wood, and Haworth}{Reed
  et~al\mbox{.}}{2010}]%
        {reed2010detection}
\bibfield{author}{\bibinfo{person}{Scott Reed}, \bibinfo{person}{Jon Wood},
  {and} \bibinfo{person}{Chris Haworth}.} \bibinfo{year}{2010}\natexlab{}.
\newblock \showarticletitle{The detection and disposal of IED devices within
  harbor regions using AUVs, smart ROVs and data processing/fusion technology}.
  In \bibinfo{booktitle}{\emph{2010 International WaterSide Security
  Conference}}. IEEE, \bibinfo{pages}{1--7}.
\newblock


\bibitem[\protect\citeauthoryear{Rehder, Nikolic, Schneider, Hinzmann, and
  Siegwart}{Rehder et~al\mbox{.}}{2016}]%
        {rehder2016extending}
\bibfield{author}{\bibinfo{person}{Joern Rehder}, \bibinfo{person}{Janosch
  Nikolic}, \bibinfo{person}{Thomas Schneider}, \bibinfo{person}{Timo
  Hinzmann}, {and} \bibinfo{person}{Roland Siegwart}.}
  \bibinfo{year}{2016}\natexlab{}.
\newblock \showarticletitle{Extending kalibr: Calibrating the extrinsics of
  multiple IMUs and of individual axes}. In \bibinfo{booktitle}{\emph{2016 IEEE
  International Conference on Robotics and Automation (ICRA)}}. IEEE,
  \bibinfo{pages}{4304--4311}.
\newblock


\bibitem[\protect\citeauthoryear{Rice, Kelmenson, and Mendelsohn}{Rice
  et~al\mbox{.}}{2004}]%
        {rice2004geophysical}
\bibfield{author}{\bibinfo{person}{Hugh Rice}, \bibinfo{person}{Steven
  Kelmenson}, {and} \bibinfo{person}{Louis Mendelsohn}.}
  \bibinfo{year}{2004}\natexlab{}.
\newblock \showarticletitle{Geophysical navigation technologies and
  applications}. In \bibinfo{booktitle}{\emph{PLANS 2004. Position Location and
  Navigation Symposium (IEEE Cat. No. 04CH37556)}}. IEEE,
  \bibinfo{pages}{618--624}.
\newblock


\bibitem[\protect\citeauthoryear{Ridao, Carreras, Ribas, Sanz, and
  Oliver}{Ridao et~al\mbox{.}}{2015}]%
        {ridao2015intervention}
\bibfield{author}{\bibinfo{person}{Pere Ridao}, \bibinfo{person}{Marc
  Carreras}, \bibinfo{person}{David Ribas}, \bibinfo{person}{Pedro~J Sanz},
  {and} \bibinfo{person}{Gabriel Oliver}.} \bibinfo{year}{2015}\natexlab{}.
\newblock \showarticletitle{Intervention AUVs: the next challenge}.
\newblock \bibinfo{journal}{\emph{Annual Reviews in Control}}
  \bibinfo{volume}{40} (\bibinfo{year}{2015}), \bibinfo{pages}{227--241}.
\newblock


\bibitem[\protect\citeauthoryear{Riisgaard and Blas}{Riisgaard and
  Blas}{2003}]%
        {riisgaard2003slam}
\bibfield{author}{\bibinfo{person}{S{\o}ren Riisgaard} {and}
  \bibinfo{person}{Morten~Rufus Blas}.} \bibinfo{year}{2003}\natexlab{}.
\newblock \showarticletitle{SLAM for Dummies}.
\newblock \bibinfo{journal}{\emph{A Tutorial Approach to Simultaneous
  Localization and Mapping}} \bibinfo{volume}{22}, \bibinfo{number}{1-127}
  (\bibinfo{year}{2003}), \bibinfo{pages}{126}.
\newblock


\bibitem[\protect\citeauthoryear{Rosten and Drummond}{Rosten and
  Drummond}{2006}]%
        {rosten2006machine}
\bibfield{author}{\bibinfo{person}{Edward Rosten} {and} \bibinfo{person}{Tom
  Drummond}.} \bibinfo{year}{2006}\natexlab{}.
\newblock \showarticletitle{Machine learning for high-speed corner detection}.
  In \bibinfo{booktitle}{\emph{European conference on computer vision}}.
  Springer, \bibinfo{pages}{430--443}.
\newblock


\bibitem[\protect\citeauthoryear{Roussillon, Gonzalez, Sola, Codol, Mansard,
  Lacroix, and Devy}{Roussillon et~al\mbox{.}}{2011}]%
        {roussillon2011rt}
\bibfield{author}{\bibinfo{person}{Cyril Roussillon}, \bibinfo{person}{Aurelien
  Gonzalez}, \bibinfo{person}{Joan Sola}, \bibinfo{person}{Jean-Marie Codol},
  \bibinfo{person}{Nicolas Mansard}, \bibinfo{person}{Simon Lacroix}, {and}
  \bibinfo{person}{Michel Devy}.} \bibinfo{year}{2011}\natexlab{}.
\newblock \showarticletitle{RT-SLAM: a generic and real-time visual SLAM
  implementation}. In \bibinfo{booktitle}{\emph{International Conference on
  Computer Vision Systems}}. Springer, \bibinfo{pages}{31--40}.
\newblock


\bibitem[\protect\citeauthoryear{Rublee, Rabaud, Konolige, and Bradski}{Rublee
  et~al\mbox{.}}{2011}]%
        {rublee2011orb}
\bibfield{author}{\bibinfo{person}{Ethan Rublee}, \bibinfo{person}{Vincent
  Rabaud}, \bibinfo{person}{Kurt Konolige}, {and} \bibinfo{person}{Gary
  Bradski}.} \bibinfo{year}{2011}\natexlab{}.
\newblock \showarticletitle{ORB: An efficient alternative to SIFT or SURF}. In
  \bibinfo{booktitle}{\emph{2011 International conference on computer vision}}.
  Ieee, \bibinfo{pages}{2564--2571}.
\newblock


\bibitem[\protect\citeauthoryear{Russell and Norvig}{Russell and
  Norvig}{2009}]%
        {russell2016artificial}
\bibfield{author}{\bibinfo{person}{Stuart~J. Russell} {and}
  \bibinfo{person}{Peter Norvig}.} \bibinfo{year}{2009}\natexlab{}.
\newblock \bibinfo{booktitle}{\emph{Artificial Intelligence: a modern approach}
  (\bibinfo{edition}{3} ed.)}.
\newblock \bibinfo{publisher}{Pearson}.
\newblock


\bibitem[\protect\citeauthoryear{S{\'a}nchez-Ib{\'a}{\~n}ez, P{\'e}rez-del
  Pulgar, and Garc{\'\i}a-Cerezo}{S{\'a}nchez-Ib{\'a}{\~n}ez
  et~al\mbox{.}}{2021}]%
        {sanchez2021path}
\bibfield{author}{\bibinfo{person}{Jos{\'e}~Ricardo
  S{\'a}nchez-Ib{\'a}{\~n}ez}, \bibinfo{person}{Carlos~J P{\'e}rez-del Pulgar},
  {and} \bibinfo{person}{Alfonso Garc{\'\i}a-Cerezo}.}
  \bibinfo{year}{2021}\natexlab{}.
\newblock \showarticletitle{Path Planning for Autonomous Mobile Robots: A
  Review}.
\newblock \bibinfo{journal}{\emph{Sensors}} \bibinfo{volume}{21},
  \bibinfo{number}{23} (\bibinfo{year}{2021}), \bibinfo{pages}{7898}.
\newblock


\bibitem[\protect\citeauthoryear{Sayre-McCord, Guerra, Antonini, Arneberg,
  Brown, Cavalheiro, Fang, Gorodetsky, McCoy, Quilter,
  et~al\mbox{.}}{Sayre-McCord et~al\mbox{.}}{2018}]%
        {sayre2018visual}
\bibfield{author}{\bibinfo{person}{Thomas Sayre-McCord},
  \bibinfo{person}{Winter Guerra}, \bibinfo{person}{Amado Antonini},
  \bibinfo{person}{Jasper Arneberg}, \bibinfo{person}{Austin Brown},
  \bibinfo{person}{Guilherme Cavalheiro}, \bibinfo{person}{Yajun Fang},
  \bibinfo{person}{Alex Gorodetsky}, \bibinfo{person}{Dave McCoy},
  \bibinfo{person}{Sebastian Quilter}, {et~al\mbox{.}}}
  \bibinfo{year}{2018}\natexlab{}.
\newblock \showarticletitle{Visual-inertial navigation algorithm development
  using photorealistic camera simulation in the loop}. In
  \bibinfo{booktitle}{\emph{2018 IEEE International Conference on Robotics and
  Automation (ICRA)}}. IEEE, \bibinfo{pages}{2566--2573}.
\newblock


\bibitem[\protect\citeauthoryear{Schaal}{Schaal}{2006}]%
        {schaal2006dynamic}
\bibfield{author}{\bibinfo{person}{Stefan Schaal}.}
  \bibinfo{year}{2006}\natexlab{}.
\newblock \showarticletitle{Dynamic movement primitives-a framework for motor
  control in humans and humanoid robotics}.
\newblock In \bibinfo{booktitle}{\emph{Adaptive motion of animals and
  machines}}. \bibinfo{publisher}{Springer}, \bibinfo{pages}{261--280}.
\newblock


\bibitem[\protect\citeauthoryear{Sch{\"a}fer and Berns}{Sch{\"a}fer and
  Berns}{2006}]%
        {schafer2006ravon}
\bibfield{author}{\bibinfo{person}{H Sch{\"a}fer} {and} \bibinfo{person}{K
  Berns}.} \bibinfo{year}{2006}\natexlab{}.
\newblock \showarticletitle{Ravon-an autonomous vehicle for risky intervention
  and surveillance}. In \bibinfo{booktitle}{\emph{International Workshop on
  Robotics for risky intervention and environmental surveillance-RISE}}.
  Citeseer.
\newblock


\bibitem[\protect\citeauthoryear{Schiele and Crowley}{Schiele and
  Crowley}{1996}]%
        {schiele1996object}
\bibfield{author}{\bibinfo{person}{Bernt Schiele} {and}
  \bibinfo{person}{James~L Crowley}.} \bibinfo{year}{1996}\natexlab{}.
\newblock \showarticletitle{Object recognition using multidimensional receptive
  field histograms}. In \bibinfo{booktitle}{\emph{European Conference on
  Computer Vision}}. Springer, \bibinfo{pages}{610--619}.
\newblock


\bibitem[\protect\citeauthoryear{Schops, Sattler, and Pollefeys}{Schops
  et~al\mbox{.}}{2019}]%
        {schops2019bad}
\bibfield{author}{\bibinfo{person}{Thomas Schops}, \bibinfo{person}{Torsten
  Sattler}, {and} \bibinfo{person}{Marc Pollefeys}.}
  \bibinfo{year}{2019}\natexlab{}.
\newblock \showarticletitle{Bad slam: Bundle adjusted direct rgb-d slam}. In
  \bibinfo{booktitle}{\emph{Proceedings of the IEEE/CVF Conference on Computer
  Vision and Pattern Recognition}}. \bibinfo{pages}{134--144}.
\newblock


\bibitem[\protect\citeauthoryear{Schulman, Wolski, Dhariwal, Radford, and
  Klimov}{Schulman et~al\mbox{.}}{2017}]%
        {schulman2017proximal}
\bibfield{author}{\bibinfo{person}{John Schulman}, \bibinfo{person}{Filip
  Wolski}, \bibinfo{person}{Prafulla Dhariwal}, \bibinfo{person}{Alec Radford},
  {and} \bibinfo{person}{Oleg Klimov}.} \bibinfo{year}{2017}\natexlab{}.
\newblock \showarticletitle{Proximal policy optimization algorithms}.
\newblock \bibinfo{journal}{\emph{arXiv preprint arXiv:1707.06347}}
  (\bibinfo{year}{2017}).
\newblock


\bibitem[\protect\citeauthoryear{Sgorbissa and Zaccaria}{Sgorbissa and
  Zaccaria}{2012}]%
        {sgorbissa2012planning}
\bibfield{author}{\bibinfo{person}{Antonio Sgorbissa} {and}
  \bibinfo{person}{Renato Zaccaria}.} \bibinfo{year}{2012}\natexlab{}.
\newblock \showarticletitle{Planning and obstacle avoidance in mobile
  robotics}.
\newblock \bibinfo{journal}{\emph{Robotics and Autonomous Systems}}
  \bibinfo{volume}{60}, \bibinfo{number}{4} (\bibinfo{year}{2012}),
  \bibinfo{pages}{628--638}.
\newblock


\bibitem[\protect\citeauthoryear{Shah, Eysenbach, Rhinehart, and Levine}{Shah
  et~al\mbox{.}}{2021}]%
        {shah2021rapid}
\bibfield{author}{\bibinfo{person}{Dhruv Shah}, \bibinfo{person}{Benjamin
  Eysenbach}, \bibinfo{person}{Nicholas Rhinehart}, {and}
  \bibinfo{person}{Sergey Levine}.} \bibinfo{year}{2021}\natexlab{}.
\newblock \showarticletitle{Rapid exploration for open-world navigation with
  latent goal models}.
\newblock \bibinfo{journal}{\emph{arXiv preprint arXiv:2104.05859}}
  (\bibinfo{year}{2021}).
\newblock


\bibitem[\protect\citeauthoryear{Shah and Levine}{Shah and Levine}{2022}]%
        {shah2022viking}
\bibfield{author}{\bibinfo{person}{Dhruv Shah} {and} \bibinfo{person}{Sergey
  Levine}.} \bibinfo{year}{2022}\natexlab{}.
\newblock \showarticletitle{ViKiNG: Vision-Based Kilometer-Scale Navigation
  with Geographic Hints}.
\newblock \bibinfo{journal}{\emph{arXiv preprint arXiv:2202.11271}}
  (\bibinfo{year}{2022}).
\newblock


\bibitem[\protect\citeauthoryear{Sharma, Shukla, Karki, Shukla, Behera, and
  K.S.}{Sharma et~al\mbox{.}}{2019}]%
        {8793911}
\bibfield{author}{\bibinfo{person}{Radhe~Shyam Sharma},
  \bibinfo{person}{Santosh Shukla}, \bibinfo{person}{Hamad Karki},
  \bibinfo{person}{Amit Shukla}, \bibinfo{person}{Laxmidhar Behera}, {and}
  \bibinfo{person}{Venkatesh K.S.}} \bibinfo{year}{2019}\natexlab{}.
\newblock \showarticletitle{DMP Based Trajectory Tracking for a Nonholonomic
  Mobile Robot With Automatic Goal Adaptation and Obstacle Avoidance}. In
  \bibinfo{booktitle}{\emph{2019 International Conference on Robotics and
  Automation (ICRA)}}. \bibinfo{pages}{8613--8619}.
\newblock
\urldef\tempurl%
\url{https://doi.org/10.1109/ICRA.2019.8793911}
\showDOI{\tempurl}


\bibitem[\protect\citeauthoryear{Shin and Ho}{Shin and Ho}{2018}]%
        {shin2018loop}
\bibfield{author}{\bibinfo{person}{Dong-Won Shin} {and}
  \bibinfo{person}{Yo-Sung Ho}.} \bibinfo{year}{2018}\natexlab{}.
\newblock \showarticletitle{Loop closure detection in simultaneous localization
  and mapping using learning based local patch descriptor}.
\newblock \bibinfo{journal}{\emph{Electronic Imaging}} \bibinfo{volume}{2018},
  \bibinfo{number}{17} (\bibinfo{year}{2018}), \bibinfo{pages}{284--1}.
\newblock


\bibitem[\protect\citeauthoryear{Simmons}{Simmons}{1996}]%
        {simmons1996curvature}
\bibfield{author}{\bibinfo{person}{Reid Simmons}.}
  \bibinfo{year}{1996}\natexlab{}.
\newblock \showarticletitle{The curvature-velocity method for local obstacle
  avoidance}. In \bibinfo{booktitle}{\emph{Proceedings of IEEE international
  conference on robotics and automation}}, Vol.~\bibinfo{volume}{4}. IEEE,
  \bibinfo{pages}{3375--3382}.
\newblock


\bibitem[\protect\citeauthoryear{Simpson, Cullip, and Revell}{Simpson
  et~al\mbox{.}}{2011}]%
        {simpson2011cheddar}
\bibfield{author}{\bibinfo{person}{R Simpson}, \bibinfo{person}{J Cullip},
  {and} \bibinfo{person}{J Revell}.} \bibinfo{year}{2011}\natexlab{}.
\newblock \showarticletitle{The cheddar gorge data set}.
\newblock \bibinfo{journal}{\emph{BAE Systems (Operations) Limited, UK, Tech.
  Rep.}} (\bibinfo{year}{2011}).
\newblock


\bibitem[\protect\citeauthoryear{Sitti}{Sitti}{2018}]%
        {sitti2018miniature}
\bibfield{author}{\bibinfo{person}{Metin Sitti}.}
  \bibinfo{year}{2018}\natexlab{}.
\newblock \showarticletitle{Miniature soft robots—road to the clinic}.
\newblock \bibinfo{journal}{\emph{Nature Reviews Materials}}
  \bibinfo{volume}{3}, \bibinfo{number}{6} (\bibinfo{year}{2018}),
  \bibinfo{pages}{74--75}.
\newblock


\bibitem[\protect\citeauthoryear{Smith, Self, and Cheeseman}{Smith
  et~al\mbox{.}}{1990}]%
        {smith1990estimating}
\bibfield{author}{\bibinfo{person}{Randall Smith}, \bibinfo{person}{Matthew
  Self}, {and} \bibinfo{person}{Peter Cheeseman}.}
  \bibinfo{year}{1990}\natexlab{}.
\newblock \showarticletitle{Estimating uncertain spatial relationships in
  robotics}.
\newblock In \bibinfo{booktitle}{\emph{Autonomous robot vehicles}}.
  \bibinfo{publisher}{Springer}, \bibinfo{pages}{167--193}.
\newblock


\bibitem[\protect\citeauthoryear{Song and Hsu}{Song and Hsu}{2021}]%
        {song2021tightly}
\bibfield{author}{\bibinfo{person}{Yang Song} {and} \bibinfo{person}{Li-Ta
  Hsu}.} \bibinfo{year}{2021}\natexlab{}.
\newblock \showarticletitle{Tightly coupled integrated navigation system via
  factor graph for UAV indoor localization}.
\newblock \bibinfo{journal}{\emph{Aerospace Science and Technology}}
  \bibinfo{volume}{108} (\bibinfo{year}{2021}), \bibinfo{pages}{106370}.
\newblock


\bibitem[\protect\citeauthoryear{Stentz}{Stentz}{1997}]%
        {stentz1997optimal}
\bibfield{author}{\bibinfo{person}{Anthony Stentz}.}
  \bibinfo{year}{1997}\natexlab{}.
\newblock \showarticletitle{Optimal and efficient path planning for partially
  known environments}.
\newblock In \bibinfo{booktitle}{\emph{Intelligent unmanned ground vehicles}}.
  \bibinfo{publisher}{Springer}, \bibinfo{pages}{203--220}.
\newblock


\bibitem[\protect\citeauthoryear{Stentz et~al\mbox{.}}{Stentz
  et~al\mbox{.}}{1995}]%
        {stentz1995focussed}
\bibfield{author}{\bibinfo{person}{Anthony Stentz} {et~al\mbox{.}}}
  \bibinfo{year}{1995}\natexlab{}.
\newblock \showarticletitle{The focussed {D}* algorithm for real-time
  replanning}. In \bibinfo{booktitle}{\emph{IJCAI}}, Vol.~\bibinfo{volume}{95}.
  \bibinfo{pages}{1652--1659}.
\newblock


\bibitem[\protect\citeauthoryear{Steux and El~Hamzaoui}{Steux and
  El~Hamzaoui}{2010}]%
        {steux2010tinyslam}
\bibfield{author}{\bibinfo{person}{Bruno Steux} {and} \bibinfo{person}{Oussama
  El~Hamzaoui}.} \bibinfo{year}{2010}\natexlab{}.
\newblock \showarticletitle{tinySLAM: A SLAM algorithm in less than 200 lines
  C-language program}. In \bibinfo{booktitle}{\emph{2010 11th International
  Conference on Control Automation Robotics \& Vision}}. IEEE,
  \bibinfo{pages}{1975--1979}.
\newblock


\bibitem[\protect\citeauthoryear{Strasdat, Montiel, and Davison}{Strasdat
  et~al\mbox{.}}{2010}]%
        {strasdat2010scale}
\bibfield{author}{\bibinfo{person}{Hauke Strasdat}, \bibinfo{person}{J
  Montiel}, {and} \bibinfo{person}{Andrew~J Davison}.}
  \bibinfo{year}{2010}\natexlab{}.
\newblock \showarticletitle{Scale drift-aware large scale monocular SLAM}.
\newblock \bibinfo{journal}{\emph{Robotics: Science and Systems VI}}
  \bibinfo{volume}{2}, \bibinfo{number}{3} (\bibinfo{year}{2010}),
  \bibinfo{pages}{7}.
\newblock


\bibitem[\protect\citeauthoryear{Stumm, Mei, Lacroix, Nieto, Hutter, and
  Siegwart}{Stumm et~al\mbox{.}}{2016}]%
        {stumm2016robust}
\bibfield{author}{\bibinfo{person}{Elena Stumm}, \bibinfo{person}{Christopher
  Mei}, \bibinfo{person}{Simon Lacroix}, \bibinfo{person}{Juan Nieto},
  \bibinfo{person}{Marco Hutter}, {and} \bibinfo{person}{Roland Siegwart}.}
  \bibinfo{year}{2016}\natexlab{}.
\newblock \showarticletitle{Robust visual place recognition with graph
  kernels}. In \bibinfo{booktitle}{\emph{Proceedings of the IEEE Conference on
  Computer Vision and Pattern Recognition}}. \bibinfo{pages}{4535--4544}.
\newblock


\bibitem[\protect\citeauthoryear{Sun, Liu, and Meng}{Sun et~al\mbox{.}}{2017}]%
        {sun2017improving}
\bibfield{author}{\bibinfo{person}{Yuxiang Sun}, \bibinfo{person}{Ming Liu},
  {and} \bibinfo{person}{Max Q-H Meng}.} \bibinfo{year}{2017}\natexlab{}.
\newblock \showarticletitle{Improving RGB-D SLAM in dynamic environments: A
  motion removal approach}.
\newblock \bibinfo{journal}{\emph{Robotics and Autonomous Systems}}
  \bibinfo{volume}{89} (\bibinfo{year}{2017}), \bibinfo{pages}{110--122}.
\newblock


\bibitem[\protect\citeauthoryear{Sun, Liu, and Meng}{Sun et~al\mbox{.}}{2018}]%
        {sun2018motion}
\bibfield{author}{\bibinfo{person}{Yuxiang Sun}, \bibinfo{person}{Ming Liu},
  {and} \bibinfo{person}{Max Q-H Meng}.} \bibinfo{year}{2018}\natexlab{}.
\newblock \showarticletitle{Motion removal for reliable RGB-D SLAM in dynamic
  environments}.
\newblock \bibinfo{journal}{\emph{Robotics and Autonomous Systems}}
  \bibinfo{volume}{108} (\bibinfo{year}{2018}), \bibinfo{pages}{115--128}.
\newblock


\bibitem[\protect\citeauthoryear{Taylor and LaValle}{Taylor and
  LaValle}{2009}]%
        {taylor2009bug}
\bibfield{author}{\bibinfo{person}{Kamilah Taylor} {and}
  \bibinfo{person}{Steven~M LaValle}.} \bibinfo{year}{2009}\natexlab{}.
\newblock \showarticletitle{I-Bug: An intensity-based bug algorithm}. In
  \bibinfo{booktitle}{\emph{2009 IEEE International Conference on Robotics and
  Automation}}. IEEE, \bibinfo{pages}{3981--3986}.
\newblock


\bibitem[\protect\citeauthoryear{Teller, Olson, and Leonard}{Teller
  et~al\mbox{.}}{2006}]%
        {teller2006fast}
\bibfield{author}{\bibinfo{person}{S Teller}, \bibinfo{person}{E Olson}, {and}
  \bibinfo{person}{J Leonard}.} \bibinfo{year}{2006}\natexlab{}.
\newblock \showarticletitle{Fast iterative optimization of pose graphs with
  poor inital estimates}. In \bibinfo{booktitle}{\emph{Proc. of IEEE
  International Conference on Robotics and Automation}}.
  \bibinfo{pages}{2262--2269}.
\newblock


\bibitem[\protect\citeauthoryear{Temizer}{Temizer}{2001}]%
        {temizer2001optical}
\bibfield{author}{\bibinfo{person}{Selim Temizer}.}
  \bibinfo{year}{2001}\natexlab{}.
\newblock \emph{\bibinfo{title}{Optical flow based local navigation}}.
\newblock \bibinfo{thesistype}{Master's\ thesis}.
  \bibinfo{school}{Massachusetts Institute of Technology}.
\newblock


\bibitem[\protect\citeauthoryear{Thakker, Alatur, Fan, Tordesillas, Paton,
  Otsu, Toupet, and Agha-mohammadi}{Thakker et~al\mbox{.}}{2020}]%
        {thakker2020autonomous}
\bibfield{author}{\bibinfo{person}{Rohan Thakker}, \bibinfo{person}{Nikhilesh
  Alatur}, \bibinfo{person}{David~D Fan}, \bibinfo{person}{Jesus Tordesillas},
  \bibinfo{person}{Michael Paton}, \bibinfo{person}{Kyohei Otsu},
  \bibinfo{person}{Olivier Toupet}, {and} \bibinfo{person}{Ali-akbar
  Agha-mohammadi}.} \bibinfo{year}{2020}\natexlab{}.
\newblock \showarticletitle{Autonomous off-road navigation over extreme
  terrains with perceptually-challenging conditions}. In
  \bibinfo{booktitle}{\emph{International Symposium on Experimental Robotics}}.
  Springer, \bibinfo{pages}{161--173}.
\newblock


\bibitem[\protect\citeauthoryear{Thorpe, Hebert, Kanade, and Shafer}{Thorpe
  et~al\mbox{.}}{1989}]%
        {10.1007/978-3-642-74585-0_6}
\bibfield{author}{\bibinfo{person}{Charles Thorpe}, \bibinfo{person}{Martial
  Hebert}, \bibinfo{person}{Takeo Kanade}, {and} \bibinfo{person}{Steven
  Shafer}.} \bibinfo{year}{1989}\natexlab{}.
\newblock \showarticletitle{Vision and Navigation for the Carnegie Mellon
  Navlab}. In \bibinfo{booktitle}{\emph{High Precision Navigation}},
  \bibfield{editor}{\bibinfo{person}{Klaus Linkwitz} {and}
  \bibinfo{person}{Ulrich Hangleiter}} (Eds.). \bibinfo{publisher}{Springer
  Berlin Heidelberg}, \bibinfo{address}{Berlin, Heidelberg},
  \bibinfo{pages}{97--122}.
\newblock
\showISBNx{978-3-642-74585-0}


\bibitem[\protect\citeauthoryear{Thrun}{Thrun}{1995}]%
        {thrun1995exploration}
\bibfield{author}{\bibinfo{person}{Sebastian Thrun}.}
  \bibinfo{year}{1995}\natexlab{}.
\newblock \showarticletitle{Exploration in active learning}.
\newblock \bibinfo{journal}{\emph{Handbook of Brain Science and Neural
  Networks}} (\bibinfo{year}{1995}), \bibinfo{pages}{381--384}.
\newblock


\bibitem[\protect\citeauthoryear{Thrun}{Thrun}{2001}]%
        {thrun2001learning}
\bibfield{author}{\bibinfo{person}{Sebastian Thrun}.}
  \bibinfo{year}{2001}\natexlab{}.
\newblock \showarticletitle{Learning occupancy grids with forward models}. In
  \bibinfo{booktitle}{\emph{Proceedings 2001 IEEE/RSJ International Conference
  on Intelligent Robots and Systems. Expanding the Societal Role of Robotics in
  the the Next Millennium (Cat. No. 01CH37180)}}, Vol.~\bibinfo{volume}{3}.
  IEEE, \bibinfo{pages}{1676--1681}.
\newblock


\bibitem[\protect\citeauthoryear{Thrun}{Thrun}{2002}]%
        {thrun2002probabilistic}
\bibfield{author}{\bibinfo{person}{Sebastian Thrun}.}
  \bibinfo{year}{2002}\natexlab{}.
\newblock \showarticletitle{Probabilistic robotics}.
\newblock \bibinfo{journal}{\emph{Commun. ACM}} \bibinfo{volume}{45},
  \bibinfo{number}{3} (\bibinfo{year}{2002}), \bibinfo{pages}{52--57}.
\newblock


\bibitem[\protect\citeauthoryear{Thrun, Montemerlo, Dahlkamp, Stavens, Aron,
  Diebel, Fong, Gale, Halpenny, Hoffmann, et~al\mbox{.}}{Thrun
  et~al\mbox{.}}{2006}]%
        {thrun2006stanley}
\bibfield{author}{\bibinfo{person}{Sebastian Thrun}, \bibinfo{person}{Mike
  Montemerlo}, \bibinfo{person}{Hendrik Dahlkamp}, \bibinfo{person}{David
  Stavens}, \bibinfo{person}{Andrei Aron}, \bibinfo{person}{James Diebel},
  \bibinfo{person}{Philip Fong}, \bibinfo{person}{John Gale},
  \bibinfo{person}{Morgan Halpenny}, \bibinfo{person}{Gabriel Hoffmann},
  {et~al\mbox{.}}} \bibinfo{year}{2006}\natexlab{}.
\newblock \showarticletitle{Stanley: The robot that won the DARPA Grand
  Challenge}.
\newblock \bibinfo{journal}{\emph{Journal of field Robotics}}
  \bibinfo{volume}{23}, \bibinfo{number}{9} (\bibinfo{year}{2006}),
  \bibinfo{pages}{661--692}.
\newblock


\bibitem[\protect\citeauthoryear{Titterton, Weston, and Weston}{Titterton
  et~al\mbox{.}}{2004}]%
        {titterton2004strapdown}
\bibfield{author}{\bibinfo{person}{David Titterton}, \bibinfo{person}{John~L
  Weston}, {and} \bibinfo{person}{John Weston}.}
  \bibinfo{year}{2004}\natexlab{}.
\newblock \bibinfo{booktitle}{\emph{Strapdown inertial navigation technology}}.
  Vol.~\bibinfo{volume}{17}.
\newblock \bibinfo{publisher}{IET}.
\newblock


\bibitem[\protect\citeauthoryear{Tomita, Zaffar, Milford, McDonald{-}Maier, and
  Ehsan}{Tomita et~al\mbox{.}}{2021}]%
        {DBLP:journals/corr/abs-2103-01994}
\bibfield{author}{\bibinfo{person}{Mihnea{-}Alexandru Tomita},
  \bibinfo{person}{Mubariz Zaffar}, \bibinfo{person}{Michael Milford},
  \bibinfo{person}{Klaus~D. McDonald{-}Maier}, {and} \bibinfo{person}{Shoaib
  Ehsan}.} \bibinfo{year}{2021}\natexlab{}.
\newblock \showarticletitle{Sequence-Based Filtering for Visual Route-Based
  Navigation: Analysing the Benefits, Trade-offs and Design Choices}.
\newblock \bibinfo{journal}{\emph{CoRR}}  \bibinfo{volume}{abs/2103.01994}
  (\bibinfo{year}{2021}).
\newblock
\showeprint[arXiv]{2103.01994}
\urldef\tempurl%
\url{https://arxiv.org/abs/2103.01994}
\showURL{%
\tempurl}


\bibitem[\protect\citeauthoryear{Triebel, Pfaff, and Burgard}{Triebel
  et~al\mbox{.}}{2006}]%
        {triebel2006multi}
\bibfield{author}{\bibinfo{person}{Rudolph Triebel}, \bibinfo{person}{Patrick
  Pfaff}, {and} \bibinfo{person}{Wolfram Burgard}.}
  \bibinfo{year}{2006}\natexlab{}.
\newblock \showarticletitle{Multi-level surface maps for outdoor terrain
  mapping and loop closing}. In \bibinfo{booktitle}{\emph{2006 IEEE/RSJ
  international conference on intelligent robots and systems}}. IEEE,
  \bibinfo{pages}{2276--2282}.
\newblock


\bibitem[\protect\citeauthoryear{Tzafestas}{Tzafestas}{2018}]%
        {tzafestas2018mobile}
\bibfield{author}{\bibinfo{person}{Spyros~G Tzafestas}.}
  \bibinfo{year}{2018}\natexlab{}.
\newblock \showarticletitle{Mobile robot control and navigation: A global
  overview}.
\newblock \bibinfo{journal}{\emph{Journal of Intelligent \& Robotic Systems}}
  \bibinfo{volume}{91}, \bibinfo{number}{1} (\bibinfo{year}{2018}),
  \bibinfo{pages}{35--58}.
\newblock


\bibitem[\protect\citeauthoryear{Ulrich and Borenstein}{Ulrich and
  Borenstein}{1998}]%
        {ulrich1998vfh+}
\bibfield{author}{\bibinfo{person}{Iwan Ulrich} {and} \bibinfo{person}{Johann
  Borenstein}.} \bibinfo{year}{1998}\natexlab{}.
\newblock \showarticletitle{VFH+: Reliable obstacle avoidance for fast mobile
  robots}. In \bibinfo{booktitle}{\emph{Proceedings. 1998 IEEE international
  conference on robotics and automation (Cat. No. 98CH36146)}},
  Vol.~\bibinfo{volume}{2}. IEEE, \bibinfo{pages}{1572--1577}.
\newblock


\bibitem[\protect\citeauthoryear{Urmson, Anhalt, Bagnell, Baker, Bittner,
  Clark, Dolan, Duggins, Galatali, Geyer, et~al\mbox{.}}{Urmson
  et~al\mbox{.}}{2008}]%
        {urmson2008autonomous}
\bibfield{author}{\bibinfo{person}{Chris Urmson}, \bibinfo{person}{Joshua
  Anhalt}, \bibinfo{person}{Drew Bagnell}, \bibinfo{person}{Christopher Baker},
  \bibinfo{person}{Robert Bittner}, \bibinfo{person}{MN Clark},
  \bibinfo{person}{John Dolan}, \bibinfo{person}{Dave Duggins},
  \bibinfo{person}{Tugrul Galatali}, \bibinfo{person}{Chris Geyer},
  {et~al\mbox{.}}} \bibinfo{year}{2008}\natexlab{}.
\newblock \showarticletitle{Autonomous driving in urban environments: Boss and
  the urban challenge}.
\newblock \bibinfo{journal}{\emph{Journal of field Robotics}}
  \bibinfo{volume}{25}, \bibinfo{number}{8} (\bibinfo{year}{2008}),
  \bibinfo{pages}{425--466}.
\newblock


\bibitem[\protect\citeauthoryear{Urmson, Bagnell, Baker, Hebert, Kelly,
  Rajkumar, Rybski, Scherer, Simmons, Singh, et~al\mbox{.}}{Urmson
  et~al\mbox{.}}{2007}]%
        {urmson2007tartan}
\bibfield{author}{\bibinfo{person}{Chris Urmson}, \bibinfo{person}{J~Andrew
  Bagnell}, \bibinfo{person}{Christopher Baker}, \bibinfo{person}{Martial
  Hebert}, \bibinfo{person}{Alonzo Kelly}, \bibinfo{person}{Raj Rajkumar},
  \bibinfo{person}{Paul~E Rybski}, \bibinfo{person}{Sebastian Scherer},
  \bibinfo{person}{Reid Simmons}, \bibinfo{person}{Sanjiv Singh},
  {et~al\mbox{.}}} \bibinfo{year}{2007}\natexlab{}.
\newblock \showarticletitle{Tartan racing: A multi-modal approach to the darpa
  urban challenge}.
\newblock  (\bibinfo{year}{2007}).
\newblock


\bibitem[\protect\citeauthoryear{Victerpaul, Saravanan, Janakiraman, and
  Pradeep}{Victerpaul et~al\mbox{.}}{2017}]%
        {victerpaul2017path}
\bibfield{author}{\bibinfo{person}{P Victerpaul}, \bibinfo{person}{D
  Saravanan}, \bibinfo{person}{S Janakiraman}, {and} \bibinfo{person}{J
  Pradeep}.} \bibinfo{year}{2017}\natexlab{}.
\newblock \showarticletitle{Path planning of autonomous mobile robots: A survey
  and comparison}.
\newblock \bibinfo{journal}{\emph{Journal of Advanced Research in Dynamical and
  Control Systems}} \bibinfo{volume}{9}, \bibinfo{number}{12}
  (\bibinfo{year}{2017}), \bibinfo{pages}{1535--1565}.
\newblock


\bibitem[\protect\citeauthoryear{Volpe, Balaram, Ohm, and Ivlev}{Volpe
  et~al\mbox{.}}{1996}]%
        {volpe1996rocky}
\bibfield{author}{\bibinfo{person}{Richard Volpe}, \bibinfo{person}{J Balaram},
  \bibinfo{person}{Timothy Ohm}, {and} \bibinfo{person}{Robert Ivlev}.}
  \bibinfo{year}{1996}\natexlab{}.
\newblock \showarticletitle{The rocky 7 mars rover prototype}. In
  \bibinfo{booktitle}{\emph{Proceedings of IEEE/RSJ International Conference on
  Intelligent Robots and Systems. IROS'96}}, Vol.~\bibinfo{volume}{3}. IEEE,
  \bibinfo{pages}{1558--1564}.
\newblock


\bibitem[\protect\citeauthoryear{Vonk}{Vonk}{2022}]%
        {vonk2022graph}
\bibfield{author}{\bibinfo{person}{Arjan Vonk}.}
  \bibinfo{year}{2022}\natexlab{}.
\newblock \showarticletitle{Graph Based LiDAR-Inertial Lo-calization with a Low
  Power Solid State LiDAR}.
\newblock  (\bibinfo{year}{2022}).
\newblock


\bibitem[\protect\citeauthoryear{Wan, Van Der~Merwe, and Haykin}{Wan
  et~al\mbox{.}}{2001}]%
        {wan2001unscented}
\bibfield{author}{\bibinfo{person}{Eric~A Wan}, \bibinfo{person}{Rudolph Van
  Der~Merwe}, {and} \bibinfo{person}{Simon Haykin}.}
  \bibinfo{year}{2001}\natexlab{}.
\newblock \showarticletitle{The unscented Kalman filter}.
\newblock \bibinfo{journal}{\emph{Kalman filtering and neural networks}}
  \bibinfo{volume}{5}, \bibinfo{number}{2007} (\bibinfo{year}{2001}),
  \bibinfo{pages}{221--280}.
\newblock


\bibitem[\protect\citeauthoryear{Wang, Zhang, Liu, and Zhang}{Wang
  et~al\mbox{.}}{2022}]%
        {wang2022high}
\bibfield{author}{\bibinfo{person}{Qingshan Wang}, \bibinfo{person}{Jun Zhang},
  \bibinfo{person}{Yuansheng Liu}, {and} \bibinfo{person}{Xinchen Zhang}.}
  \bibinfo{year}{2022}\natexlab{}.
\newblock \showarticletitle{High-Precision and Fast LiDAR Odometry and Mapping
  Algorithm}.
\newblock \bibinfo{journal}{\emph{Journal of Advanced Computational
  Intelligence and Intelligent Informatics}} \bibinfo{volume}{26},
  \bibinfo{number}{2} (\bibinfo{year}{2022}), \bibinfo{pages}{206--216}.
\newblock


\bibitem[\protect\citeauthoryear{Welch, Bishop, et~al\mbox{.}}{Welch
  et~al\mbox{.}}{1995}]%
        {welch1995introduction}
\bibfield{author}{\bibinfo{person}{Greg Welch}, \bibinfo{person}{Gary Bishop},
  {et~al\mbox{.}}} \bibinfo{year}{1995}\natexlab{}.
\newblock \showarticletitle{An introduction to the Kalman filter}.
\newblock  (\bibinfo{year}{1995}).
\newblock


\bibitem[\protect\citeauthoryear{Wenzel, Sch{\"o}n, Leal-Taix{\'e}, and
  Cremers}{Wenzel et~al\mbox{.}}{2021}]%
        {wenzel2021vision}
\bibfield{author}{\bibinfo{person}{Patrick Wenzel}, \bibinfo{person}{Torsten
  Sch{\"o}n}, \bibinfo{person}{Laura Leal-Taix{\'e}}, {and}
  \bibinfo{person}{Daniel Cremers}.} \bibinfo{year}{2021}\natexlab{}.
\newblock \showarticletitle{Vision-based mobile robotics obstacle avoidance
  with deep reinforcement learning}. In \bibinfo{booktitle}{\emph{2021 IEEE
  International Conference on Robotics and Automation (ICRA)}}. IEEE,
  \bibinfo{pages}{14360--14366}.
\newblock


\bibitem[\protect\citeauthoryear{Xiang, Jouvencel, and Parodi}{Xiang
  et~al\mbox{.}}{2010}]%
        {xiang2010coordinated}
\bibfield{author}{\bibinfo{person}{Xianbo Xiang}, \bibinfo{person}{Bruno
  Jouvencel}, {and} \bibinfo{person}{Olivier Parodi}.}
  \bibinfo{year}{2010}\natexlab{}.
\newblock \showarticletitle{Coordinated formation control of multiple
  autonomous underwater vehicles for pipeline inspection}.
\newblock \bibinfo{journal}{\emph{International Journal of Advanced Robotic
  Systems}} \bibinfo{volume}{7}, \bibinfo{number}{1} (\bibinfo{year}{2010}),
  \bibinfo{pages}{3}.
\newblock


\bibitem[\protect\citeauthoryear{Xu, Chavez, B{\"u}low, Birk, and
  Schwertfeger}{Xu et~al\mbox{.}}{2019}]%
        {xu2019improved}
\bibfield{author}{\bibinfo{person}{Qingwen Xu}, \bibinfo{person}{Arturo~Gomez
  Chavez}, \bibinfo{person}{Heiko B{\"u}low}, \bibinfo{person}{Andreas Birk},
  {and} \bibinfo{person}{S{\"o}ren Schwertfeger}.}
  \bibinfo{year}{2019}\natexlab{}.
\newblock \showarticletitle{Improved fourier mellin invariant for robust
  rotation estimation with omni-cameras}. In \bibinfo{booktitle}{\emph{2019
  IEEE International Conference on Image Processing (ICIP)}}. IEEE,
  \bibinfo{pages}{320--324}.
\newblock


\bibitem[\protect\citeauthoryear{Xu, Kuang, Kneip, and Schwertfeger}{Xu
  et~al\mbox{.}}{2021}]%
        {xu2021rethinking}
\bibfield{author}{\bibinfo{person}{Qingwen Xu}, \bibinfo{person}{Haofei Kuang},
  \bibinfo{person}{Laurent Kneip}, {and} \bibinfo{person}{S{\"o}ren
  Schwertfeger}.} \bibinfo{year}{2021}\natexlab{}.
\newblock \showarticletitle{Rethinking the Fourier-Mellin Transform: Multiple
  Depths in the Camera’s View}.
\newblock \bibinfo{journal}{\emph{Remote Sensing}} \bibinfo{volume}{13},
  \bibinfo{number}{5} (\bibinfo{year}{2021}), \bibinfo{pages}{1000}.
\newblock


\bibitem[\protect\citeauthoryear{Yamauchi}{Yamauchi}{2004}]%
        {packbot}
\bibfield{author}{\bibinfo{person}{Brian~M. Yamauchi}.}
  \bibinfo{year}{2004}\natexlab{}.
\newblock \showarticletitle{{PackBot: a versatile platform for military
  robotics}}. In \bibinfo{booktitle}{\emph{Unmanned Ground Vehicle Technology
  VI}}, \bibfield{editor}{\bibinfo{person}{Grant~R. Gerhart},
  \bibinfo{person}{Chuck~M. Shoemaker}, {and} \bibinfo{person}{Douglas~W.
  Gage}} (Eds.), Vol.~\bibinfo{volume}{5422}. International Society for Optics
  and Photonics, \bibinfo{publisher}{SPIE}, \bibinfo{pages}{228 -- 237}.
\newblock
\urldef\tempurl%
\url{https://doi.org/10.1117/12.538328}
\showDOI{\tempurl}


\bibitem[\protect\citeauthoryear{Yang}{Yang}{2012}]%
        {yang2012efficient}
\bibfield{author}{\bibinfo{person}{Po Yang}.} \bibinfo{year}{2012}\natexlab{}.
\newblock \showarticletitle{Efficient particle filter algorithm for ultrasonic
  sensor-based 2D range-only simultaneous localisation and mapping
  application}.
\newblock \bibinfo{journal}{\emph{IET Wireless Sensor Systems}}
  \bibinfo{volume}{2}, \bibinfo{number}{4} (\bibinfo{year}{2012}),
  \bibinfo{pages}{394--401}.
\newblock


\bibitem[\protect\citeauthoryear{Yang, Qu, Xu, Zou, He, and Sun}{Yang
  et~al\mbox{.}}{2018}]%
        {yang2018mobile}
\bibfield{author}{\bibinfo{person}{Qifeng Yang}, \bibinfo{person}{Daokui Qu},
  \bibinfo{person}{Fang Xu}, \bibinfo{person}{Fengshan Zou},
  \bibinfo{person}{Guojian He}, {and} \bibinfo{person}{Mingze Sun}.}
  \bibinfo{year}{2018}\natexlab{}.
\newblock \showarticletitle{Mobile robot motion control and autonomous
  navigation in GPS-denied outdoor environments using 3D laser scanning}.
\newblock \bibinfo{journal}{\emph{Assembly Automation}} (\bibinfo{year}{2018}).
\newblock


\bibitem[\protect\citeauthoryear{Yin and Berger}{Yin and Berger}{2017}]%
        {yin2017use}
\bibfield{author}{\bibinfo{person}{Hang Yin} {and} \bibinfo{person}{Christian
  Berger}.} \bibinfo{year}{2017}\natexlab{}.
\newblock \showarticletitle{When to use what data set for your self-driving car
  algorithm: An overview of publicly available driving datasets}. In
  \bibinfo{booktitle}{\emph{2017 IEEE 20th International Conference on
  Intelligent Transportation Systems (ITSC)}}. IEEE, \bibinfo{pages}{1--8}.
\newblock


\bibitem[\protect\citeauthoryear{Yousif, Bab-Hadiashar, and
  Hoseinnezhad}{Yousif et~al\mbox{.}}{2015}]%
        {yousif2015overview}
\bibfield{author}{\bibinfo{person}{Khalid Yousif}, \bibinfo{person}{Alireza
  Bab-Hadiashar}, {and} \bibinfo{person}{Reza Hoseinnezhad}.}
  \bibinfo{year}{2015}\natexlab{}.
\newblock \showarticletitle{An overview to visual odometry and visual SLAM:
  Applications to mobile robotics}.
\newblock \bibinfo{journal}{\emph{Intelligent Industrial Systems}}
  \bibinfo{volume}{1}, \bibinfo{number}{4} (\bibinfo{year}{2015}),
  \bibinfo{pages}{289--311}.
\newblock


\bibitem[\protect\citeauthoryear{Yu, Shao, Chen, Wu, Fan, Mo, and Dong}{Yu
  et~al\mbox{.}}{2021}]%
        {yu2021roboassembly}
\bibfield{author}{\bibinfo{person}{Mingxin Yu}, \bibinfo{person}{Lin Shao},
  \bibinfo{person}{Zhehuan Chen}, \bibinfo{person}{Tianhao Wu},
  \bibinfo{person}{Qingnan Fan}, \bibinfo{person}{Kaichun Mo}, {and}
  \bibinfo{person}{Hao Dong}.} \bibinfo{year}{2021}\natexlab{}.
\newblock \showarticletitle{RoboAssembly: Learning Generalizable Furniture
  Assembly Policy in a Novel Multi-robot Contact-rich Simulation Environment}.
\newblock \bibinfo{journal}{\emph{arXiv preprint arXiv:2112.10143}}
  (\bibinfo{year}{2021}).
\newblock


\bibitem[\protect\citeauthoryear{Zahran, Moussa, Sesay, and El-Sheimy}{Zahran
  et~al\mbox{.}}{2019}]%
        {8594610}
\bibfield{author}{\bibinfo{person}{Shady Zahran}, \bibinfo{person}{Adel~M.
  Moussa}, \bibinfo{person}{Abu~B. Sesay}, {and} \bibinfo{person}{Naser
  El-Sheimy}.} \bibinfo{year}{2019}\natexlab{}.
\newblock \showarticletitle{A New Velocity Meter Based on Hall Effect Sensors
  for UAV Indoor Navigation}.
\newblock \bibinfo{journal}{\emph{IEEE Sensors Journal}} \bibinfo{volume}{19},
  \bibinfo{number}{8} (\bibinfo{year}{2019}), \bibinfo{pages}{3067--3076}.
\newblock
\urldef\tempurl%
\url{https://doi.org/10.1109/JSEN.2018.2890094}
\showDOI{\tempurl}


\bibitem[\protect\citeauthoryear{Zghair and Al-Araji}{Zghair and
  Al-Araji}{2021}]%
        {zghair2021one}
\bibfield{author}{\bibinfo{person}{Noor Abdul~Khaleq Zghair} {and}
  \bibinfo{person}{Ahmed~S Al-Araji}.} \bibinfo{year}{2021}\natexlab{}.
\newblock \showarticletitle{A one decade survey of autonomous mobile robot
  systems}.
\newblock \bibinfo{journal}{\emph{International Journal of Electrical and
  Computer Engineering}} \bibinfo{volume}{11}, \bibinfo{number}{6}
  (\bibinfo{year}{2021}), \bibinfo{pages}{4891}.
\newblock


\bibitem[\protect\citeauthoryear{Zhang and Hsu}{Zhang and Hsu}{2018}]%
        {zhang2018intelligent}
\bibfield{author}{\bibinfo{person}{Guohao Zhang} {and} \bibinfo{person}{Li-Ta
  Hsu}.} \bibinfo{year}{2018}\natexlab{}.
\newblock \showarticletitle{Intelligent GNSS/INS integrated navigation system
  for a commercial UAV flight control system}.
\newblock \bibinfo{journal}{\emph{Aerospace science and technology}}
  \bibinfo{volume}{80} (\bibinfo{year}{2018}), \bibinfo{pages}{368--380}.
\newblock


\bibitem[\protect\citeauthoryear{Zhang, Zhang, Wang, Liu, Yang, Zhou, and
  Bian}{Zhang et~al\mbox{.}}{2021}]%
        {zhang2021subsea}
\bibfield{author}{\bibinfo{person}{Hongwei Zhang}, \bibinfo{person}{Shitong
  Zhang}, \bibinfo{person}{Yanhui Wang}, \bibinfo{person}{Yuhong Liu},
  \bibinfo{person}{Yanan Yang}, \bibinfo{person}{Tian Zhou}, {and}
  \bibinfo{person}{Hongyu Bian}.} \bibinfo{year}{2021}\natexlab{}.
\newblock \showarticletitle{Subsea pipeline leak inspection by autonomous
  underwater vehicle}.
\newblock \bibinfo{journal}{\emph{Applied Ocean Research}}
  \bibinfo{volume}{107} (\bibinfo{year}{2021}), \bibinfo{pages}{102321}.
\newblock


\bibitem[\protect\citeauthoryear{Zhang, Su, and Zhu}{Zhang
  et~al\mbox{.}}{2017}]%
        {zhang2017loop}
\bibfield{author}{\bibinfo{person}{Xiwu Zhang}, \bibinfo{person}{Yan Su}, {and}
  \bibinfo{person}{Xinhua Zhu}.} \bibinfo{year}{2017}\natexlab{}.
\newblock \showarticletitle{Loop closure detection for visual SLAM systems
  using convolutional neural network}. In \bibinfo{booktitle}{\emph{2017 23rd
  International Conference on Automation and Computing (ICAC)}}. IEEE,
  \bibinfo{pages}{1--6}.
\newblock


\bibitem[\protect\citeauthoryear{Zhang, Kayacan, Thompson, and Chowdhary}{Zhang
  et~al\mbox{.}}{2020}]%
        {zhang2020high}
\bibfield{author}{\bibinfo{person}{Zhongzhong Zhang}, \bibinfo{person}{Erkan
  Kayacan}, \bibinfo{person}{Benjamin Thompson}, {and} \bibinfo{person}{Girish
  Chowdhary}.} \bibinfo{year}{2020}\natexlab{}.
\newblock \showarticletitle{High precision control and deep learning-based corn
  stand counting algorithms for agricultural robot}.
\newblock \bibinfo{journal}{\emph{Autonomous Robots}} \bibinfo{volume}{44},
  \bibinfo{number}{7} (\bibinfo{year}{2020}), \bibinfo{pages}{1289--1302}.
\newblock


\bibitem[\protect\citeauthoryear{Zhou, Wei, and Tan}{Zhou
  et~al\mbox{.}}{2003}]%
        {1241767}
\bibfield{author}{\bibinfo{person}{Chao Zhou}, \bibinfo{person}{Yucheng Wei},
  {and} \bibinfo{person}{Tieniu Tan}.} \bibinfo{year}{2003}\natexlab{}.
\newblock \showarticletitle{Mobile robot self-localization based on global
  visual appearance features}. In \bibinfo{booktitle}{\emph{2003 IEEE
  International Conference on Robotics and Automation (Cat. No.03CH37422)}},
  Vol.~\bibinfo{volume}{1}. \bibinfo{pages}{1271--1276 vol.1}.
\newblock
\urldef\tempurl%
\url{https://doi.org/10.1109/ROBOT.2003.1241767}
\showDOI{\tempurl}


\bibitem[\protect\citeauthoryear{Zhou, Shiju, Cao, and Dong}{Zhou
  et~al\mbox{.}}{2016}]%
        {zhou2016review}
\bibfield{author}{\bibinfo{person}{Wu Zhou}, \bibinfo{person}{E Shiju},
  \bibinfo{person}{Zhenxin Cao}, {and} \bibinfo{person}{Ying Dong}.}
  \bibinfo{year}{2016}\natexlab{}.
\newblock \showarticletitle{Review of slam data association study}. In
  \bibinfo{booktitle}{\emph{Proceedings of the 2016 International Conference on
  Sensor Network and Computer Engineering, Xian, China}}.
  \bibinfo{pages}{8--10}.
\newblock


\bibitem[\protect\citeauthoryear{Zhu, Zhang, Song, and Li}{Zhu
  et~al\mbox{.}}{2012}]%
        {zhu2012new}
\bibfield{author}{\bibinfo{person}{Yi Zhu}, \bibinfo{person}{Tao Zhang},
  \bibinfo{person}{Jingyan Song}, {and} \bibinfo{person}{Xiaqin Li}.}
  \bibinfo{year}{2012}\natexlab{}.
\newblock \showarticletitle{A new bug-type navigation algorithm for mobile
  robots in unknown environments containing moving obstacles}.
\newblock \bibinfo{journal}{\emph{Industrial Robot: An International Journal}}
  \bibinfo{volume}{39}, \bibinfo{number}{1} (\bibinfo{year}{2012}),
  \bibinfo{pages}{27--39}.
\newblock


\bibitem[\protect\citeauthoryear{Zubizarreta, Aguinaga, and
  Montiel}{Zubizarreta et~al\mbox{.}}{2020}]%
        {zubizarreta2020direct}
\bibfield{author}{\bibinfo{person}{Jon Zubizarreta}, \bibinfo{person}{Iker
  Aguinaga}, {and} \bibinfo{person}{Jose Maria~Martinez Montiel}.}
  \bibinfo{year}{2020}\natexlab{}.
\newblock \showarticletitle{Direct sparse mapping}.
\newblock \bibinfo{journal}{\emph{IEEE Transactions on Robotics}}
  \bibinfo{volume}{36}, \bibinfo{number}{4} (\bibinfo{year}{2020}),
  \bibinfo{pages}{1363--1370}.
\newblock


\end{thebibliography}



\end{document}